\documentclass[11pt]{article}
\usepackage[preprint]{acl}
\usepackage{times}
\usepackage{latexsym}

\usepackage[T1]{fontenc}
\usepackage[utf8]{inputenc}
\usepackage{microtype}
\usepackage{inconsolata}
\usepackage{graphicx}
\usepackage{amsmath}
\usepackage{amsfonts}
\usepackage{multirow}
\usepackage{booktabs}
\usepackage{tabularx}
\usepackage{adjustbox}
\usepackage{subcaption}
\usepackage{enumitem}
\usepackage{float}
\usepackage{bm}

\usepackage{mathtools}
\usepackage{amssymb}

\newcommand\blfootnote[1]{
  \begingroup
  \renewcommand\thefootnote{}\footnote{#1}
  \addtocounter{footnote}{-1}
  \endgroup
}

\usepackage{comment}

\usepackage[whole]{bxcjkjatype}

\usepackage{tikz}

\newcommand{\bgblue}[1]{
  \tikz[baseline=(X.base)]{
    \node(X)[rectangle, fill=blue!25, rounded corners,
      inner xsep=0.55pt, inner ysep=1.4pt,
      text height=1.8ex, text depth=0.2ex, draw=white]{#1};}}

\newcommand{\bggray}[1]{
  \tikz[baseline=(X.base)]{
    \node(X)[rectangle, fill=gray!25, rounded corners,
      inner xsep=0.55pt, inner ysep=1.4pt,
      text height=1.8ex, text depth=0.2ex, draw=white]{#1};}}

\newcommand{\bgred}[1]{
  \tikz[baseline=(X.base)]{
    \node(X)[rectangle, fill=red!25, rounded corners,
      inner xsep=0.55pt, inner ysep=1.4pt,
      text height=1.8ex, text depth=0.2ex, draw=white]{#1};}}

\newcommand{\bggreen}[1]{
  \tikz[baseline=(X.base)]{
    \node(X)[rectangle, fill=green!25, rounded corners,
      inner xsep=0.55pt, inner ysep=1.4pt,
      text height=1.8ex, text depth=0.2ex, draw=white]{#1};}}

\title{Measuring Affinity between Attention-Head Weight Subspaces\\via the Projection Kernel}

\author{
  Hiroaki Yamagiwa${}^{1}$\thanks{Work done while at Kyoto University.} \qquad 
  Yusuke Takase${}^{2}$ \qquad
  Hidetoshi Shimodaira${}^{2,3}$ \\
  ${}^1\,$SB Intuitions \qquad ${}^2\,$Kyoto University \qquad ${}^3\,$RIKEN AIP\\
  \texttt{hiroaki.yamagiwa@sbintuitions.co.jp, y.takase@sys.i.kyoto-u.ac.jp,}\\
  \texttt{shimo@i.kyoto-u.ac.jp}
}

\begin{document}
\maketitle

\begin{abstract}
Understanding relationships between attention heads is essential for interpreting the internal structure of Transformers, yet existing metrics do not capture this structure well. 
We focus on the subspaces spanned by attention-head weight matrices and quantify head-to-head relationships using the Projection Kernel (PK), a principal-angle-based measure of subspace similarity. 
Experiments on GPT2-small, BLOOM-560M, and Qwen2-0.5B-Instruct show that PK reproduces known head-to-head interactions on the IOI task more clearly than prior metrics such as the Composition Score. 
We further introduce a framework to quantify the informativeness of PK distributions by comparing them with a reference distribution derived from random orthogonal subspaces. 
As an application, we analyze a directed graph constructed from PK and show that, in GPT2-small, L4H7 acts as a hub by functioning as an Identity Head.
\blfootnote{
Our code is available at \url{https://github.com/ymgw55/proj-kernel-attn-subspace}.
}
\end{abstract}

\section{Introduction}
\begin{figure}[t!]
    \centering
    \includegraphics[width=0.97\columnwidth]{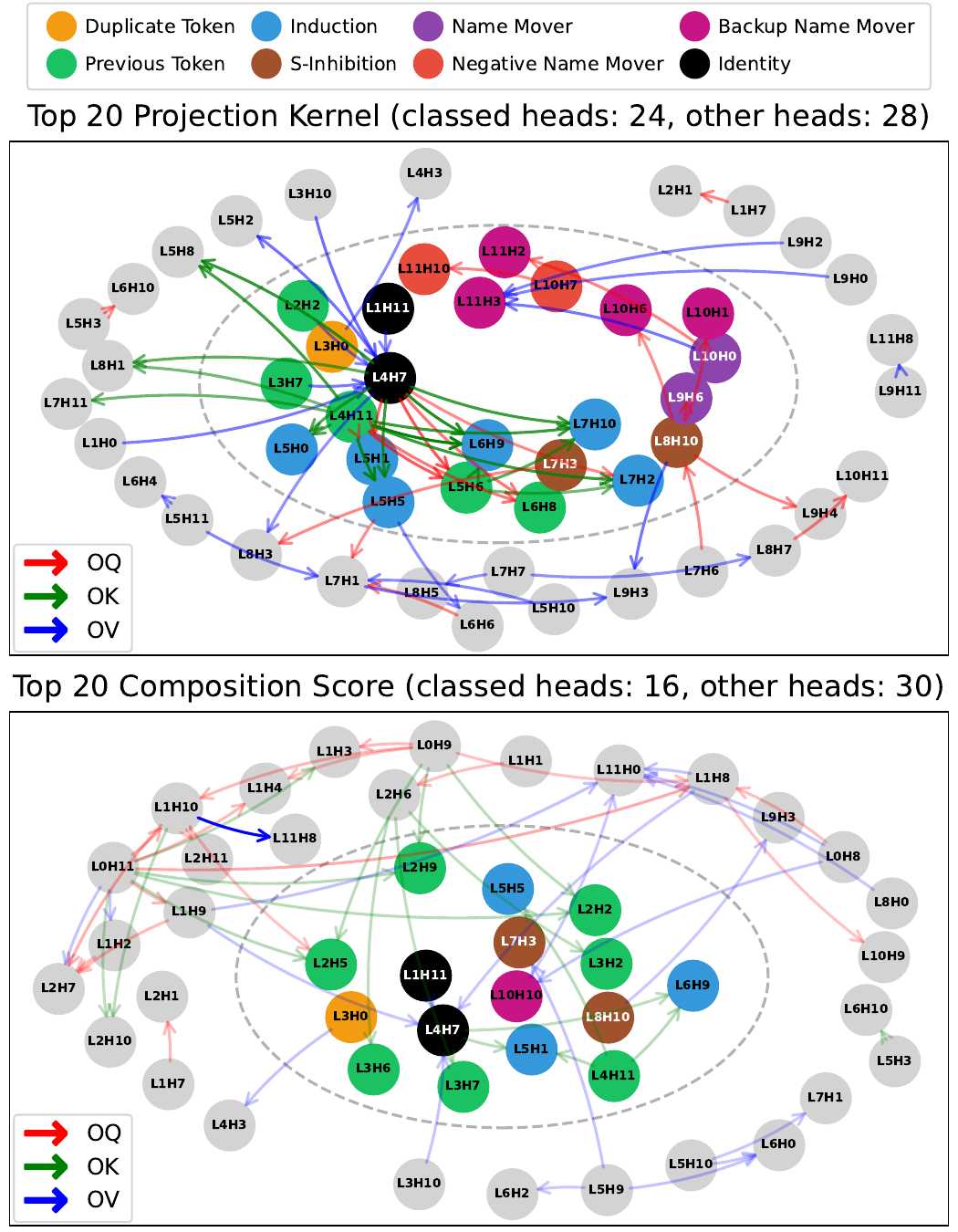}
\caption{
For the attention-head weights of GPT2-small, we compute similarities between heads using (top) the Projection Kernel (PK) and (bottom) the Composition Score (CS), and visualize the top 20 scores as edges for each weight pairing (OQ, OK, and OV).
Edge opacity corresponds to the score normalized by the maximum score.
The annotations for the seven head classes other than Identity Heads follow \citet{wang2023interpretability}.
With PK, many heads from IOI-relevant head classes appear, and edges concentrate among these heads.
In contrast, this tendency is weaker with CS; instead, many edges connect heads that do not belong to any head class.
}
\label{fig:intro}
\end{figure}
Mechanistic interpretability~\cite{mech-interp-essay} is a research field that aims to understand the internal computations of language models. 
Recently, many studies have identified functional modules by probing models with specific input sentences, exemplified by the Indirect Object Identification (IOI) task~\cite{wang2023interpretability} (see Appendix~\ref{app:related-work} for related work). 
Such analyses capture model behavior at a microscopic level, but the resulting insights depend on the particular inputs.

In contrast, weight-based analyses can reveal global properties of a model without relying on any specific input~\cite{ameisen2025circuit}. 
In particular, \citet{elhage2021mathematical} introduced the Composition Score (CS), which quantifies the similarity between heads in different layers based on the Frobenius norm of the attention-head weight matrices. 
Since their method is derived from the attention computation, CS tends to be large for pairs of heads that are functionally related. 
However, CS is a correlation-like measure that normalizes the Frobenius norm of a matrix product by the product of the Frobenius norms.
This normalization removes the overall matrix scale, but it retains direction-dependent scaling effects and relatively emphasizes directions that dominate the product, namely directions associated with large singular values.
As a result, CS can emphasize the alignment of a few dominant directions more than the overall overlap between the subspaces associated with the two heads. This emphasis can hinder the comparison and interpretation of similarities between heads.

\begin{figure}[t!]
    \centering
    \includegraphics[width=0.83\columnwidth]{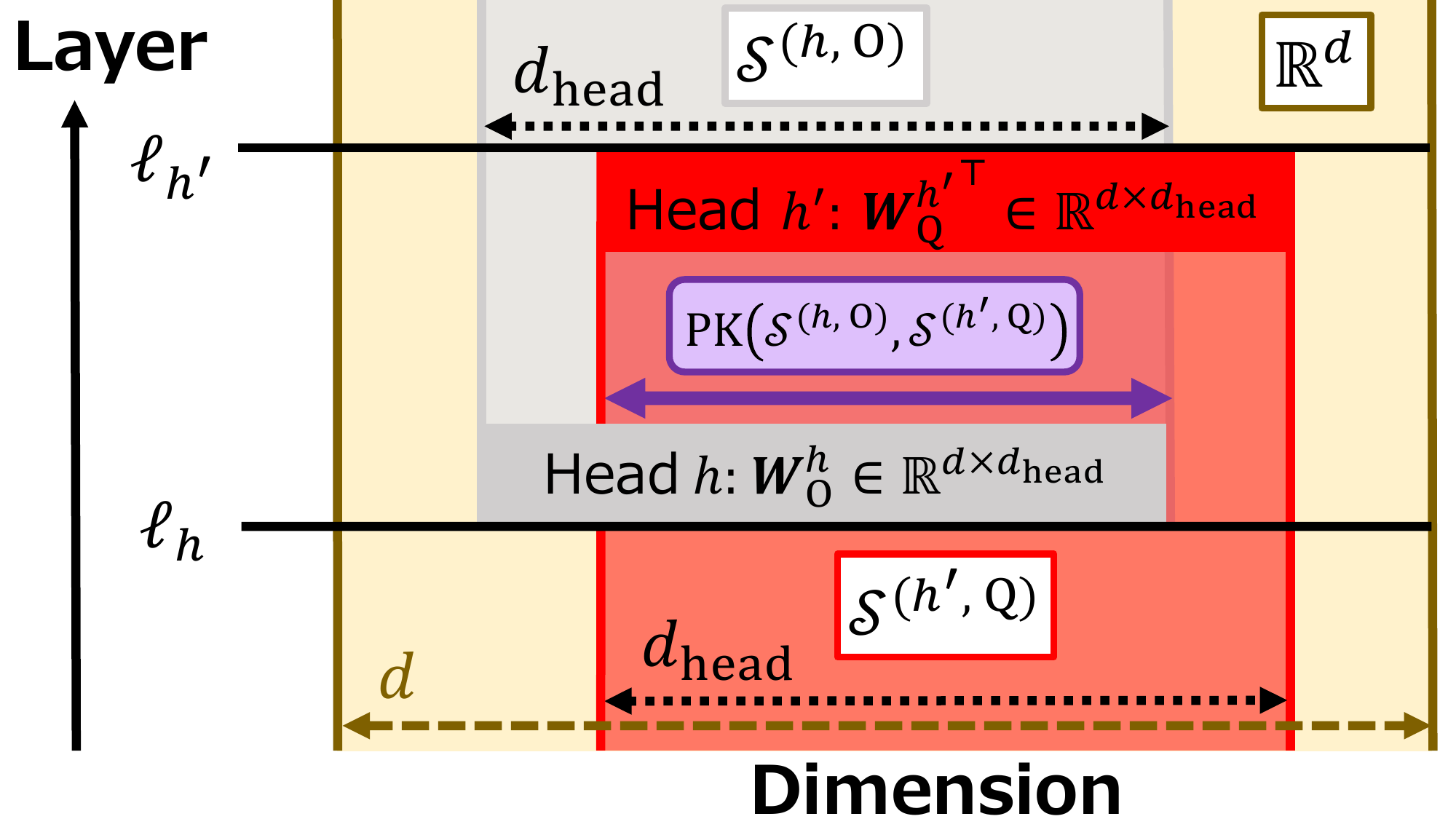}
    \caption{
Illustration of measuring similarity using the Projection Kernel (PK) in (\ref{eq:pk}) between the $d_\text{head}$-dimensional subspace $\mathcal{S}^{(h, \text{O})}\subset\mathbb{R}^d$ spanned by the output weights $\bm{W}_\text{O}^h\in\mathbb{R}^{d\times d_\text{head}}$ of head $h$ and the $d_\text{head}$-dimensional subspace $\mathcal{S}^{(h', \text{Q})}\subset\mathbb{R}^d$ spanned by the query weights ${\bm{W}_\text{Q}^{h'}}^\top \in\mathbb{R}^{d\times d_\text{head}}$ of head $h'$.
The layer indices of heads $h$ and $h'$ are denoted by $\ell_h$ and $\ell_{h'}$ with $\ell_h < \ell_{h'}$.
A larger $\operatorname{PK}(\mathcal{S}^{(h, \text{O})}, \mathcal{S}^{(h', \text{Q})})$ indicates that the subspaces share more directions with small principal angles in (\ref{eq:pa-def}), and thus that the output of head $h$ is more strongly incorporated into the query of head $h'$.
}
    \label{fig:attn_subspace}
\end{figure}

Therefore, in this work, we focus on the $d_\text{head}$-dimensional subspace spanned by the weights of each head, and characterize relationships between heads not by the magnitudes of the weights but solely by the directions of these subspaces.
Specifically, we interpret the amount of information that attention heads can share during reading and writing as the \emph{overlap between their subspaces}, and quantify it using the principal-angle-based Projection Kernel~\cite{DBLP:conf/icml/HamL08} (PK) (Fig.~\ref{fig:attn_subspace}). 
Figure~\ref{fig:intro} compares PK with CS in analyzing relationships between attention heads in GPT2-small~\cite{radford2019language}.
The contrast observed in this figure suggests that PK has properties distinct from CS when capturing functional relationships between heads.

We now outline the paper. 
Section~\ref{sec:bg} reviews CS, a weight-based similarity metric, and summarizes its limitations. 
Section~\ref{sec:pk} defines PK and examines its behavior using a toy example and a two-layer attention-only model.
Section~\ref{sec:unembedding-proj} introduces a method for interpreting heads based on projection onto head-specific subspaces.
Section~\ref{sec:exp-setup} describes the experimental setup using GPT2-small, and Section~\ref{sec:pk-vs-cs} compares PK and CS. Section~\ref{sec:eval-metric} evaluates similarity metrics using functional groups of heads identified in the IOI task, hereafter referred to as \emph{head classes}.
Section~\ref{sec:disc} shows that the hub head L4H7 in Fig.~\ref{fig:intro} is an Identity Head and introduces a method for comparing PK distributions.
It also shows that PK is effective for comparing annotated heads on the IOI task in BLOOM-560M~\cite{workshop2023bloom176bparameteropenaccessmultilingual} and Qwen2-0.5B-Instruct~\cite{yang2024qwen2technicalreport}, which have architectures different from that of GPT2-small.
Finally, Section~\ref{sec:conclusion} concludes the paper.

\section{Background}\label{sec:bg}
This section briefly reviews the QK and OV matrices introduced by \citet{elhage2021mathematical} and the Composition Score (CS), which quantifies relationships between attention heads.
We then summarize the limitations of CS.
See Appendix~\ref{app:cs} for details.

\subsection{QK and OV matrices}
Let $d$ be the model dimension and $d_{\text{head}}$ the head dimension.
For head $h$, let $\bm{W}_{\text{Q}}^{h},\bm{W}_{\text{K}}^{h},\bm{W}_{\text{V}}^{h}\in\mathbb{R}^{d_{\text{head}}\times d}$ be the query, key, and value weight matrices, and let $\bm{W}_{\text{O}}^{h}\in\mathbb{R}^{d\times d_{\text{head}}}$ be the output weight matrix.

Based on the attention computation, \citet{elhage2021mathematical} defined the QK and OV matrices as
\begin{align}
    \bm{W}_{\text{QK}}^{h} &= {\bm{W}_\text{Q}^{h}}^\top\bm{W}_\text{K}^{h}\in\mathbb{R}^{d\times d},\label{eq:QK-matrix}\\
    \bm{W}_{\text{OV}}^{h} &= \bm{W}_{\text{O}}^{h}\bm{W}_{\text{V}}^{h}\in\mathbb{R}^{d\times d}.\label{eq:OV-matrix}
\end{align}
Intuitively, the QK matrix captures \emph{where} information is moved from, whereas the OV matrix captures \emph{what} information is moved~\cite{nanda2022transformerlens}.

\subsection{Composition Score}
Using the QK and OV matrices, \citet{elhage2021mathematical} quantified how the output of head $h$ contributes to the query, key, and value computations of a head $h'$ in a subsequent layer.
They refer to these quantities as the scores for Q-, K-, and V-composition, respectively.
In this paper, following \citet{merullo2024talking}, we refer to this metric as the Composition Score (CS), as we use the same form to capture head-to-head relationships more generally, rather than restricting it to the relationships between the output and the query, key, or value.

For $\bm{W}^{h'}, \bm{W}^{h}\in\mathbb{R}^{d\times d}$, we define
\begin{align}
    \operatorname{CS}(\bm{W}^{h'}, \bm{W}^{h}) = \frac{\|\bm{W}^{h'}\bm{W}^{h}\|_{\text{F}}}{\|\bm{W}^{h'}\|_{\text{F}}\|\bm{W}^{h}\|_{\text{F}}},\label{eq:cs}
\end{align}
where $\|\bm{A}\|_{\text{F}}=\sqrt{\operatorname{tr}(\bm{A}^\top \bm{A})}$ denotes the Frobenius norm.
Thus, CS in (\ref{eq:cs}) normalizes the Frobenius norm of the product $\bm{W}^{h'}\bm{W}^{h}$ by the Frobenius norms of $\bm{W}^{h'}$ and $\bm{W}^{h}$. 

Using the definition in (\ref{eq:cs}), we define $\operatorname{CS}_{\,\text{OQ}}$, $\operatorname{CS}_{\,\text{OK}}$, and $\operatorname{CS}_{\,\text{OV}}$ as the respective counterparts of the scores for Q-, K-, and V-composition described by \citet{elhage2021mathematical}\footnote{Here, OQ, OK, and OV indicate that the source side is the output weight of head $h$, while the target side is the query, key, or value weight of head $h'$, respectively.}:
\begin{align}
    &\operatorname{CS}_{\text{\bggray{O}\bgred{Q}}}(h,h')=\operatorname{CS}\left({\bm{W}_{\text{\bgred{Q}K}}^{h'}}^\top, \bm{W}_{\text{\bggray{O}V}}^{h}\right),\label{eq:OQ-CS}\\
    &\operatorname{CS}_{\text{\bggray{O}\bggreen{K}}}(h,h')=\operatorname{CS}\left(\bm{W}_{\text{Q\bggreen{K}}}^{h'}, \bm{W}_{\text{\bggray{O}V}}^{h}\right),\label{eq:OK-CS}\\
    &\operatorname{CS}_{\text{\bggray{O}\bgblue{V}}}(h,h')=\operatorname{CS}\left(\bm{W}_{\text{O\bgblue{V}}}^{h'}, \bm{W}_{\text{\bggray{O}V}}^{h}\right).\label{eq:OV-CS}
\end{align}
For example, in (\ref{eq:OQ-CS}), replacing $\bm{W}_{\text{OV}}^{h}$ with $\bm{W}_{\text{QK}}^{h}$ yields analogous definitions such as $\operatorname{CS}_{\,\text{QQ}}$.
We use CS as a normalized measure of head-to-head interaction, which tends to take larger values for pairs of heads that are more strongly related.

\subsection{Limitations}\label{sec:cs-limitations}
While CS provides a reasonable way to compare heads using the QK and OV matrices that arise in the attention computation, we argue that it has several limitations.
First, CS retains the direction-dependent scaling present in the weight matrices.
Specifically, in (\ref{eq:cs}), the overlap between the output singular directions of $\bm{W}^{h}$ and the input singular directions of $\bm{W}^{h'}$ is weighted according to the magnitudes of the corresponding singular values (see Appendix~\ref{app:cs-direction-dependent-scaling}).
Consequently, rather than capturing the geometric overlap of the entire subspaces uniformly, CS places relatively greater emphasis on directions associated with large singular values.

Second, although any invertible $\bm{A}\in\mathbb{R}^{d\times d}$ yields the factorization $\bm{W}^{h'}\bm{W}^h=\bm{W}^{h'}\bm{A}\bm{A}^{-1}\bm{W}^h$, in general
$\|\bm{W}^{h'}\|_\text{F}\|\bm{W}^h\|_\text{F}\neq \|\bm{W}^{h'}\bm{A}\|_\text{F}\|\bm{A}^{-1}\bm{W}^h\|_\text{F}$.
Therefore, CS is not invariant to this factorization, and typically
$\operatorname{CS}(\bm{W}^{h'},\bm{W}^h)\neq \operatorname{CS}(\bm{W}^{h'}\bm{A}, \bm{A}^{-1}\bm{W}^h)$.

Finally, we observe layerwise variation in the Frobenius norms of QK and OV matrices, a tendency for $\operatorname{CS}_{\,\text{OV}}$ to decrease as its denominator increases (see Appendix~\ref{app:gpt2-frobenius-norms}), and substantial changes in CS under weight preprocessing~\cite{nanda2022transformerlens} that simplifies the weights (see Appendix~\ref{app:weight-preproc}).
Taken together, these observations suggest that CS can be sensitive to variations in weight scaling and parameterization.

\section{Quantifying overlap between subspaces spanned by weights}\label{sec:pk}
In this section, we describe how to quantify the overlap between the subspaces spanned by attention-head weights using the Projection Kernel (PK) based on principal angles.
We then examine the behavior and usefulness of PK through experiments on a toy subspace example and a two-layer attention-only Transformer.

\begin{figure}[t!]
    \centering
    \includegraphics[width=0.6\columnwidth]{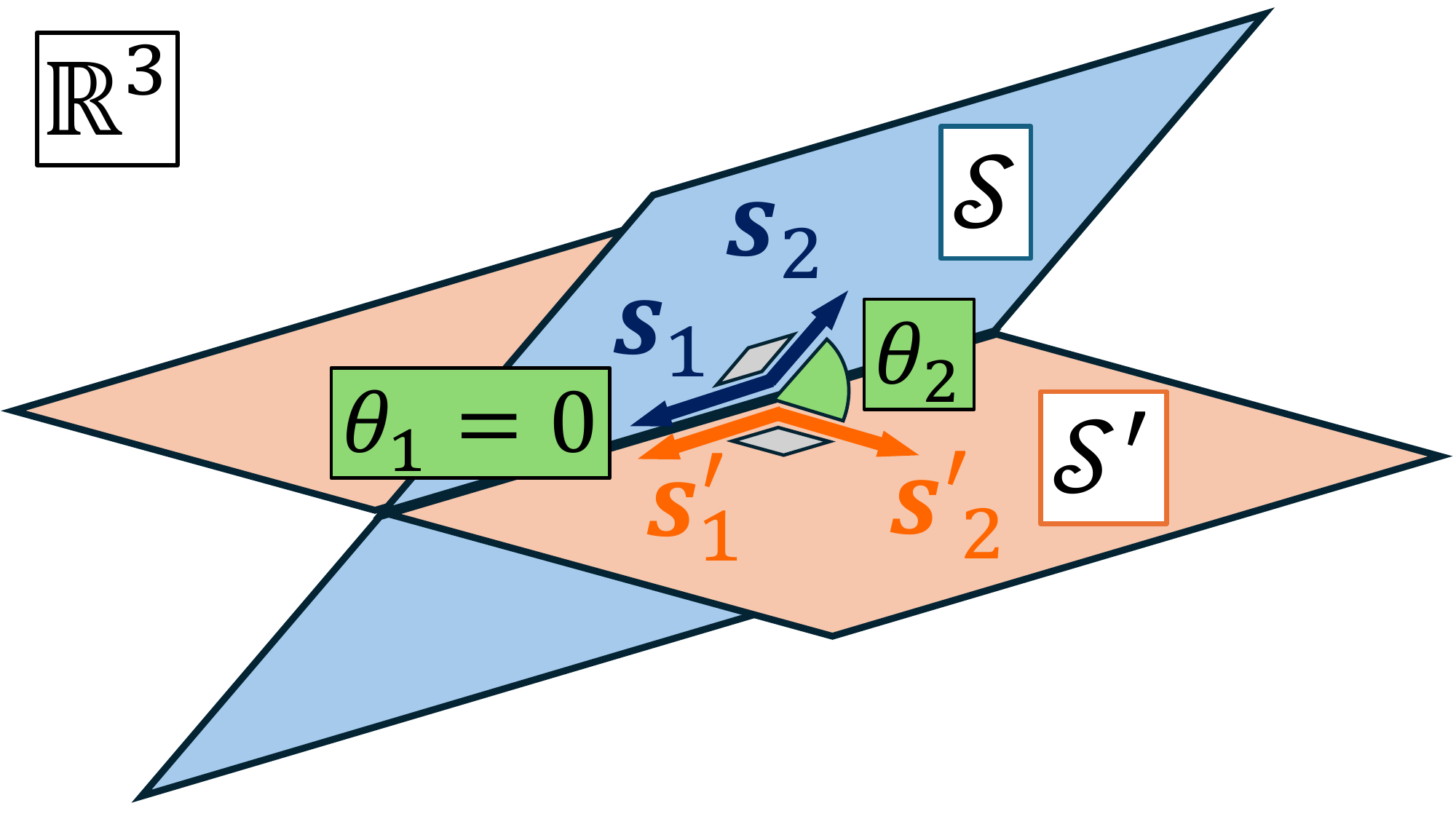}
    \caption{
Principal angles $\theta_1=0$ and $\theta_2$ between two planes $\mathcal{S},\mathcal{S}'\subset \mathbb{R}^3$ with $\dim \mathcal{S}=\dim \mathcal{S}' = 2$, and the corresponding principal vectors $\bm{s}_1,\bm{s}_2\in\mathcal{S}$ and $\bm{s}_1',\bm{s}_2'\in\mathcal{S}'$.
Inspired by \citet[Fig.~1]{mandolesi2021blade}.
}
    \label{fig:pa}
\end{figure}

\subsection{Principal angles}\label{sec:pa}
Let $\mathcal{S},\mathcal{S}'$ be $m$-dimensional subspaces of $\mathbb{R}^d$.
For $i=1,\ldots,m$, we define
\begin{equation}
    \cos\theta_{i} =
    \max_{\substack{
        \bm{s}\in\mathcal{S},\ \|\bm{s}\|=1, \bm{s} \perp \bm{s}_1,\ldots,\bm{s}_{i-1}\\
        \bm{s}'\in\mathcal{S}',\ \|\bm{s}'\|=1, \bm{s}' \perp \bm{s}_1',\ldots,\bm{s}_{i-1}'
    }}
    \bm{s}^\top\bm{s}'\label{eq:pa-def}
\end{equation}
and call $\theta_i$ the $i$-th principal angle~\cite{bjorck1973numerical}.
The vectors $\bm{s}_{i}\in\mathcal{S}$ and $\bm{s}_{i}'\in\mathcal{S}'$ that attain the maximum are called the $i$-th principal vectors~\cite{dahlquist2008numerical}.
As an example, Fig.~\ref{fig:pa} illustrates principal angles and principal vectors between two planes in $\mathbb{R}^3$.

Let $\bm{U},\bm{U}'\in\mathbb{R}^{d\times m}$ be orthonormal basis matrices of $\mathcal{S}$ and $\mathcal{S}'$, and let $\sigma_1\geq \ldots \geq \sigma_m$ be the singular values of $\bm{U}^\top\bm{U}'\in\mathbb{R}^{m\times m}$.
Then the $i$-th singular value $\sigma_i$ corresponds to $\cos\theta_{i}$~\cite{dahlquist2008numerical} (see Appendix~\ref{app:pk}).

\subsection{Projection Kernel}
To quantify the overlap between subspaces $\mathcal{S}$ and $\mathcal{S}'$ based on principal angles, we use the Projection Kernel (PK)~\cite{DBLP:conf/icml/HamL08}\footnote{
Variants of PK include $\sqrt{\operatorname{PK}(\mathcal{S}, \mathcal{S}')}$~\cite{10.1214/12-AOS1034} and $\sqrt{\operatorname{PK}(\mathcal{S}, \mathcal{S}')/m}$, a normalized version~\cite{10.1214/13-AOS1199}.
}
\begin{equation}
\operatorname{PK}(\mathcal{S}, \mathcal{S}') = \sum_{i=1}^m\cos^2\theta_{i},\label{eq:pk-def}
\end{equation}
which satisfies\footnote{Since $-1\leq \cos\theta_i \leq 1$, we have $0\leq \cos^2\theta_i \leq 1$.} $0\leq \operatorname{PK}(\mathcal{S}, \mathcal{S}') \leq m$.
In particular, $\operatorname{PK}(\mathcal{S}, \mathcal{S}')=m$ when $\mathcal{S}=\mathcal{S}'$.
By algebraic manipulation (see Appendix~\ref{app:pk}), we obtain
\begin{equation}
\operatorname{PK}(\mathcal{S},\mathcal{S}') = \left\|\bm{U}^\top\bm{U}'\right\|_{\text{F}}^2
= \operatorname{tr}(\bm{P}\bm{P}'),
\label{eq:pk}
\end{equation}
where $\bm{P}=\bm{U}\bm{U}^\top\in\mathbb{R}^{d\times d}$ and $\bm{P}'=\bm{U}'{\bm{U}'}^\top\in\mathbb{R}^{d\times d}$ are the orthogonal projection matrices onto $\mathcal{S}$ and $\mathcal{S}'$.
From~(\ref{eq:pk}), PK can be computed directly from the basis matrices.
From the definitions of PK and CS (see Appendix~\ref{app:pk}), we have
\begin{equation}
\operatorname{PK}(\mathcal{S},\mathcal{S}') = (m\operatorname{CS}(\bm{P}',\bm{P}))^2.\label{eq:PK-CS}
\end{equation}
Therefore, PK is CS applied to projection matrices, which have no direction-dependent scaling.

\subsection{Toy example of subspaces}\label{sec:toy-example-of-subspaces}
As a toy example, let $\bm{b}_1,\ldots,\bm{b}_d\in\mathbb{R}^d$ be an orthonormal basis of $\mathbb{R}^d$.
Choose $m$ basis vectors indexed by $\mathcal{I},\mathcal{I}'\subset\{1,\ldots,d\}$ to form orthonormal basis matrices $\bm{F},\bm{F}'\in\mathbb{R}^{d\times m}$, and use PK to measure the overlap between the subspaces $\mathcal{F},\mathcal{F}'\subset\mathbb{R}^d$ they span.
By construction, the dimension of the intersection $\mathcal{F}\cap\mathcal{F}'$ is $\left|\mathcal{I}\cap \mathcal{I}'\right|$.

Using the Kronecker delta $\delta_{ii'}$, we have ${\bm{b}_i}^\top\bm{b}_{i'}=\delta_{ii'}$, and therefore
\begin{align}
    &\operatorname{PK}(\mathcal{F},\mathcal{F}')
    =\left\|\bm{F}^\top \bm{F}'\right\|_{\text{F}}^2 
    =\,\sum_{i\in\mathcal{I},\,{i'}\in\mathcal{I}'}({\bm{b}_i}^\top\bm{b}_{i'})^2 \notag \\
    =\,&\sum_{i\in\mathcal{I}\cap \mathcal{I}'} 1 + \sum_{i\in\mathcal{I},\,{i'}\in\mathcal{I}',\,i \neq i'} 0
    =\left|\mathcal{I}\cap \mathcal{I}'\right|.
\end{align}
The PK score therefore coincides with the dimension of the intersection and gives a natural measure of overlap between the subspaces.

\subsection{Measuring overlap between weight subspaces with PK}\label{sec:proposed}
We use PK to measure overlap between attention-head weight subspaces.
We define the set of weight matrices
\begin{equation}
\mathcal{W}=\bigcup_{h\in \mathcal{H}}\left\{{\bm{W}_{\text{Q}}^h}^\top,{\bm{W}_{\text{K}}^h}^\top,{\bm{W}_{\text{V}}^h}^\top, {\bm{W}_{\text{O}}^h}\right\},\label{eq:mathcalW}
\end{equation}
where $\mathcal{W} \subset\mathbb{R}^{d\times d_{\text{head}}}$ and $\mathcal{H}$ is the set of all heads.
For weights $\bm{W}^r,\bm{W}^{r'}\in\mathcal{W}$, let $\mathcal{S}^r$ and $\mathcal{S}^{r'}$ denote the subspaces spanned by their column vectors\footnote{For example, when $r=(h, \text{Q})$ we have $\bm{W}^r={\bm{W}_{\text{Q}}^h}^\top$.}.
In this paper, we assume $\dim \mathcal{S}^r=\dim \mathcal{S}^{r'}=d_{\text{head}}$, and measure similarity between attention-head weight subspaces using $\operatorname{PK}(\mathcal{S}^r,\mathcal{S}^{r'})$ (see Fig.~\ref{fig:attn_subspace}).
As with CS, for example, we denote the PK score between $\bm{W}^{(h, \text{\bggray{O}})}$ and $\bm{W}^{(h', \text{\bgred{Q}})}$ by $\operatorname{PK}_{\text{\bggray{O}\bgred{Q}}}(h, h')$.

As an example, consider the OQ pairing to compare PK and CS.
Unlike $\operatorname{CS}_{\,\text{OQ}}$, which uses the QK and OV matrices in (\ref{eq:QK-matrix}) and (\ref{eq:OV-matrix}), $\operatorname{PK}_{\,\text{OQ}}$ does not involve $\bm{W}^{(h, \text{K})}$ or $\bm{W}^{(h',\text{V})}$.
Moreover, since PK is computed from orthonormal bases, it is not directly affected by weight norms.

\subsection{Measuring similarity in a toy model}\label{sec:attn-only-2l}
\citet{elhage2021mathematical} showed that, in a two-layer attention-only Transformer, $\operatorname{CS}_{\,\text{OK}}$ takes large values between Previous Token Heads and Induction Heads.
We therefore run the same experiment with PK and CS using the \texttt{attn-only-2l} model in TransformerLens~\cite{nanda2022transformerlens}\footnote{
This model includes pre-norm and final LayerNorm.
}.

\begin{figure}[t!]
\centering
\begin{subfigure}{\columnwidth}
\centering
    \includegraphics[width=0.85\textwidth]{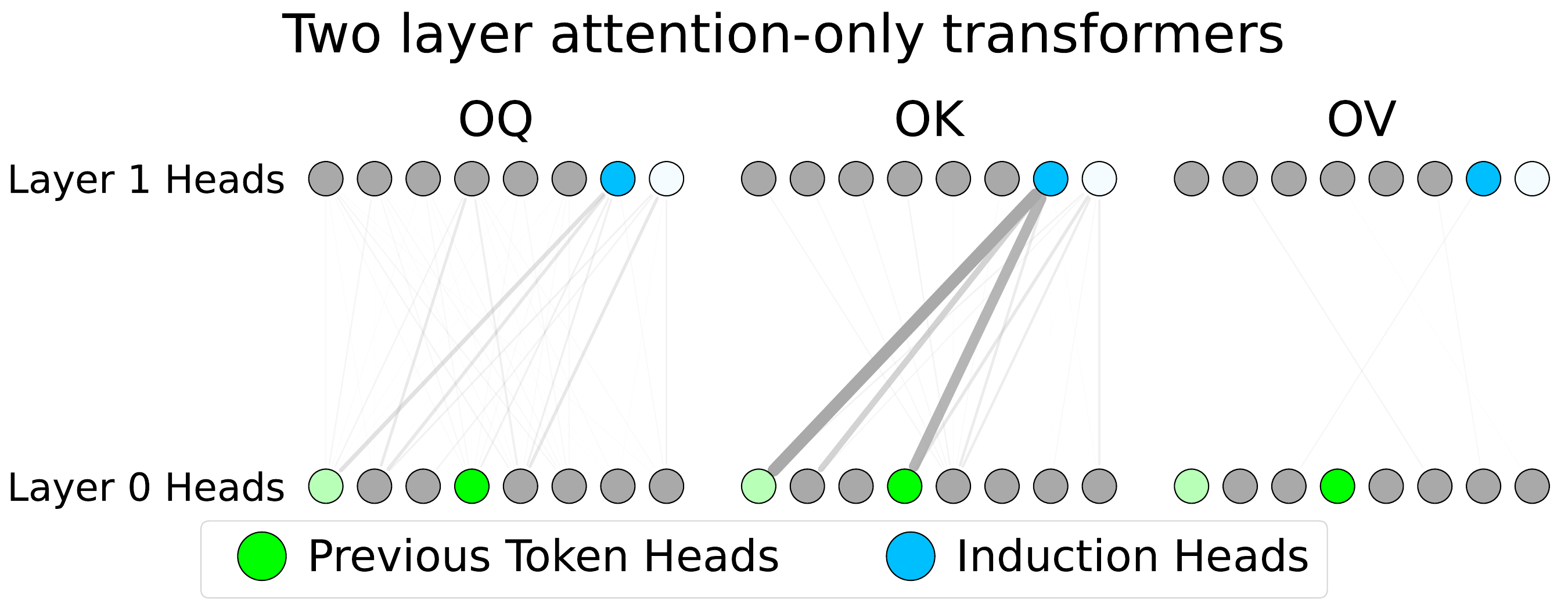}
    \subcaption{Projection Kernel (PK)}
    \label{fig:attn-only-2l-pk}
\end{subfigure}
\par\bigskip
\begin{subfigure}{\columnwidth}
\centering
    \includegraphics[width=0.85\textwidth]{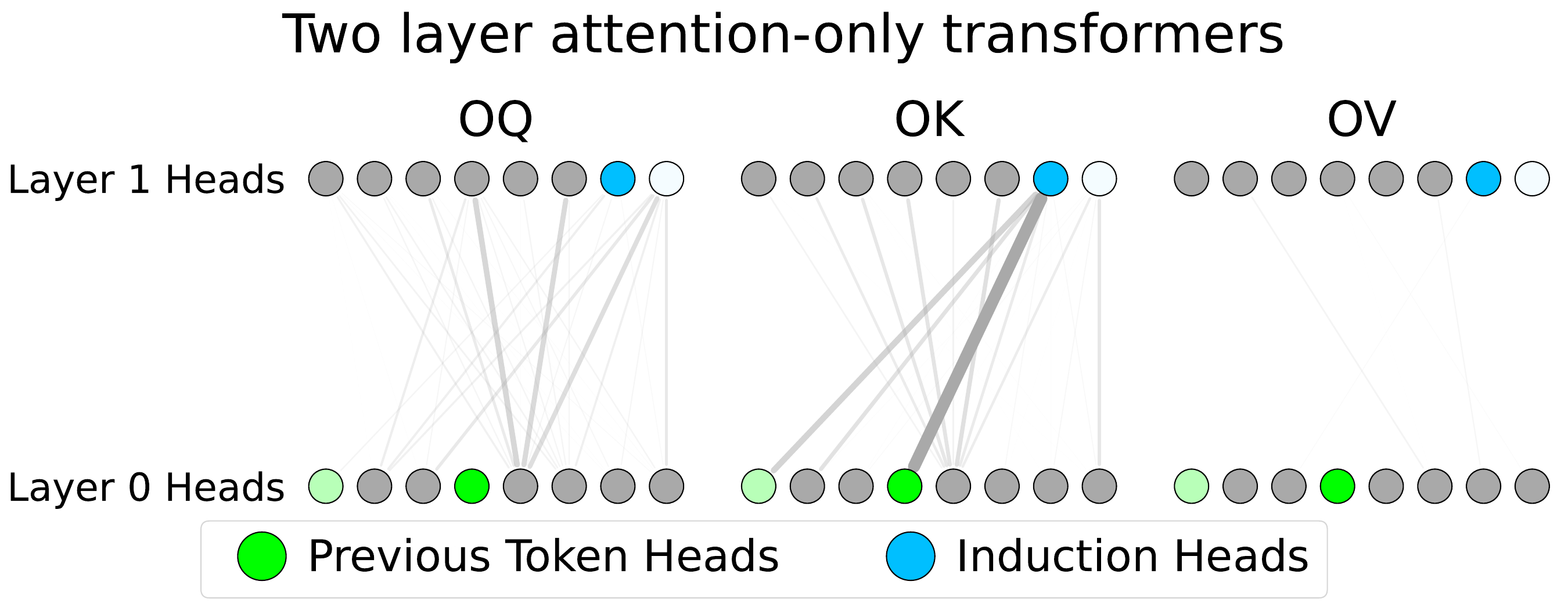}
    \subcaption{Composition Score (CS)}
    \label{fig:attn-only-2l-cs}
\end{subfigure}
    \caption{
For a two-layer attention-only Transformer, we visualize head-to-head similarity strengths using (a) PK and (b) CS.
We show the top two Previous Token Heads in the first layer and the top two Induction Heads in the second layer, coloring them by their head scores (see Appendix~\ref{app:attn-only-2l} for details).
}
\label{fig:attn-only-2l}
\end{figure}

Figure~\ref{fig:attn-only-2l} shows the results.
In Fig.~\ref{fig:attn-only-2l-pk}, $\operatorname{PK}_{\,\text{OK}}$ is large only for the two pairs consisting of the two Previous Token Heads and the top Induction Head, and is small for all other pairs.
In contrast, in Fig.~\ref{fig:attn-only-2l-cs}, among these two pairs only one has a large $\operatorname{CS}_{\,\text{OK}}$ score and the other has a small score.
Moreover, CS is noisier and tends to assign relatively larger values to many other pairs compared to PK.
These results indicate that PK captures the relationships between Previous Token Heads and Induction Heads more clearly than CS.

\section{Interpreting head subspaces}\label{sec:unembedding-proj}
Section~\ref{sec:pk} introduced PK to measure the similarity between attention-head weight subspaces.
In this section, we identify which vocabulary tokens are strongly associated with each attention head from a subspace perspective.
Specifically, we project the unembedding matrix onto the subspace spanned by each head, and identify tokens whose projected norms are large as representative vocabulary items that are strongly read and written by that head.

In the Logit Lens~\cite{logit-lens}, logits are computed by applying the unembedding matrix after the final layer normalization (LN), denoted $\operatorname{LN}_{\text{final}}$.
Following this idea, for $\bm{W}^r\in\mathcal{W}$ we define
$\widetilde{\bm{W}}^r := \operatorname{LN}_{\text{final}}(\bm{W}^r)\in\mathbb{R}^{d\times d_\text{head}}$ and let $\widetilde{\mathcal{S}}^r$ be the subspace spanned by the columns of $\widetilde{\bm{W}}^r$.
The orthogonal projection matrix onto $\widetilde{\mathcal{S}}^r$ is
$\widetilde{\bm{P}}^{r}
= \widetilde{\bm{W}}^{r}
\left( {\widetilde{\bm{W}}^{r\,\top}}\widetilde{\bm{W}}^{r} \right)^{-1}
\widetilde{\bm{W}}^{r\,\top}\in\mathbb{R}^{d\times d}$.

Let $\mathcal{T}$ be the vocabulary.
For the unembedding matrix $\bm{E}_{\text{out}}\in\mathbb{R}^{d \times |\mathcal{T}|}$, let $\bm{e}_t\in\mathbb{R}^d$ denote its column for token $t$, which we call the unembedding vector.
After preprocessing $\bm{e}_t$ with a function $f$, we project it onto $\widetilde{\mathcal{S}}^r$ using $\widetilde{\bm{P}}^{r}$, and treat the norm of the projected vector $\widetilde{\bm{P}}^{r}f(\bm{e}_t)\in\mathbb{R}^{d}$ as a logit.

Empirically, we find that normalization shrinks the norms of less interpretable tokens, and that centering before normalization preserves the intrinsic statistical properties related to token generation probabilities (see Appendix~\ref{app:proj-details}).
Accordingly, for $\bm{W}^r\in\mathcal{W}$, letting $\bar{\bm{e}}\in\mathbb{R}^d$ be the mean unembedding vector, we compute the logit for token $t$ as
\begin{equation}
\left\|\widetilde{\bm{P}}^{r}\frac{\bm{e}_t - \bar{\bm{e}}}{\|\bm{e}_t - \bar{\bm{e}}\|}\right\|.
\end{equation}

\section{Experimental setup}\label{sec:exp-setup}
Following \citet{wang2023interpretability}, we use GPT2-small~\cite{radford2019language} ($d=768$, $d_\text{head}=64$) as the base model.
As in Fig.~\ref{fig:attn-only-2l}, to analyze similarities between heads based on their functions, we need to specify annotations for head classes.

We adopt the IOI-task annotations from \citet{wang2023interpretability}.
However, we found that both Previous Token Heads and Induction Heads are substantially more numerous than Duplicate Token Heads\footnote{Moreover, \citet{wang2023interpretability} identify L0H1, L0H10, and L3H0 as Duplicate Token Heads, but in our replication we replace L0H10 with L0H5, which has the largest head score.} (see Fig.~\ref{fig:head_scores} in Appendix~\ref{app:head-scores}).
We therefore recompute head scores for these two classes and assign the Previous Token Head or Induction Head label to unannotated heads among the top 10 heads in each class.
The full list of annotations and details are provided in Appendix~\ref{app:head-scores}.

\begin{figure}[t!]
    \centering
    \includegraphics[width=\linewidth]{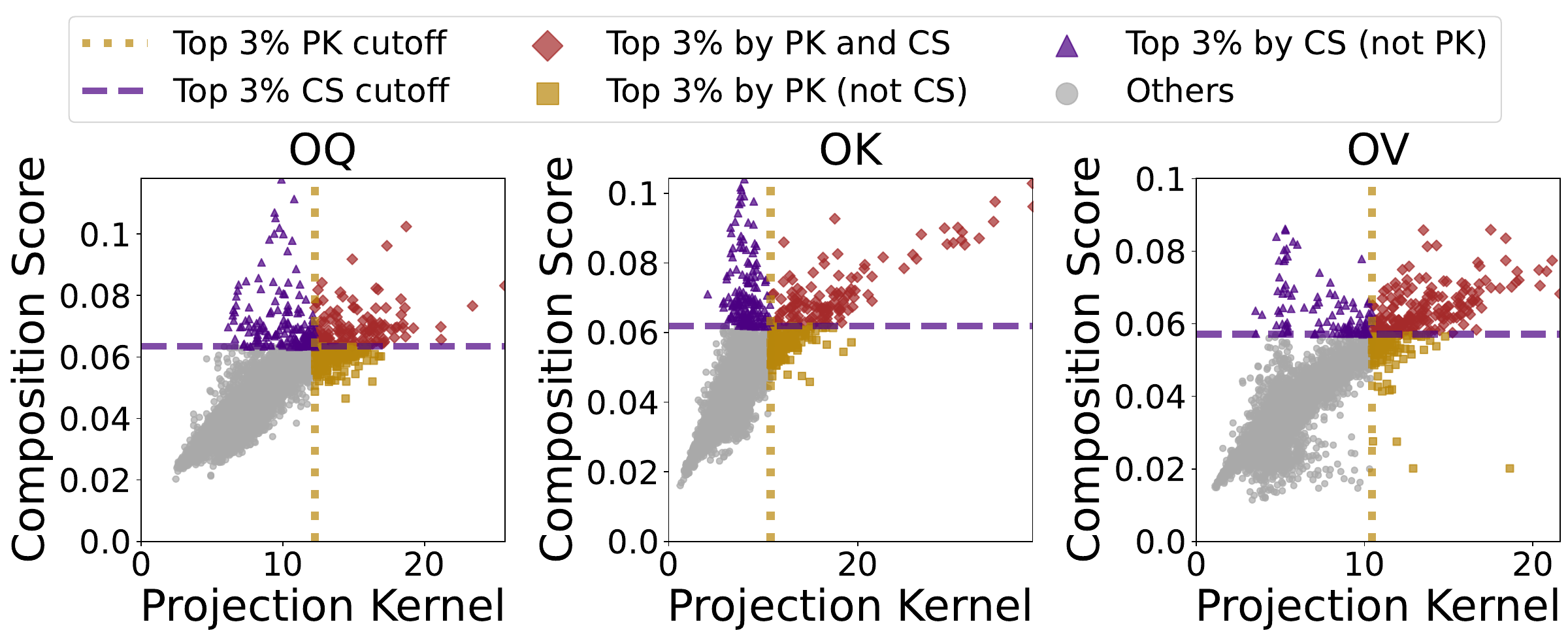}
    \caption{
We compute PK and CS scores for OQ, OK, and OV and plot them as scatter plots.
Spearman's $\rho$ values over all pairs are $0.885$, $0.844$, and $0.866$, indicating strong correlations.
In contrast, when we restrict to pairs that belong to the top 3\% in either metric (the {\color[rgb]{0.647, 0.165, 0.165}$\blacklozenge$}, {\color[rgb]{0.722, 0.525, 0.043}$\blacksquare$}, and {\color[rgb]{0.294, 0.000, 0.510}$\blacktriangle$} symbols), the coefficients drop to $-0.251$, $-0.110$, and $0.186$, and the correlations become weak.
In particular, for pairs that belong only to the top 3\% in CS (the {\color[rgb]{0.294, 0.000, 0.510}$\blacktriangle$} symbol), many have small PK scores.
For visibility, we plot values up to the $99.99$th percentile.
}
    \label{fig:gpt2-small_PK_vs_CS_scatter}
\end{figure}

\begin{figure*}[p!]
    \centering
    \includegraphics[width=0.96\linewidth]{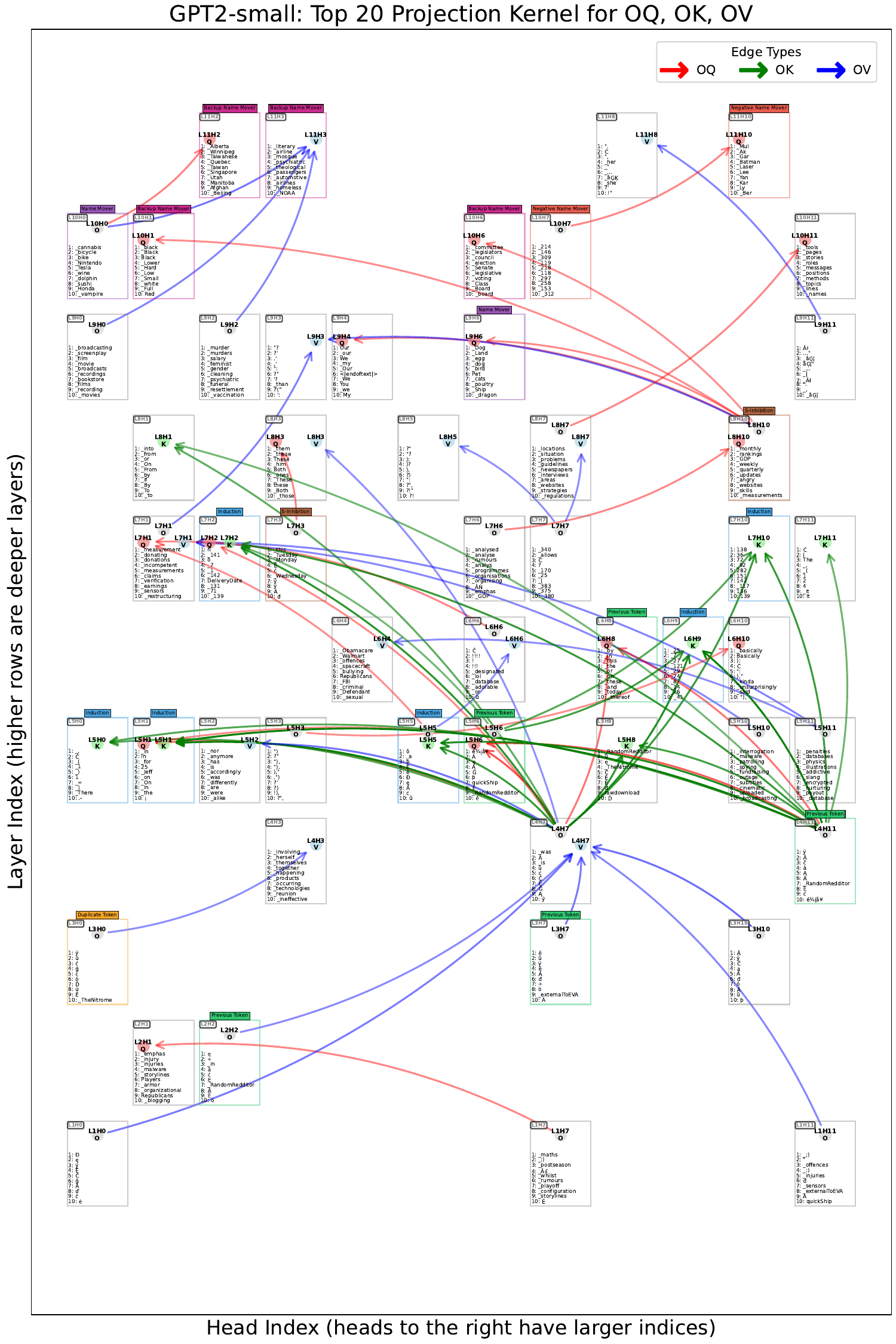}
    \caption{
Visualization of the top 20 scores of $\operatorname{PK}_{\,\text{OQ}}$, $\operatorname{PK}_{\,\text{OK}}$, and $\operatorname{PK}_{\,\text{OV}}$.
For each head $h$, we show the top 10 tokens that represent $\bm{W}_\text{O}^h$ using the projection method in Section~\ref{sec:unembedding-proj}.
For readability, we replace Ġ in tokens with \_.
The L4H7 head is an Identity Head (see Section~\ref{sec:disc}).
Appendix~\ref{app:additional-wiring-diagrams} shows results for the top 10 and top 30 scores.
}
    \label{fig:wd-pk-top20}
\end{figure*}

\section{Comparing PK and CS}\label{sec:pk-vs-cs}
For each pair consisting of a head in an earlier layer and a head in a later layer, we compare PK and CS scores for OQ, OK, and OV.

\subsection{Scatter plots}
Figure~\ref{fig:gpt2-small_PK_vs_CS_scatter} shows scatter plots of PK and CS scores for OQ, OK, and OV.
The head pairs ranked highly by PK differ substantially from those ranked highly by CS.
Although the overall correlation is strong, it is weaker when we consider only head pairs in the top 3\% of either metric.
This result indicates that PK and CS behave differently in the high-score regime.

\subsection{Wiring diagrams of attention heads}\label{sec:wiring-diagram}
In Fig.~\ref{fig:attn-only-2l}, we compared PK and CS scores between attention heads in a two-layer model.
We perform the same analysis for GPT2-small, which has 12 layers with 12 heads per layer.
We compare directed graphs constructed from head-to-head similarity scores, which we call ``wiring diagrams''.

\subsubsection{Setup}
For each of OQ, OK, and OV, we visualize the top 20 scores.
We also annotate each head $h$ with the top 10 tokens that represent $\bm{W}_\text{O}^h$ using the projection method in Section~\ref{sec:unembedding-proj}.

\subsubsection{Results for PK}\label{sec:pk-results}
Figure~\ref{fig:wd-pk-top20} shows the PK wiring diagram.

\paragraph{OQ.}
We observe high scores from the S-Inhibition Head L8H10 to the Name Mover Head L9H6 and the Backup Name Mover Heads L10H1 and L10H6.
This is consistent with the findings of \citet{wang2023interpretability}.
In addition, scores are large among Name-Mover-related heads, such as from L10H0 to L11H2 and from L10H7 to L11H10.

\paragraph{OK.}
As in Fig.~\ref{fig:attn-only-2l}, we observe high scores from Previous Token Heads, such as L4H11 and L5H6, to Induction Heads.
Moreover, when L4H7 or L4H11 is the source head, scores are large for many target heads.
We discuss such outlet heads in Section~\ref{sec:disc}.

\paragraph{OV.}
Except for the edge from L10H0 to L11H3, we find no edges where both the source and target belong to head classes, which makes interpretation difficult.
Notably, many source heads yield high scores with L4H7 or L11H3 as the target.
We discuss such inlet heads in Section~\ref{sec:disc}.

\subsubsection{Results for CS}\label{sec:cs-results}
In the CS wiring diagram (Fig.~\ref{fig:wd-cs-top20} in Appendix~\ref{app:additional-wiring-diagrams}), many unannotated heads appear in early layers. 
As a result, the Name-Mover-related heads, which are prominent under PK and are located mainly in late layers, become less visible.
Moreover, CS often assigns large scores to head pairs that are widely separated across layers.
For example, $\operatorname{CS}_{\,\text{OV}}$ scores from layer $1$ to layer $11$ are high, which makes the layerwise flow of information harder to see than with PK.
As in Fig.~\ref{fig:attn-only-2l}, CS also captures relationships between Previous Token Heads and Induction Heads, although fewer such pairs appear than with PK.

\subsubsection{Interpreting top tokens of heads}
In middle and late layers, the top tokens of the heads are relatively easy to interpret.
For example, those of L10H6 include many politics-related tokens.
Across head classes, they are mainly nouns for Name-Mover-related heads and often numbers for Induction Heads.
Some heads outside the head classes are also interpretable, such as L9H4, corresponding to personal pronouns, and L8H5, corresponding to punctuation such as ``?'' and ``!''.
Heads in early layers are harder to interpret, consistent with observations from the Logit Lens.

\begin{table}[t!]
\scriptsize
\centering
\begin{tabular}{l|l|rrr|r}
\toprule
& Method & OQ & OK & OV & Avg. \\
\midrule
\multirow{5}{*}{\rotatebox{90}{PR-AUC}}  & PK & \textbf{0.446} & \textbf{0.451} & 0.289 & \textbf{0.395} \\
 & CS & 0.227 & 0.352 & 0.215 & 0.265 \\
 & Simple-CS & 0.245 & \underline{0.359} & 0.305 & 0.303 \\
 & Linear CKA & \underline{0.337} & 0.323 & \underline{0.409} & \underline{0.356} \\
 & Proc. Sim. & 0.328 & 0.269 & \textbf{0.428} & 0.342 \\
\bottomrule
\end{tabular}
\caption{
PR-AUC scores for detecting annotated heads in GPT2-small. 
We treat heads in head classes for the IOI task (36 out of 144) as positives, rank head pairs by similarity, and compute PR-AUC from head-level precision and recall.
The best score is shown in bold and the second best score is underlined.
}
\label{tab:PRAUC-OQ-OK-OV}
\end{table}

\begin{table}[t!]
\scriptsize
\centering
\begin{tabular}{l|rr}
\toprule
Method & PR-AUC & ROC-AUC \\
\midrule
PK & \textbf{0.047} & \textbf{0.809} \\
CS & 0.013 & 0.785 \\
Simple-CS & 0.016 & \underline{0.792} \\
Linear CKA & 0.021 & 0.715 \\
Proc. Sim. & \underline{0.038} & 0.787 \\
\bottomrule
\end{tabular}
\caption{
Mean PR-AUC and ROC-AUC, averaged over the same-type pairings (QQ, KK, VV, and OO) and head classes for the IOI task in GPT2-small.
For each pairing type and head class, we treat within-class head pairs as positives and compute PR-AUC and ROC-AUC based on similarities, then average the scores.
The best score is shown in bold and the second best is underlined.
See Appendix~\ref{app:eval-clustering} for details.
}
\label{tab:classification-summary}
\end{table}

\section{Evaluating similarity metrics}\label{sec:eval-metric}
Using the annotated heads, we evaluate PK against baseline methods.
We consider the OQ, OK, and OV pairings from Section~\ref{sec:pk-vs-cs}, and also compare the same-type pairings QQ, KK, VV, and OO.

\subsection{Baselines}\label{sec:eval-baseline}
\citet{elhage2021mathematical} used $\operatorname{CS}\left({\bm{W}_{\text{QK}}^{h'}}^\top, \bm{W}_{\text{OV}}^{h}\right)$ to measure, for example, the relationship between outputs and queries.
We also include as a baseline a simpler method that directly computes $\operatorname{CS}\left({\bm{W}_\text{Q}^{h'}}^\top,\bm{W}_\text{O}^h\right)$, which we call Simple-CS.
In addition, as weight-based similarity methods, we adopt Linear CKA (Centered Kernel Alignment)~\cite{DBLP:conf/icml/Kornblith0LH19} and a similarity based on the normalized Procrustes distance~\cite{rohlf1999shape}, which we refer to as Procrustes Similarity.
See Appendix~\ref{app:eval-baselines} for details.

\subsection{Head detection for OQ, OK, and OV}\label{sec:eval-OQ-OK-OV}
As in Section~\ref{sec:pk-vs-cs}, we compute similarity for each pair of a head in an earlier layer and a head in a later layer ($9{,}504$ pairs).
We treat heads in head classes (36 out of 144) as positives and compare detection performance of the similarity metrics.
We evaluate each weight pairing type OQ, OK, and OV.
Specifically, we rank head pairs by similarity, compute precision and recall based on whether the heads are positive, and report PR-AUC.

Table~\ref{tab:PRAUC-OQ-OK-OV} shows the results.
PK outperforms CS for all pairing types, and achieves the best performance for OQ and OK.
The relatively low performance of PK for OV is consistent with the difficulty of interpreting OV in Fig.~\ref{fig:wd-pk-top20}.

\subsection{Head classification for same-type pairings QQ, KK, VV, and OO}\label{sec:eval-QQ-KK-VV-OO}
We expect that, when comparing weights of the same type, similarity will be higher for heads within the same head class.
We therefore evaluate classwise classification performance for each similarity metric.
We consider the $\binom{144}{2}=10{,}296$ directed head pairs whose source is in an earlier layer than the target or in the same layer as the target.
We treat head pairs that belong to the same head class as positives, rank pairs by similarity, and compute PR-AUC and ROC-AUC.
We compare results for the pairing types QQ, KK, VV, and OO.

Table~\ref{tab:classification-summary} reports the mean scores over QQ, KK, VV, and OO.
In this highly imbalanced setting, where the number of positive pairs is very small\footnote{For example, there are only 15 pairs of Induction Heads.}, PK achieves the highest mean PR-AUC and ROC-AUC and shows strong classification performance.
These results suggest that heads in the same head class have large overlaps between the subspaces spanned by their weights and therefore handle similar information in the residual stream.

\section{Discussion}\label{sec:disc}
\begin{figure}[t!]
    \centering
    \includegraphics[width=\columnwidth]{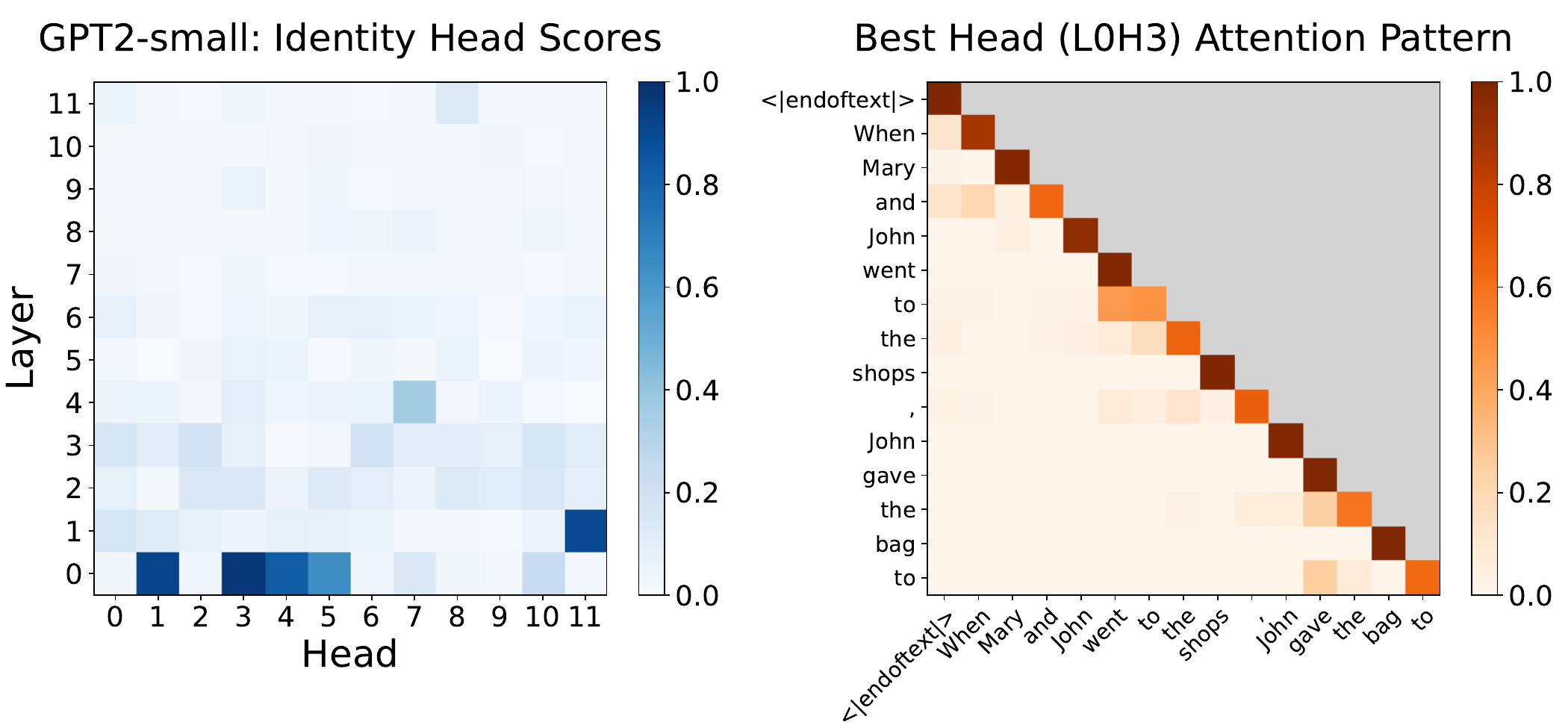}
    \caption{
For GPT2-small, (left) Identity Head scores for each head and (right) an example attention pattern of the L0H3 head, which has the largest score.
In particular, several heads in early layers and L4H7 have large scores.
See Appendix~\ref{app:head-scores} for details.
}
\label{fig:identity}
\end{figure}

\begin{table}[t!]
\scriptsize
\centering
\begin{tabular}{@{\hspace{0.5em}}l@{\hspace{0.5em}}|@{\hspace{0.5em}}l@{\hspace{0.5em}}|@{\hspace{0.5em}}l@{\hspace{0.5em}}l@{\hspace{0.5em}}l@{\hspace{0.5em}}l@{\hspace{0.5em}}l@{\hspace{0.5em}}}
\toprule
 & & Top 1 & Top 2 & Top 3 & Top 4 & Top 5 \\
\midrule
\multirow{3}{*}{\rotatebox{90}{Inlet}} & OQ & L10H7$^{\text{N-NM}}$ & L5H6$^{\text{Prev}}$ & L7H1 & L8H10$^{\text{S-Inh}}$ & L10H11 \\
 & OK & L5H1$^{\text{Ind}}$ & L7H7 & L5H5$^{\text{Ind}}$ & L11H3$^{\text{B-NM}}$ & L11H8 \\
 & OV & \textbf{L4H7}$^{\text{Id}}$ & L11H3$^{\text{B-NM}}$ & L9H3 & L5H2 & L11H8 \\
\midrule
\multirow{3}{*}{\rotatebox{90}{Outlet}} & OQ & L7H6 & L1H1 & \textbf{L4H7}$^{\text{Id}}$ & L8H10$^{\text{S-Inh}}$ & L5H2 \\
 & OK & \textbf{L4H7}$^{\text{Id}}$ & L4H11$^{\text{Prev}}$ & L0H9 & \textbf{L1H11}$^{\text{Id}}$ & L5H10 \\
 & OV & \textbf{L0H1}$^{\text{Dup, Id}}$ & \textbf{L3H0}$^{\text{Dup, Id}}$ & \textbf{L4H7}$^{\text{Id}}$ & L5H10 & L5H11 \\
\bottomrule
\end{tabular}
\caption{
Top five heads in GPT2-small by inlet and outlet scores for OQ, OK, and OV, defined in (\ref{eq:inlet}) and (\ref{eq:outlet}).
We treat the top 10 heads by Identity Head score in Fig.~\ref{fig:identity} as Identity Heads and highlight them in bold.
For heads that belong to head classes for the IOI task (including Identity Heads), we also show the class labels.
See Appendix~\ref{app:inlet-outlet} for details.
}
\label{tab:inlet_outlet_identity}
\end{table}

In this section, we first examine the function of L4H7 in GPT2-small, whose hub-like behavior was observed in Section~\ref{sec:pk-results}, and then discuss heads that act as hubs.
We then compare PK distributions across weight pairing types in GPT2-small.
Finally, we show that PK is also effective for comparing annotated heads on the IOI task in BLOOM-560M and Qwen2-0.5B-Instruct, which have architectures different from that of GPT2-small.

\subsection{L4H7 is an Identity Head}
In Fig.~\ref{fig:wd-pk-top20}, L4H7 in GPT2-small plays the role of a hub.
Our experiments show that L4H7 is an Identity Head~\cite{gopalani2024abrupt} whose attention pattern is close to a diagonal matrix.
Figure~\ref{fig:identity} shows that several heads in early layers and L4H7 have large Identity Head scores (see Appendix~\ref{app:head-scores}).

\subsection{Detecting hub heads}\label{sec:detecting-hub-heads}
To detect hub heads such as L4H7, we focus on (i) target heads that receive the maximum PK from source heads in earlier layers and (ii) source heads that send the maximum PK to target heads in later layers.
For each of OQ, OK, and OV, we quantify these as inlet and outlet scores, defined in (\ref{eq:inlet}) and (\ref{eq:outlet}) in Appendix~\ref{app:inlet-outlet}.

Table~\ref{tab:inlet_outlet_identity} lists the top five heads by inlet and outlet scores.
For inlet scores, heads in middle and late layers tend to rank highly, and for OV, L4H7 has the largest value.
In contrast, for outlet scores, heads in early to middle layers tend to rank highly, and for OK, L4H7 has the largest value.
Many Identity Heads also appear among the top-ranked heads.

Although Identity Heads are widely known~\cite{bonino2025geometrybert,queipodellano2025attentionsinkscompressionvalleys}, their simplicity means they have received less scrutiny than Previous Token Heads or Induction Heads.
Our results may therefore offer a new perspective on Identity Heads.
A more detailed analysis of Identity Heads, including whether L4H7 is special, is left for future work.

\begin{figure}[t!]
    \centering
    \includegraphics[width=\linewidth]{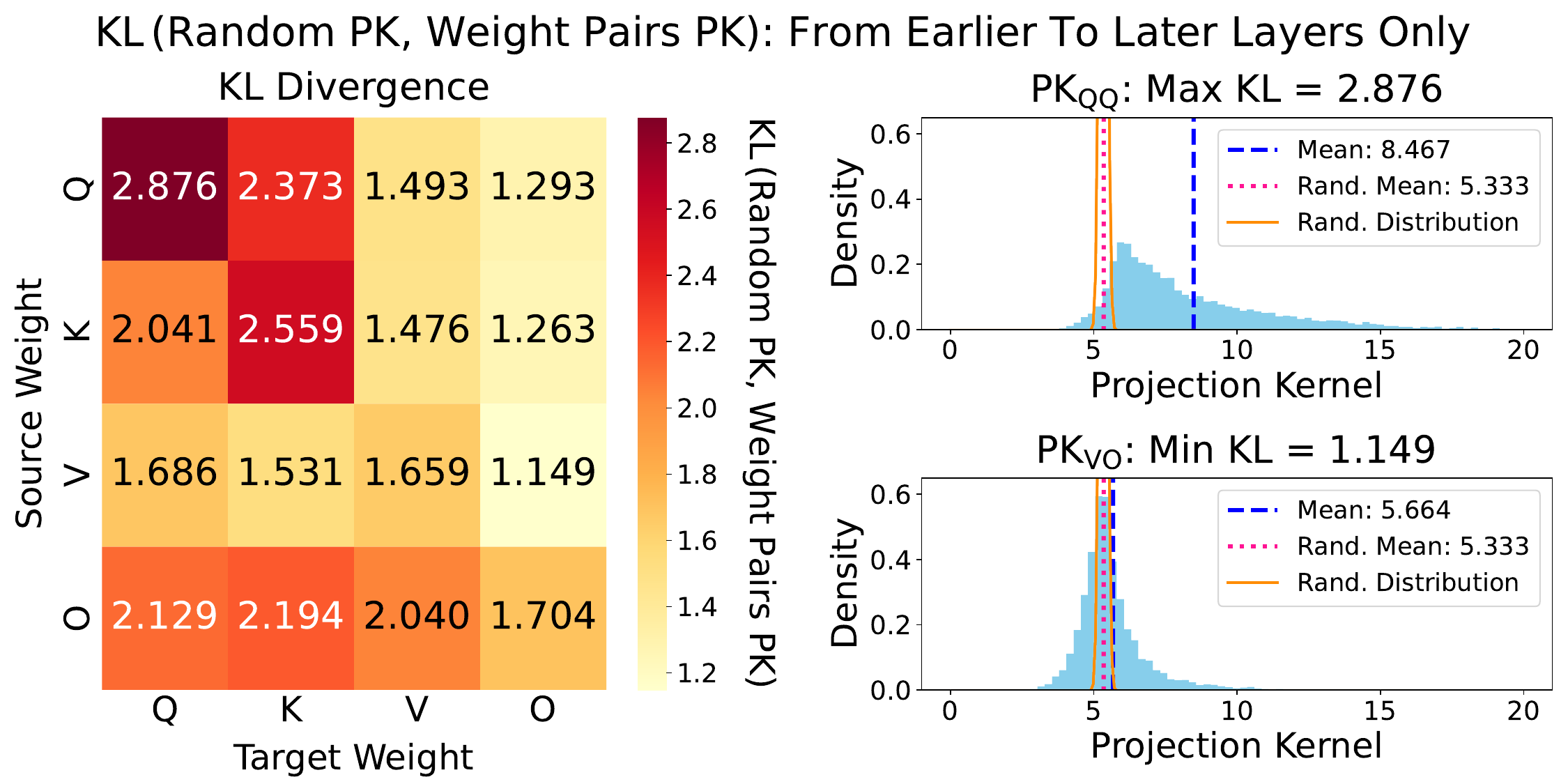}
    \caption{
(Left) KL divergence between the PK distribution for random orthogonal matrices in (\ref{eq:pk-random}) and the PK distributions measured for each weight pairing type in GPT2-small.
(Right) The PK distributions for QQ, which has the largest KL divergence, and VO, which has the smallest KL divergence.
In the left panel, the KL scores are large for OQ, OK, and OV, for same-type pairings, and for query and key pairings, and are small otherwise.
In the right panel, the QQ distribution differs substantially from the random distribution, whereas the VO distribution is close to the random one.
}
    \label{fig:kl-heatmap}
\end{figure}

\subsection{Comparing PK distributions}\label{sec:comparing-pk-distributions}
To compare weight pairing types, we quantify the informativeness of PK distributions.
For random orthogonal matrices, using expectations of products of their entries~\cite{meckes2019random}, we derive the following normal distribution as an approximation to the PK distribution between such matrices (see Appendix~\ref{app:pk-random} for details):
\begin{equation}
  \mathcal{N}\left(\frac{d_\text{head}^2}{d},
     \frac{2d_\text{head}^2(d-d_\text{head})^2}{d^2(d-1)(d+2)}\right).\label{eq:pk-random}
\end{equation}
The Kullback--Leibler divergence (KL) between this distribution and an empirical PK distribution can be viewed as a measure of how informative the PK distribution is relative to the random baseline.

Figure~\ref{fig:kl-heatmap} shows a heatmap of KL scores for the 16 weight pairing types.
The KL score is large for (i) pairings with output weights as sources, (ii) same-type pairings, and (iii) query and key pairings used to compute attention patterns (QK and KQ).
For other pairing types, such as those with output weights as targets, the KL scores are small.
These results indicate that our framework captures meaningful differences across weight pairing types.

\subsection{PK for comparing annotated heads on the IOI task in other models}\label{sec:other-models}

\begin{table}[t!]
\scriptsize
\centering
\begin{subtable}{0.95\linewidth}
\centering
\begin{tabular}{l|l|rrr|r}
\toprule
& Method & OQ & OK & OV & Avg. \\
\midrule
\multirow{5}{*}{\rotatebox{90}{PR-AUC}}
& PK & \textbf{0.358} & \textbf{0.283} & 0.164 & \textbf{0.268} \\
& CS & 0.151 & 0.157 & 0.114 & 0.141 \\
& Simple-CS & \underline{0.266} & \underline{0.261} & 0.145 & \underline{0.224} \\
& Linear CKA & 0.135 & 0.136 & \textbf{0.193} & 0.155 \\
& Proc. Sim. & 0.133 & 0.140 & \underline{0.169} & 0.147 \\
\bottomrule
\end{tabular}
\caption{BLOOM-560M}
\label{tab:PRAUC-OQ-OK-OV-bloom}
\end{subtable}

\vspace{0.5em}

\begin{subtable}{0.95\linewidth}
\centering
\begin{tabular}{l|l|rrr|r}
\toprule
& Method & OQ & OK & OV & Avg. \\
\midrule
\multirow{5}{*}{\rotatebox{90}{PR-AUC}}
& PK & \textbf{0.247} & \underline{0.369} & \underline{0.137} & \textbf{0.251} \\
& CS & 0.140 & 0.275 & \textbf{0.156} & 0.190 \\
& Simple-CS & \underline{0.152} & \textbf{0.379} & 0.135 & \underline{0.222} \\
& Linear CKA & 0.096 & 0.093 & 0.088 & 0.092 \\
& Proc. Sim. & 0.089 & 0.101 & 0.084 & 0.091 \\
\bottomrule
\end{tabular}
\caption{Qwen2-0.5B-Instruct}
\label{tab:PRAUC-OQ-OK-OV-qwen2}
\end{subtable}
\caption{
Detection performance for annotated heads on the IOI task in BLOOM-560M and Qwen2-0.5B-Instruct, evaluated by PR-AUC as in Table~\ref{tab:PRAUC-OQ-OK-OV}.
}
\label{tab:PRAUC-OQ-OK-OV-two-models}
\end{table}

\begin{table}[t!]
\scriptsize
\centering
\begin{tabular}{l|rr|rr}
\toprule
& \multicolumn{2}{c|}{BLOOM-560M} & \multicolumn{2}{c}{Qwen2-0.5B-Instruct} \\
Method & PR-AUC & ROC-AUC & PR-AUC & ROC-AUC \\
\midrule
PK & \textbf{0.027} & \underline{0.866} & \textbf{0.032} & 0.781 \\
CS & 0.003 & 0.844 & 0.005 & \textbf{0.804} \\
Simple-CS & \underline{0.015} & \textbf{0.869} & 0.004 & \underline{0.794} \\
Linear CKA & 0.001 & 0.533 & \underline{0.024} & 0.509 \\
Proc. Sim. & 0.001 & 0.568 & 0.024 & 0.557 \\
\bottomrule
\end{tabular}
\caption{
Classification performance for annotated heads on the IOI task in BLOOM-560M and Qwen2-0.5B-Instruct, evaluated by PR- and ROC-AUC as in Table~\ref{tab:classification-summary}.
}
\label{tab:classification-summary-by-model}
\end{table}

To examine whether our analysis based on GPT2-small generalizes to other models, we apply the same analysis to BLOOM-560M and Qwen2-0.5B-Instruct, using the IOI task head classes from \citet{zhang2025the}. These two models differ from GPT2-small in both size and architecture. See Appendix~\ref{app:bloom-qwen2} for experimental details.

For both models, PK achieves the best performance in the annotated-head detection task (Table~\ref{tab:PRAUC-OQ-OK-OV-two-models}) and in the classification task evaluated by PR-AUC (Table~\ref{tab:classification-summary-by-model})\footnote{Table~\ref{tab:classification-summary-by-model} corresponds to a highly imbalanced setting with very few positive examples, so we emphasize PR-AUC rather than ROC-AUC.}.
Furthermore, in BLOOM-560M, as in GPT2-small in Table~\ref{tab:inlet_outlet_identity}, we find Identity Heads with large inlet and outlet scores\footnote{We do not analyze Qwen2-0.5B-Instruct here because it uses grouped-query attention (GQA)~\cite{ainslie-etal-2023-gqa}, which produces many ties in the inlet scores in (\ref{eq:inlet}).}.

\section{Conclusion}\label{sec:conclusion}
We use the Projection Kernel (PK) to quantify overlap between attention-head subspaces and capture head relationships.
We reveal structure within head classes, identify L4H7 as an Identity Head, and interpret heads via unembedding projections.
We also introduce hub detection scores and a framework to quantify PK-distribution informativeness.
We expect these results to contribute to a deeper understanding of language models.

\clearpage

\section*{Limitations}

\begin{itemize}
\item Our analysis is limited to small language models with fewer than 1B parameters. We have not analyzed large language models (LLMs) with more modern architectures. Specifically, we mainly focus on GPT2-small~\cite{radford2019language}, and Section~\ref{sec:other-models} demonstrates the effectiveness of PK on the IOI task for BLOOM-560M~\cite{workshop2023bloom176bparameteropenaccessmultilingual} and Qwen2-0.5B-Instruct~\cite{yang2024qwen2technicalreport} (see Appendix~\ref{app:model-comparison} for differences among these model architectures).

\item This study implicitly assumes a multi-head attention (MHA) architecture~\cite{NIPS2017_3f5ee243} in which each head has its own query, key, value, and output weights. For example, for Qwen2-0.5B-Instruct, which uses grouped-query attention (GQA)~\cite{ainslie-etal-2023-gqa}, the inlet and outlet scores can still be computed by repeating the key and value weights, but this produces many ties in the inlet scores in~(\ref{eq:inlet})\footnote{For this reason, in Section~\ref{sec:other-models}, we analyze the inlet and outlet scores only for BLOOM-560M to avoid complicating the interpretation of the hub-head analysis for Qwen2-0.5B-Instruct.}. Recent LLMs commonly use attention variants such as GQA, multi-query attention (MQA)~\cite{DBLP:journals/corr/abs-1911-02150}, and multi-head latent attention (MLA)~\cite{deepseekv2}, rather than MHA, to improve inference speed. Therefore, as in the case of Qwen2-0.5B-Instruct, our approach may not be directly applicable to LLMs with these attention architectures.

\item In addition, our evaluation of similarity metrics is based on how well they reproduce known head-to-head relationships reported for the IOI task~\cite{wang2023interpretability}. 
Although PK captures many of these relationships, CS and other baselines may capture other types of head-to-head relationships that are not represented by the IOI-based annotations.
Future work should therefore evaluate these metrics on a broader range of tasks with known circuits, such as the greater-than task~\cite{hanna2023how} and the docstring completion task~\cite{heimersheim2023circuit}.

\item As in \citet{elhage2021mathematical}, our method ignores feed-forward layers, and therefore cannot fully account for the internal computations of language models.

\item Except for the interpretation of head subspaces proposed in Section~\ref{sec:unembedding-proj}, our method does not account for layer normalization (LN)~\cite{ba2016layernormalization}, as in \citet{elhage2021mathematical}. Moreover, recent LLMs often use root mean square layer normalization (RMSNorm) without bias~\cite{NEURIPS2019_1e8a1942}. A careful comparison that accounts for the presence and type of normalization layers is an important direction for future work.

\item For preprocessing before projecting the unembedding matrix, we restrict our comparison to four choices, namely the identity map, centering, normalization, and normalization after centering, focusing on the bias term of the final-layer LN (see Appendix~\ref{app:proj-details}). We adopt normalization after centering, but other preprocessing choices may be more appropriate. Moreover, recent LLMs often use RMSNorm without bias, so our analysis based on the bias term of the final-layer LN may not be directly transferable to such architectures.

\item PK provides a similarity measure between attention heads for each weight pairing, but it does not directly identify head functions.
Nonetheless, using wiring diagrams based on PK, we discovered that L4H7 acts as a hub head. Guided by this observation, we further verified experimentally that L4H7 is an Identity Head. In this sense, PK can be useful not only for validating known head-to-head relationships but also for discovering heads with previously unnoticed functions.

\item We do not provide a detailed analysis of how Identity Heads relate to hub heads. However, the following observations led us to identify L4H7 in GPT2-small as an Identity Head. In the early stage of our experiments, we noticed from Fig.~\ref{fig:wd-pk-top20} that L4H7 has a hub-like structure, with a high inlet score for OV and high outlet scores for OQ, OK, and OV. 
This structure suggests that the value weight $\bm{W}_V^{\text{L4H7}}$ is aligned with directions in the residual stream to which many heads in preceding layers write information. It further suggests that the output weight $\bm{W}_O^{\text{L4H7}}$ writes to directions in the residual stream that are likely to be read by subsequent heads in their query, key, and value computations. This matches the role of an Identity Head. 
Such a head attends to its own position, reading the residual stream only
at that position through its value weights, and can then relay this
information to subsequent layers.
We therefore hypothesized that L4H7 might be an Identity Head, and confirmed this by measuring the Identity Head score. 
However, information can also be transmitted across layers through the residual stream itself. Thus, further analysis and discussion are needed to distinguish the contribution of L4H7 from information transmission through the residual stream itself.

\item In Fig.~\ref{fig:wd-pk-top20}, we visualize only the top 20 edges for each of OQ, OK, and OV. From the perspective of interpretability, there may exist better strategies for selecting edges. In practice, reducing the number of edges to the top 10 (PK: Fig.~\ref{fig:wd-pk-top10}, CS: Fig.~\ref{fig:wd-cs-top10} in Appendix~\ref{app:additional-wiring-diagrams}) yields a sparser diagram, whereas increasing it to the top 30 (PK: Fig.~\ref{fig:wd-pk-top30}, CS: Fig.~\ref{fig:wd-cs-top30} in Appendix~\ref{app:additional-wiring-diagrams}) makes the diagram much denser.
Moreover, as the numbers of layers and heads in LLMs increase, selecting edges for visualization may become increasingly difficult when interpreting wiring diagrams of attention heads.

\item PK between the subspaces spanned by head weights captures global interactions between heads, but it cannot capture token-level relationships. For example, on the IOI task, it is known that the output of an S-Inhibition Head affects Name-Mover-related heads via the query vector of the final token~\cite{wang2023interpretability}. Such behavior is not directly visible from Fig.~\ref{fig:wd-pk-top20}.

\item Even with the PK wiring diagram showing the top 30 edges for each pairing (Fig.~\ref{fig:wd-pk-top30} in Appendix~\ref{app:additional-wiring-diagrams}), some head-to-head relationships reported by \citet{wang2023interpretability} are not visualized. For example, the PK values between the output weights of Duplicate Token Heads or Induction Heads and the key weights of the S-Inhibition Head are small, and therefore do not appear as edges.

\item Although many heads exhibit multiple functions~\cite{ahmad2025beyond}, we assign each head to a single head class in our analysis. For example, we label L10H7 as a Negative Name Mover Head, but it also has a high Induction Head score, as shown in Table~\ref{tab:induction} in Appendix~\ref{app:head-scores}\footnote{In addition, L10H7 also has a copy suppression function~\cite{mcdougall-etal-2024-copy}.}.

\item The inlet and outlet scores defined in (\ref{eq:inlet}) and (\ref{eq:outlet}) in Appendix~\ref{app:inlet-outlet} do not correct for frequency biases due to layer position.
For example, in inlet scores, heads in layer 1 have only heads in layer 0 as source candidates, which can lead to relatively smaller scores. Normalizing by the number of candidate heads is possible, but we did not adopt it because it inflated inlet scores for early-layer heads.

\item The token logits obtained by projecting the unembedding matrix onto each head subspace are partly interpretable in terms of the meanings of the top tokens, especially in middle and late layers, but it remains difficult to infer the specific function of a head from these tokens. 
For example, L10H6 has many politics-related tokens among the top tokens, but it is not obvious whether this head actually increases the generation probabilities of such tokens. 
To address this, we aggregated token generation probabilities for each head using the OpenWebText Corpus~\cite{Gokaslan2019OpenWeb}, based on the attention output in (\ref{eq:ohX}) in Appendix~\ref{app:cs}, in a manner analogous to the Logit Lens\footnote{We used \url{https://huggingface.co/datasets/NeelNanda/openwebtext-tokenized-9b}.}.
However, since many contexts are unrelated to the top tokens of each head, the results did not yield meaningful interpretations. Using carefully designed inputs, as in the IOI task, may improve interpretability, but a detailed analysis is left for future work.

\item In GPT2-small, the embedding and unembedding matrices are tied, but we do not provide a detailed analysis of how weight tying affects the projection results.

\item In this work, we adopt a theoretical approximation of the PK distribution between random orthogonal matrices as a common baseline that does not depend on the weight pairing types or the training states. 
This baseline provides a consistent reference distribution for comparison. 
However, as shown in Fig.~\ref{fig:kl-heatmap} (right), even the VO distribution with the smallest KL divergence, while closer to the theoretical distribution in shape than the QQ distribution with the largest KL divergence, still shows a noticeable mismatch in spread. Therefore, significance tests that treat this theoretical distribution as the null may not directly assess differences across weight pairing types, and should be interpreted with caution. To test differences across weight pairing types more directly, it is desirable to introduce an appropriate baseline based on the observed weight matrices.

\item In our discussion of PK, we implicitly assume that each weight matrix $\bm{W}^r\in\mathcal{W}$ has full rank\footnote{Equivalently, the subspace $\mathcal{S}^r\subset\mathbb{R}^d$ spanned by $\bm{W}^r$ has dimension $\dim \mathcal{S}^r=d_{\text{head}}$.}, where $\mathcal{W}\subset \mathbb{R}^{d \times d_{\text{head}}}$ is defined in (\ref{eq:mathcalW}).
In Appendix~\ref{app:check-d-head}, we assess the validity of this assumption by estimating the rank of each weight matrix from its singular values, and find that the assumption is reasonably well supported.

\end{itemize}

\section*{Acknowledgments}
This work was partially supported by JSPS KAKENHI JP22H05106, JP23K28045, JP26H02495 (to HS), JP25K24366 (to HY), JST CREST JPMJCR21N3 (to HS), JST Moonshot JPMJMS263I (to HS), and JST BOOST JPMJBS2407 (to YT).
We thank the developers of TransformerLens~\cite{nanda2022transformerlens} for making their implementation publicly available, which facilitated our analysis.

\section*{Use of AI assistants}
In the camera-ready version, we used generative AI assistants for language polishing, coding support, drafting parts of the manuscript based on experimental code and results, and checking mathematical expressions. All outputs were reviewed and verified by the authors.

\bibliography{custom}

\onecolumn
\appendix

\section{Overview of the appendices}\label{app:Structure-of-the-Appendix}
Appendix~\ref{app:Structure-of-the-Appendix} provides an overview of the subsequent appendices.
Since each appendix in this paper contains relatively independent material, this overview is intended to help readers directly identify and refer to the sections relevant to their interests.
In addition, some appendices provide detailed derivations of the equations required for the discussion, together with the necessary background, so that they can be read independently.

Appendix~\ref{app:related-work} discusses related work.
Appendix~\ref{app:related-work-circuit-analysis} reviews prior work on circuit analysis, Appendix~\ref{app:related-work-model-weight-analysis} reviews analyses of model weights, and Appendix~\ref{app:related-work-attention-pattern-analysis} reviews analyses of attention patterns.

Appendix~\ref{app:cs} provides details of the Composition Score (CS).
As background, following \citet{elhage2021mathematical}, we first derive the QK and OV matrices in Appendix~\ref{app:one-layer-attn} and then describe products of these matrices in Appendix~\ref{app:two-layer-attn}.
Appendix~\ref{app:cs-matrices} describes the matrices used to compute CS.
Next, Appendix~\ref{app:cs-direction-dependent-scaling} discusses the direction-dependent scaling inherent in CS, and Appendix~\ref{app:gpt2-frobenius-norms} examines the properties of the Frobenius norms of the QK and OV matrices used in CS for GPT2-small.
Finally, Appendix~\ref{app:cs-range} derives the range of CS.

Appendix~\ref{app:pk} provides details of the Projection Kernel (PK).
First, Appendix~\ref{app:solving-pk} shows that the optimization problem for the cosine similarities corresponding to the principal angles can be solved analytically.
Next, Appendix~\ref{app:pk-from-orthonormal-bases} derives PK using orthonormal bases, and Appendix~\ref{app:pk-from-orthogonal-projections} derives it using orthogonal projections.
Appendix~\ref{app:rel-bw-pk-and-cs} then establishes the relationship between PK and CS in (\ref{eq:PK-CS}), and finally, Appendix~\ref{app:pk-rotation-invariance} shows that PK is invariant to rotations.

Appendix~\ref{app:pk-cs2} shows that PK depends only on directions by fixing the weight pairing type and computing PK from the matrices used to compute CS\footnote{
For example, for the OQ weight pairing type, CS uses the matrices ${\bm{W}_\text{QK}^{h'}}^\top$ and $\bm{W}_\text{OV}^h$, whereas PK uses ${\bm{W}_\text{Q}^{h'}}^\top$ and $\bm{W}_\text{O}^h$.
Appendix~\ref{app:pk-cs2} derives the subspaces spanned by ${\bm{W}_\text{Q}^{h'}}^\top$ and $\bm{W}_\text{O}^h$ from ${\bm{W}_\text{QK}^{h'}}^\top$ and $\bm{W}_\text{OV}^h$.
}.
Appendix~\ref{app:direction-pk-cs} rewrites PK in a form that allows direct comparison with the square of CS.
Appendix~\ref{app:pk-cs2-summand} then compares the summands in this form for the top five PK and CS pairs of each of the OQ, OK, and OV weight pairing types in GPT2-small, empirically illustrating the distinct characteristics of CS and PK.

Appendix~\ref{app:attn-only-2l} provides details of the wiring diagram for the two-layer attention-only Transformer shown in Fig.~\ref{fig:attn-only-2l} in Section~\ref{sec:attn-only-2l}.

Appendix~\ref{app:proj-details} provides details of the projections of the unembedding matrix in GPT2-small.
First, Appendix~\ref{app:large-norms-trend} shows that tokens with lower generation probabilities, which we regard as less interpretable, tend to have larger unembedding norms.
Next, Appendix~\ref{app:normalization-reason} shows that normalizing the unembedding vectors before projecting them onto the subspace $\widetilde{\mathcal{S}}^r$ reduces the projected norms of less interpretable tokens.
Finally, Appendix~\ref{app:centering-reason} explains why centering is necessary before normalization.

Appendix~\ref{app:head-scores} first provides details of the computation of the head scores in Appendix~\ref{app:head-scores-details}, and then describes the head class annotations for GPT2-small used in our experiments in Appendix~\ref{app:class-annotation}.

Appendix~\ref{app:eval-baselines} provides details of the baselines used in this work: Simple-CS (Appendix~\ref{app:simple-cs}), Linear CKA (Appendix~\ref{app:linear-cka}), and Procrustes Similarity (Appendix~\ref{app:procrustes-similarity}).

Appendix~\ref{app:additional-experiments} presents additional experimental results for each method on GPT2-small.
Appendix~\ref{app:additional-wiring-diagrams} presents wiring diagrams for PK and CS with varying numbers of displayed edges, as well as wiring diagrams for the baseline methods.
Next, Appendix~\ref{app:heatmap-for-all-methods} shows heatmaps for each method for the 16 weight pairing types.
Finally, Appendix~\ref{app:layerwise-metric-trend} compares the layerwise mean values of each method for the QQ, KK, VV, and OO weight pairing types.

Appendix~\ref{app:eval-details} provides details of head detection and head classification for GPT2-small in Appendix~\ref{app:eval-detecting} and Appendix~\ref{app:eval-clustering}, respectively.

Appendix~\ref{app:inlet-outlet} provides details of the inlet and outlet scores for GPT2-small.

Appendix~\ref{app:pk-random} describes approximate distributions of PK between random orthogonal matrices.
First, Appendix~\ref{app:simplified-random-settings} simplifies the setting using the rotation invariance of PK.
Next, Appendix~\ref{app:loose-approx} derives the PK distribution under a loose approximation in which the entries of random orthogonal matrices are approximated by Gaussian distributions.
Finally, Appendix~\ref{app:tight-approx} derives the PK distribution under a tighter approximation using the distributions of the entries of random orthogonal matrices.

Appendix~\ref{app:bloom-qwen2} provides details of additional experiments using BLOOM-560M and Qwen2-0.5B-Instruct.
First, Appendix~\ref{app:model-comparison} compares the architectures of these two models with that of GPT2-small, and Appendix~\ref{app:bloom-qwen2-head-class} reports their head scores and describes the resulting head class annotations.
Next, Appendix~\ref{app:bloom-qwen2-eval-details} provides details of head detection and classification.
Finally, Appendix~\ref{app:bloom-hub} provides details of hub heads in BLOOM-560M.

Appendix~\ref{app:weight-preproc} compares PK and CS using preprocessed weights of GPT2-small.
First, Appendix~\ref{app:wf-bg} provides background on weight preprocessing.
For both PK and CS, we compare the original and preprocessed weights using wiring diagrams in Appendix~\ref{app:wp-pk-vs-cs} and quantitatively evaluate their differences using MSE and Pearson's $r$ in Appendix~\ref{app:eval-wp-vs-org}.
For the preprocessed weights, Appendix~\ref{app:wp-heatmap} shows heatmaps of PK and CS for the 16 weight pairing types, and Appendix~\ref{app:wp-layerwise-metric-trend} compares their layerwise mean values for the QQ, KK, VV, and OO weight pairing types.

Appendix~\ref{app:check-d-head} revisits the assumption made in Section~\ref{sec:proposed} that each attention weight matrix $\bm{W}^r\in \mathcal{W}$ of GPT2-small has full column rank.

\section{Related work}\label{app:related-work}

\subsection{Circuit analysis}\label{app:related-work-circuit-analysis}
Mechanistic interpretability~\cite{mech-interp-essay,sharkey2025open} aims to interpret the internal computations of language models, and circuit analysis is a common approach.
As foundational work, \citet{elhage2021mathematical} formalized the attention computation and, in a two-layer attention-only model, identified a circuit consisting of Previous Token Heads and Induction Heads.
For Induction Heads, \citet{olsson2022context} connected them to in-context learning, and \citet{DBLP:conf/icml/SinghMHCS24} studied how they emerge during training.

As a standard benchmark for circuit analysis, \citet{wang2023interpretability} proposed the Indirect Object Identification (IOI) task and analyzed head classes and circuits in GPT2-small~\cite{radford2019language} in detail.
Subsequent work studies more fine-grained functions of a single head (e.g., L10H7)~\cite{mcdougall-etal-2024-copy}, compares similarities and differences of internal structures in multilingual models~\cite{zhang2025the}, and identifies head-to-head circuits from a single forward pass~\cite{DBLP:journals/corr/abs-2410-00340}.

To scale circuit identification, recent work proposes frameworks that automatically search for and extract important paths~\cite{NEURIPS2023_34e1dbe9,ferrando-voita-2024-information,hsu2025efficient}.
Libraries that support feature-based circuit tracing have also been proposed to facilitate verification~\cite{hanna-etal-2025-circuit}.

There is also work examining the extent to which circuits and head functions are universal across model families~\cite{wang2025towards}.
In addition, \citet{gould2024successor} showed that successor heads, which map tokens in ordered sequences such as numbers and days of the week to their immediate successors, appear in multiple large language models.

\subsection{Model weight analysis}\label{app:related-work-model-weight-analysis}
In contrast to input-dependent analyses, some studies characterize model properties directly from weights.
\citet{elhage2021mathematical} introduced \mbox{Q-,} \mbox{K-,} and V-composition to characterize interactions between attention heads through compositions of the QK and OV matrices.
Weight preprocessing methods that simplify weights without changing the output have been proposed~\cite{nanda2022transformerlens}, and are useful when implementing methods that remove layer normalization~\cite{heimersheim2024you}.
Moreover, \citet{ameisen2025circuit} analyzed the global weights of a replacement model using interference-corrected virtual weights.
In addition, \citet{kamoda-etal-2025-weight} analyzed the 0th-layer QK matrices of GPT2-small to explain detokenization without running inference.

Another active line of work interprets mechanisms within Transformers by decomposing weight matrices using singular value decomposition (SVD).
\citet{merullo2024talking} applied SVD to the QK and OV matrices and analyzed information transmission between attention heads via a small number of singular directions.
\citet{ahmad2025beyond} decomposed the QK and OV matrices as well as MLP weights using SVD, and analyzed multiple functions encoded in these weights in terms of individual singular directions.
\citet{pan2024dissecting} analyzed cosine similarities between singular vectors of QK matrices in Vision Transformers~\cite{dosovitskiy2021an}, while \citet{franco2026singular} theoretically and empirically studied the correspondence between singular vectors of QK matrices and feature directions in activation space.
Relatedly, \citet{wang2026dimensional} applied SVD to attention output activations rather than weights and analyzed their low-dimensional structure.

Another line of work interprets tokens by relating the unembedding matrix to singular vectors of OV matrices~\cite{millidge2022singular} or more generally to weight matrices~\cite{millidge2022singular,dar-etal-2023-analyzing}.
Our method is related to these approaches, but identifies representative tokens by orthogonally projecting the unembedding matrix onto the subspace associated with a specific attention head.

\subsection{Attention pattern analysis}\label{app:related-work-attention-pattern-analysis}
Besides circuit identification, analyses based on visualizations and statistics of attention matrices (patterns) have long been used to characterize attention-head functions.
For example, prior work analyzed attention patterns in BERT~\cite{devlin-etal-2019-bert} and examined their correspondence to linguistic properties~\cite{clark-etal-2019-bert,marecek-rosa-2019-balustrades}.
It has also been reported that many heads are redundant, and pruning them causes little performance degradation~\cite{NEURIPS2019_2c601ad9}.

However, treating attention patterns as explanations requires caution.
The validity of explanations based on attention patterns has been extensively discussed~\cite{wiegreffe-pinter-2019-attention,serrano-smith-2019-attention}, and explanations can be intentionally manipulated~\cite{pruthi-etal-2020-learning}.
Moreover, excessive attention to specific tokens, especially the first token, has been reported, and mitigation methods have been proposed~\cite{miller2023attentionoffbyone,xiao2024efficient,gu2025when,kaul2025from}.

\section{Details of the Composition Score (CS)}\label{app:cs}
As background, following \citet{elhage2021mathematical}, we first derive the QK and OV matrices from the attention computation in a single layer and then explain the matrix products that arise from the attention computation across two layers.
We then describe the matrices used to compute CS and discuss the direction-dependent scaling inherent in CS.
Next, using GPT2-small, we examine layerwise variation in the QK and OV Frobenius norms and the relationship between CS scores and their denominators.
Finally, we show that CS lies in $[0,1]$.

\subsection{Background: deriving the QK and OV matrices from single-layer attention}\label{app:one-layer-attn}
We represent $n$ token embeddings as a matrix $\bm{X}\in\mathbb{R}^{d\times n}$.
Following \citet{elhage2021mathematical}, we describe the attention computation in a single layer with input $\bm{X}$.

For head $h$, the attention pattern $a^h(\bm{X})\in[0,1]^{n\times n}$ for input $\bm{X}$ is computed from the query matrix $\bm{W}_{\text{Q}}^{h}\bm{X}\in\mathbb{R}^{d_\text{head}\times n}$ and the key matrix $\bm{W}_{\text{K}}^{h}\bm{X}\in\mathbb{R}^{d_\text{head}\times n}$ as
\begin{align}
    a^h(\bm{X})
    =\sigma\left(\left(\bm{W}_{\text{Q}}^{h}\bm{X}\right)^\top{\bm{W}_{\text{K}}^{h}}\bm{X}\right)
    =\sigma\left(\bm{X}^\top{\bm{W}_{\text{Q}}^{h}}^\top{\bm{W}_{\text{K}}^{h}}\bm{X}\right)
    =\sigma\left(\bm{X}^\top\bm{W}_{\text{QK}}^{h} \bm{X}\right),\label{eq:ahX}
\end{align}
where $\sigma(\cdot)$ denotes the softmax that includes normalization by $\sqrt{d_{\text{head}}}$ and the causal mask\footnote{
Here $\sigma(\bm{Y})$ denotes the row-wise softmax of $\bm{Y}/\sqrt{d_{\text{head}}}+\bm{M}_{\text{mask}}$ for $\bm{Y}\in\mathbb{R}^{n\times n}$, where $\bm{M}_{\text{mask}}\in\mathbb{R}^{n\times n}$ satisfies $(\bm{M}_{\text{mask}})_{ij}=-\infty$ for $j>i$ and $(\bm{M}_{\text{mask}})_{ij}=0$ otherwise.
}.
The expression in (\ref{eq:ahX}) contains the QK matrix $\bm{W}_{\text{QK}}^{h}\in\mathbb{R}^{d\times d}$ defined in (\ref{eq:QK-matrix}).
For simplicity, we write $\bm{A}^h=a^h(\bm{X})\in[0,1]^{n\times n}$.

Next, we consider the write from head $h$ to the residual stream.
The residual stream accumulates the outputs written by the model components through residual connections~\cite{elhage2021mathematical}.
We denote the contribution written by head $h$ as $o^{h}(\bm{X})\in\mathbb{R}^{d\times n}$.
Using the attention pattern $\bm{A}^h$, the value matrix $\bm{W}_{\text{V}}^{h}\bm{X}\in\mathbb{R}^{d_\text{head}\times n}$, and the output weight matrix $\bm{W}_{\text{O}}^{h}\in\mathbb{R}^{d \times d_\text{head}}$, it is computed as
\begin{equation}
    o^{h}(\bm{X})
    = \bm{W}_{\text{O}}^{h}\bm{W}_{\text{V}}^{h}\bm{X}{\bm{A}^h}^\top
    = \bm{W}_{\text{OV}}^{h}\bm{X}{\bm{A}^h}^\top.\label{eq:ohX}
\end{equation}
The expression in (\ref{eq:ohX}) contains the OV matrix $\bm{W}_{\text{OV}}^{h}\in\mathbb{R}^{d\times d}$ defined in (\ref{eq:OV-matrix}).

\subsection{Background: products of QK and OV matrices from two-layer attention}\label{app:two-layer-attn}
Following \citet{elhage2021mathematical}, we describe the attention computation across two layers\footnote{Note that we ignore the layer normalization and feed-forward layers.} with input $\bm{X}$.
Let $\mathcal{H}_\ell$ be the set of heads in layer $\ell$.
The contribution of the attention heads in layer $\ell$ to the residual stream is
\begin{equation}
    \bm{O}^\ell = \sum_{h\in \mathcal{H}_\ell}o^{h}(\bm{X})\in\mathbb{R}^{d\times n}.\label{eq:Oell}
\end{equation}
We consider the input to a head $h'$ in layer $\ell'(>\ell)$ to be the residual stream after the attention layer in layer $\ell$, namely $\bm{X}+\bm{O}^\ell$.

As in (\ref{eq:ahX}), the attention pattern $\bm{A}^{h'}\in[0,1]^{n\times n}$ is computed as
\begin{align}
    \bm{A}^{h'} = a^{h'}(\bm{X}+\bm{O}^\ell)
    =\sigma\left((\bm{X}+\bm{O}^\ell)^\top\bm{W}_{\text{QK}}^{h'} (\bm{X}+\bm{O}^\ell)\right)\in[0,1]^{n\times n}.\label{eq:Ah-}
\end{align}
The argument of $\sigma(\cdot)$ in (\ref{eq:Ah-}),
$(\bm{X}+\bm{O}^\ell)^\top\bm{W}_{\text{QK}}^{h'} (\bm{X}+\bm{O}^\ell)\in\mathbb{R}^{n\times n}$,
can be expanded as
\begin{align}
&(\bm{X}+\bm{O}^\ell)^\top\bm{W}_{\text{QK}}^{h'} (\bm{X}+\bm{O}^\ell) \notag\\
    =\,&\left(\bm{X}+\sum_{h\in \mathcal{H}_\ell}o^{h}(\bm{X})\right)^\top\bm{W}_{\text{QK}}^{h'} \left(\bm{X}+\sum_{h\in \mathcal{H}_\ell}o^{h}(\bm{X})\right) \quad(\because(\ref{eq:Oell}))\notag\\
    =\,&\bm{X}^\top\bm{W}_{\text{QK}}^{h'}\bm{X}
    + \sum_{h\in \mathcal{H}_\ell}{o^{h}(\bm{X})}^\top\bm{W}_{\text{QK}}^{h'}\bm{X}\notag\\
    &\quad+\sum_{h\in \mathcal{H}_\ell}\bm{X}^\top\bm{W}_{\text{QK}}^{h'}o^{h}(\bm{X})
    +\left(\sum_{h\in \mathcal{H}_\ell}o^{h}(\bm{X})\right)^\top\bm{W}_{\text{QK}}^{h'} \left(\sum_{h\in \mathcal{H}_\ell}o^{h}(\bm{X})\right).\label{eq:XOWQKXO}
\end{align}
To examine weight relationships between heads $h$ and $h'$, we focus on the summand in the second term, which uses $o^{h}(\bm{X})$ and $\bm{X}$ as inputs to the query and key computations, respectively, and the summand in the third term, which uses $\bm{X}$ and $o^{h}(\bm{X})$.

For the summand ${o^{h}(\bm{X})}^\top\bm{W}_{\text{QK}}^{h'} \bm{X}\in\mathbb{R}^{n\times n}$ in the second term of (\ref{eq:XOWQKXO}), we have
\begin{align}
    {o^{h}(\bm{X})}^\top\bm{W}_{\text{QK}}^{h'} \bm{X}
    &=\left( \bm{W}_{\text{OV}}^h \bm{X} {\bm{A}^h}^\top \right)^\top \bm{W}_{\text{QK}}^{h'} \bm{X} \quad(\because(\ref{eq:ohX}))\notag \\
    &={\bm{A}^h} \bm{X}^\top
    \underbrace{{\bm{W}_{\text{OV}}^{h}}^\top\bm{W}_{\text{QK}}^{h'}}_{\mathclap{\text{head-$h$ \bggray{O} $\rightarrow$ head-$h'$ \bgred{Q}}}}
    \bm{X}.
\end{align}
Thus, the relationship from the output of head $h$ to the query of head $h'$ is represented by the matrix product
$\left({\bm{W}_{\text{OV}}^{h}}^\top{\bm{W}_{\text{QK}}^{h'}}\right)^\top
= {\bm{W}_{\text{QK}}^{h'}}^\top{\bm{W}_{\text{OV}}^{h}}$.
 
For the summand $\bm{X}^\top \bm{W}_{\text{QK}}^{h'} o^{h}(\bm{X})\in\mathbb{R}^{n\times n}$ in the third term of (\ref{eq:XOWQKXO}), we have
\begin{align}
    \bm{X}^\top \bm{W}_{\text{QK}}^{h'} o^{h}(\bm{X})
    =\bm{X}^\top
    \underbrace{\bm{W}_{\text{QK}}^{h'}{\bm{W}_{\text{OV}}^{h}}}_{\mathclap{\text{head-$h'$ \bggreen{K} $\leftarrow$ head-$h$ \bggray{O}}}}
    \bm{X}{\bm{A}^h}^\top\quad(\because(\ref{eq:ohX})).
\end{align}
Thus, the relationship from the output of head $h$ to the key of head $h'$ is represented by the matrix product ${\bm{W}_{\text{QK}}^{h'}}{\bm{W}_{\text{OV}}^{h}}$.

As in (\ref{eq:ohX}), the write of head $h'$ to the residual stream, $o^{h'}(\bm{X}+\bm{O}^\ell)\in\mathbb{R}^{d\times n}$, is computed as
\begin{align}
    o^{h'}(\bm{X}+\bm{O}^\ell)
    &=\bm{W}_{\text{OV}}^{h'}\left(\bm{X}+\sum_{h\in \mathcal{H}_\ell}o^{h}(\bm{X})\right){\bm{A}^{h'}}^\top\quad(\because(\ref{eq:Oell}))\notag\\
    &=\bm{W}_{\text{OV}}^{h'}\bm{X}{\bm{A}^{h'}}^\top
    +\sum_{h\in \mathcal{H}_\ell}
    \underbrace{\bm{W}_{\text{OV}}^{h'}\bm{W}_{\text{OV}}^{h}}_{\mathclap{\quad\quad\quad\quad\text{head-$h'$ \bgblue{V} $\leftarrow$ head-$h$ \bggray{O}}}}
    {\bm{X}\bm{A}^{h}}^\top{\bm{A}^{h'}}^\top\quad(\because(\ref{eq:ohX})).
    \label{eq:Oh'OhX}
\end{align}
Thus, the relationship from the output of head $h$ to the value of head $h'$ is represented by the matrix product $\bm{W}_{\text{OV}}^{h'}\bm{W}_{\text{OV}}^{h}$.

\subsection{Matrices used to compute CS}\label{app:cs-matrices}
In Appendix~\ref{app:two-layer-attn}, we saw that the matrix products
${\bm{W}_{\text{QK}}^{h'}}^\top{\bm{W}_{\text{OV}}^{h}}$ for OQ,
${\bm{W}_{\text{QK}}^{h'}}{\bm{W}_{\text{OV}}^{h}}$ for OK,
and $\bm{W}_{\text{OV}}^{h'}\bm{W}_{\text{OV}}^{h}$ for OV
represent weight relationships.
Based on these products, \citet{elhage2021mathematical} compute CS in (\ref{eq:cs}) as the Frobenius norm of each product normalized by the product of the Frobenius norms of the original matrices, which yields $\operatorname{CS}_{\,\text{OQ}}$ in (\ref{eq:OQ-CS}), $\operatorname{CS}_{\,\text{OK}}$ in (\ref{eq:OK-CS}), and $\operatorname{CS}_{\,\text{OV}}$ in (\ref{eq:OV-CS}).

While \citet{elhage2021mathematical} compute CS only for the OQ, OK, and OV pairings, we compute CS for all $4\times 4=16$ combinations of query, key, value, and output weights for heads $h$ and $h'$ to enable a direct comparison with PK.
As shown in Appendix~\ref{app:two-layer-attn}, whether a transpose appears depends on the multiplication order in the attention computation.
Accordingly, we define the matrices associated with the query, key, value, and output of head $h$ as
$\bm{W}_{\text{\bgred{Q}K}}^{h}$,
${\bm{W}_{\text{Q\bggreen{K}}}^{h}}^\top$,
${\bm{W}_{\text{O\bgblue{V}}}^{h}}^\top$,
and $\bm{W}_{\text{\bggray{O}V}}^{h}$,
and the matrices associated with the query, key, value, and output of head $h'$ as
${\bm{W}_{\text{\bgred{Q}K}}^{h'}}^\top$,
${\bm{W}_{\text{Q\bggreen{K}}}^{h'}}$,
${\bm{W}_{\text{O\bgblue{V}}}^{h'}}$,
and ${\bm{W}_{\text{\bggray{O}V}}^{h'}}^\top$.
We compute CS between head weights based on these matrices\footnote{Our notation represents inputs as $d\times n$ matrices whose columns are token vectors, whereas many implementations use $n\times d$ inputs.
Accordingly, the multiplication order of weight matrices and the use of transposes depend on these conventions.}.

\subsection{Details of direction-dependent scaling in CS}\label{app:cs-direction-dependent-scaling}

Section~\ref{sec:cs-limitations} identified direction-dependent scaling as one of the limitations of CS.
In this section, we first rewrite the square of CS in (\ref{eq:cs}) using SVD to show explicitly that CS retains direction-dependent scaling.
We then interpret the Frobenius norm in terms of isotropic inputs and show that CS does not account for directional biases in the actual input distribution.

\subsubsection{Interpretation of CS using SVD}\label{app:cs-svd}

For the matrices\footnote{
In our setting, in particular, $\bm{W}^h\in \left\{\bm{W}_\text{QK}^h,{\bm{W}_\text{QK}^h}^\top, \bm{W}_\text{OV}^h,{\bm{W}_\text{OV}^h}^\top\right\}$.
From (\ref{eq:QK-matrix}) and (\ref{eq:OV-matrix}), each of $\bm{W}_\text{QK}^h$ and $\bm{W}_\text{OV}^h$ can be expressed as the product of a $d\times d_\text{head}$ matrix and a $d_\text{head}\times d$ matrix.
Furthermore, in Section~\ref{sec:proposed}, we assume that these $d\times d_\text{head}$ and $d_\text{head}\times d$ matrices have rank $d_\text{head}$, so $\bm{W}^h$ also has rank $d_\text{head}$.
The same argument applies to $\bm{W}^{h'}$.
} $\bm{W}^{h}, \bm{W}^{h'}\in\mathbb{R}^{d\times d}$ used to compute CS in (\ref{eq:cs}), consider their compact SVDs
\begin{align}
    \bm{W}^{h}&=\bm{U}^h\bm{\Sigma}^h{\bm{V}^h}^\top,\label{eq:Wh-svd}\\
    \bm{W}^{h'}&=\bm{U}^{h'}\bm{\Sigma}^{h'}{\bm{V}^{h'}}^\top.\label{eq:Wh'-svd}
\end{align}
Here, $\bm{U}^h,\,\bm{U}^{h'}\in\mathbb{R}^{d\times d_\text{head}}$ and
$\bm{V}^h,\,\bm{V}^{h'}\in\mathbb{R}^{d\times d_\text{head}}$ are column-orthonormal matrices whose columns are the left and right singular vectors, corresponding to the output-side and input-side directions, respectively.
The matrices $\bm{\Sigma}^h,\,\bm{\Sigma}^{h'}\in\mathbb{R}^{d_\text{head}\times d_\text{head}}$ are diagonal matrices with nonnegative singular values on their diagonals.
We denote these singular values by
$\sigma_1^{h}\geq\ldots\geq\sigma_{d_\text{head}}^{h}\geq 0$ and
$\sigma_1^{h'}\geq\ldots\geq\sigma_{d_\text{head}}^{h'}\geq 0$.
In particular, let
$\bm{u}^h_1,\ldots,\bm{u}^h_{d_\text{head}}\in\mathbb{R}^d$
denote the columns of $\bm{U}^h$, and let
$\bm{v}^{h'}_1,\ldots,\bm{v}^{h'}_{d_\text{head}}\in\mathbb{R}^d$
denote the columns of $\bm{V}^{h'}$.

Substituting (\ref{eq:Wh-svd}) and (\ref{eq:Wh'-svd}) into the square of CS in (\ref{eq:cs}), and using the column orthonormality of the singular-vector matrices, we obtain
\begin{align}
    \operatorname{CS}(\bm{W}^{h'}, \bm{W}^{h})^2
    &=\frac{\left\|\bm{\Sigma}^{h'}{\bm{V}^{h'}}^\top\bm{U}^h\bm{\Sigma}^h\right\|_{\text{F}}^2}{\left\|\bm{\Sigma}^{h'}\right\|_{\text{F}}^2\left\|\bm{\Sigma}^{h}\right\|_{\text{F}}^2}\notag\\
    &=\frac{\sum_{i=1}^{d_\text{head}}\sum_{j=1}^{d_\text{head}} {\sigma_i^{h'}}^2 {\sigma_j^{h}}^2 \left({\bm{v}^{h'}_i}^{\top}\bm{u}^{h}_j\right)^2 }
    {\left(\sum_{i'=1}^{d_\text{head}}{\sigma_{i'}^{h'}}^2\right)\left(\sum_{j'=1}^{d_\text{head}}{\sigma_{j'}^{h}}^2\right)}\notag\\
    &=\sum_{i=1}^{d_\text{head}}\sum_{j=1}^{d_\text{head}}
\left\{\left(\frac{\sigma_i^{h'}}{\sqrt{\sum_{i'=1}^{d_\text{head}}{\sigma_{i'}^{h'}}^2}}\bm{v}^{h'}_i\right)^\top
\left(\frac{\sigma_j^{h}}{\sqrt{\sum_{j'=1}^{d_\text{head}}{\sigma_{j'}^{h}}^2}}\bm{u}^{h}_j\right)\right\}^2.
\label{eq:CS-weighted}
\end{align}
In (\ref{eq:CS-weighted}), each singular vector is weighted by a coefficient proportional to its corresponding singular value.
Consequently, directions associated with larger singular values contribute more strongly to CS.
This is precisely the direction-dependent scaling retained by CS.

\subsubsection{Interpretation of the Frobenius norm using isotropic inputs}
The Frobenius norm can be interpreted as measuring the magnitude of the output of a linear map averaged over isotropic reference inputs.
Let $\bm{I}_d\in\mathbb{R}^{d\times d}$ denote the identity matrix, and let a random vector $\bm{x}\in\mathbb{R}^d$ satisfy $\mathbb{E}_{\bm{x}}[\bm{x}\bm{x}^{\top}]=\bm{I}_d$.
Then, for a matrix $\bm{A}$,
\begin{align}
\mathbb{E}_{\bm{x}}
\left[
\left\|\bm{A}\bm{x}\right\|^2
\right]
=
\operatorname{tr}\left(
\bm{A}^{\top}\bm{A}\,
\mathbb{E}_{\bm{x}}[\bm{x}\bm{x}^{\top}]
\right)
=
\operatorname{tr}\left(\bm{A}^{\top}\bm{A}\right)
= \left\|\bm{A}\right\|_{\text{F}}^2.
\end{align}
Thus, $\left\|\bm{A}\right\|_{\text{F}}$ can be interpreted as the root mean square of $\left\|\bm{A}\bm{x}\right\|$ over isotropic reference inputs $\bm{x}$.

Applying this interpretation to CS, the effect of the direction-dependent scaling retained in $\|\bm{W}^{h'}\bm{W}^{h}\|_{\text{F}}$ can be interpreted in terms of the magnitude of $\|\bm{W}^{h'}\bm{W}^{h}\bm{x}\|$ for isotropic reference inputs $\bm{x}$.
However, contextualized representations in Transformers are known to exhibit anisotropy~\cite{ethayarajh-2019-contextual,yamagiwa-shimodaira-2025-norm}.
Therefore, this interpretation based on isotropic reference inputs does not necessarily reflect the actual activation distribution.
Thus, in addition to retaining direction-dependent scaling, CS does not explicitly account for directional biases in the actual activations.

\subsection{Properties of the Frobenius norms of the QK and OV matrices used in CS for GPT2-small}\label{app:gpt2-frobenius-norms}
In this section, we examine the properties of the Frobenius norms of the QK and OV matrices used in CS for GPT2-small.
Specifically, we first examine their layerwise trends (Appendix~\ref{app:layerwise-qk-ov}) and then investigate the relationship between CS scores and their denominators (Appendix~\ref{app:cs-denominator-vs-cs}).

\subsubsection{Layerwise trends in Frobenius norms}\label{app:layerwise-qk-ov}
\begin{figure}[t!]
    \centering
    \includegraphics[width=0.5\columnwidth]{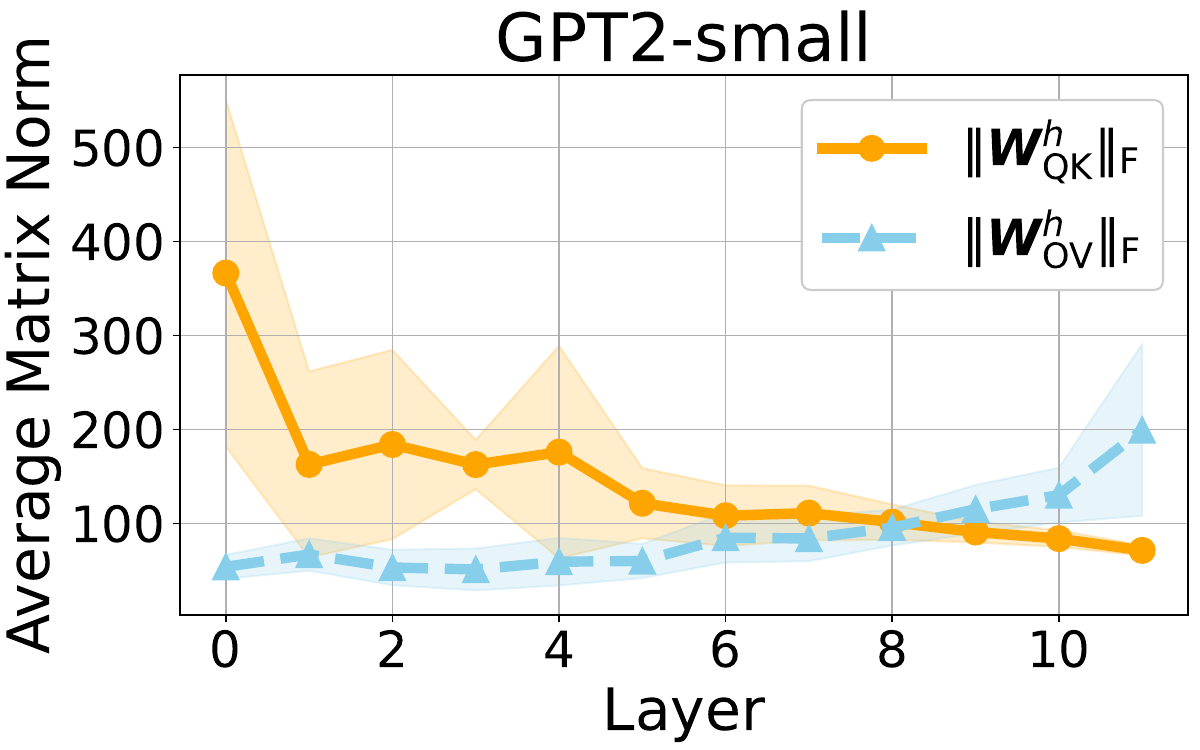}
    \caption{
For the QK matrix in (\ref{eq:QK-matrix}) and the OV matrix in (\ref{eq:OV-matrix}), we plot the Frobenius norms across layers in GPT2-small.
For each layer, we also show the $1\sigma$ range.
}
    \label{fig:layerwise_matrix_norm}
\end{figure}

\begin{figure}[t!]
    \centering
    \includegraphics[width=\columnwidth]{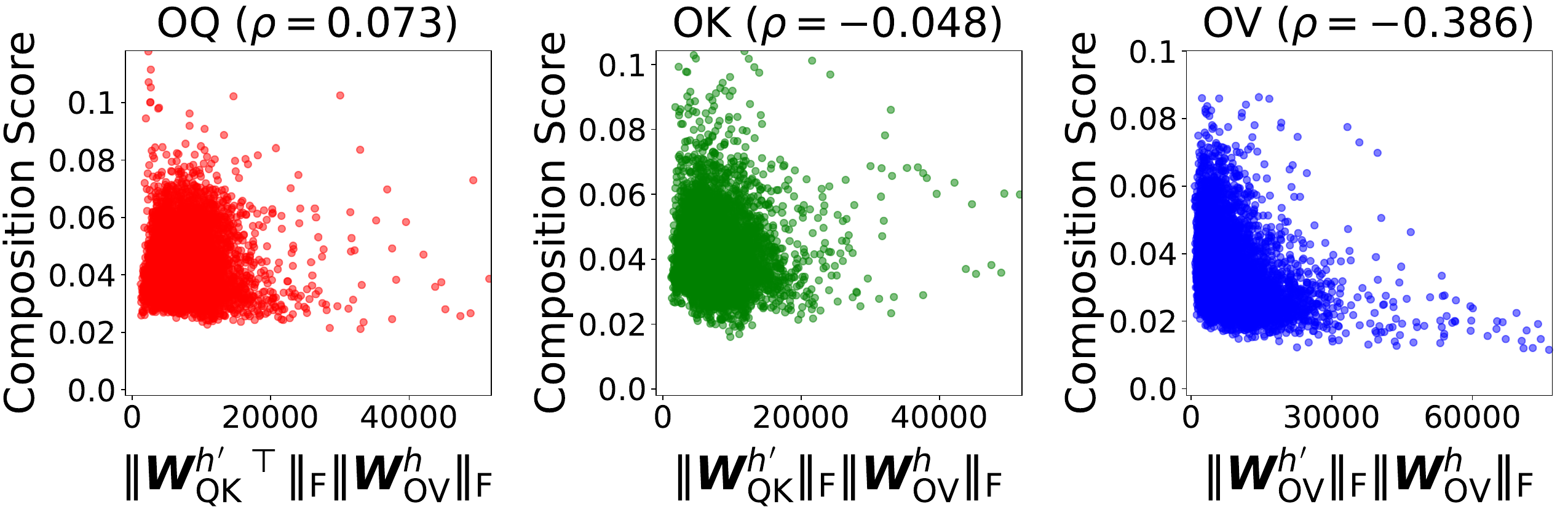}
    \caption{
For each pair consisting of a head in an earlier layer and a head in a later layer in GPT2-small, totaling $9{,}504$ pairs, we plot the OQ, OK, and OV CS scores against the corresponding denominator values.
Each panel also reports Spearman's $\rho$ between these quantities.
}
    \label{fig:norm_vs_cs}
\end{figure}

Figure~\ref{fig:layerwise_matrix_norm} plots the Frobenius norms of the QK and OV matrices across layers in GPT2-small.
The QK norm is large at layer 0 and decreases with depth.
In contrast, the OV norm gradually increases with depth.
We conjecture that these large layerwise changes in weight norms are one reason why CS in (\ref{eq:cs}) can be difficult to use for quantifying differences between heads.

\subsubsection{CS scores can vary with the magnitude of the denominator}\label{app:cs-denominator-vs-cs}
Figure~\ref{fig:norm_vs_cs} plots the OQ, OK, and OV CS scores computed from head pairs in GPT2-small against their corresponding denominator values.
For OV, the CS score tends to decrease as the denominator increases, with a Spearman's rank correlation coefficient of $\rho=-0.386$.

CS in (\ref{eq:cs}) is designed to remove differences in overall matrix scale by normalizing the Frobenius norm of the matrix product in the numerator by the product of the Frobenius norms in the denominator.
However, as observed for OV in GPT2-small, the numerator does not necessarily increase proportionally with the denominator.
In such cases, larger denominator values can suppress the CS score, making the score sensitive to the magnitudes of the Frobenius norms.

\subsection{Range of CS}\label{app:cs-range}
In this section, assuming $\|\bm{W}^{h'}\|_{\text{F}}\|\bm{W}^{h}\|_{\text{F}}\neq 0$, we show that CS in (\ref{eq:cs}) satisfies
\begin{equation}
    0 \leq
    \operatorname{CS}(\bm{W}^{h'}, \bm{W}^{h}) = \frac{\|\bm{W}^{h'}\bm{W}^{h}\|_{\text{F}}}{\|\bm{W}^{h'}\|_{\text{F}}\|\bm{W}^{h}\|_{\text{F}}}
    \leq 1.\label{eq:cs-range}
\end{equation}
When $\|\bm{W}^{h'}\|_{\text{F}} = 0$ or $\|\bm{W}^{h}\|_{\text{F}} = 0$, we define $\operatorname{CS}(\bm{W}^{h'}, \bm{W}^{h})=0$.

The vectors $\bm{v}^{h'}_i$ and $\bm{u}^{h}_j$ defined in Appendix~\ref{app:cs-svd} are columns of the column-orthonormal matrices $\bm{V}^{h'}$ and $\bm{U}^{h}$, respectively, and therefore have unit norm.
Thus, by the Cauchy--Schwarz inequality,
\begin{align}
    0 \leq \left|{\bm{v}^{h'}_i}^{\top}\bm{u}^{h}_j\right|
    \leq \left\|\bm{v}^{h'}_i\right\| \left\|\bm{u}^{h}_j\right\| = 1.\label{eq:Cauchy--Schwarz}
\end{align}
Applying (\ref{eq:Cauchy--Schwarz}) to (\ref{eq:CS-weighted}) gives
\begin{equation}
    0 \leq
    \operatorname{CS}(\bm{W}^{h'}, \bm{W}^{h})^2 \leq 1.
\end{equation}
Since $\operatorname{CS}(\bm{W}^{h'}, \bm{W}^{h})$ is nonnegative, taking the square root yields (\ref{eq:cs-range}).

\section{Details of the Projection Kernel (PK)}\label{app:pk}
In this section, we first solve the optimization problem that defines principal angles~\cite{bjorck1973numerical} following \citet{dahlquist2008numerical}.
Using this result, we show that PK can be expressed in two equivalent forms, one using orthonormal bases and the other using orthogonal projections.
We then derive (\ref{eq:PK-CS}), which characterizes the relationship between PK and CS, and also present the rotation invariance of PK, which will be used in Appendix~\ref{app:pk-random}.

\subsection{Solving the cosine similarity optimization for principal angles}\label{app:solving-pk}
Let $\bm{U},\bm{U}'\in\mathbb{R}^{d\times m}$ be orthonormal basis matrices of $m$-dimensional subspaces $\mathcal{S},\mathcal{S}'\subset\mathbb{R}^d$, and let
\begin{align}
\bm{M}=\bm{U}^\top\bm{U}'\in\mathbb{R}^{m\times m}\label{eq:MUU}
\end{align}
have singular values $\sigma_1\geq \ldots \geq \sigma_m \geq 0$.
Based on \citet[Theorem 8.1.20]{dahlquist2008numerical}, we show that the cosine similarity $\cos\theta_{i}$ of the $i$-th principal angle $\theta_i$ between $\mathcal{S}$ and $\mathcal{S}'$ corresponds to the $i$-th singular value $\sigma_i$ of $\bm{M}$.

We begin with the optimization problem for the first principal angle.
The first principal angle $\theta_1$ is defined by
\begin{equation}
    \cos\theta_{1} =
    \max_{\substack{
        \bm{s}\in\mathcal{S},\ \|\bm{s}\|=1\\
        \bm{s}'\in\mathcal{S}',\ \|\bm{s}'\|=1
    }}
    \bm{s}^\top\bm{s}'.\label{eq:cos-theta1}
\end{equation}
Since $\bm{U}$ and $\bm{U}'$ are orthonormal basis matrices of $\mathcal{S}$ and $\mathcal{S}'$, any $\bm{s}\in\mathcal{S}$ and $\bm{s}'\in\mathcal{S}'$ can be written using coefficients $\bm{g},\bm{g}'\in\mathbb{R}^{m}$ as
\begin{equation}
    \bm{s} = \bm{U}\bm{g},\,\bm{s}' = \bm{U}'\bm{g}'.\label{eq:s-Ug}
\end{equation}
For $\bm{s}$ in (\ref{eq:s-Ug}),
\begin{align}
    \|\bm{s}\|^2
    = \bm{s}^\top \bm{s} 
    = (\bm{U}\bm{g})^\top \bm{U}\bm{g}
    =\bm{g}^\top\bm{U}^\top \bm{U}\bm{g}
    =\bm{g}^\top\bm{g}
    = \|\bm{g}\|^2,
\end{align}
where we used $\bm{U}^\top \bm{U}=\bm{I}_{m}\in\mathbb{R}^{m\times m}$ since the columns of $\bm{U}$ are orthonormal.
Therefore, $\|\bm{s}\|=1$ implies $\|\bm{g}\|=1$, and similarly $\|\bm{s}'\|=1$ implies $\|\bm{g}'\|=1$.
For $\bm{s}, \bm{s}'$ in (\ref{eq:s-Ug}), we obtain
\begin{align}
    \bm{s}^\top\bm{s}' 
    = ({\bm{U}\bm{g}})^\top\bm{U}'\bm{g}'
    =\bm{g}^\top\bm{U}^\top\bm{U}'\bm{g}'
    = \bm{g}^\top\bm{M}\bm{g}'.\label{eq:ssgg}
\end{align}
Using $\|\bm{g}\|=\|\bm{g}'\|=1$ and (\ref{eq:ssgg}), (\ref{eq:cos-theta1}) can be rewritten as
\begin{align}
    \cos\theta_{1} &=
    \max_{\substack{
        \|\bm{g}\| = \|\bm{g}'\|=1
    }}
    \bm{g}^\top\bm{M}\bm{g}'.\label{eq:cos-theta1-gMg}
\end{align}

Let $\bm{M}$ in (\ref{eq:MUU}) be written as $\bm{M}=\bm{U}_M\bm{\Sigma}_M{\bm{V}_M}^\top$ via the SVD.
Here, the diagonal entries of $\bm{\Sigma}_M$ are the singular values $\sigma_1\geq \ldots\geq \sigma_m\geq 0$, and $\bm{U}_M,\,\bm{V}_M\in\mathbb{R}^{m\times m}$ are orthogonal matrices.
For $\bm{g}^\top\bm{M}\bm{g}'$ in (\ref{eq:cos-theta1-gMg}),
\begin{align}
    \bm{g}^\top\bm{M}\bm{g}'
    = \bm{g}^\top\bm{U}_M\bm{\Sigma}_M{\bm{V}_M}^\top\bm{g}'
    = ({\bm{U}_M}^\top \bm{g})^\top \bm{\Sigma}_M({\bm{V}_M}^\top\bm{g}').\label{eq:gMgUmgSigmaMVmg}
\end{align}
Define
\begin{equation}
    \bm{\xi} := {\bm{U}_M}^\top \bm{g},\,
    \bm{\xi}' := {\bm{V}_M}^\top\bm{g}'\in\mathbb{R}^m. \label{eq:xi-UMg}
\end{equation}
Since $\bm{U}_M$ and $\bm{V}_M$ are orthogonal and $\|\bm{g}\|=\|\bm{g}'\|=1$, we have $\|\bm{\xi}\|=\|\bm{\xi}'\|=1$.
Writing $\bm{\xi}=(\xi_1,\ldots,\xi_m)^\top$ and $\bm{\xi}'=(\xi_1',\ldots,\xi_m')^\top$, (\ref{eq:gMgUmgSigmaMVmg}) becomes
\begin{align}
    \bm{g}^\top\bm{M}\bm{g}'
    = \bm{\xi}^\top\bm{\Sigma}_M\bm{\xi}'
    = \sum_{j=1}^{m}\sigma_j \xi_j \xi_j'.\label{eq:gMgxiSigmaxi}
\end{align}

Since $\sigma_1$ is the largest singular value and $\|\bm{\xi}\|=\|\bm{\xi}'\|=1$, the Cauchy--Schwarz inequality gives
\begin{align}
    \sum_{j=1}^{m}\sigma_j \xi_j \xi_j'\notag
    &\leq\left(\sum_{j=1}^{m}\sigma_j^2 {\xi_j}^2 \right)^{1/2}
    \left(\sum_{j=1}^{m}{\xi_j'}^2\right)^{1/2} \notag \\
    &\leq\sigma_1\left(\sum_{j=1}^{m} {\xi_j}^2 \right)^{1/2}
    \left(\sum_{j=1}^{m}{\xi_j'}^2\right)^{1/2}
    = \sigma_1.\label{eq:CSineq}
\end{align}
Equality holds when $\bm{\xi}$ and $\bm{\xi}'$ equal the first standard basis vector $\bm{e}_1^\text{Std}=(1,0,\ldots,0)^\top\in\mathbb{R}^m$.
Therefore, for (\ref{eq:cos-theta1-gMg}) we obtain
\begin{align}
    \cos\theta_{1} = \max_{\substack{
        \|\bm{g}\| = \|\bm{g}'\|=1
    }}
    \bm{g}^\top\bm{M}\bm{g}'
    &= \sigma_1.\label{eq:cos-theta1-sigma1}
\end{align}

Let $\bm{\xi}_1=\bm{\xi}_1'=\bm{e}^\text{Std}_1\in\mathbb{R}^m$ be the maximizers, and let $\bm{g}_1,\bm{g}_1'\in\mathbb{R}^m$ be the corresponding coefficients.
From (\ref{eq:xi-UMg}), we have
\begin{align}
    &\bm{g}_1 =\bm{U}_M\bm{\xi}_1=\bm{U}_M\bm{e}_1^\text{Std}=\bm{u}_{M,1},\label{eq:g1-u1M}\\
    &\bm{g}_1'=\bm{V}_M\bm{\xi}_1'=\bm{V}_M\bm{e}_1^\text{Std}=\bm{v}_{M,1},\label{eq:g-1-u1M}
\end{align}
where $\bm{u}_{M,1}$ and $\bm{v}_{M,1}$ are the first columns of $\bm{U}_M$ and $\bm{V}_M$.
We call the maximizers $\bm{s}=\bm{s}_1$ and $\bm{s}'=\bm{s}_1'\in\mathbb{R}^d$ in (\ref{eq:cos-theta1}) the principal vectors~\cite{dahlquist2008numerical}.
By (\ref{eq:s-Ug}), the principal vectors are
\begin{align}
    &\bm{s}_1 =\bm{U}\bm{g}_1=\bm{U}\bm{u}_{M,1},\label{eq:s1-Ug1}\\
    &\bm{s}_1'=\bm{U}'\bm{g}_1'=\bm{U}'\bm{v}_{M,1}.\label{eq:s-1-Ug1}
\end{align}

Next, the second principal angle $\theta_2$ is defined by
\begin{equation}
    \cos\theta_{2} =
    \max_{\substack{
        \bm{s}\in\mathcal{S},\ \|\bm{s}\|=1,\bm{s}^\top\bm{s}_1=0\\
        \bm{s}'\in\mathcal{S}',\ \|\bm{s}'\|=1,{\bm{s}'}^\top\bm{s}_1' = 0
    }}
    \bm{s}^\top\bm{s}'.\label{eq:cos-theta2}
\end{equation}
From (\ref{eq:s-Ug}) we have $\bm{s}=\bm{U}\bm{g}$, and from (\ref{eq:s1-Ug1}) we have $\bm{s}_1=\bm{U}\bm{g}_1$; therefore,
\begin{align}
    \bm{s}^\top\bm{s}_1 
    = (\bm{U}\bm{g})^\top \bm{U}\bm{g}_1
    =\bm{g}^\top \bm{U}^\top \bm{U}\bm{g}_1
    = \bm{g}^\top \bm{g}_1.
\end{align}
Similarly, ${\bm{s}'}^\top\bm{s}_1'={\bm{g}'}^\top \bm{g}_1'$.
Combining these with $\|\bm{g}\|=\|\bm{g}'\|=1$ and (\ref{eq:ssgg}), we can rewrite (\ref{eq:cos-theta2}) as
\begin{equation}
    \cos\theta_{2} =
    \max_{\substack{
        \|\bm{g}\|=1,\bm{g}^\top\bm{g}_1=0\\
        \|\bm{g}'\|=1,{\bm{g}'}^\top\bm{g}_1'=0
    }}
    \bm{g}^\top\bm{M}\bm{g}'.\label{eq:cos-theta2-g}
\end{equation}
Moreover, from (\ref{eq:xi-UMg}) and (\ref{eq:g1-u1M}),
\begin{align}
    \bm{g}^\top\bm{g}_1
    = (\bm{U}_M\bm{\xi})^\top \bm{U}_M\bm{\xi}_1
    = \bm{\xi}^\top {\bm{U}_M}^\top\bm{U}_M\bm{\xi}_1
    =\bm{\xi}^\top \bm{\xi}_1
    = \bm{\xi}^\top\bm{e}^\text{Std}_1=\xi_1.\label{eq:gg1xi1}
\end{align}
Similarly, ${\bm{g}'}^\top\bm{g}_1'=\xi_1'$.
Using $\|\bm{\xi}\|=\|\bm{\xi}'\|=1$, (\ref{eq:gMgxiSigmaxi}), and (\ref{eq:gg1xi1}), we can rewrite (\ref{eq:cos-theta2-g}) as
\begin{align}
    \cos\theta_{2} =
    \max_{\substack{
        \|\bm{\xi}\|=1,\xi_1=0\\
        \|\bm{\xi}'\|=1,\xi_1'=0
    }}
    \bm{\xi}^\top\bm{\Sigma}_M\bm{\xi}'
    = \max_{\substack{
        \|\bm{\xi}\|=1,\xi_1=0\\
        \|\bm{\xi}'\|=1,\xi_1'=0
    }}\sum_{j=2}^{m}\sigma_j \xi_j \xi_j'.
\end{align}
By the same argument as in (\ref{eq:CSineq}), the maximum is attained at $\bm{\xi}=\bm{\xi}'=\bm{e}_2^\text{Std}$, and we obtain $\cos\theta_{2}=\sigma_2$.
Repeating the same steps for the $i$-th principal angle\footnote{
It is known that the $i$-th canonical correlation is equal to the cosine of the $i$-th principal angle between the column spaces of the data matrices~\cite{bjorck1973numerical}.
We verify this correspondence below.
Let the data matrices be $\bm{X}\in\mathbb{R}^{n\times p}$ and $\bm{Y}\in\mathbb{R}^{n\times q}$, and assume that their columns are centered.
Define
\begin{align*}
\bm{S}_{XX}=\frac{1}{n}\bm{X}^\top\bm{X},\quad
\bm{S}_{YY}=\frac{1}{n}\bm{Y}^\top\bm{Y},\quad
\bm{S}_{XY}=\frac{1}{n}\bm{X}^\top\bm{Y}.
\end{align*}
The $i$-th canonical correlation is defined as
\begin{align*}
\rho_i
:=
\max_{\substack{
\bm{a}^\top\bm{S}_{XX}\bm{a}=1,\,
\bm{a}^\top\bm{S}_{XX}\bm{a}_1=\cdots=
\bm{a}^\top\bm{S}_{XX}\bm{a}_{i-1}=0,\\
\bm{b}^\top\bm{S}_{YY}\bm{b}=1,\,
\bm{b}^\top\bm{S}_{YY}\bm{b}_1=\cdots=
\bm{b}^\top\bm{S}_{YY}\bm{b}_{i-1}=0
}}
\bm{a}^\top\bm{S}_{XY}\bm{b},
\end{align*}
where $\bm{a}_i\in\mathbb{R}^p$ and $\bm{b}_i\in\mathbb{R}^q$ are the vectors that attain the $i$-th canonical correlation.
We assume $\operatorname{rank}(\bm{X})=p$ and $\operatorname{rank}(\bm{Y})=q$.
For rank-deficient matrices, it suffices to restrict the analysis to the subspaces corresponding to nonzero singular values.

Let the SVDs of $\bm{X}$ and $\bm{Y}$ be
\begin{align*}
\bm{X}=\bm{U}_X\bm{\Sigma}_X\bm{V}_X^\top,\quad\bm{Y}=\bm{U}_Y\bm{\Sigma}_Y\bm{V}_Y^\top,
\end{align*}
where
$\bm{U}_X\in\mathbb{R}^{n\times p}$, $\bm{\Sigma}_X\in\mathbb{R}^{p\times p}$, $\bm{V}_X\in\mathbb{R}^{p\times p}$, $\bm{U}_Y\in\mathbb{R}^{n\times q}$, $\bm{\Sigma}_Y\in\mathbb{R}^{q\times q}$, and $\bm{V}_Y\in\mathbb{R}^{q\times q}$.
Under the change of variables
\begin{align*}
&\bm{a}=\sqrt{n}\bm{V}_X\bm{\Sigma}_X^{-1}\bm{\alpha},\quad
\bm{a}_j=\sqrt{n}\bm{V}_X\bm{\Sigma}_X^{-1}\bm{\alpha}_j,\\
&\bm{b}=\sqrt{n}\bm{V}_Y\bm{\Sigma}_Y^{-1}\bm{\beta},\quad
\bm{b}_j=\sqrt{n}\bm{V}_Y\bm{\Sigma}_Y^{-1}\bm{\beta}_j,
\end{align*}
we have
\begin{align*}
&\bm{a}^\top\bm{S}_{XX}\bm{a}=\|\bm{\alpha}\|^2,\quad
\bm{a}^\top\bm{S}_{XX}\bm{a}_j=\bm{\alpha}^\top\bm{\alpha}_j,\\
&\bm{b}^\top\bm{S}_{YY}\bm{b}=\|\bm{\beta}\|^2,\quad
\bm{b}^\top\bm{S}_{YY}\bm{b}_j=\bm{\beta}^\top\bm{\beta}_j,
\end{align*}
and
\begin{align*}
\bm{a}^\top\bm{S}_{XY}\bm{b}=\bm{\alpha}^\top\bm{U}_X^\top\bm{U}_Y\bm{\beta}.
\end{align*}
Therefore,
\begin{align*}
\rho_i
=
\max_{\substack{
\|\bm{\alpha}\|=1,\,
\bm{\alpha}\perp\bm{\alpha}_1,\ldots,\bm{\alpha}_{i-1},\\
\|\bm{\beta}\|=1,\,
\bm{\beta}\perp\bm{\beta}_1,\ldots,\bm{\beta}_{i-1}
}}
\bm{\alpha}^\top
\bm{U}_X^\top\bm{U}_Y
\bm{\beta}.
\end{align*}
Since $\bm{U}_X$ and $\bm{U}_Y$ are orthonormal bases of $\operatorname{col}(\bm{X})$ and $\operatorname{col}(\bm{Y})$, respectively, the right-hand side is equal to the cosine of the $i$-th principal angle.
}
yields
\begin{equation}
\cos\theta_{i}=\sigma_i.\label{eq:cos-theta-i-sigma}
\end{equation}
Moreover, as in (\ref{eq:s1-Ug1}) and (\ref{eq:s-1-Ug1}), if $\bm{S}, \bm{S}'\in\mathbb{R}^{d\times m}$ are matrices whose columns are the principal vectors, then
\begin{equation}
    \bm{S} = \bm{U}\bm{U}_M,\,\bm{S}' = \bm{U}'\bm{V}_M.
\end{equation}

\subsection{Deriving PK using orthonormal bases}\label{app:pk-from-orthonormal-bases}
For $\operatorname{PK}(\mathcal{S},\mathcal{S}')$ in (\ref{eq:pk-def}), using $\cos\theta_i = \sigma_i$ in (\ref{eq:cos-theta-i-sigma}), we can derive $\operatorname{PK}(\mathcal{S}, \mathcal{S}')=\|\bm{U}^\top\bm{U}'\|_{\text{F}}^2$ in (\ref{eq:pk}) as follows:
\begin{align}
\operatorname{PK}(\mathcal{S},\mathcal{S}')
&=\sum_{i=1}^m \cos^2\theta_i= \sum_{i=1}^m \sigma_i^2 
= \operatorname{tr}\left(\bm{\Sigma}_M^\top\bm{\Sigma}_M\right) \notag \\
&=\operatorname{tr}\left({\bm{V}_M}^\top{\bm{V}_M}\bm{\Sigma}_M^\top{\bm{U}_M}^\top\bm{U}_M\bm{\Sigma}_M\right)
=\operatorname{tr}\left({\bm{V}_M}\bm{\Sigma}_M^\top{\bm{U}_M}^\top\bm{U}_M\bm{\Sigma}_M{\bm{V}_M}^\top\right)\notag\\
&=\operatorname{tr}\left(\left(\bm{U}_M\bm{\Sigma}_M{\bm{V}_M}^\top\right)^\top\bm{U}_M\bm{\Sigma}_M{\bm{V}_M}^\top\right)
=\operatorname{tr}\left(\bm{M}^\top\bm{M}\right)
=\left\|\bm{M}\right\|_{\text{F}}^2=\|\bm{U}^\top\bm{U}'\|_{\text{F}}^2.\label{eq:PK-UU}
\end{align}

\subsection{Deriving PK using orthogonal projections}\label{app:pk-from-orthogonal-projections}
For $\operatorname{PK}(\mathcal{S},\mathcal{S}')$ in (\ref{eq:pk-def}), using $\operatorname{PK}(\mathcal{S},\mathcal{S}')=\|\bm{U}^\top\bm{U}'\|_{\text{F}}^2$ in (\ref{eq:PK-UU}), we can derive $\operatorname{PK}(\mathcal{S},\mathcal{S}')=\operatorname{tr}\left(\bm{P}\bm{P}'\right)$ in (\ref{eq:pk}) as follows:
\begin{align}
\operatorname{PK}(\mathcal{S},\mathcal{S}')
&=\|\bm{U}^\top\bm{U}'\|_{\text{F}}^2 = \left\|\bm{M}\right\|_{\text{F}}^2 
= \operatorname{tr}\left(\bm{M}^\top\bm{M}\right)
=\operatorname{tr}\left(({\bm{U}}^\top\bm{U}')^\top {\bm{U}}^\top\bm{U}'\right)\notag \\
&= \operatorname{tr}\left({\bm{U}'}^\top \bm{U} {\bm{U}}^\top\bm{U}'\right)
=\operatorname{tr}\left(\bm{U} {\bm{U}}^\top\bm{U}'{\bm{U}'}^\top \right)
= \operatorname{tr}\left(\bm{P}\bm{P}'\right).\label{eq:PK-Proj}
\end{align}

Furthermore, let $\mathcal{S}_\perp,{\mathcal{S}'_\perp}\subset\mathbb{R}^d$ denote the orthogonal complements of $\mathcal{S}$ and $\mathcal{S}'$, respectively. 
Then, $\operatorname{PK}(\mathcal{S}_\perp,{\mathcal{S}'_\perp})$ can be readily computed from $\operatorname{PK}(\mathcal{S},\mathcal{S}')$.
Since the orthogonal projection matrices onto $\mathcal{S}_\perp$ and ${\mathcal{S}'_\perp}$ are $\bm{I}_d-\bm{P}$ and $\bm{I}_d-\bm{P}'\in\mathbb{R}^{d\times d}$, respectively, we have
\begin{align}
\operatorname{PK}(\mathcal{S}_\perp,{\mathcal{S}'_\perp})
&=\operatorname{tr}\left((\bm{I}_d-\bm{P})(\bm{I}_d-\bm{P}')\right) \notag \\
&=\operatorname{tr}(\bm{I}_d) - \operatorname{tr}(\bm{P}) - \operatorname{tr}(\bm{P}') + \operatorname{tr}\left(\bm{P}\bm{P}'\right) \notag \\
&=d - 2d_\text{head} + \operatorname{PK}(\mathcal{S},\mathcal{S}').
\end{align}
Here, we used the result\footnote{
Let $\bm{U}\in\mathbb{R}^{d\times d_\text{head}}$ and $\bm{U}_\perp\in\mathbb{R}^{d\times (d-d_\text{head})}$ be orthonormal bases of the $d_\text{head}$-dimensional subspace $\mathcal{S}$ and the $(d-d_\text{head})$ dimensional subspace $\mathcal{S}_\perp$, respectively. The orthogonal matrix $\bm{\Upsilon} = [\bm{U}, \bm{U}_\perp]\in\mathbb{R}^{d\times d}$ satisfies $\bm{P}\bm{\Upsilon}=\bm{U}\bm{U}^\top[\bm{U}, \bm{U}_\perp]=[\bm{U}, \bm{O}_{d\times (d - d_\text{head})}]$.
Therefore,
\begin{align*}
\bm{\Upsilon}^\top\bm{P}\bm{\Upsilon} 
=
\begin{bmatrix}
\bm{U}^\top\\
(\bm{U}_{\perp})^\top
\end{bmatrix}
\begin{bmatrix}
\bm{U} & \bm{O}_{d\times(d-d_{\text{head}})}
\end{bmatrix}
=
\begin{bmatrix}
\bm{U}^\top\bm{U}
&
\bm{O}_{d_{\text{head}}\times(d-d_{\text{head}})}
\\
(\bm{U}_{\perp})^\top\bm{U}
&
\bm{O}_{(d-d_{\text{head}})\times(d-d_{\text{head}})}
\end{bmatrix}
=
\begin{bmatrix}
\bm{I}_{d_{\text{head}}}
&
\bm{O}_{d_{\text{head}}\times(d-d_{\text{head}})}
\\
\bm{O}_{(d-d_{\text{head}})\times d_{\text{head}}}
&
\bm{O}_{(d-d_{\text{head}})\times(d-d_{\text{head}})}
\end{bmatrix}.
\end{align*}
Thus, $\operatorname{tr}(\bm{P}) = \operatorname{tr}\left(\bm{P}\bm{\Upsilon}\bm{\Upsilon}^\top\right) = \operatorname{tr}\left(\bm{\Upsilon}^\top\bm{P}\bm{\Upsilon}\right) = \operatorname{tr}(\bm{I}_{d_\text{head}}) = d_\text{head}$. The same argument applies to $\bm{P}'$.
} $\operatorname{tr}(\bm{P})=\operatorname{tr}(\bm{P}')=d_\text{head}$.

\subsection{Relationship between PK and CS in (\ref{eq:PK-CS})}\label{app:rel-bw-pk-and-cs}
Based on the definition of CS in (\ref{eq:cs}), we show (\ref{eq:PK-CS}).

First, for the orthogonal projection matrix $\bm{P}=\bm{U}\bm{U}^\top$ onto $\mathcal{S}$, we have
\begin{align}
    \|\bm{P}\|_\text{F}^2 
    &= \operatorname{tr}\left(\bm{P}^\top \bm{P}\right)
    = \operatorname{tr}\left({(\bm{U}\bm{U}^\top)}^\top\bm{U}\bm{U}^\top\right)
    = \operatorname{tr}\left(\bm{U}\bm{U}^\top\bm{U}\bm{U}^\top\right)\notag\\
    &= \operatorname{tr}\left(\bm{U}\bm{I}_m\bm{U}^\top\right)
    = \operatorname{tr}\left(\bm{U}^\top\bm{U}\right)
    = \operatorname{tr}(\bm{I}_m)
    = m.\label{eq:CS-deno}
\end{align}
Here we used $\bm{U}^\top\bm{U}=\bm{I}_m$ since $\bm{U}$ is an orthonormal basis matrix of $\mathcal{S}$.
Similarly, for the projection matrix $\bm{P}'$ onto $\mathcal{S}'$, we have $\|\bm{P}'\|_\text{F}^2=m$.

Next, for $\bm{P}$ and $\bm{P}'$,
\begin{align}
    \|\bm{P}'\bm{P}\|_\text{F}^2 
    & = \operatorname{tr}(({\bm{P}'\bm{P}})^\top \bm{P}'\bm{P})
    = \operatorname{tr}(\bm{P}^\top {\bm{P}'}^\top\bm{P}'\bm{P})
    = \operatorname{tr}(\bm{P}{\bm{P}'}\bm{P}'\bm{P})\notag\\ 
    &= \operatorname{tr}(\bm{P}\bm{P}\bm{P}'\bm{P}')
    = \operatorname{tr}(\bm{P}\bm{P}')
    = \operatorname{PK}(\mathcal{S}, \mathcal{S}')\quad(\because(\ref{eq:PK-Proj})).\label{eq:CS-nume}
\end{align}
Here we used $\bm{P}^\top = \bm{P}$ and $\bm{P}^2 = \bm{P}$ for an orthogonal projection matrix $\bm{P}$, and the same holds for $\bm{P}'$.

By (\ref{eq:CS-deno}) and (\ref{eq:CS-nume}),
\begin{align}
    \operatorname{CS}(\bm{P}', \bm{P})^2 = \frac{\|\bm{P}'\bm{P}\|_\text{F}^2}{\|\bm{P}'\|_\text{F}^2\|\bm{P}\|_\text{F}^2} 
    = \frac{\operatorname{PK}(\mathcal{S}, \mathcal{S}')}{m^2},
\end{align}
which proves (\ref{eq:PK-CS}).

\subsection{Rotation invariance of PK}\label{app:pk-rotation-invariance}
Let $\bm{\Psi}\in\mathbb{R}^{d\times d}$ be an orthogonal matrix, and let $\mathcal{S}_\Psi$ and $\mathcal{S}_\Psi'$ be the subspaces spanned by $\bm{\Psi}\bm{U}$ and $\bm{\Psi}\bm{U}'$.
The PK between $\mathcal{S}_\Psi$ and $\mathcal{S}_\Psi'$ satisfies
\begin{align}
\operatorname{PK}(\mathcal{S}_\Psi, \mathcal{S}_\Psi') 
&=\operatorname{tr}\left(\bm{\Psi}\bm{U}(\bm{\Psi}\bm{U})^\top\bm{\Psi}\bm{U}'(\bm{\Psi}\bm{U}')^\top\right)
=\operatorname{tr}\left(\bm{\Psi}\bm{U}\bm{U}^\top\bm{\Psi}^\top\bm{\Psi}\bm{U}'{\bm{U}'}^\top\bm{\Psi}^\top\right) \notag\\
&=\operatorname{tr}\left(\bm{U}\bm{U}^\top\bm{U}'{\bm{U}'}^\top\right)
=\operatorname{tr}(\bm{P}\bm{P}') = \operatorname{PK}(\mathcal{S},\mathcal{S}')\quad(\because(\ref{eq:PK-Proj})).
\end{align}
Therefore, PK is unchanged by left multiplication of $\bm{U}$ and $\bm{U}'$ by any orthogonal matrix $\bm{\Psi}\in\mathbb{R}^{d\times d}$.

In Appendix~\ref{app:pk-random}, leveraging this rotation invariance, we simplify the setting of the PK distribution between random orthogonal matrices and derive an approximate distribution of PK.

\section{Computing PK from the matrices used for CS with a fixed weight pairing type}\label{app:pk-cs2}
For a fixed weight pairing type, we derive the subspaces required to compute PK from the matrices used to compute CS, and show that the PK between these subspaces depends only on their directions.
We then examine the squared inner products that constitute the summands of PK and squared CS, where PK has no direction-dependent scaling whereas CS does.

\subsection{Deriving PK from the matrices used to compute CS}\label{app:direction-pk-cs}

We first derive the subspaces $\mathcal{S}'$ and $\mathcal{S}$ spanned by the matrices $\bm{W}^{r'}, \bm{W}^{r}\in\mathcal{W}\subset\mathbb{R}^{d\times d_\text{head}}$ used to compute PK, starting from the matrices $\bm{W}^{h'}, \bm{W}^{h}\in\mathbb{R}^{d\times d}$ used to compute CS in (\ref{eq:cs}).
We then compute $\operatorname{PK}(\mathcal{S}', \mathcal{S})$ in (\ref{eq:pk-def}) between these subspaces\footnote{
From (\ref{eq:pk}), $\operatorname{PK}(\mathcal{S}, \mathcal{S}')=\operatorname{PK}(\mathcal{S}', \mathcal{S})$.
For simplicity, we consider $\operatorname{PK}(\mathcal{S}', \mathcal{S})$ throughout this section.
}.

\subsubsection{Example setting for OQ}

As an example, consider OQ as the weight pairing type.
From (\ref{eq:OQ-CS}), the matrices $\bm{W}^{h'}, \bm{W}^{h}\in\mathbb{R}^{d\times d}$ used to compute CS in (\ref{eq:cs}) are
\begin{align}
    \bm{W}^{h'} &= {\bm{W}_\text{QK}^{h'}}^\top = {\bm{W}_\text{K}^{h'}}^\top{\bm{W}_\text{Q}^{h'}},\\
    \bm{W}^{h} &= \bm{W}_\text{OV}^h = \bm{W}_\text{O}^h \bm{W}_\text{V}^h.
\end{align}
In contrast, the matrices $\bm{W}^{r'}, \bm{W}^{r}\in\mathcal{W}\subset\mathbb{R}^{d\times d_\text{head}}$ used to compute PK are
\begin{align}
\bm{W}^{r'}&={\bm{W}_\text{Q}^{h'}}^\top,\\
\bm{W}^{r}&=\bm{W}_\text{O}^h.
\end{align}

\subsubsection{Deriving the subspaces used to compute PK for OQ}
Let $\operatorname{col}(\bm{A})$ denote the subspace spanned by the columns of $\bm{A}$ and $\operatorname{row}(\bm{A})$ denote the subspace spanned by its rows.
From the SVDs in (\ref{eq:Wh-svd}) and (\ref{eq:Wh'-svd}), for $\mathcal{S}'=\operatorname{col}(\bm{W}^{r'})$, we have\footnote{
The equality
$\operatorname{row}(\bm{W}_{\text{Q}}^{h'})
=
\operatorname{row}({\bm{W}_{\text{QK}}^{h'}}^\top)$
follows from the fact that ${\bm{W}_{\text{K}}^{h'}}^\top\in\mathbb{R}^{d\times d_\text{head}}$ has full column rank.
Indeed, since
${\bm{W}_{\text{QK}}^{h'}}^\top
=
{\bm{W}_{\text{K}}^{h'}}^\top
\bm{W}_{\text{Q}}^{h'}$,
we have
$\operatorname{row}({\bm{W}_{\text{QK}}^{h'}}^\top)
\subseteq
\operatorname{row}(\bm{W}_{\text{Q}}^{h'})$.
Conversely, because ${\bm{W}_{\text{K}}^{h'}}^\top$ has full column rank, there exists a left inverse $\bm{L}$ satisfying
$\bm{L}{\bm{W}_{\text{K}}^{h'}}^\top=\bm{I}_{d_{\text{head}}}$.
Therefore,
$\bm{W}_{\text{Q}}^{h'}
=
\bm{L}{\bm{W}_{\text{QK}}^{h'}}^\top$,
which gives
$\operatorname{row}(\bm{W}_{\text{Q}}^{h'})
\subseteq
\operatorname{row}({\bm{W}_{\text{QK}}^{h'}}^\top)$.
Thus, the two row spaces are identical.
Similarly, from (\ref{eq:Wh'-svd}), since $\bm{U}^{h'}\bm{\Sigma}^{h'}$ has full column rank, we have $\operatorname{row}(\bm{W}^{h'}) = \operatorname{row}({\bm{V}^{h'}}^\top)$.
}
\begin{align}
\operatorname{col}(\bm{W}^{r'}) 
= \operatorname{col}({\bm{W}_\text{Q}^{h'}}^\top)
= \operatorname{row} (\bm{W}_\text{Q}^{h'})
= \operatorname{row} ({\bm{W}_\text{QK}^{h'}}^\top)
= \operatorname{row} (\bm{W}^{h'})
= \operatorname{row} ({\bm{V}^{h'}}^\top) 
= \operatorname{col}(\bm{V}^{h'}).\label{eq:OQ-eg-eq1}
\end{align}
Similarly, for $\mathcal{S}=\operatorname{col}(\bm{W}^{r})$,
\begin{align}
\operatorname{col}(\bm{W}^{r}) 
= \operatorname{col}({\bm{W}_\text{O}^{h}})
=\operatorname{col}({\bm{W}_\text{O}^{h}}{\bm{W}_\text{V}^{h}})
= \operatorname{col}(\bm{W}_\text{OV}^{h})
= \operatorname{col}(\bm{W}^{h})
= \operatorname{col}(\bm{U}^{h}).\label{eq:OQ-eg-eq2}
\end{align}
Thus, from (\ref{eq:OQ-eg-eq1}) and (\ref{eq:OQ-eg-eq2}), for OQ, the subspaces used to compute PK can be derived from the matrices used to compute CS.

\subsubsection{Computing PK between the derived subspaces}
The same argument applies to the other weight pairing types, yielding
\begin{align}
\operatorname{col}(\bm{W}^{r'}) &= \operatorname{col}(\bm{V}^{h'}), \\
\operatorname{col}(\bm{W}^{r}) &= \operatorname{col}(\bm{U}^{h}).
\end{align}
Therefore,
\begin{align}
\operatorname{PK}(\mathcal{S}', \mathcal{S})
=\operatorname{PK}(\operatorname{col}(\bm{V}^{h'}), \operatorname{col}(\bm{U}^{h}))
=\left\|{\bm{V}^{h'}}^\top\bm{U}^{h} \right\|_\text{F}^2
=\sum_{i=1}^{d_\text{head}}\sum_{j=1}^{d_\text{head}} \left({\bm{v}^{h'}_i}^{\top}\bm{u}^{h}_j\right)^2.\label{eq:PK-uniform}
\end{align}
Unlike the squared CS in (\ref{eq:CS-weighted}), which scales the singular vectors in proportion to their singular values, PK in (\ref{eq:PK-uniform}) can be interpreted as a score that depends only on the directions of the singular vectors.

\subsection{Comparison of summands in PK and squared CS}\label{app:pk-cs2-summand}
In this section, we examine the summands of PK and squared CS using empirical data and observe the effect of direction-dependent scaling.
Specifically, we focus on the summand of PK in (\ref{eq:PK-uniform}),
\begin{align}
    \left({\bm{v}^{h'}_i}^{\top}\bm{u}^{h}_j\right)^2,\label{eq:pk-summand}
\end{align}
and the summand of squared CS in (\ref{eq:CS-weighted}),
\begin{align}
\left\{\left(\frac{\sigma_i^{h'}\bm{v}^{h'}_i}{\sqrt{\sum_{i'=1}^{d_\text{head}}{\sigma_{i'}^{h'}}^2}}\right)^\top \left(\frac{\sigma_j^{h}\bm{u}^{h}_j}{\sqrt{\sum_{j'=1}^{d_\text{head}}{\sigma_{j'}^{h}}^2}}\right)\right\}^2.\label{eq:cs2-summand}
\end{align}

\begin{figure*}[p!]
    \centering
    \includegraphics[width=\linewidth]{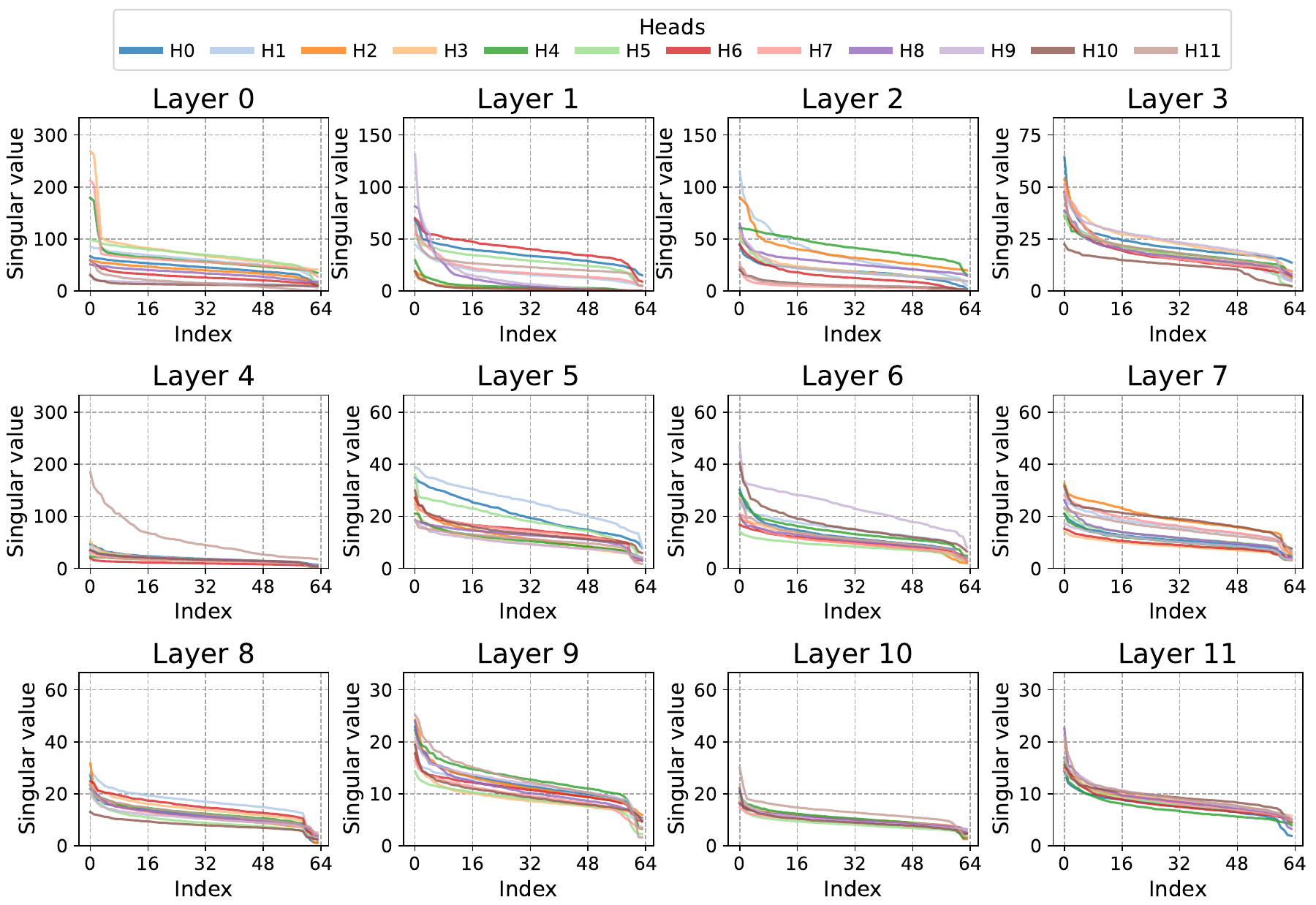}
    \caption{Singular values of GPT2-small QK matrices $\bm{W}_\text{QK}^h$.}
    \label{fig:singular-qk}
\end{figure*}

\begin{figure*}[p!]
    \centering
    \includegraphics[width=\linewidth]{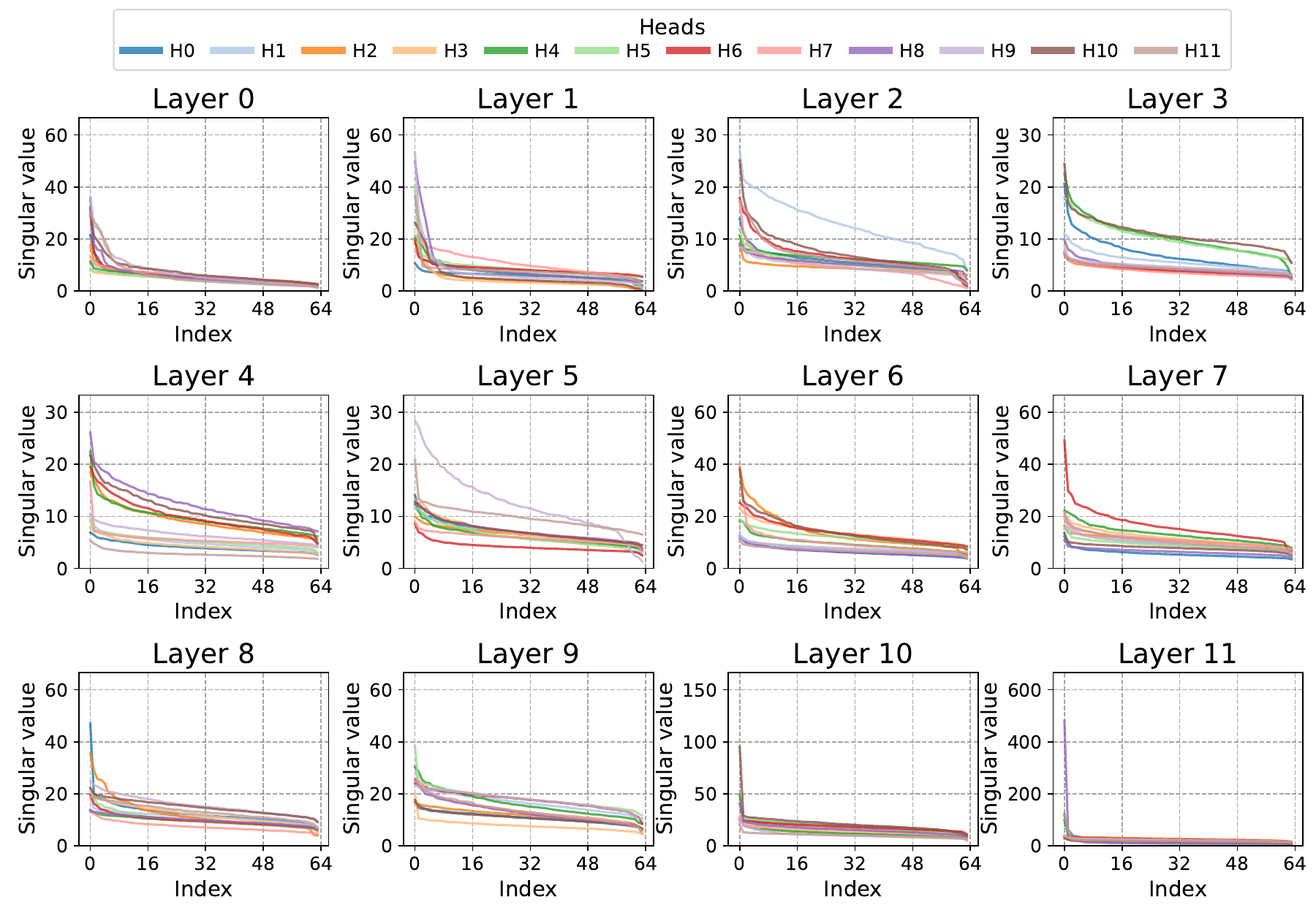}
    \caption{Singular values of GPT2-small OV matrices $\bm{W}_\text{OV}^h$.}
    \label{fig:singular-ov}
\end{figure*}

\begin{figure*}[p!]
    \centering
    \includegraphics[width=\linewidth]{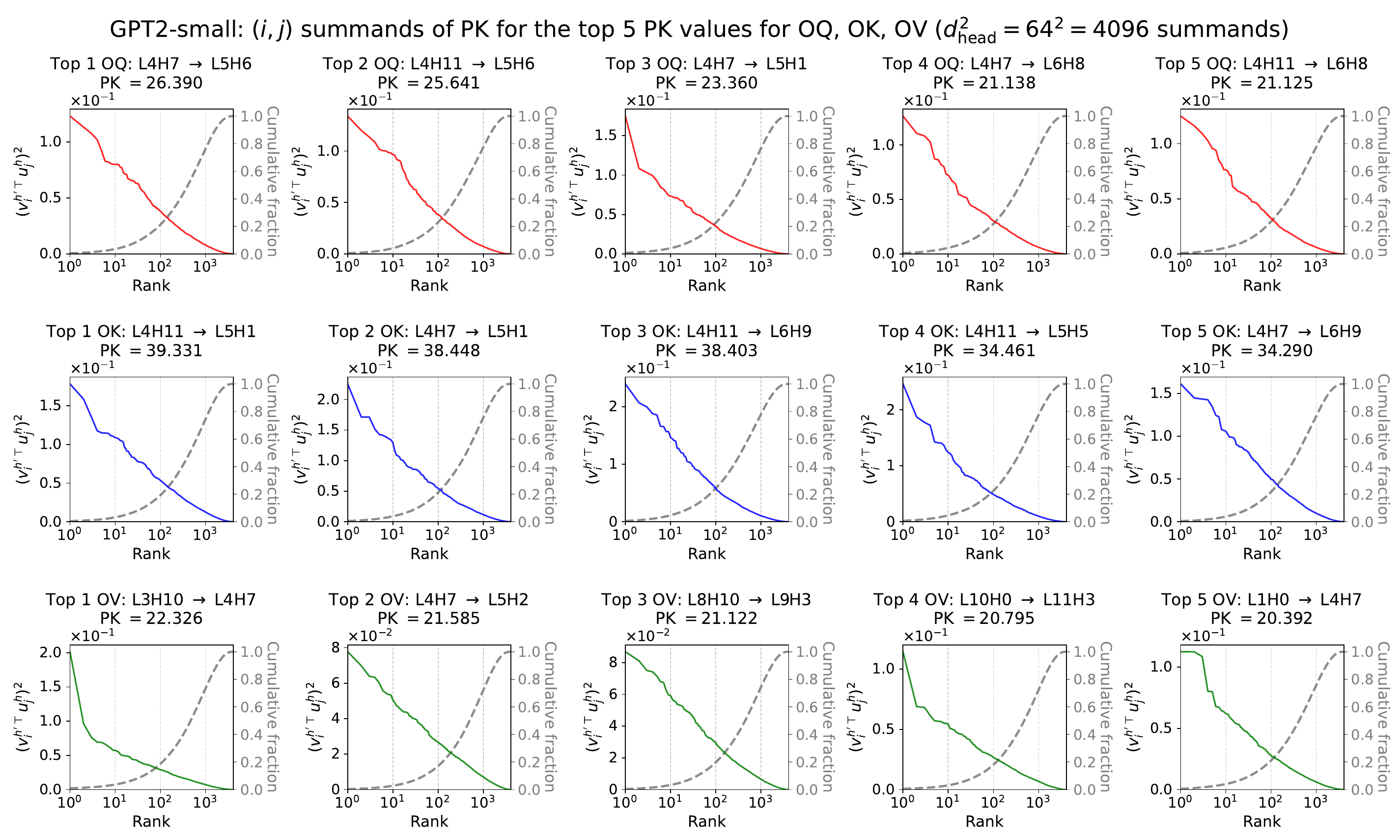}
\caption{Summands of PK for the top 5 PK head pairs for each of OQ, OK, and OV (rows) in GPT2-small; the columns correspond to the top five head pairs. Each panel shows the $d_{\text{head}}^2 = 64^2 = 4096$ summands $({\bm{v}_i^{h'}}^{\top}\bm{u}_j^{h})^2$, which add up to the PK value. Here $\bm{v}_i^{h'}$ and $\bm{u}_j^{h}$ are the leading $d_{\text{head}}$ right and left singular vectors of the matrices used to compute the corresponding CS. The summands are sorted in descending order (solid line, left axis), with their cumulative fraction of the total shown by the dashed line (right axis).}
    \label{fig:pk-components}
\end{figure*}

\begin{figure*}[p!]
    \centering
    \includegraphics[width=\linewidth]{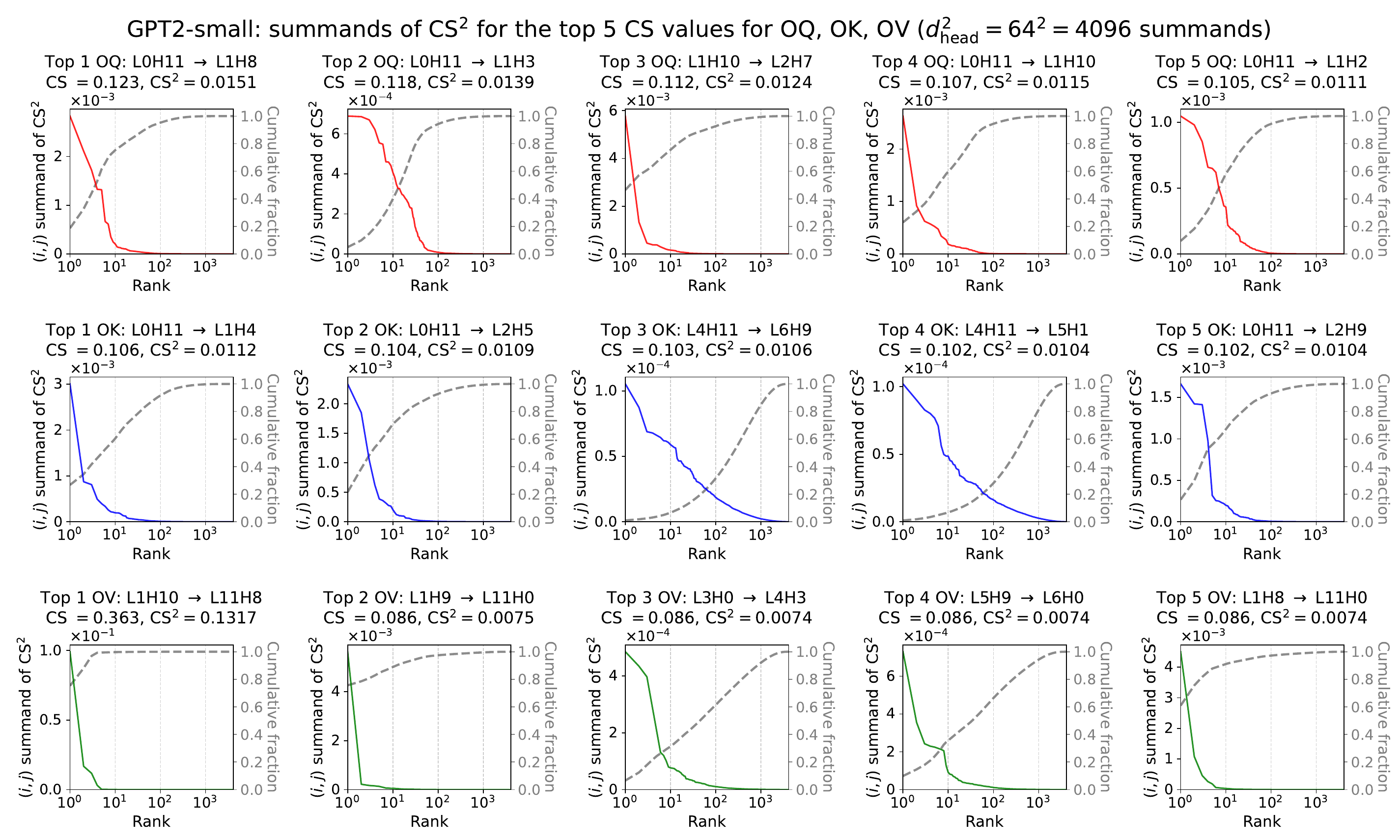}
\caption{Summands of squared CS for the top 5 CS head pairs for each of OQ, OK, and OV (rows) in GPT2-small; the columns correspond to the top five head pairs. Each panel shows the $d_{\text{head}}^2 = 64^2 = 4096$ summands $\{(\sigma_i^{h'}\bm{v}_i^{h'}/\|\bm{\Sigma}^{h'}\|_\text{F})^{\top}(\sigma_j^{h}\bm{u}_j^{h}/\|\bm{\Sigma}^{h}\|_\text{F})\}^2$. Here $\sigma_i^{h'}, \bm{v}_i^{h'}$ and $\sigma_j^{h}, \bm{u}_j^{h}$ are the singular values and the right and left singular vectors of the source and target matrices of CS, respectively. The summands are sorted in descending order (solid line, left axis), with their cumulative fraction of the total shown by the dashed line (right axis).}
    \label{fig:cs-components}
\end{figure*}

We first plot the singular values of the QK and OV matrices for all 144 heads in GPT2-small in Figs.~\ref{fig:singular-qk} and~\ref{fig:singular-ov} to examine their overall trends.
The scales of the singular values vary substantially across layers and heads.
In particular, some heads have large singular values in early layers for the QK matrices and in late layers for the OV matrices\footnote{
Since $\left\|\bm{W}^h\right\|_\text{F}=\left\|\bm{\Sigma}^h\right\|_\text{F}$, the tendency for the norms of the QK matrices to be large in layer 0 and gradually decrease with depth, while those of the OV matrices gradually increase with depth, is consistent with Fig.~\ref{fig:layerwise_matrix_norm}.
}.

To select head pairs for comparing the summands, we follow the same procedure as for the wiring diagrams in Section~\ref{sec:pk-vs-cs} and select the top five head pairs according to PK and CS for each of OQ, OK, and OV.
For these pairs, we sort the summands of PK in (\ref{eq:pk-summand}) and squared CS in (\ref{eq:cs2-summand}) in descending order and plot their values and the cumulative fraction of their sums in Figs.~\ref{fig:pk-components} and~\ref{fig:cs-components}, respectively.
The horizontal axis is shown on a logarithmic scale.

For the summands of squared CS in (\ref{eq:cs2-summand}), which include direction-dependent scaling by the singular values, the summands at higher ranks, typically within the first $10^1$ to $10^2$ ranks, are large for many head pairs, whereas those at lower ranks become nearly negligible.
Consequently, a small number of summands at higher ranks account for most of the cumulative fraction\footnote{
For squared CS, however, the summands decrease more gradually for the third- and fourth-ranked OK pairs involving the QK matrices of L6H9 and L5H11, and for the third- and fourth-ranked OV pairs involving the OV matrices of L4H3 and L6H0.
We attribute this to the more uniform singular values of QK matrices in the middle layers than in early layers, and of OV matrices in the middle layers than in late layers, as shown in Figs.~\ref{fig:layerwise_matrix_norm},~\ref{fig:singular-qk}, and~\ref{fig:singular-ov}.
Furthermore, \citet{merullo2024talking} showed that CS can be high between specific heads that function in the IOI task, such as from S-Inhibition Heads to Name Mover Heads, even when only a small number of singular vectors are used.
}.
In contrast, the summands of PK in (\ref{eq:pk-summand}), which do not include such scaling, decrease more gradually than those of squared CS.
Consequently, the cumulative fraction also increases more gradually.
These results show that CS strongly reflects the influence of a small number of directions associated with large singular values, whereas PK treats the directions of the singular vectors uniformly.

\section{Details of the wiring diagram for a two-layer attention-only Transformer}\label{app:attn-only-2l}
In Fig.~\ref{fig:attn-only-2l} in Section~\ref{sec:attn-only-2l}, using the \texttt{attn-only-2l} model in TransformerLens~\cite{nanda2022transformerlens} (two layers with eight heads per layer), we showed that PK captures the relationship between Previous Token Heads and Induction Heads more clearly than CS.
In this section, we briefly describe how we constructed Fig.~\ref{fig:attn-only-2l}.

For visibility, following \citet{elhage2021mathematical}, we subtract a random bias from the similarity computed for each weight pairing.
This is because PK and CS can be nonzero even for unrelated weights, which shifts the baseline.
Specifically, we generate $8^2 = 64$ pairs of random matrices whose entries follow the standard normal distribution, estimate the bias as the mean similarity over these pairs, and subtract it\footnote{
\citet{elhage2021mathematical} describe this step as \emph{subtract off the empirical expected amount for random matrices of the same shapes}, so the exact distribution is unclear.
However, our CS results (Fig.~\ref{fig:attn-only-2l-cs}) roughly reproduce their findings.}.
After subtraction, we clip values at $0$ and normalize by the maximum value.
In the figures, edge width and opacity are proportional to this normalized similarity.

\section{Details of projections of the unembedding matrix in GPT2-small}\label{app:proj-details}

In Section~\ref{sec:unembedding-proj}, to interpret heads in GPT2-small, we project preprocessed unembedding vectors onto the subspace $\widetilde{\mathcal{S}}^r$ spanned by the weights $\widetilde{\bm{W}}^r$ after the final LN, treat the norm of the projected vector as a logit, and interpret representative tokens.
In particular, we use normalization after centering as preprocessing for the unembedding vectors.
In this section, we first discuss properties of unembedding vectors, then describe the effect of normalization, and finally show why centering before normalization is necessary.

\subsection{Tokens with lower generation probabilities (and lower interpretability) tend to have larger unembedding norms}\label{app:large-norms-trend}
We treat the norm of the projected vector as a logit.
Therefore, for less interpretable tokens, it is preferable that the projected norm is small.
We accordingly regard tokens with low generation probabilities as less interpretable.
For this purpose, we need a tokenwise quantity that represents how easy a token is to generate, independent of the input prompt.

To define such a quantity, we first review the standard logit computation.
In language models such as GPT2, logits are computed by applying layer normalization (LN) to the final layer output and then multiplying by the unembedding matrix.
Let this final LN be $\operatorname{LN}_{\text{final}}$.
Using the scaling weights $\bm{\gamma}_\text{final}\in\mathbb{R}^d$ and the bias $\bm{\beta}_\text{final}\in\mathbb{R}^d$, we write
\begin{equation}
    \operatorname{LN}_{\text{final}}(\bm{x}) = \bm{\gamma}_\text{final} \odot \frac{\bm{x} - \operatorname{Mean}(\bm{x})\bm{1}_d}{\operatorname{Std}(\bm{x})} + \bm{\beta}_\text{final},\label{eq:LN-final}
\end{equation}
where $\odot$ denotes the Hadamard product, $\bm{1}_d\in\mathbb{R}^d$ is the $d$-dimensional all-ones vector, and $\operatorname{Mean}(\cdot)$ and $\operatorname{Std}(\cdot)$ denote the mean and standard deviation.

Based on $\operatorname{LN}_{\text{final}}$, we use the inner product between the unembedding vector $\bm{e}_t$ and the bias term $\bm{\beta}_{\text{final}}$ in (\ref{eq:LN-final}), namely $\bm{e}_t^\top \bm{\beta}_{\text{final}}$, as a token-specific statistic related to generation probabilities\footnote{See \url{https://transformerlensorg.github.io/TransformerLens/generated/demos/Main_Demo.html} for details.}~\cite{nanda2022transformerlens}.
In fact, the logit for token $t$ is computed as the inner product between the unembedding vector $\bm{e}_t$ and $\operatorname{LN}_{\text{final}}(\bm{x})$ in (\ref{eq:LN-final}),
\begin{align}
    \bm{e}_t^\top\operatorname{LN}_{\text{final}}(\bm{x})
    =\bm{e}_t^\top\left(\bm{\gamma}_\text{final} \odot \frac{\bm{x} - \operatorname{Mean}(\bm{x})\bm{1}_d}{\operatorname{Std}(\bm{x})}\right) + \bm{e}_t^\top\bm{\beta}_\text{final},\label{eq:et-LN-final}
\end{align}
where the first term on the right-hand side of (\ref{eq:et-LN-final}) depends on the final-layer output $\bm{x}$, whereas the second term $\bm{e}_t^\top\bm{\beta}_\text{final}$ is an input-independent, token-specific statistic (i.e., a scalar bias term).
In particular, tokens with more negative $\bm{e}_t^\top \bm{\beta}_{\text{final}}$ tend to have smaller probabilities under the softmax.

\begin{figure}[t!]
    \centering
    \includegraphics[width=\linewidth]{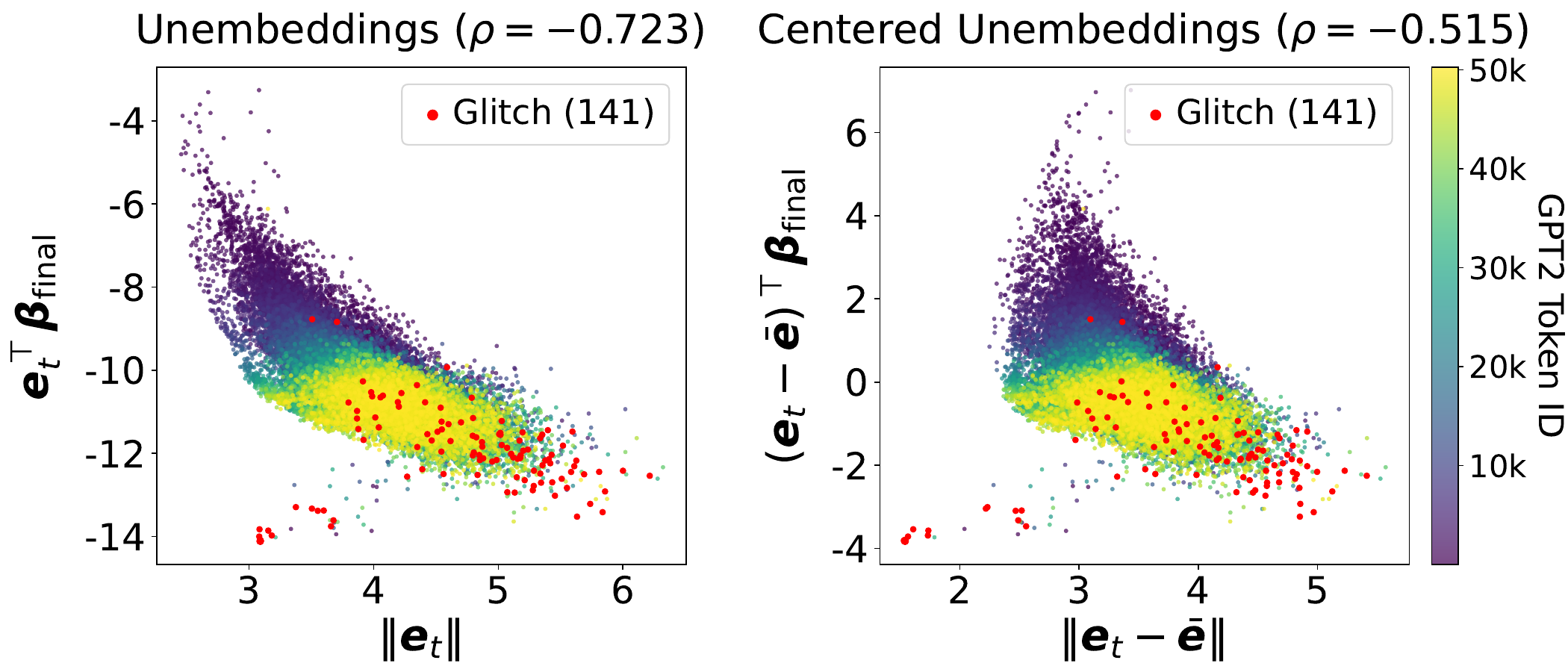}
    \caption{
Scatter plots of the norm of unembedding vectors in GPT2-small and the inner product with the bias term $\bm{\beta}_\text{final}$ of the final layer normalization in (\ref{eq:LN-final}), (left) without centering and (right) with centering.
Since GPT2 uses byte pair encoding (BPE), the token ID can be viewed as a proxy for frequency, where larger token IDs tend to be less frequent~\cite{DBLP:conf/nips/HayaseL0OS24} (see Fig.~\ref{fig:gpt2-small_token_id_vs_freq_scatter}).
Without centering (left), the norm and the inner product with $\bm{\beta}_\text{final}$ show a strong negative correlation.
With centering (right), the correlation becomes slightly weaker but the same trend remains.
}
    \label{fig:norms-vs-b-final-inner-product}
\end{figure}

\begin{figure}[t!]
    \centering
    \includegraphics[width=0.75\linewidth]{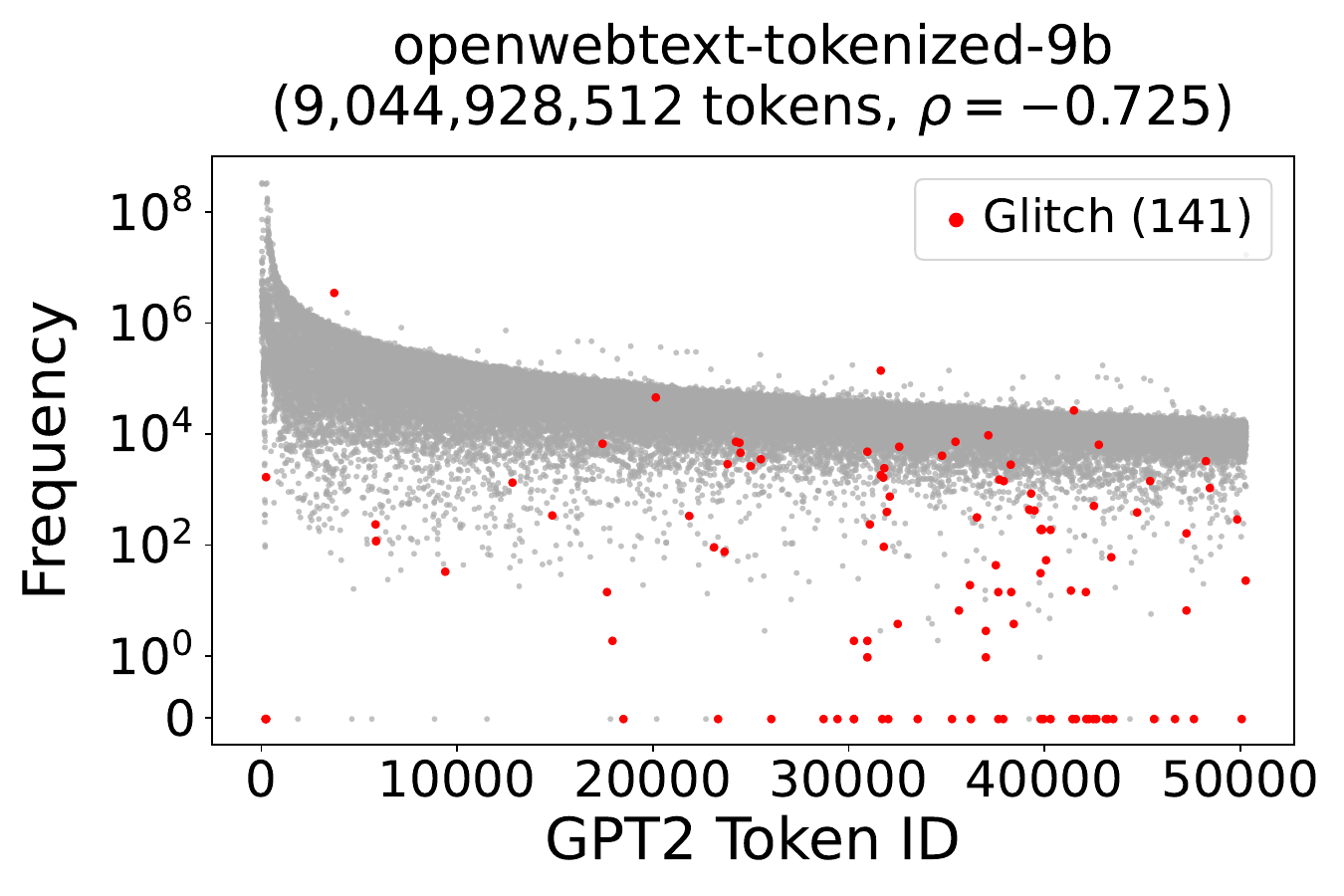}
    \caption{
Scatter plot of GPT2 token IDs versus token frequencies in the OpenWebText Corpus~\cite{Gokaslan2019OpenWeb} (9B tokens).
Despite differences between OpenWebText and the undisclosed GPT2 tokenizer training corpus, larger token IDs tend to be less frequent, consistent with \citet{DBLP:conf/nips/HayaseL0OS24} (Spearman's $\rho=-0.725$).
Since token frequencies vary across datasets, we use token ID as a dataset-agnostic proxy for frequency.
Many of the 141 glitch tokens identified by \citet{rumbelow2023solidgoldmagikarp} in Table~\ref{tab:gpt2_small_glitch_tokens} have zero frequency in OpenWebText.
}
    \label{fig:gpt2-small_token_id_vs_freq_scatter}
\end{figure}

To understand the behavior of $\bm{e}_t^\top \bm{\beta}_{\text{final}}$, we examine its relationship with the unembedding norm.
For comparison, we also consider centering the unembedding vectors using their mean $\bar{\bm{e}}\in\mathbb{R}^d$.
Figure~\ref{fig:norms-vs-b-final-inner-product} shows scatter plots of the unembedding norm and the inner product, with and without centering.
In Fig.~\ref{fig:norms-vs-b-final-inner-product}, as examples of less interpretable tokens, we label glitch tokens~\cite{rumbelow2023solidgoldmagikarp,rumbelow2023solidgoldmagikarp2}, which are extremely low-frequency tokens whose embedding parameters are poorly trained\footnote{We use the list of 141 glitch tokens provided by \citet{rumbelow2023solidgoldmagikarp} in Table~\ref{tab:gpt2_small_glitch_tokens}.}.
With or without centering, larger unembedding norms correspond to more negative inner products with $\bm{\beta}_{\text{final}}$, suggesting smaller generation probabilities and lower interpretability.
In fact, many tokens with large norms are low-frequency tokens, including glitch tokens\footnote{Note that some glitch tokens also have small norms.}.

\begin{table*}[t]
\centering
\scriptsize
\begin{tabular}{rl rl rl}
\toprule
ID & Token & ID & Token & ID & Token \\
\midrule
188 & \texttt{\textbackslash{}x00} & 29372 & \texttt{\_guiActiveUn} & 39177 & \texttt{ItemThumbnailImage} \\
189 & \texttt{\textbackslash{}x01} & 30202 & \texttt{\_guiName} & 39253 & \texttt{\_UCHIJ} \\
190 & \texttt{\textbackslash{}x02} & 30208 & \texttt{\_externalTo} & 39446 & \texttt{\_SetFontSize} \\
191 & \texttt{\textbackslash{}x03} & 30209 & \texttt{\_unfocusedRange} & 39749 & \texttt{DeliveryDate} \\
192 & \texttt{\textbackslash{}x04} & 30210 & \texttt{\_guiActiveUnfocused} & 39752 & \texttt{quickShip} \\
193 & \texttt{\textbackslash{}x05} & 30211 & \texttt{\_guiIcon} & 39753 & \texttt{quickShipAvailable} \\
194 & \texttt{\textbackslash{}x06} & 30212 & \texttt{\_externalToEVA} & 39755 & \texttt{isSpecialOrderable} \\
195 & \texttt{\textbackslash{}x07} & 30213 & \texttt{\_externalToEVAOnly} & 39756 & \texttt{inventoryQuantity} \\
196 & \texttt{\textbackslash{}x08} & 30897 & \texttt{reportprint} & 39757 & \texttt{channelAvailability} \\
202 & \texttt{\textbackslash{}x0E} & 30898 & \texttt{embedreportprint} & 39803 & \texttt{soType} \\
203 & \texttt{\textbackslash{}x0F} & 30899 & \texttt{cloneembedreportprint} & 39811 & \texttt{soDeliveryDate} \\
204 & \texttt{\textbackslash{}x10} & 30905 & \texttt{rawdownload} & 39821 & \texttt{龍契士} \\
205 & \texttt{\textbackslash{}x11} & 30906 & \texttt{rawdownloadcloneembedreportprint} & 39906 & \texttt{EStream} \\
206 & \texttt{\textbackslash{}x12} & 31032 & \texttt{SpaceEngineers} & 40012 & \texttt{uyomi} \\
207 & \texttt{\textbackslash{}x13} & 31576 & \texttt{externalActionCode} & 40240 & \texttt{oreAndOnline} \\
208 & \texttt{\textbackslash{}x14} & 31583 & \texttt{к} & 40241 & \texttt{InstoreAndOnline} \\
209 & \texttt{\textbackslash{}x15} & 31666 & \texttt{?????-?????-} & 40242 & \texttt{BuyableInstoreAndOnline} \\
210 & \texttt{\textbackslash{}x16} & 31708 & \texttt{ーン} & 41297 & \texttt{\_TAMADRA} \\
211 & \texttt{\textbackslash{}x17} & 31727 & \texttt{cffff} & 41380 & \texttt{natureconservancy} \\
212 & \texttt{\textbackslash{}x18} & 31765 & \texttt{MpServer} & 41383 & \texttt{assetsadobe} \\
213 & \texttt{\textbackslash{}x19} & 31886 & \texttt{\_gmaxwell} & 41441 & \texttt{\textbackslash{}-} \\
214 & \texttt{\textbackslash{}x1A} & 31957 & \texttt{cffffcc} & 41551 & \texttt{Downloadha} \\
215 & \texttt{\textbackslash{}x1B} & 32047 & \texttt{\_"\$:/} & 42066 & \texttt{Nitrome} \\
221 & \texttt{\textbackslash{}x7F} & 32437 & \texttt{\_Smartstocks} & 42089 & \texttt{\_TheNitrome} \\
3693 & \texttt{.[} & 32509 & \texttt{":[\{"} & 42090 & \texttt{\_TheNitromeFan} \\
5808 & \texttt{ÃÂ} & 33454 & \texttt{龍喚士} & 42202 & \texttt{GoldMagikarp} \\
5815 & \texttt{ÃÂÃÂ} & 34713 & \texttt{":"","} & 42424 & \texttt{DragonMagazine} \\
9364 & \texttt{ÃÂÃÂÃÂÃÂ} & 35207 & \texttt{\_attRot} & 42470 & \texttt{TextColor} \\
12781 & \texttt{wcsstore} & 35384 & \texttt{''.} & 42586 & \texttt{\_srfN} \\
14827 & \texttt{ÃÂÃÂÃÂÃÂÃÂÃÂÃÂÃÂ} & 35579 & \texttt{\_Mechdragon} & 42728 & \texttt{\_largeDownload} \\
17405 & \texttt{\textbackslash{}.} & 36130 & \texttt{\_PsyNet} & 43065 & \texttt{\_srfAttach} \\
17629 & \texttt{\_practition} & 36173 & \texttt{\_RandomRedditor} & 43177 & \texttt{EStreamFrame} \\
17900 & \texttt{\_Dragonbound} & 36174 & \texttt{\_RandomRedditorWithNo} & 43361 & \texttt{ゼウス} \\
18472 & \texttt{\_guiActive} & 36481 & \texttt{ertodd} & 43453 & \texttt{\_SolidGoldMagikarp} \\
20126 & \texttt{\_\textbackslash{}u200B} & 36938 & \texttt{\_sqor} & 44686 & \texttt{ーティ} \\
21807 & \texttt{\textbackslash{}\textbackslash{}\textbackslash{}\textbackslash{}\textbackslash{}\textbackslash{}\textbackslash{}\textbackslash{}} & 36940 & \texttt{\_istg} & 45335 & \texttt{aterasu} \\
23090 & \texttt{ÃÂÃÂÃÂÃÂÃÂÃÂÃÂÃÂÃÂÃÂÃÂÃÂÃÂÃÂÃÂÃÂ} & 37082 & \texttt{\_"\textbackslash{}} & 45544 & \texttt{\_サーティ} \\
23282 & \texttt{\_davidjl} & 37444 & \texttt{\_petertodd} & 45545 & \texttt{\_サーティワン} \\
23614 & \texttt{覚醒} & 37574 & \texttt{StreamerBot} & 46600 & \texttt{\_Adinida} \\
23785 & \texttt{"]=>} & 37579 & \texttt{TPPStreamerBot} & 47182 & \texttt{":""\},\{"} \\
24200 & \texttt{\_--------} & 37631 & \texttt{FactoryReloaded} & 47198 & \texttt{ItemTracker} \\
24398 & \texttt{\_\textbackslash{}u200E} & 37842 & \texttt{\_partName} & 47571 & \texttt{\_DevOnline} \\
24440 & \texttt{ュ} & 37858 & \texttt{ヤ} & 48193 & \texttt{@\#\&} \\
24934 & \texttt{ForgeModLoader} & 38214 & \texttt{\textbackslash{}">} & 48396 & \texttt{ÛÛ} \\
25465 & \texttt{天} & 38250 & \texttt{\_Skydragon} & 49781 & \texttt{EngineDebug} \\
25992 & \texttt{\_裏覚醒} & 38370 & \texttt{iHUD} & 50009 & \texttt{\_strutConnector} \\
28666 & \texttt{PsyNetMessage} & 39165 & \texttt{catentry} & 50216 & \texttt{\_Leilan} \\
\bottomrule
\end{tabular}
\caption{
The 141 glitch tokens in GPT2 identified by \citet{rumbelow2023solidgoldmagikarp}. Here, \_ denotes a leading space, and \texttt{\textbackslash{}xNN} a non-printable byte with hexadecimal code \texttt{NN} (e.g., \texttt{\textbackslash{}x00} for ID 188 and \texttt{\textbackslash{}x7F} for ID 221).
}
\label{tab:gpt2_small_glitch_tokens}
\end{table*}

\begin{table*}[t!]
\scriptsize
\centering
\begin{tabular}{@{\hspace{0.2em}}llrrrrr@{\hspace{0.2em}}}
  \toprule
  $f(\bm{e}_t)$ & Unembed. Norm & Query & Key & Value & Output & All \\
  \midrule
  $\bm{e}_t$ & $\|\bm{e}_t\|$ & $0.750 \pm 0.049$ & $0.757\pm 0.045$ & $0.745\pm 0.063$ & $0.670\pm 0.091$ & $0.731 \pm 0.074$ \\
  $\bm{e}_t - \bar{\bm{e}} $ & $\|\bm{e}_t - \bar{\bm{e}}\|$  & $0.605\pm 0.084$ & $0.625\pm 0.078$ & $0.670\pm 0.108$ & $0.547\pm 0.147$ & $0.612 \pm 0.116$ \\
  $\bm{e}_t/\|\bm{e}_t\|$ & $\|\bm{e}_t\|$  & $-0.392\pm 0.092$ & $-0.363\pm 0.086$ & $-0.382\pm 0.127$ & $-0.402\pm 0.145$ & $-0.385 \pm 0.116$ \\
  $(\bm{e}_t - \bar{\bm{e}})/\|\bm{e}_t - \bar{\bm{e}}\|$ & $\|\bm{e}_t - \bar{\bm{e}}\|$ & $-0.198\pm 0.138$ & $-0.168\pm 0.139$ & $-0.054\pm0.226$ & $-0.163\pm 0.185$ & $-0.146 \pm 0.184$\\
  \bottomrule
\end{tabular}
\caption{
For the 144 heads in GPT2-small, we report the mean and standard deviation of Spearman's $\rho$ between the unembedding norm and the norm of the vector projected onto $\widetilde{\mathcal{S}}^r$ after preprocessing $f$, separately for each weight type.
We also report the results when ignoring weight types (All).
Without normalization ($f(\bm{e}_t) = \bm{e}_t$ or $f(\bm{e}_t) = \bm{e}_t-\bar{\bm{e}}$), the pre- and post-projection norms show a strong positive correlation.
With normalization ($f(\bm{e}_t) = \bm{e}_t/\|\bm{e}_t\|$ or $f(\bm{e}_t) = (\bm{e}_t-\bar{\bm{e}})/\|\bm{e}_t - \bar{\bm{e}}\|$), we instead observe a weak negative correlation.
}
\label{tab:norms-vs-projected-norms}
\end{table*}

\subsection{Normalization before projection reduces projected norms for less interpretable tokens}
\label{app:normalization-reason}
We consider projecting preprocessed unembedding vectors onto the subspace $\widetilde{\mathcal{S}}^r$ spanned by $\widetilde{\bm{W}}^r$, where $\widetilde{\bm{W}}^r$ is obtained by applying $\operatorname{LN}_{\text{final}}$ to $\bm{W}^r\in\mathcal{W}$.
As candidates for preprocessing $f$, we consider the identity map, centering, normalization, and normalization after centering.
Table~\ref{tab:norms-vs-projected-norms} reports, for these four choices, the correlation between the norm of the (centered or uncentered) unembedding vector and the norm of its projection onto $\widetilde{\mathcal{S}}^r$.
Without normalization, the pre- and post-projection norms show a strong positive correlation, whereas with normalization we observe a weak negative correlation.

From Fig.~\ref{fig:norms-vs-b-final-inner-product} and Table~\ref{tab:norms-vs-projected-norms}, we obtain the following.
Without normalization, less interpretable tokens that have large unembedding norms and strongly negative inner products with $\bm{\beta}_{\text{final}}$ tend to also have large projected norms.
With normalization, such tokens instead tend to have small projected norms.
Therefore, to treat the projected norm as a logit, normalization appears necessary.

\begin{figure}[t!]
    \centering
    \includegraphics[width=\linewidth]{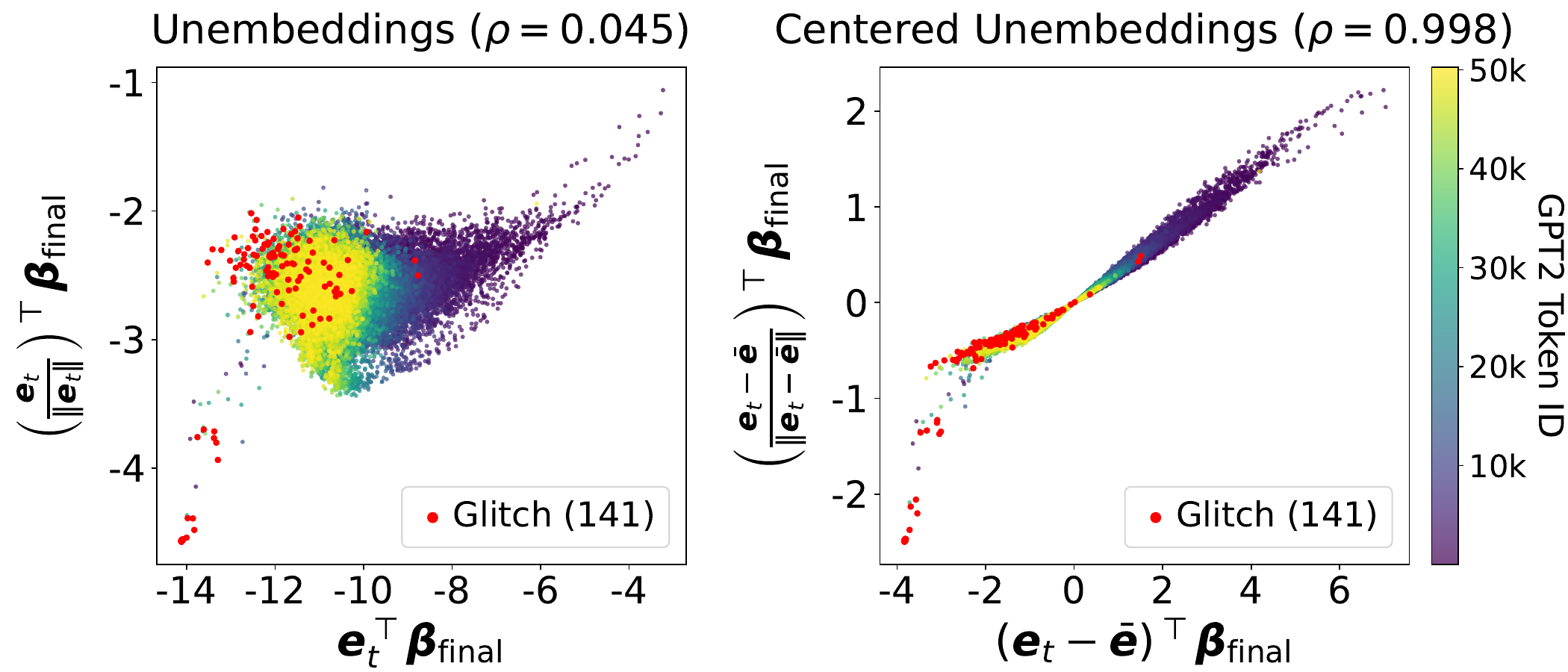}
    \caption{
Scatter plots of the inner product with $\bm{\beta}_{\text{final}}$ before and after normalization, (left) without centering and (right) with centering.
As in Fig.~\ref{fig:norms-vs-b-final-inner-product}, we indicate token IDs as a proxy for frequency, where larger token IDs tend to be less frequent, and we label glitch tokens.
Without centering (left), the inner products become nearly uncorrelated after normalization.
With centering (right), the inner products remain strongly correlated before and after normalization.
}
\label{fig:normed-vs-centered-normed}
\end{figure}

\subsection{Why centering before normalization is necessary}\label{app:centering-reason}
Since the inner product between an unembedding vector and $\bm{\beta}_{\text{final}}$ can be regarded as a token-specific statistic related to generation probabilities, regardless of whether we center\footnote{The inner product between the centered unembedding vector $\bm{e}_t-\bar{\bm{e}}$ and $\bm{\beta}_{\text{final}}$ equals $\bm{e}_t^\top \bm{\beta}_{\text{final}}-\bar{\bm{e}}^\top \bm{\beta}_{\text{final}}$, and $\bar{\bm{e}}^\top \bm{\beta}_{\text{final}}$ is a token-independent constant.
Moreover, when computing token probabilities, the softmax is invariant to adding the same constant shift to all logits, and thus is unaffected by centering the unembedding matrix. See Appendix~\ref{app:wf-bg} for details.}, it is desirable that this inner product remains correlated even after normalizing the unembedding vectors.
Figure~\ref{fig:normed-vs-centered-normed} plots how the inner product with $\bm{\beta}_{\text{final}}$ changes before and after normalization, with and without centering.
With normalization alone, the correlation almost vanishes.
With normalization after centering, the inner products remain strongly correlated.

To analyze the effect of centering, we examine cosine similarities of unembedding vectors.
The mean cosine similarity between $\bm{e}_t$ and the mean vector $\bar{\bm{e}}\in\mathbb{R}^d$, namely $\cos(\bm{e}_t,\bar{\bm{e}})$, is $0.518$.
The mean cosine similarity with $\bm{\beta}_{\text{final}}$, namely $\cos(\bm{e}_t,\bm{\beta}_{\text{final}})$, is $-0.225$.
These results\footnote{For reference, $\cos(\bar{\bm{e}},\bm{\beta}_{\text{final}})=-0.431$.} suggest that the unembedding vectors are anisotropic\footnote{\citet{bis-etal-2021-much} link the anisotropy of the unembedding vectors to training dynamics induced by optimizing the softmax cross-entropy loss.} toward the direction of $\bar{\bm{e}}$, and that this dominant direction is somewhat opposite to $\bm{\beta}_{\text{final}}$.

Therefore, centering can be viewed as an operation analogous to all-but-the-top~\cite{DBLP:conf/iclr/MuV18}, which removes a dominant direction that is somewhat opposite to $\bm{\beta}_{\text{final}}$.
Based on these results\footnote{After centering, the mean cosine similarities of $\bm{e}_t-\bar{\bm{e}}$ with $\bar{\bm{e}}$ and $\bm{\beta}_{\text{final}}$ are $-0.005$ and $0.001$, respectively.} and the above discussion, when interpreting top tokens of attention heads, we use unembedding vectors that are normalized after centering.

\section{Details of head scores and head class annotations in GPT2-small}\label{app:head-scores}
In this section, we describe how we compute head scores using TransformerLens\footnote{v2.16.1;
  \url{https://github.com/TransformerLensOrg/TransformerLens/blob/v2.16.1/demos/Exploratory_Analysis_Demo.ipynb}} following \citet{wang2023interpretability}, and provide details of our head class annotations.

\begin{table*}[t!]
\centering

\begin{subtable}{.245\textwidth}
  \centering
 \begin{tabular}{lr}
  \toprule
  Head & Score \\
  \midrule
L0H3 & 0.966 \\
L0H1 & 0.918 \\
L1H11 & 0.900 \\
L0H4 & 0.826 \\
L0H5 & 0.639 \\
L4H7 & 0.358 \\
L0H10 & 0.235 \\
L3H2 & 0.197 \\
L3H6 & 0.192 \\
L3H0 & 0.163 \\
\bottomrule
  \end{tabular}
\caption{Identity}
\label{tab:identity}
\end{subtable}
\begin{subtable}{.245\textwidth}
  \centering
 \begin{tabular}{lr}
  \toprule
  Head & Score \\
  \midrule
L0H5 & 0.703 \\
L3H0 & 0.660 \\
L0H1 & 0.133 \\
L7H1 & 0.045 \\
L5H10 & 0.042 \\
L7H2 & 0.039 \\
L1H11 & 0.030 \\
L8H1 & 0.026 \\
L7H7 & 0.025 \\
L9H0 & 0.023 \\
\bottomrule
  \end{tabular}
\caption{Duplicate Token}
\label{tab:duplicate}
\end{subtable}
\begin{subtable}{.245\textwidth}
  \centering
 \begin{tabular}{lr}
  \toprule
  Head & Score \\
  \midrule
L4H11 & 0.988 \\
L3H7 & 0.533 \\
L6H8 & 0.490 \\
L2H2 & 0.485 \\
L5H6 & 0.422 \\
L3H2 & 0.408 \\
L2H9 & 0.362 \\
L3H8 & 0.344 \\
L2H5 & 0.339 \\
L3H6 & 0.330 \\
\bottomrule
  \end{tabular}
\caption{Previous Token}
\label{tab:previous}
\end{subtable}
\begin{subtable}{.245\textwidth}
  \centering
 \begin{tabular}{lr}
  \toprule
  Head & Score \\
  \midrule
L5H1 & 0.936 \\
L5H5 & 0.927 \\
L6H9 & 0.925 \\
L7H10 & 0.914 \\
L7H2 & 0.849 \\
L10H1 & 0.474 \\
L9H9 & 0.458 \\
L9H6 & 0.437 \\
L5H0 & 0.425 \\
L10H7 & 0.367 \\
\bottomrule
  \end{tabular}
\caption{Induction}
\label{tab:induction}
\end{subtable}

  \caption{
For Identity, Duplicate Token, Previous Token, and Induction Heads in GPT2-small, we list the top 10 heads and their head scores.
See Appendix~\ref{app:head-scores-details} for how we compute these scores.
  }
\label{tab:table2_4columns}
\end{table*}

\begin{table*}[t!]
\small
\centering
\begin{tabular}{lrl}
\toprule
Head Classes & Count & Heads \\
\midrule
Duplicate Token Heads & 3 & L0H1, L0H5$^*$, L3H0\\
Previous Token Heads & 10 & L2H2, L2H5$^*$, L2H9, L3H2$^*$, L3H6$^*$, L3H7$^*$, L3H8$^*$, L4H11, L5H6$^*$, L6H8$^*$ \\
Induction Heads & 6 & L5H0$^*$, L5H1$^*$, L5H5, L6H9, L7H2$^*$, L7H10$^*$ \\
Name Mover Heads & 3 & L9H6, L9H9, L10H0\\
Negative Name Mover Heads & 2 & L10H7, L11H10 \\
Backup Name Mover Heads & 8 & L9H7, L10H1, L10H2, L10H6, L10H10, L11H2, L11H3, L11H9 \\
S-Inhibition Heads & 4 & L7H3, L7H9, L8H6, L8H10 \\
\bottomrule
\end{tabular}
\caption{
Head classes and their corresponding heads in our experimental setup for GPT2-small.
Heads that are not included in \citet[Fig.~17]{wang2023interpretability} are marked with ``$*$''.
The total number of annotated heads is 36.
}
\label{tab:head-class}
\end{table*}

\subsection{Computing head scores}\label{app:head-scores-details}
We first describe how we construct the dataset.
Specifically, we sample 100 tokens uniformly at random from token IDs 0 to 20{,}000 to form a sentence\footnote{TransformerLens~\cite{nanda2022transformerlens} samples token IDs from 100 to 20{,}000.
Since GPT2 has token IDs from 0 to 50{,}256 and higher IDs tend to be rarer~\cite{DBLP:conf/nips/HayaseL0OS24}, we also restrict the range to 20{,}000.
In addition, since token IDs from 0 to 100 include alphabetic characters and digits, we include 0 to 20{,}000 for simplicity.}, and repeat this sentence to obtain a length-200 token sequence.
We create 128 such sequences, compute head scores for each sequence, and average them to obtain the final head scores.

For Identity Heads, we define the head score as the mean of the diagonal entries of the attention pattern.
For example, in L0H3, which attains the largest Identity Head score and is shown in the right panel of Fig.~\ref{fig:identity}, attention weights concentrate on the diagonal.
For the Identity Head scores shown in the left panel of Fig.~\ref{fig:identity}, Table~\ref{tab:identity} lists the top 10 Identity Heads.
Except for L4H7, which we discussed in Section~\ref{sec:disc}, these heads mainly appear in early layers.

\begin{figure*}[p!]
\centering
\begin{subfigure}{0.9\linewidth}
    \centering
    \includegraphics[width=\textwidth]{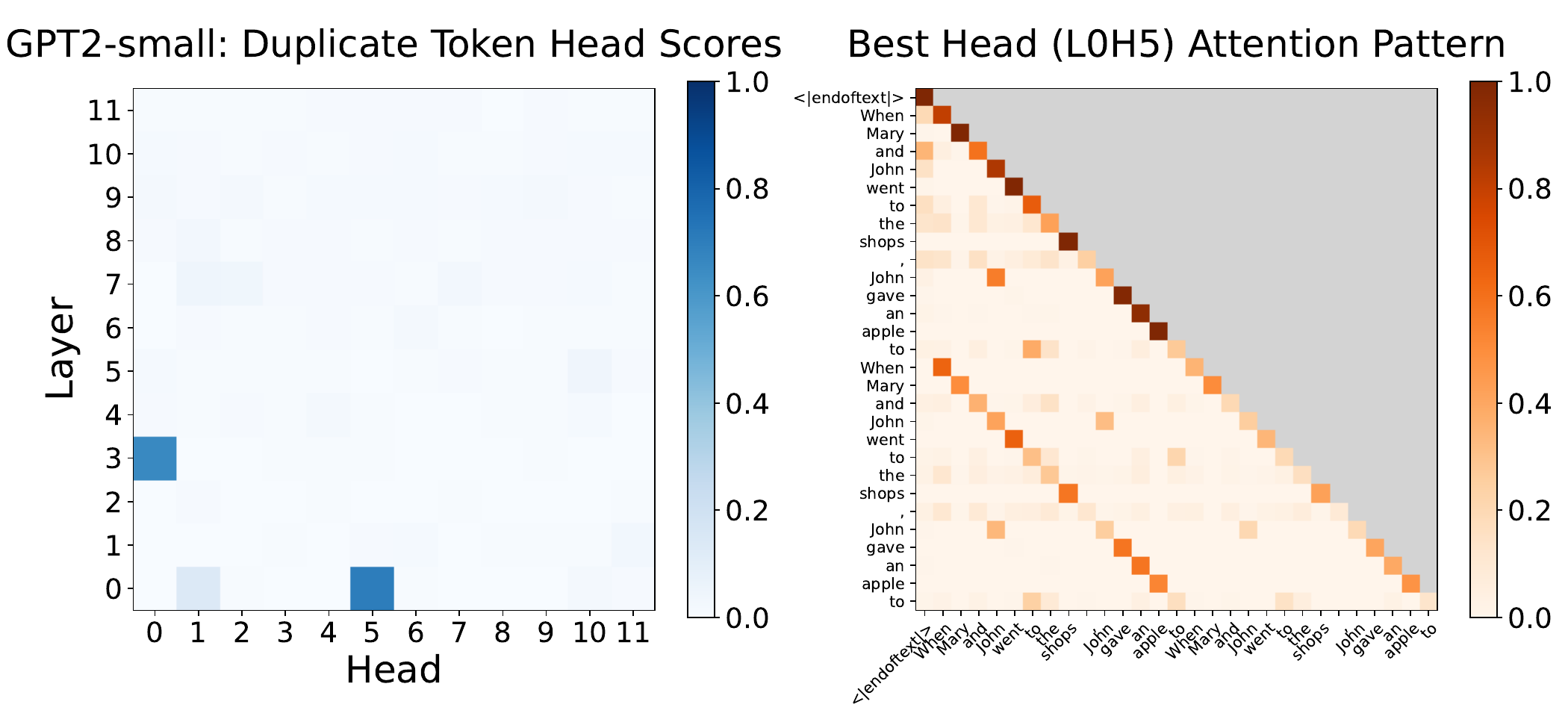}
    \subcaption{Duplicate Token Heads}
    \label{fig:duplicate_token_scores}
\end{subfigure}
\par\bigskip
\begin{subfigure}{0.9\linewidth}
    \centering
    \includegraphics[width=\textwidth]{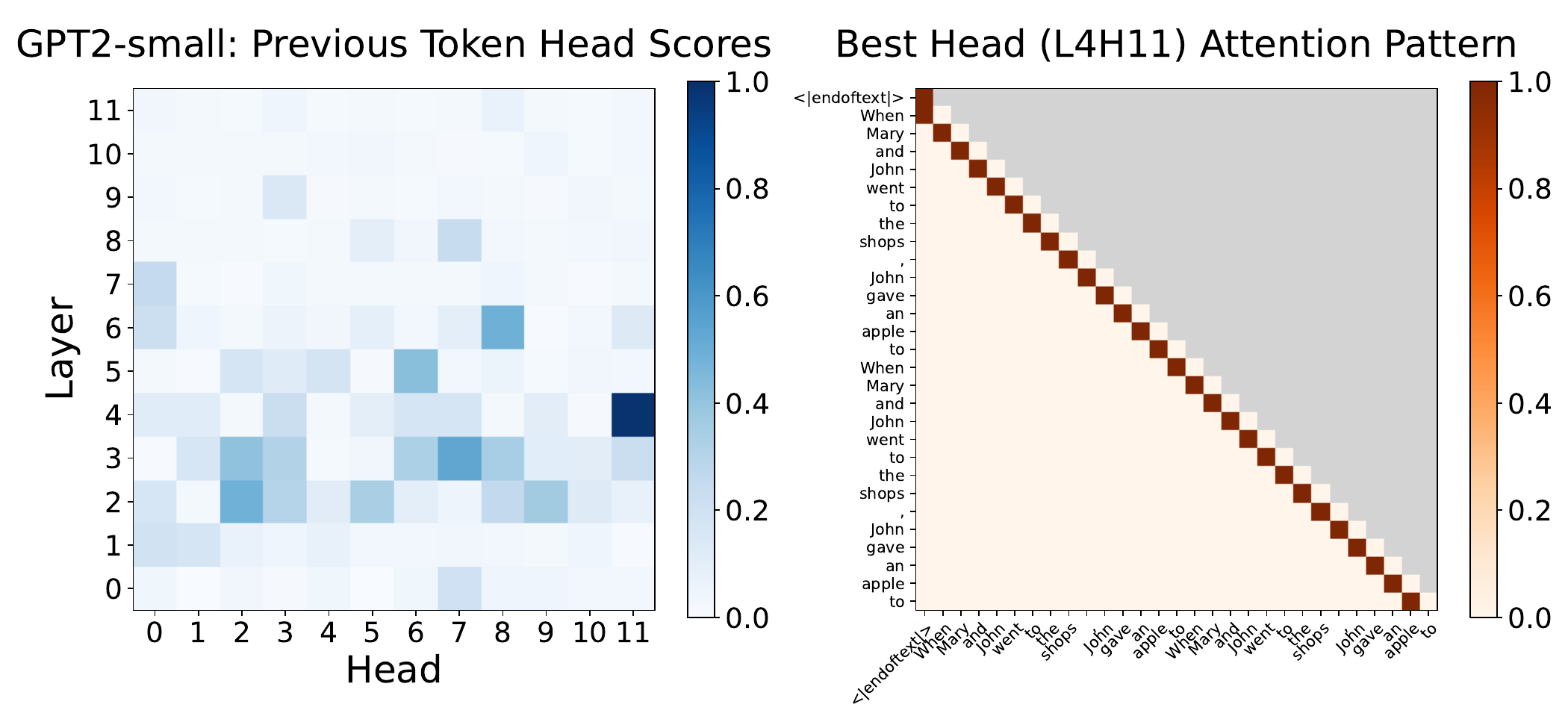}
    \subcaption{Previous Token Heads}
    \label{fig:previous_token_scores}
\end{subfigure}
\par\bigskip
\begin{subfigure}{0.9\linewidth}
    \centering
    \includegraphics[width=\textwidth]{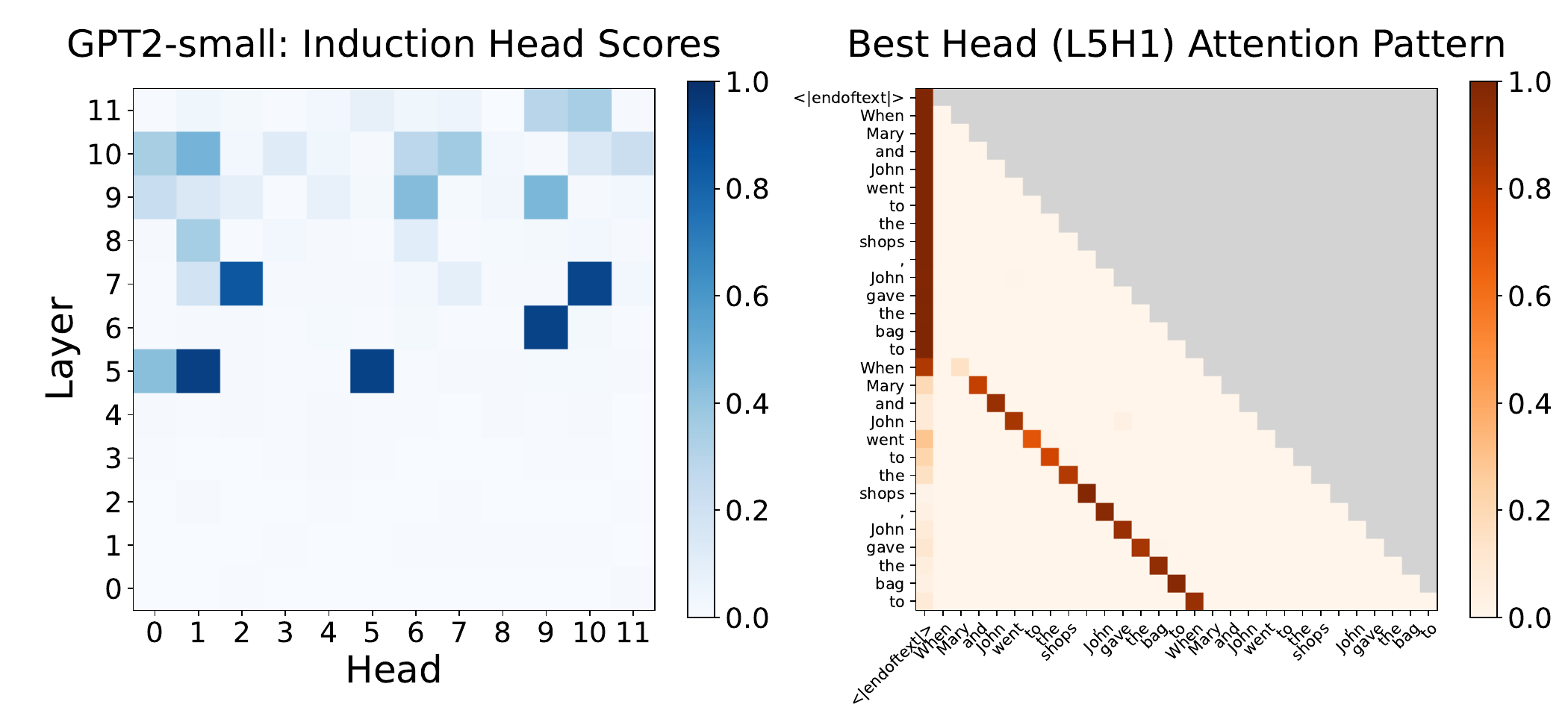}
    \subcaption{Induction Heads}
    \label{fig:induction_head_scores}
\end{subfigure}
\caption{
For Duplicate Token, Previous Token, and Induction Heads, we show (left) head scores for all heads in GPT2-small and (right) an example attention pattern for the head with the largest score.
The head-score patterns broadly reproduce the results of \citet[Fig.~12]{wang2023interpretability}.
For each head class, the attention pattern of the top-scoring head captures the characteristic behavior of that class.
}
\label{fig:head_scores}
\end{figure*}

To revisit the head class annotations, we recompute head scores for Duplicate Token, Previous Token, and Induction Heads following \citet[Appendix F]{wang2023interpretability}, and report the results in Fig.~\ref{fig:head_scores}.
Duplicate Token Heads mainly appear in early layers, whereas Previous Token Heads mainly appear in middle layers, and Induction Heads span middle to late layers.
Moreover, for each head class, the attention pattern of the top-scoring head captures the characteristic behavior of that class, which supports the validity of these scores.
Tables~\ref{tab:duplicate},~\ref{tab:previous}, and~\ref{tab:induction} list the top 10 heads and their head scores for each class.
For Duplicate Token Heads, the top three heads have much larger scores than the rest, whereas for Previous Token and Induction Heads, the scores decrease more gradually.
Based on these results, we next describe our head class annotations.

\subsection{Head class annotations}\label{app:class-annotation}

We adopt the IOI task head class annotations from \citet[Fig.~17]{wang2023interpretability}.
Based on Appendix~\ref{app:head-scores-details}, we revisit the annotations for Duplicate Token, Previous Token, and Induction Heads.

For Duplicate Token Heads, \citet{wang2023interpretability} annotate L0H1, L0H10, and L3H0.
Based on Table~\ref{tab:duplicate}, we replace L0H10, which does not appear in the table, with L0H5, which attains the largest score\footnote{In fact, \citet[Fig.~12]{wang2023interpretability} show the head scores for Duplicate Token Heads, where L0H5 has a large score and L0H10 has a small score.
}.

As shown in Tables~\ref{tab:previous} and~\ref{tab:induction}, there are many more Previous Token Heads and Induction Heads than Duplicate Token Heads.
We therefore assign the Previous Token Head or Induction Head label to unannotated heads among the top 10 heads for each of these two classes.

Table~\ref{tab:head-class} lists the heads in each head class under our setup.
All Previous Token Heads annotated by \citet{wang2023interpretability}, namely L2H2, L2H9, and L4H11, are included in Table~\ref{tab:head-class}.
In contrast, among the Induction Heads annotated by \citet{wang2023interpretability}, namely L5H5, L5H8, L5H9, and L6H9, only L5H5 and L6H9 are included in Table~\ref{tab:head-class}.

\section{Details of baseline methods}\label{app:eval-baselines}
In this section, we define and describe Simple-CS, Linear CKA~\cite{DBLP:conf/icml/Kornblith0LH19}, and Procrustes Similarity.

\subsection{Simple-CS}\label{app:simple-cs}
We explain Simple-CS using the example of the weights associated with an output and a query.
In $\operatorname{CS}_{\,\text{OQ}}$ in (\ref{eq:OQ-CS}), we compute $\operatorname{CS}({\bm{W}_\text{QK}^{h'}}^\top,\bm{W}_\text{OV}^{h})$ using $\bm{W}_\text{OV}^{h},{\bm{W}_\text{QK}^{h'}}^\top\in\mathbb{R}^{d\times d}$.
Thus, even when measuring similarity between weights associated with an output and a query, CS uses the OV matrix $\bm{W}_\text{OV}^{h}$ as the output side and the QK matrix ${\bm{W}_\text{QK}^{h'}}^\top$ as the query side.
As a result, $\bm{W}_\text{OV}^{h}$ in (\ref{eq:OV-matrix}) includes the value weight matrix $\bm{W}_\text{V}^{h}$, and $\bm{W}_\text{QK}^{h'}$ in (\ref{eq:QK-matrix}) includes the key weight matrix $\bm{W}_\text{K}^{h'}$.

As a baseline, we therefore consider directly comparing the output weight $\bm{W}_\text{O}^{h}$ and the query weight ${\bm{W}_\text{Q}^{h'}}^\top$ by computing $\operatorname{CS}({\bm{W}_\text{Q}^{h'}}^\top,\bm{W}_\text{O}^{h})$, and refer to this method as Simple-CS.
That is, for $\bm{W}^r,\bm{W}^{r'}\in\mathcal{W}$, we define
\begin{equation}
\operatorname{Simple\text{-}CS}(r,r') = \operatorname{CS}(\bm{W}^{r'}, \bm{W}^{r}).
\end{equation}

\subsection{Linear CKA}\label{app:linear-cka}
We define Linear CKA following \citet{DBLP:conf/icml/Kornblith0LH19}.
For $\bm{W}^r=(\bm{w}_1^r,\ldots,\bm{w}_{d_\text{head}}^r)\in\mathcal{W}\subset\mathbb{R}^{d\times d_\text{head}}$, we define the columnwise mean vector as
\begin{equation}
  \bm{w}_\text{Mean}^r
  = \frac{1}{d_\text{head}} \sum_{i=1}^{d_\text{head}} \bm{w}_i^r\in\mathbb{R}^{d}.
\end{equation}
We then define the centered matrix $\bm{W}_c^r$ as\footnote{Using the centering matrix $\bm{C}_{d_\text{head}}$ in (\ref{eq:C_d}), we have $\bm{W}_c^r = \bm{W}^r\bm{C}_{d_\text{head}}$.}
\begin{equation}
  \bm{W}_c^r
  = \bm{W}^r - \bm{w}_\text{Mean}^r {\bm{1}_{d_\text{head}}}^\top\in\mathbb{R}^{d\times d_\text{head}}.
\end{equation}
Similarly, for $\bm{W}^{r'}\in\mathcal{W}$ we define $\bm{W}_c^{r'}$.
We define Linear CKA for $\bm{W}_c^r$ and $\bm{W}_c^{r'}$ as
\begin{align}
    \operatorname{Linear\text{-}CKA}(r, r') =\frac{\left\|\bm{W}_c^{r}{\bm{W}_c^{r'}}^\top\right\|_{\text{F}}^2}{\left\|\bm{W}_c^{r}{\bm{W}_c^{r}}^\top\right\|_{\text{F}}\left\|\bm{W}_c^{r'}{\bm{W}_c^{r'}}^\top\right\|_{\text{F}}}.
\end{align}

To verify that $0 \le \operatorname{Linear\text{-}CKA}(r,r') \le 1$, we note the following inequality\footnote{
By the definition of the Frobenius norm and the cyclic invariance of the trace, we have
\begin{align*}
\|{\bm{W}_c^{r}}^\top \bm{W}_c^{r}\|_{\text{F}}^2
=
\operatorname{tr}\left(
{\bm{W}_c^{r}}^\top \bm{W}_c^{r}
{\bm{W}_c^{r}}^\top \bm{W}_c^{r}
\right) 
=
\operatorname{tr}\left(
\bm{W}_c^{r}{\bm{W}_c^{r}}^\top
\bm{W}_c^{r}{\bm{W}_c^{r}}^\top
\right) 
=
\|\bm{W}_c^{r}{\bm{W}_c^{r}}^\top\|_{\text{F}}^2.
\end{align*}
Therefore, $\|{\bm{W}_c^{r}}^\top \bm{W}_c^{r}\|_{\text{F}}
=
\|\bm{W}_c^{r}{\bm{W}_c^{r}}^\top\|_{\text{F}}$.
Similarly, $\|{\bm{W}_c^{r'}}^\top \bm{W}_c^{r'}\|_{\text{F}}
=
\|\bm{W}_c^{r'}{\bm{W}_c^{r'}}^\top\|_{\text{F}}$.
}
\begin{align}
    &\left\|\bm{W}_c^{r}{\bm{W}_c^{r'}}^\top\right\|_{\text{F}}^2
    =\operatorname{tr}\left(\left(\bm{W}_c^{r}{\bm{W}_c^{r'}}^\top\right)^\top\bm{W}_c^{r}{\bm{W}_c^{r'}}^\top\right)
    = \operatorname{tr}\left(\bm{W}_c^{r'}{\bm{W}_c^{r}}^\top{\bm{W}_c^{r}}{\bm{W}_c^{r'}}^\top\right)\notag \\
    =\,& \operatorname{tr}\left({\bm{W}_c^{r}}^\top\bm{W}_c^{r}{\bm{W}_c^{r'}}^\top{\bm{W}_c^{r'}}\right) 
    =\operatorname{tr}\left({\left({\bm{W}_c^{r}}^\top\bm{W}_c^{r}\right)}^\top{\bm{W}_c^{r'}}^\top{\bm{W}_c^{r'}}\right)
    =\langle {\bm{W}_c^{r}}^\top\bm{W}_c^{r}, {\bm{W}_c^{r'}}^\top{\bm{W}_c^{r'}} \rangle_\text{F} \notag \\
    \leq\,& \|{\bm{W}_c^{r}}^\top\bm{W}_c^{r}\|_\text{F} \|{\bm{W}_c^{r'}}^\top{\bm{W}_c^{r'}}\|_\text{F}
    =\|\bm{W}_c^{r}{\bm{W}_c^{r}}^\top\|_\text{F} \|{\bm{W}_c^{r'}}{\bm{W}_c^{r'}}^\top\|_\text{F},
\end{align}
where the inequality follows from the Cauchy--Schwarz inequality for the Frobenius inner product, and equality holds if and only if
${\bm{W}_c^{r}}^\top\bm{W}_c^{r} = \lambda\,{\bm{W}_c^{r'}}^\top\bm{W}_c^{r'}$ for some $\lambda \ge 0$.
Therefore, when $\|\bm{W}_c^{r}{\bm{W}_c^{r}}^\top\|_\text{F} \|{\bm{W}_c^{r'}}{\bm{W}_c^{r'}}^\top\|_\text{F}\neq 0$, we have
\begin{align}
    0 \leq \operatorname{Linear\text{-}CKA}(r, r') \leq 1.
\end{align}

\begin{table*}[t!]
\centering
\scriptsize
\begin{tabular}{llcrrrrrrl}
\toprule
& & & \multicolumn{3}{c}{Heads} & \multicolumn{3}{c}{Edges} & \\
\cmidrule(lr){4-6} \cmidrule(lr){7-9}
& Method & WP & \#Total & \#Annot. & \#Unannot. & \#Total & \#Both Annot. & \#Either Annot. & Figure \\
\midrule
\multirow{2}{*}{Top10} & PK &  & 30 & 15 (50.0\%) & 15 (50.0\%) & 30 & 10 (33.3\%) & 22 (73.3\%) & Fig.~\ref{fig:wd-pk-top10} \\
 & CS &  & 30 & 9 (30.0\%) & 21 (70.0\%) & 30 & 3 (10.0\%) & 9 (30.0\%) & Fig.~\ref{fig:wd-cs-top10} \\
\midrule
\multirow{7}{*}{Top20} & \multirow{2}{*}{PK} &  & 52 & 22 (42.3\%) & 30 (57.7\%) & 60 & 18 (30.0\%) & 42 (70.0\%) & Fig.~\ref{fig:wd-pk-top20} \\
 &  & \checkmark & 51 & 21 (41.2\%) & 30 (58.8\%) & 60 & 17 (28.3\%) & 41 (68.3\%) & Fig.~\ref{fig:wd-pk_wp-top20} \\
 & \multirow{2}{*}{CS} &  & 46 & 14 (30.4\%) & 32 (69.6\%) & 60 & 3 (5.0\%) & 20 (33.3\%) & Fig.~\ref{fig:wd-cs-top20} \\
 &  & \checkmark & 34 & 6 (17.6\%) & 28 (82.4\%) & 60 & 0 (0.0\%) & 9 (15.0\%) & Fig.~\ref{fig:wd-cs_wp-top20} \\
 & Simple-CS &  & 51 & 16 (31.4\%) & 35 (68.6\%) & 60 & 5 (8.3\%) & 26 (43.3\%) & Fig.~\ref{fig:wd-simple-cs-top20} \\
 & Linear CKA &  & 24 & 12 (50.0\%) & 12 (50.0\%) & 60 & 22 (36.7\%) & 51 (85.0\%) & Fig.~\ref{fig:wd-linearcka-top20} \\
 & Procrustes Similarity &  & 31 & 12 (38.7\%) & 19 (61.3\%) & 60 & 19 (31.7\%) & 47 (78.3\%) & Fig.~\ref{fig:wd-procrustes-top20} \\
\midrule
\multirow{2}{*}{Top30} & PK &  & 58 & 24 (41.4\%) & 34 (58.6\%) & 90 & 22 (24.4\%) & 61 (67.8\%) & Fig.~\ref{fig:wd-pk-top30} \\
 & CS &  & 61 & 20 (32.8\%) & 41 (67.2\%) & 90 & 10 (11.1\%) & 37 (41.1\%) & Fig.~\ref{fig:wd-cs-top30} \\
\bottomrule
\end{tabular}
\caption{
Number of heads and edges appearing in the wiring diagrams of GPT2-small. 
For each of OQ, OK, and OV, a score is computed for each pair of a head in an earlier layer and a head in a later layer ($9{,}504$ pairs), and the top-$k$ pairs are drawn as edges. 
A head is counted if it is an endpoint of at least one edge. 
Annotated heads are those registered in the head classes, and an edge is counted in the ``Both Annot.'' column when both of its endpoints are annotated and in the ``Either Annot.'' column when at least one of its endpoints is annotated. 
The symbol \checkmark in the ``WP'' column indicates that the method is applied to the preprocessed weights described in Appendix~\ref{app:weight-preproc}.
}
\label{tab:gpt2_small_wiring_diagram_heads}
\end{table*}

\subsection{Procrustes Similarity}\label{app:procrustes-similarity}
For weights $\bm{W}^r,\bm{W}^{r'}\in\mathcal{W}\subset\mathbb{R}^{d\times d_\text{head}}$, the orthogonal Procrustes problem~\cite{schonemann1966generalized} is
\begin{equation}
    \operatorname*{minimize}_{\substack{\bm{\Phi}\in\mathbb{R}^{d\times d},\,\\
     \bm{\Phi}^\top\bm{\Phi}=\bm{\Phi}\bm{\Phi}^\top=\bm{I}_d}}
    \|\bm{\Phi}\bm{W}^r-\bm{W}^{r'}\|_\text{F}.\label{eq:proc}
\end{equation}
This problem admits an analytic solution.
Let $\bm{W}^{r'}{\bm{W}^{r}}^\top=\bm{U}^{r,r'}\bm{\Sigma}^{r,r'}{\bm{V}^{r,r'}}^\top$ be the SVD, where $\bm{U}^{r,r'},{\bm{V}^{r,r'}}\in\mathbb{R}^{d\times d}$ are orthogonal and $\bm{\Sigma}^{r,r'}\in\mathbb{R}^{d\times d}$ is diagonal with nonnegative entries.
Then (\ref{eq:proc}) is minimized by $\bm{\Phi}^\ast=\bm{U}^{r,r'}{\bm{V}^{r,r'}}^\top\in\mathbb{R}^{d\times d}$~\cite{schonemann1966generalized}.
We refer to $\|\bm{\Phi}^\ast\bm{W}^r-\bm{W}^{r'}\|_\text{F}$ as the Procrustes distance~\cite{rohlf1999shape}.

To compare with PK and CS, we convert the Procrustes distance into a normalized similarity.
For $\bm{\Phi}^\ast\bm{W}^r-\bm{W}^{r'}\in\mathbb{R}^{d\times d_\text{head}}$, we have
\begin{align}
\|\bm{\Phi}^\ast\bm{W}^r-\bm{W}^{r'}\|_\text{F}^2
&=\operatorname{tr}\left(\left(\bm{\Phi}^\ast\bm{W}^r-\bm{W}^{r'}\right)^\top\left(\bm{\Phi}^\ast\bm{W}^r-\bm{W}^{r'}\right)\right) \notag \\
&=\operatorname{tr}\left(\left(\bm{\Phi}^\ast\bm{W}^r\right)^\top\bm{\Phi}^\ast\bm{W}^r\right)
+\operatorname{tr}\left({\bm{W}^{r'}}^\top{\bm{W}^{r'}}\right)
- 2 \operatorname{tr}\left({\bm{W}^{r'}}^\top\bm{\Phi}^\ast\bm{W}^r\right)\notag \\
&=\|\bm{\Phi}^\ast\bm{W}^r\|_\text{F}^2 + \|\bm{W}^{r'}\|_\text{F}^2 - 2 \operatorname{tr}\left({\bm{W}^{r'}}^\top\bm{\Phi}^\ast\bm{W}^r\right) \notag \\
&\leq\|\bm{\Phi}^\ast\bm{W}^r\|_\text{F}^2 + \|\bm{W}^{r'}\|_\text{F}^2,
\end{align}
where the inequality uses
\begin{align}
&\operatorname{tr}\left({\bm{W}^{r'}}^\top \bm{\Phi}^\ast\bm{W}^r\right)
=\operatorname{tr}\left({\left({\bm{W}^{r'}}^\top \bm{\Phi}^\ast\bm{W}^r
    \right)}^\top\right)
=\operatorname{tr}\left({\bm{W}^r}^\top{\bm{\Phi}^\ast}^\top {\bm{W}^{r'}}
    \right)
=\operatorname{tr}\left({\bm{W}^{r'}}{\bm{W}^r}^\top{\bm{\Phi}^\ast}^\top \right)\notag\\
=\,&\operatorname{tr}\left(\bm{U}^{r,r'}\bm{\Sigma}^{r,r'}{\bm{V}^{r,r'}}^\top\left(\bm{U}^{r,r'}{\bm{V}^{r,r'}}^\top\right)^\top\right)
=\operatorname{tr}\left(\bm{U}^{r,r'}\bm{\Sigma}^{r,r'}{\bm{V}^{r,r'}}^\top\bm{V}^{r,r'}{\bm{U}^{r,r'}}^\top\right)
=\operatorname{tr}\left(\bm{\Sigma}^{r,r'}\right)\ \ge\ 0,
\end{align}
where we use ${\bm{U}^{r,r'}}^\top\bm{U}^{r,r'}={\bm{V}^{r,r'}}^\top\bm{V}^{r,r'}=\bm{I}_d$ and the cyclic invariance of the trace.
In addition, the inequality is tight if and only if $\operatorname{tr}(\bm{\Sigma}^{r,r'})=0$, equivalently, $\bm{\Sigma}^{r,r'}=\bm{O}$ (that is, $\bm{W}^r{\bm{W}^{r'}}^\top=\bm{O}$).
Therefore, when $\|\bm{\Phi}^\ast\bm{W}^r\|_\text{F}^2 + \|\bm{W}^{r'}\|_\text{F}^2 \neq 0$, we have
\begin{equation}
0 \leq \frac{\|\bm{\Phi}^\ast\bm{W}^r-\bm{W}^{r'}\|_\text{F}^2}{\|\bm{\Phi}^\ast\bm{W}^r\|_\text{F}^2 + \|\bm{W}^{r'}\|_\text{F}^2} \leq 1.
\end{equation}
We then define Procrustes Similarity as
\begin{align}
    \operatorname{Procrustes\text{-}Similarity}(r, r')
    =1 - \frac{\|\bm{\Phi}^\ast\bm{W}^r-\bm{W}^{r'}\|_\text{F}^2}{\|\bm{\Phi}^\ast\bm{W}^r\|_\text{F}^2 + \|\bm{W}^{r'}\|_\text{F}^2}.
\end{align}

\section{Additional experimental details for each method}\label{app:additional-experiments}

\subsection{Additional wiring diagrams}\label{app:additional-wiring-diagrams}
In this section, we show wiring diagrams for PK and CS while varying the number of top edges displayed.
We also show wiring diagrams with the top 20 edges for the baseline methods in Section~\ref{sec:eval-metric}, namely Simple-CS, Linear CKA, and Procrustes Similarity.
Table~\ref{tab:gpt2_small_wiring_diagram_heads} summarizes the numbers of heads and edges appearing in these wiring diagrams\footnote{
Identity Heads are excluded from the head classes used to define annotated heads in Table~\ref{tab:gpt2_small_wiring_diagram_heads}, but are included among the head classes in Fig.~\ref{fig:intro}.
}.

\begin{figure*}[p!]
    \centering
    \includegraphics[width=0.96\linewidth]{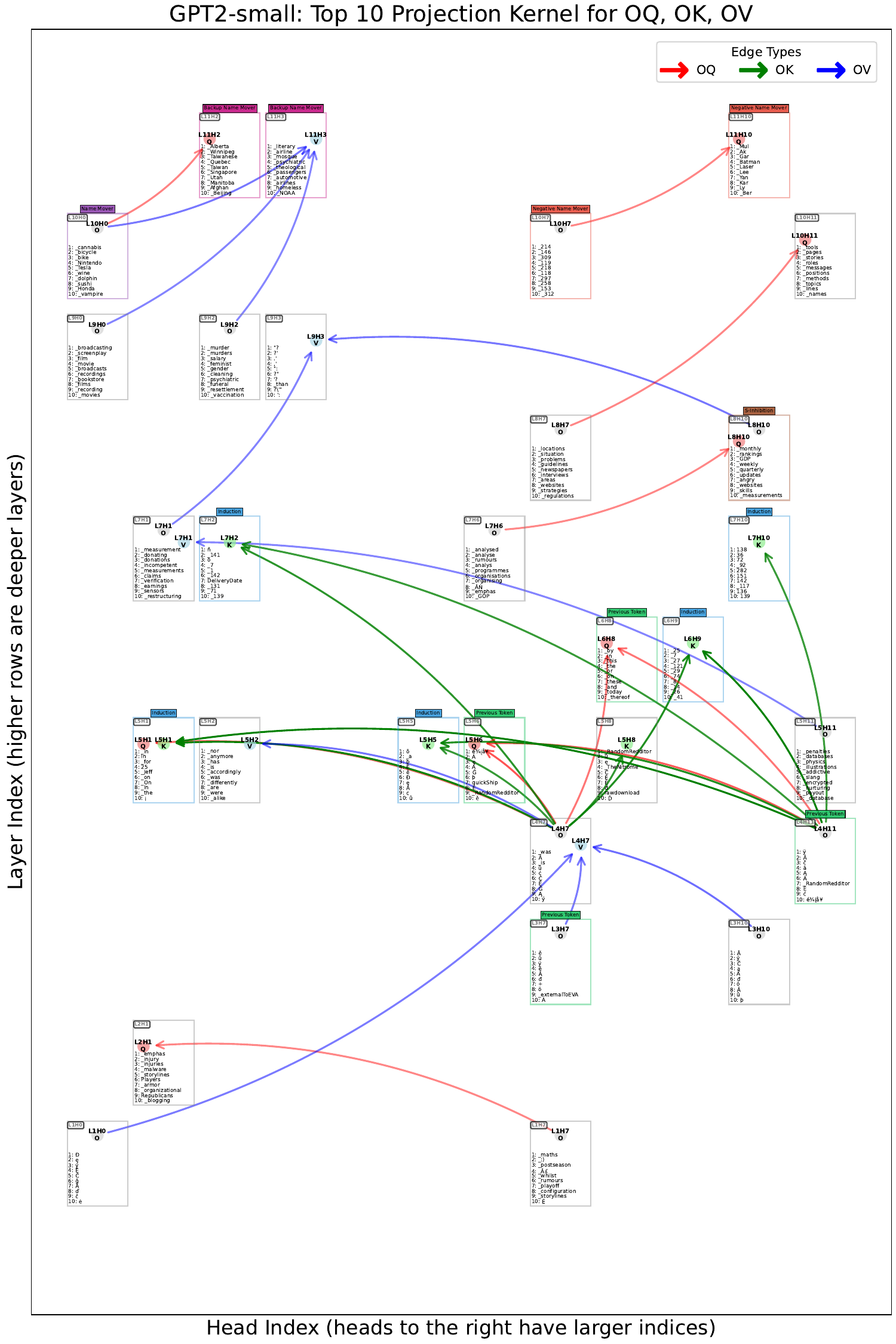}
    \caption{
For GPT2-small, we visualize the top 10 scores of $\operatorname{PK}_{\,\text{OQ}}$, $\operatorname{PK}_{\,\text{OK}}$, and $\operatorname{PK}_{\,\text{OV}}$.
For each head $h$, we show the top 10 tokens that represent $\bm{W}_\text{O}^h$ using the projection method in Section~\ref{sec:unembedding-proj}.
For readability, we replace Ġ in tokens with \_.
}
    \label{fig:wd-pk-top10}
\end{figure*}

\begin{figure*}[p!]
    \centering
    \includegraphics[width=0.96\linewidth]{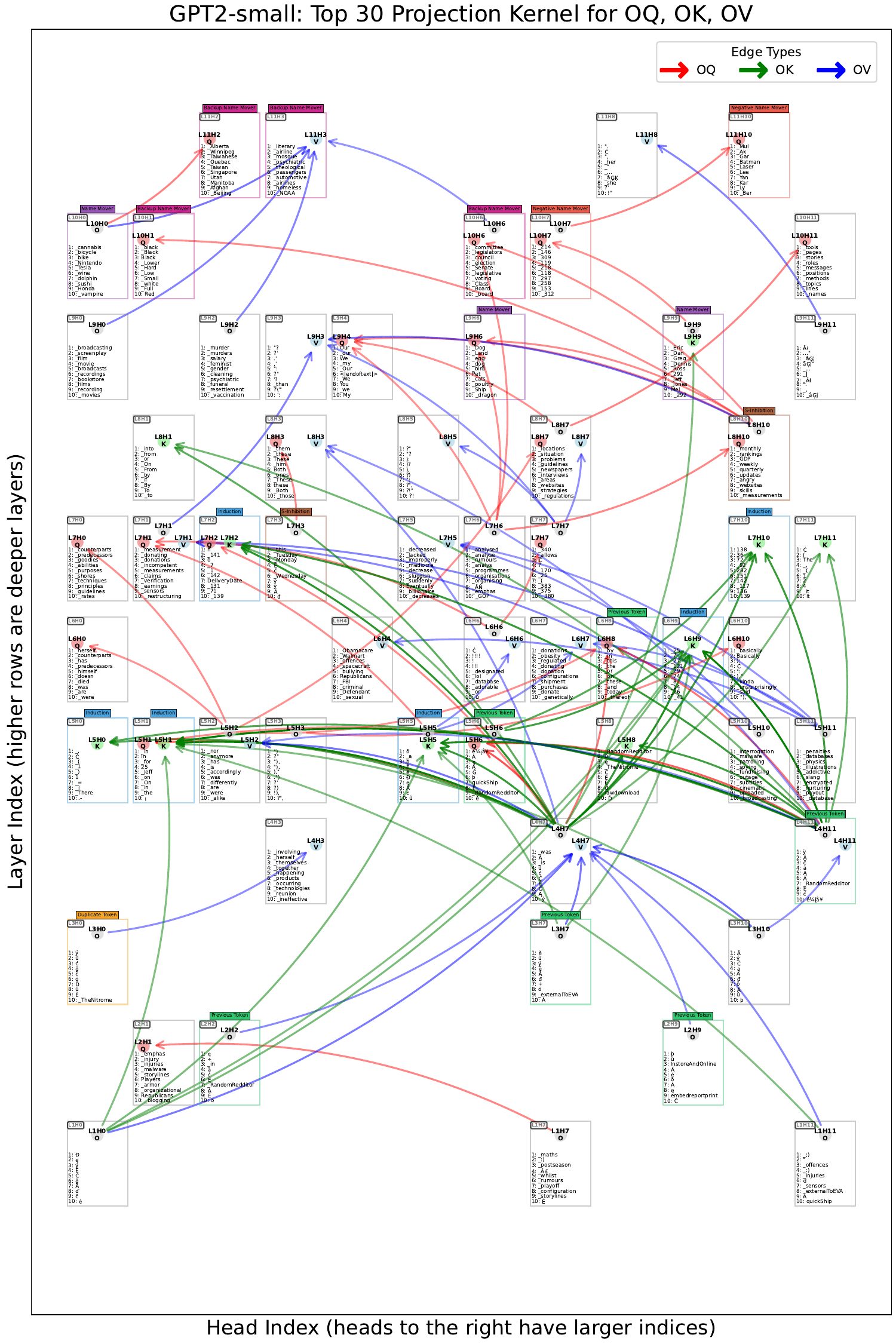}
    \caption{
For GPT2-small, we visualize the top 30 scores of $\operatorname{PK}_{\,\text{OQ}}$, $\operatorname{PK}_{\,\text{OK}}$, and $\operatorname{PK}_{\,\text{OV}}$.
For each head $h$, we show the top 10 tokens that represent $\bm{W}_\text{O}^h$ using the projection method in Section~\ref{sec:unembedding-proj}.
For readability, we replace Ġ in tokens with \_.
}
    \label{fig:wd-pk-top30}
\end{figure*}

Figures~\ref{fig:wd-pk-top10} and~\ref{fig:wd-pk-top30} show the PK wiring diagrams with the top 10 and top 30 edges, respectively.
Although Fig.~\ref{fig:wd-pk-top10} is sparse, the OK PK scores remain large between Previous Token Heads and Induction Heads.
It also shows large OQ PK scores among Name-Mover-related heads and confirms that L4H7 acts as a hub.
In contrast, Fig.~\ref{fig:wd-pk-top30} is denser and contains more information, but is harder to interpret.
Compared with Fig.~\ref{fig:wd-pk-top20}, which shows the top 20 edges, it also includes more heads in the middle layers that do not belong to any head class.
For these reasons, in Section~\ref{sec:pk-vs-cs}, we set the number of displayed edges to 20.

\begin{figure*}[p!]
    \centering
    \includegraphics[width=0.96\linewidth]{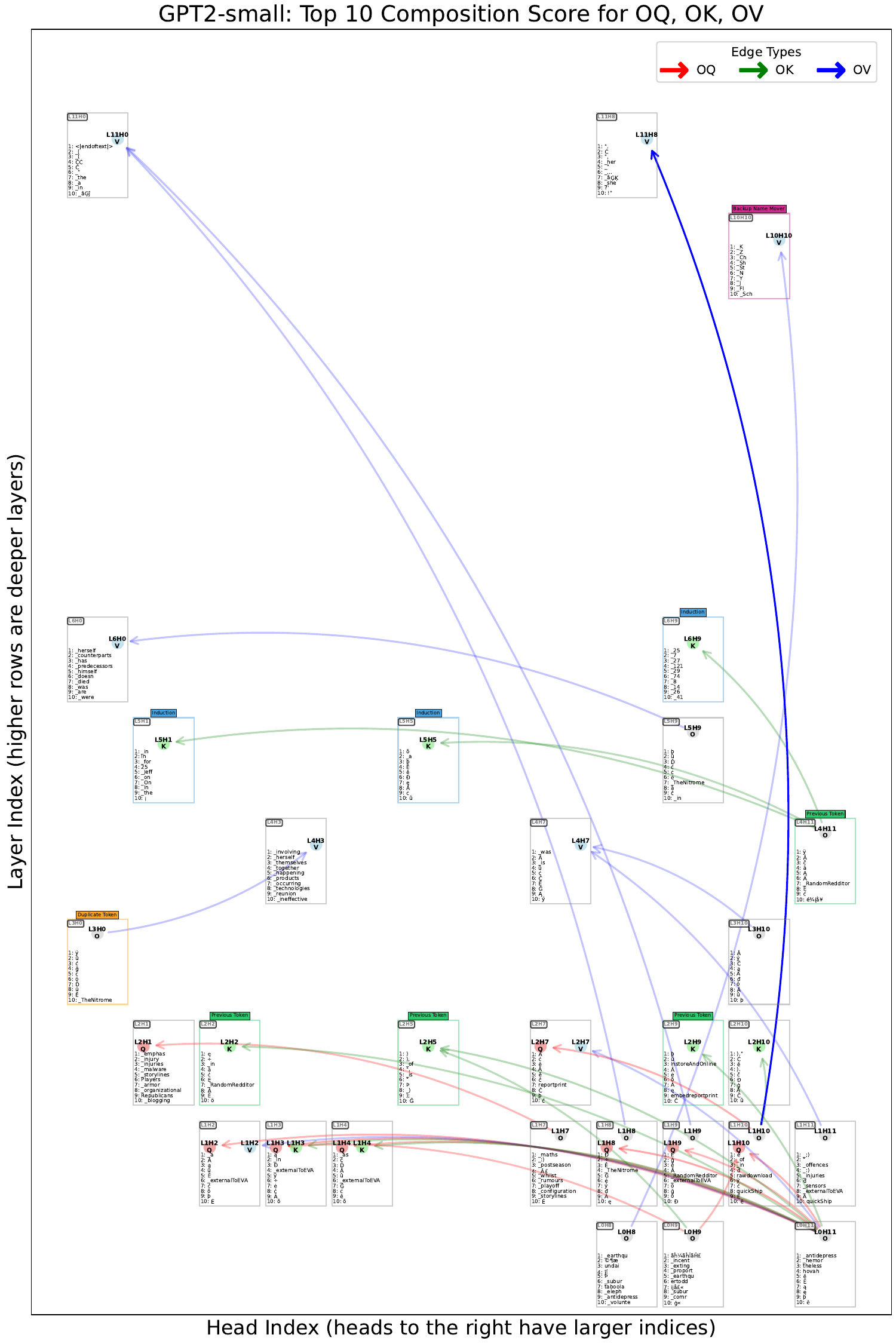}
    \caption{
For GPT2-small, we visualize the top 10 scores of $\operatorname{CS}_{\,\text{OQ}}$, $\operatorname{CS}_{\,\text{OK}}$, and $\operatorname{CS}_{\,\text{OV}}$.
For each head $h$, we show the top 10 tokens that represent $\bm{W}_\text{O}^h$ using the projection method in Section~\ref{sec:unembedding-proj}.
For readability, we replace Ġ in tokens with \_.
}
    \label{fig:wd-cs-top10}
\end{figure*}

\begin{figure*}[p!]
    \centering
    \includegraphics[width=0.96\linewidth]{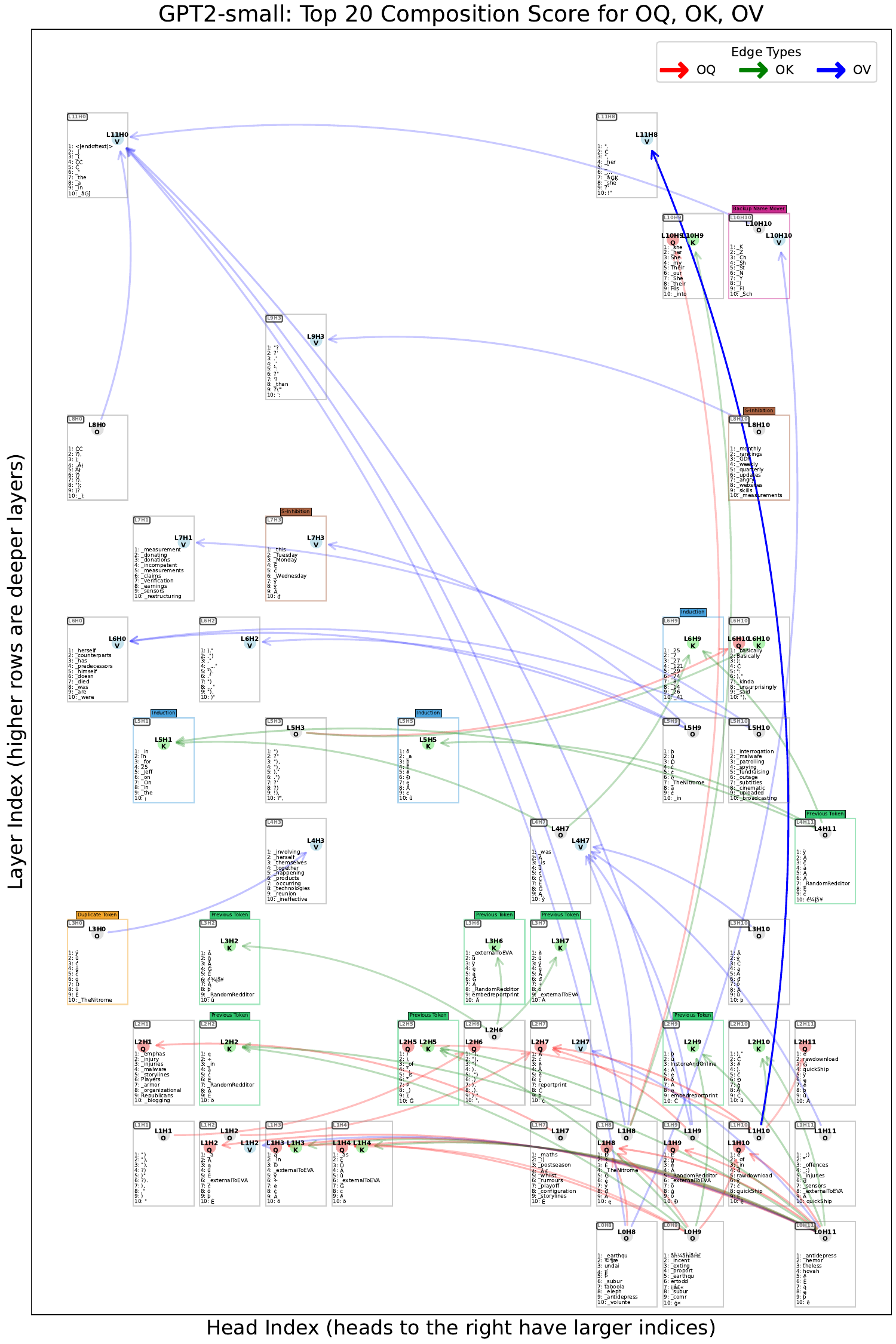}
    \caption{
For GPT2-small, we visualize the top 20 scores of $\operatorname{CS}_{\,\text{OQ}}$, $\operatorname{CS}_{\,\text{OK}}$, and $\operatorname{CS}_{\,\text{OV}}$.
For each head $h$, we show the top 10 tokens that represent $\bm{W}_\text{O}^h$ using the projection method in Section~\ref{sec:unembedding-proj}.
For readability, we replace Ġ in tokens with \_.
}
    \label{fig:wd-cs-top20}
\end{figure*}

\begin{figure*}[p!]
    \centering
    \includegraphics[width=0.96\linewidth]{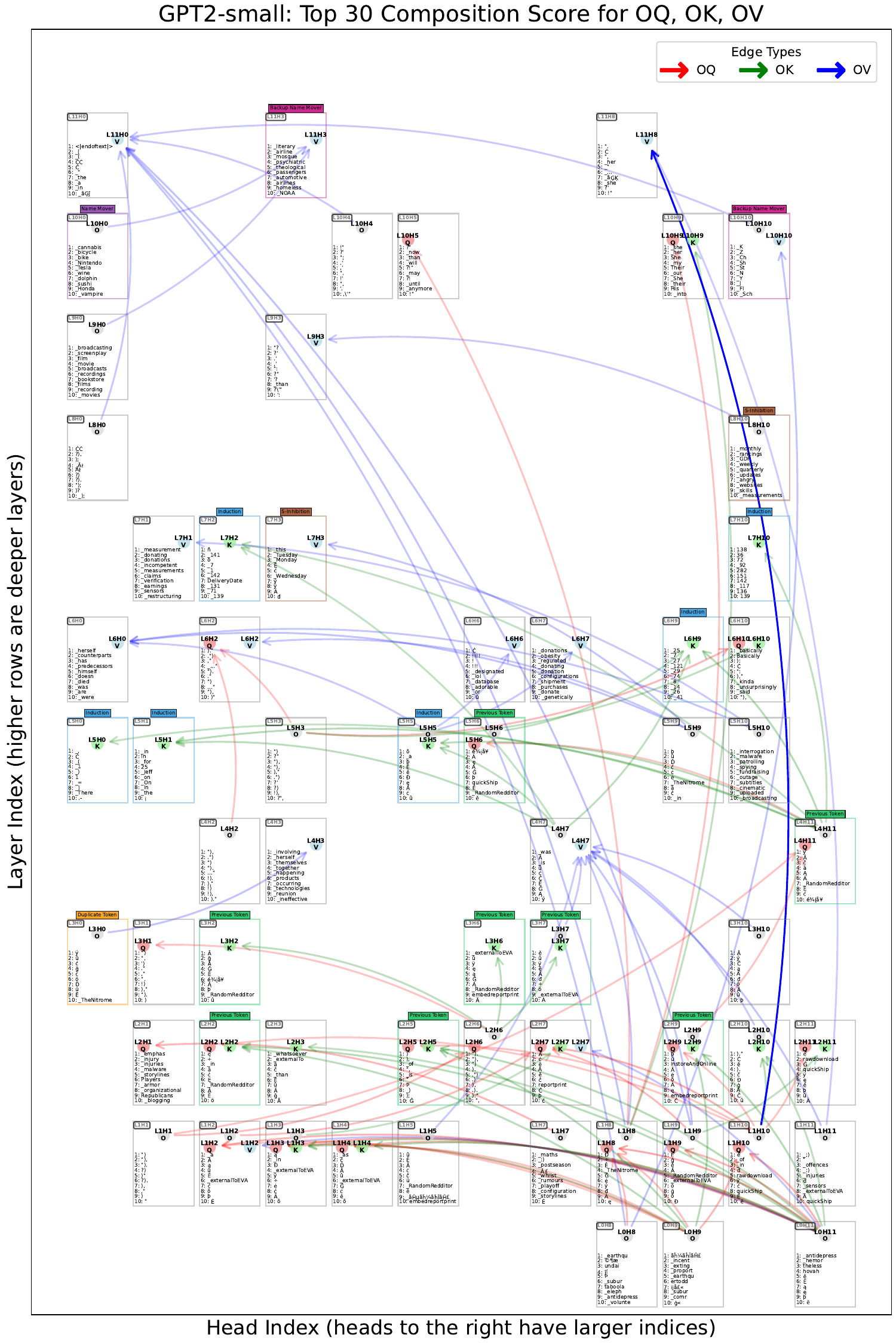}
    \caption{
For GPT2-small, we visualize the top 30 scores of $\operatorname{CS}_{\,\text{OQ}}$, $\operatorname{CS}_{\,\text{OK}}$, and $\operatorname{CS}_{\,\text{OV}}$.
For each head $h$, we show the top 10 tokens that represent $\bm{W}_\text{O}^h$ using the projection method in Section~\ref{sec:unembedding-proj}.
For readability, we replace Ġ in tokens with \_.
}
    \label{fig:wd-cs-top30}
\end{figure*}

Figures~\ref{fig:wd-cs-top10},~\ref{fig:wd-cs-top20}, and~\ref{fig:wd-cs-top30} show the CS wiring diagrams with the top 10, top 20, and top 30 edges, respectively.
Compared with the PK wiring diagrams, more heads in early layers appear, and many edges connect early and late layers.
Even in the top 30 diagram, Name-Mover-related heads appear less frequently than under PK.
Moreover, the OQ edges from the S-Inhibition Head L8H10 to Name-Mover-related heads, which appear under PK, are absent under CS.

\begin{figure*}[p!]
    \centering
    \includegraphics[width=0.96\linewidth]{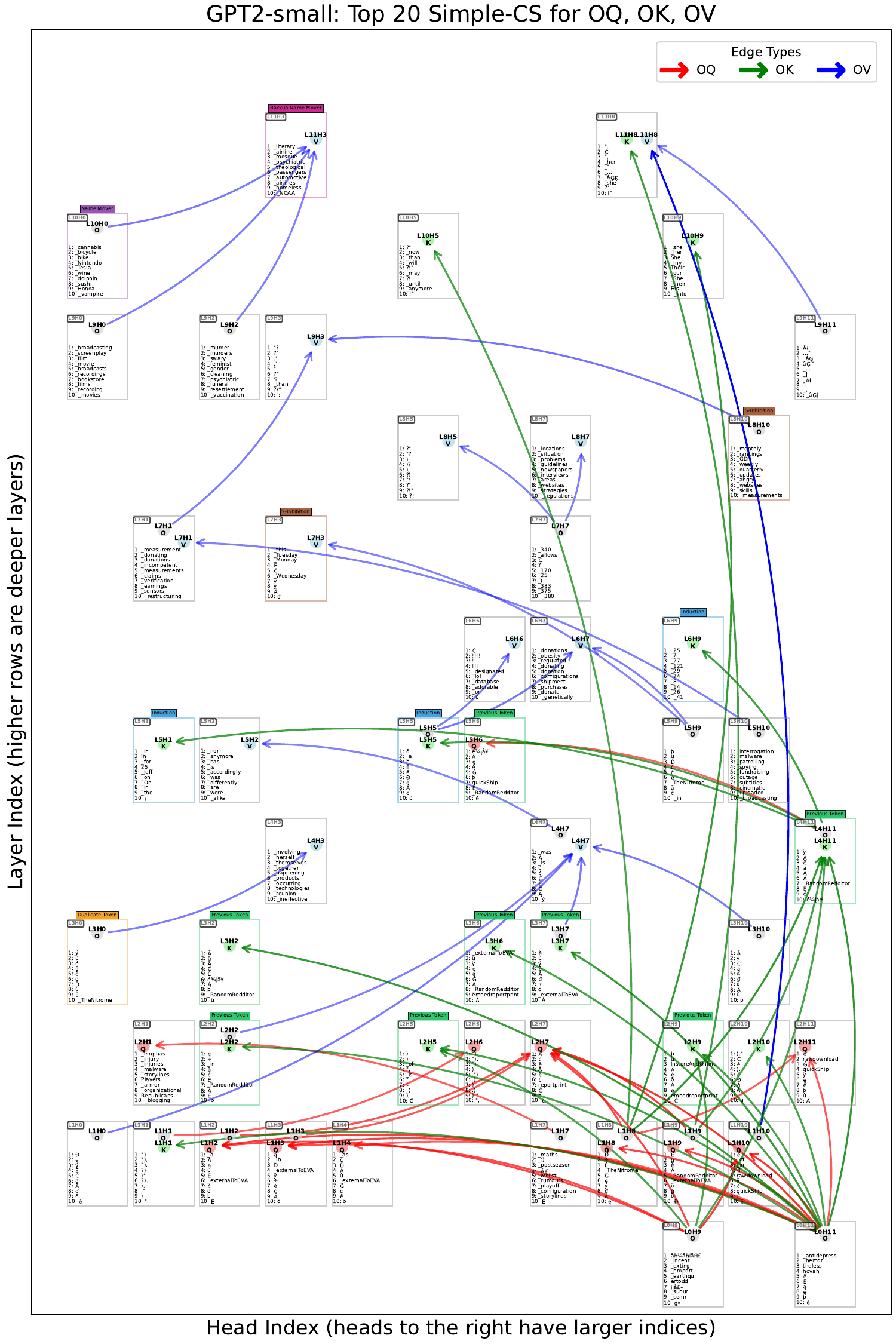}
    \caption{For GPT2-small, we visualize the top 20 scores of Simple-CS for the pairing types OQ, OK, and OV.
For each head $h$, we show the top 10 tokens that represent $\bm{W}_\text{O}^h$ using the projection method in Section~\ref{sec:unembedding-proj}. For readability, we replace Ġ in tokens with \_.}
\label{fig:wd-simple-cs-top20}
\end{figure*}

\begin{figure*}[p!]
    \centering
    \includegraphics[width=0.96\linewidth]{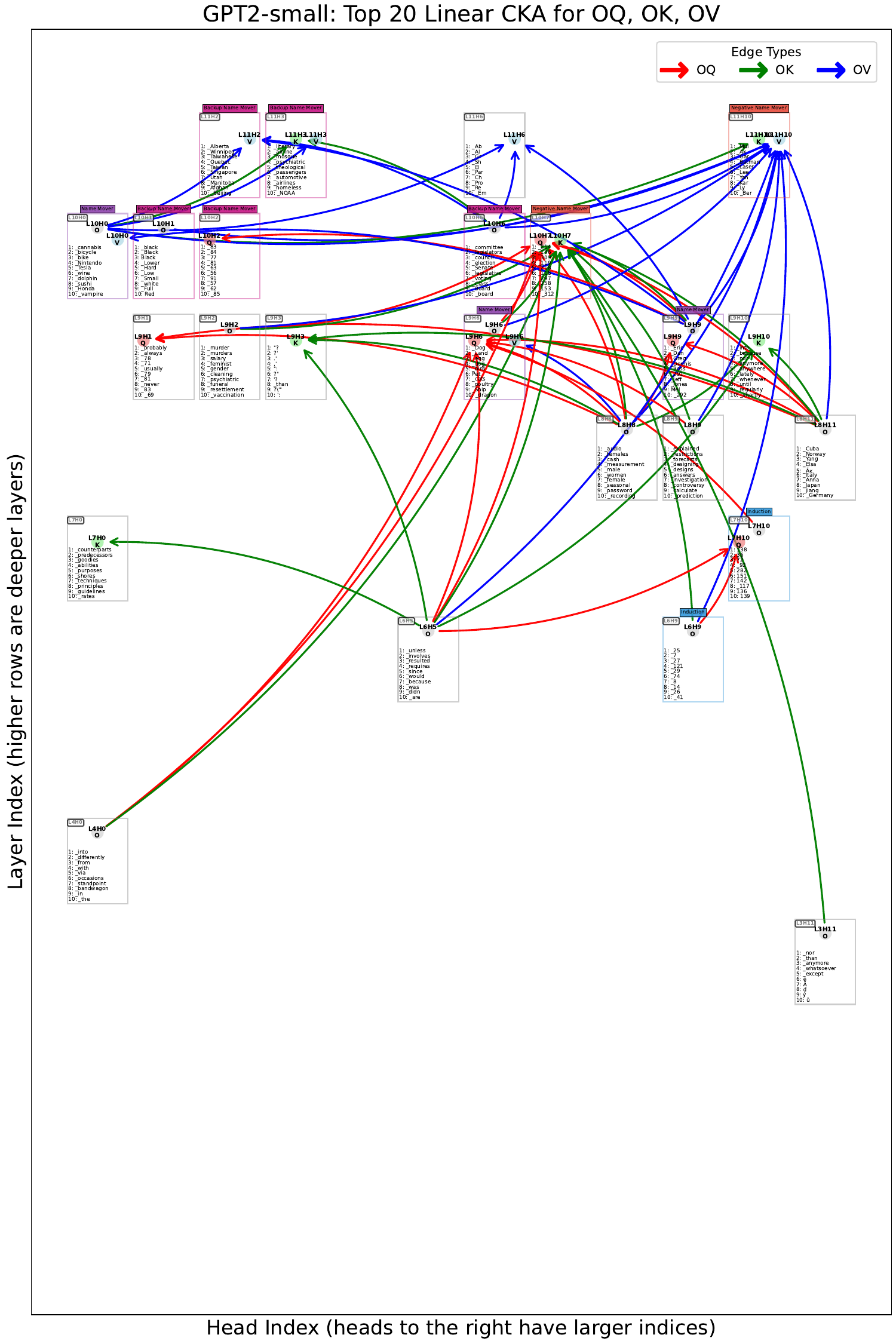}
    \caption{For GPT2-small, we visualize the top 20 scores of Linear CKA for the pairing types OQ, OK, and OV. 
For each head $h$, we show the top 10 tokens that represent $\bm{W}_\text{O}^h$ using the projection method in Section~\ref{sec:unembedding-proj}. For readability, we replace Ġ in tokens with \_.}
\label{fig:wd-linearcka-top20}
\end{figure*}

\begin{figure*}[p!]
    \centering
    \includegraphics[width=0.96\linewidth]{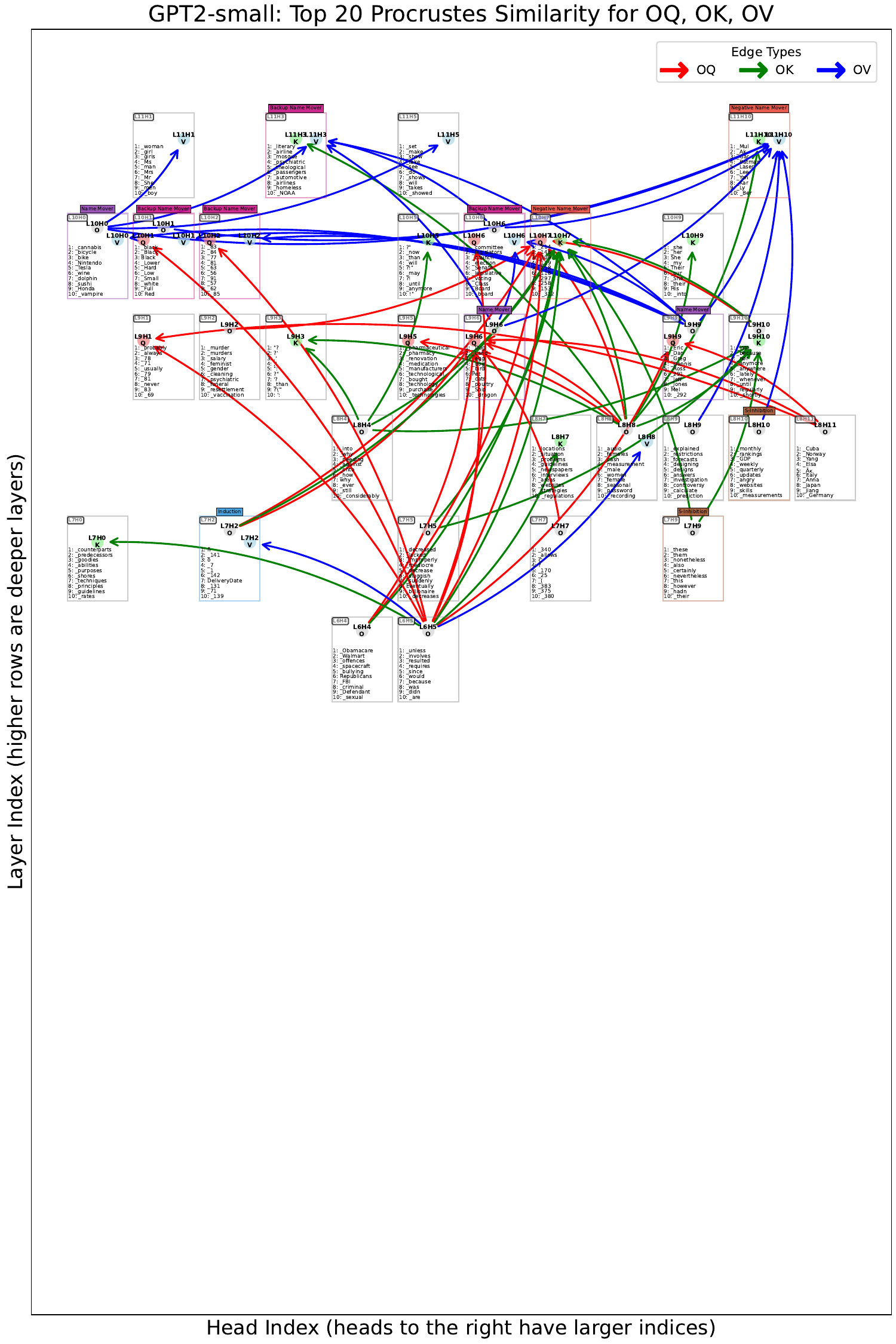}
    \caption{For GPT2-small, we visualize the top 20 scores of Procrustes Similarity for the pairing types OQ, OK, and OV.
For each head $h$, we show the top 10 tokens that represent $\bm{W}_\text{O}^h$ using the projection method in Section~\ref{sec:unembedding-proj}. For readability, we replace Ġ in tokens with \_.}
\label{fig:wd-procrustes-top20}
\end{figure*}

Figures~\ref{fig:wd-simple-cs-top20},~\ref{fig:wd-linearcka-top20}, and~\ref{fig:wd-procrustes-top20} show the wiring diagrams with the top 20 edges for Simple-CS, Linear CKA, and Procrustes Similarity, respectively.
For Simple-CS in Fig.~\ref{fig:wd-simple-cs-top20}, as with CS, many heads that do not belong to any head class appear in early layers.
In particular, Simple-CS yields the largest number of such heads among the methods under the top-20 setting (see Table~\ref{tab:gpt2_small_wiring_diagram_heads}).
For Linear CKA in Fig.~\ref{fig:wd-linearcka-top20} and Procrustes Similarity in Fig.~\ref{fig:wd-procrustes-top20}, the displayed heads are concentrated in late layers, with dense connections among them.
Consequently, unlike PK in Fig.~\ref{fig:wd-pk-top20}, these methods do not capture relationships involving heads in early layers.

\begin{figure*}[p!]
    \centering
    \includegraphics[width=0.8\linewidth]{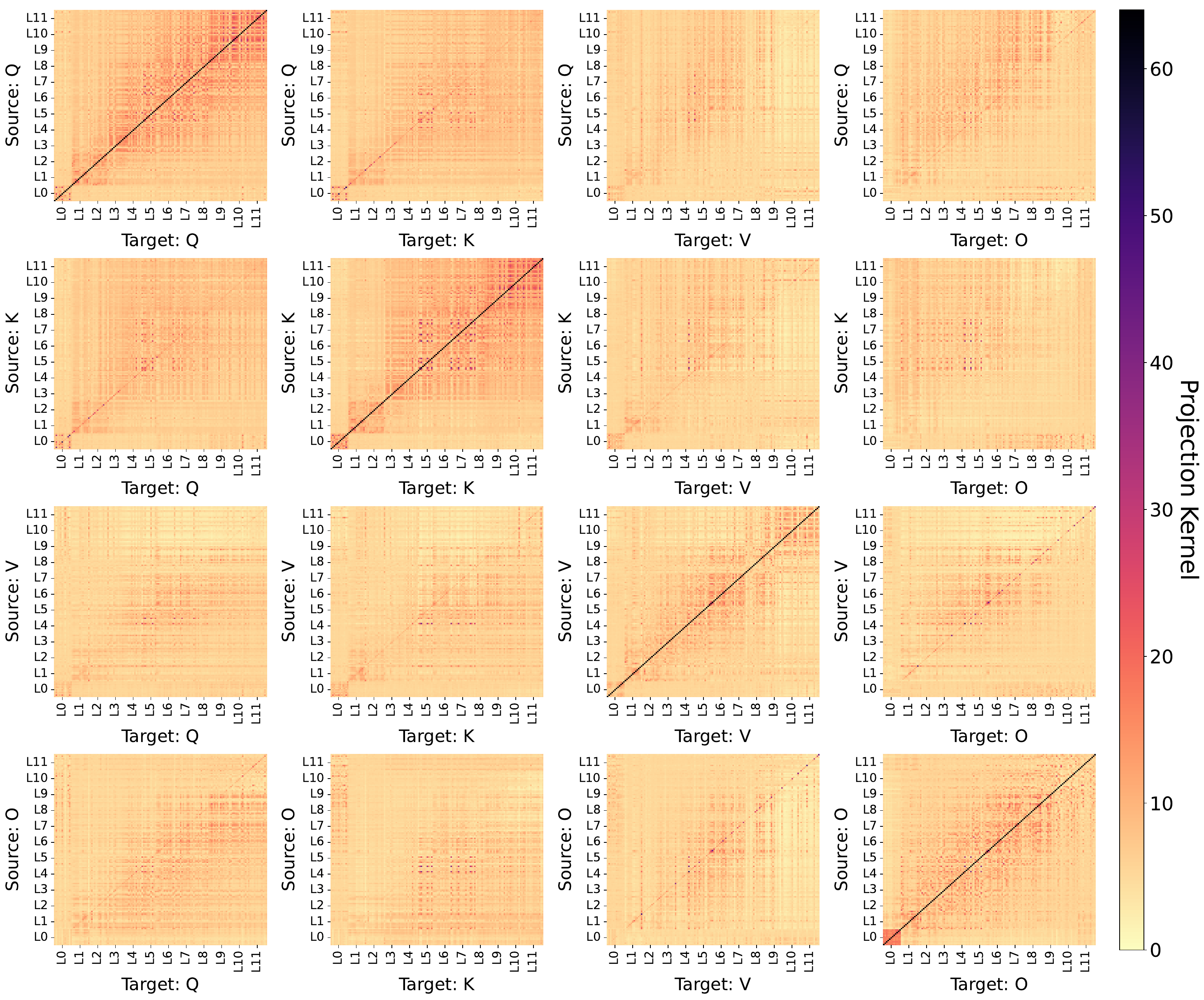}
    \caption{Projection Kernel (PK) between all head pairs of GPT2-small for each weight pairing type (source, target) of Q, K, V, and O, where the tick label L$\ell$ denotes layer $\ell$.}
\label{fig:gpt2-small_heatmap_pk}
\end{figure*}

\begin{figure*}[p!]
    \centering
    \includegraphics[width=0.8\linewidth]{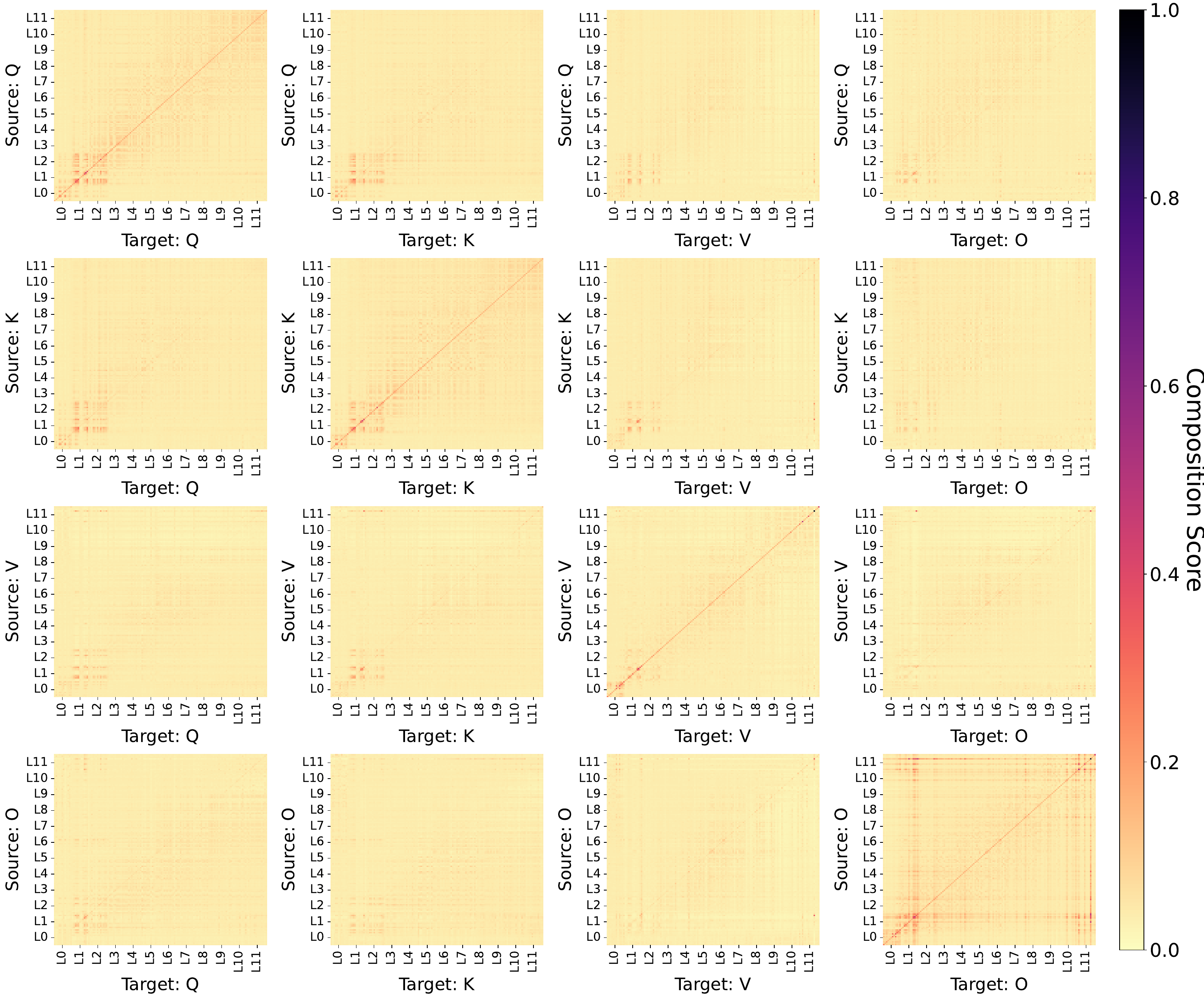}
    \caption{Composition Score (CS) between all head pairs of GPT2-small for each weight pairing type (source, target) of Q, K, V, and O, where the tick label L$\ell$ denotes layer $\ell$.}
\label{fig:gpt2-small_heatmap_cs}
\end{figure*}

\begin{figure*}[p!]
    \centering
    \includegraphics[width=0.8\linewidth]{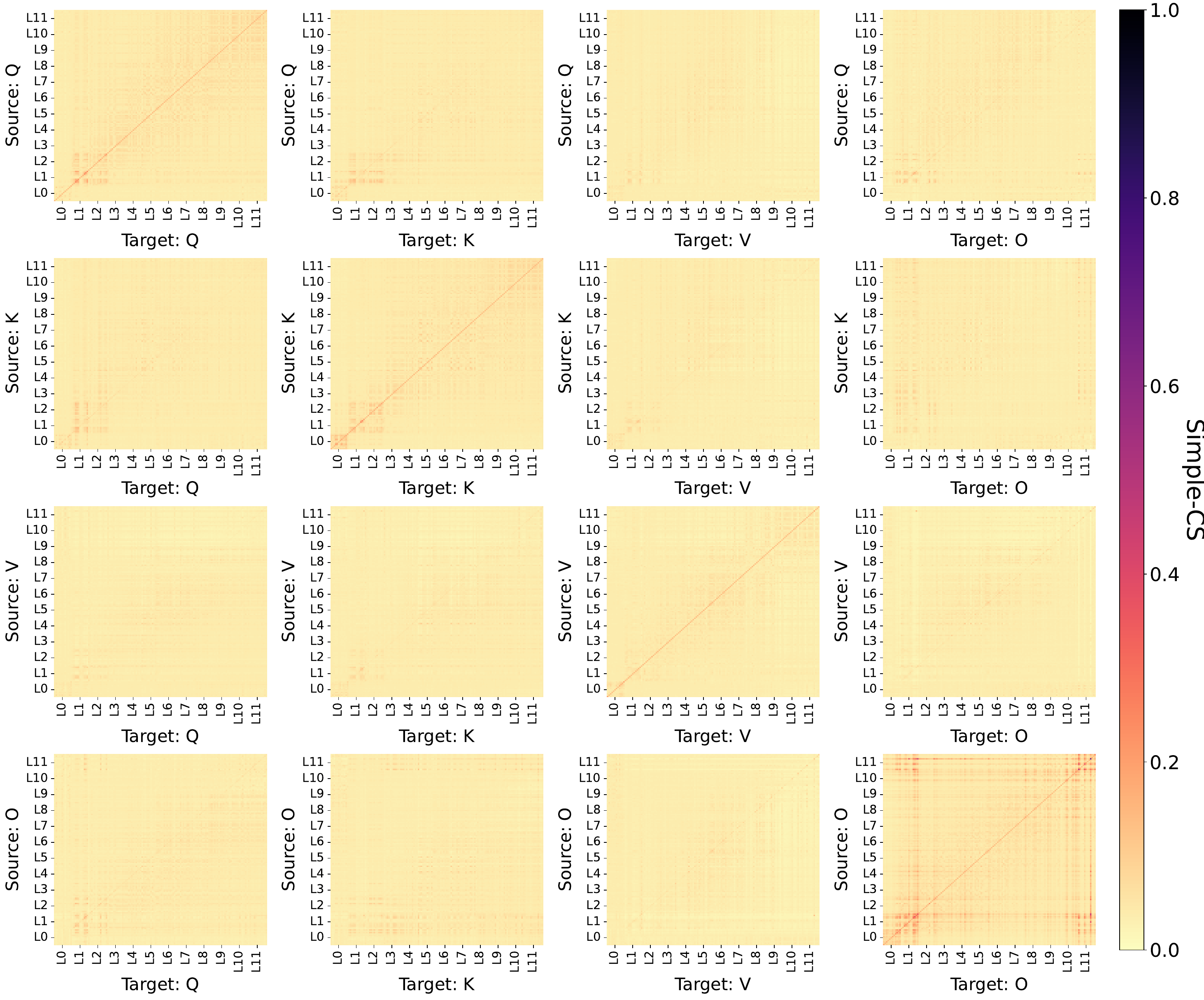}
    \caption{Simple-CS between all head pairs of GPT2-small for each weight pairing type (source, target) of Q, K, V, and O, where the tick label L$\ell$ denotes layer $\ell$.}
\label{fig:gpt2-small_heatmap_simple_cs}
\end{figure*}

\begin{figure*}[p!]
    \centering
    \includegraphics[width=0.8\linewidth]{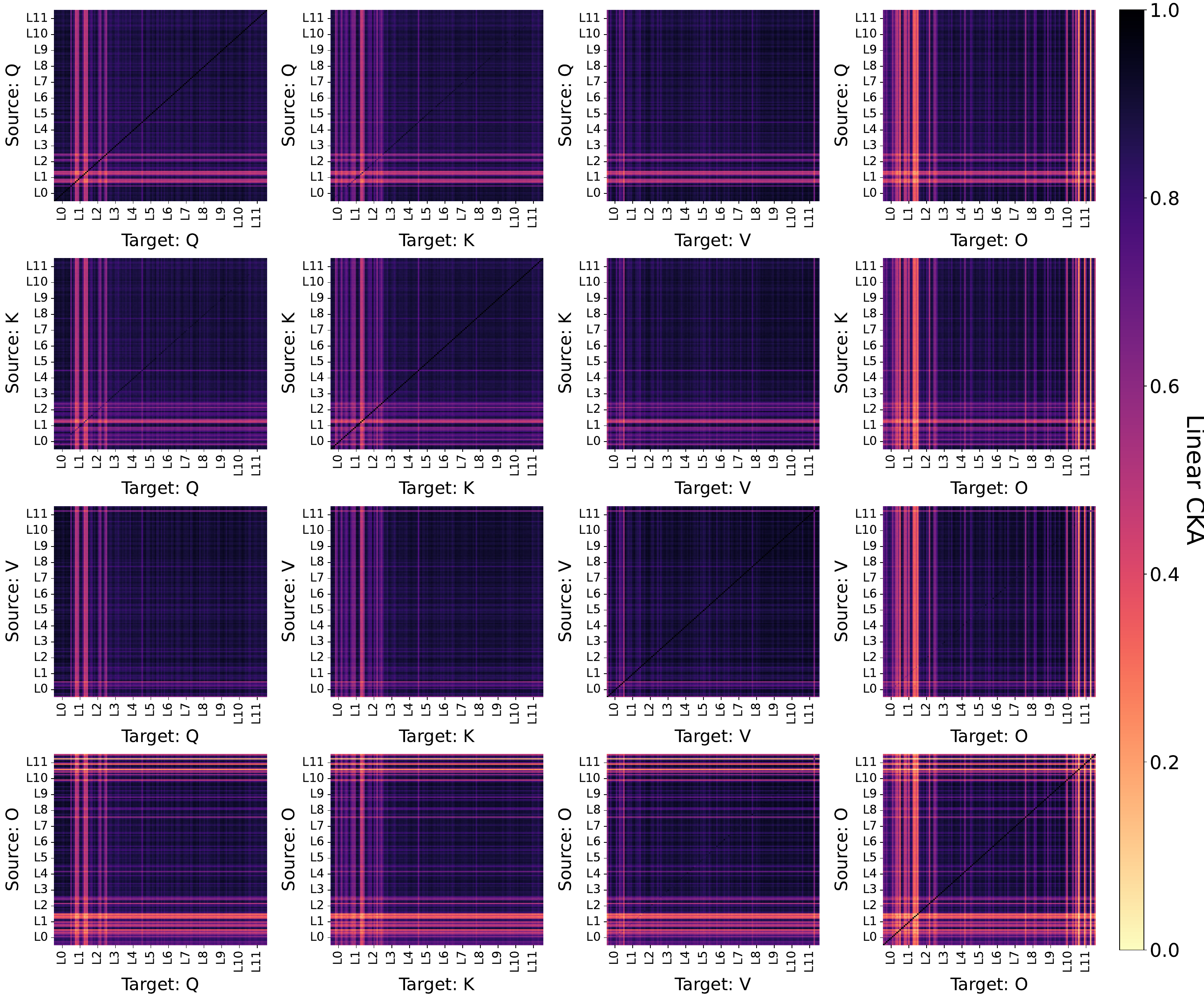}
    \caption{Linear CKA between all head pairs of GPT2-small for each weight pairing type (source, target) of Q, K, V, and O, where the tick label L$\ell$ denotes layer $\ell$.}
\label{fig:gpt2-small_heatmap_linear_cka}
\end{figure*}

\begin{figure*}[p!]
    \centering
    \includegraphics[width=0.8\linewidth]{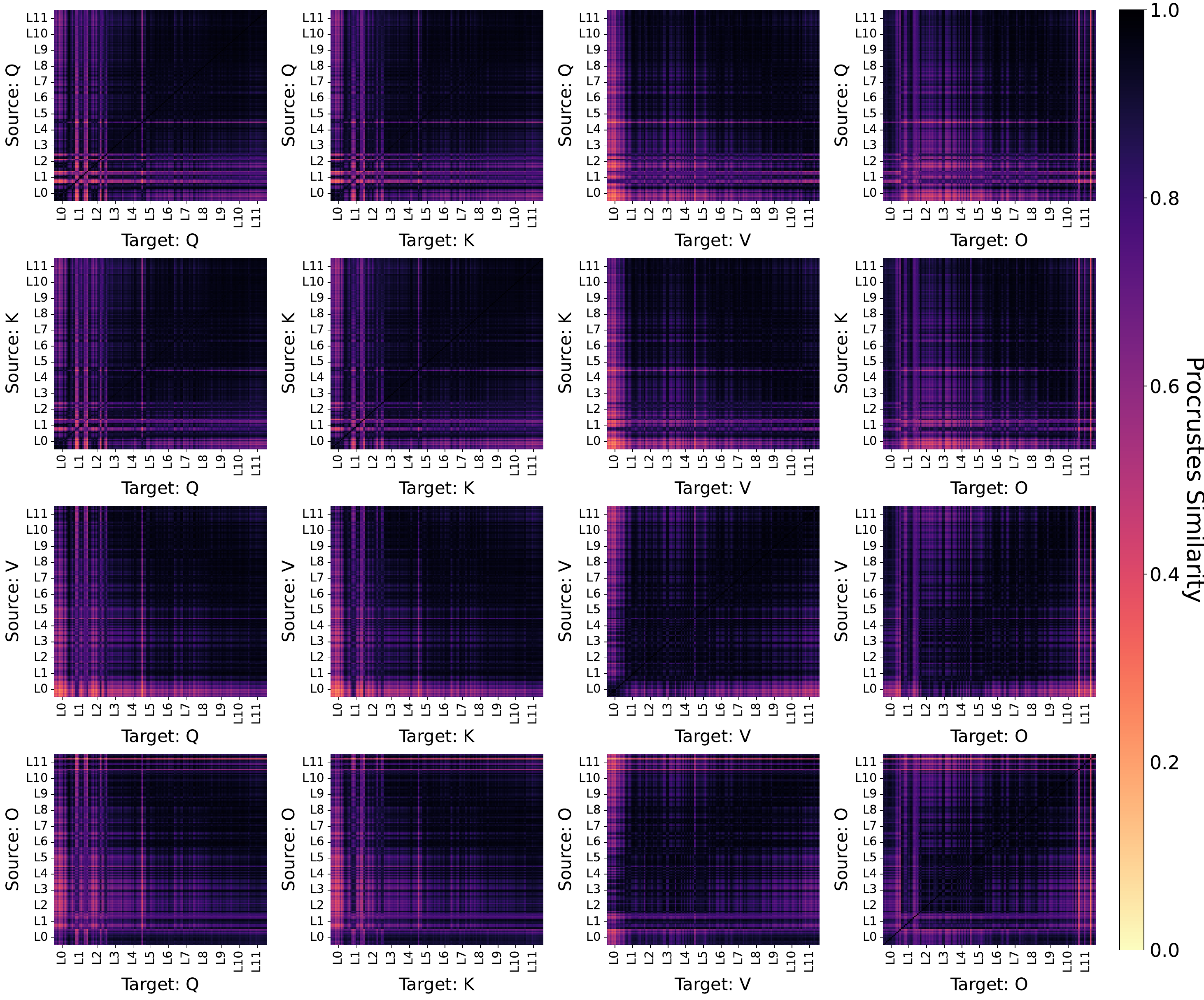}
    \caption{Procrustes Similarity between all head pairs of GPT2-small for each weight pairing type (source, target) of Q, K, V, and O, where the tick label L$\ell$ denotes layer $\ell$.}
\label{fig:gpt2-small_heatmap_procrustes_similarity}
\end{figure*}

\begin{figure*}[p!]
    \centering
    \includegraphics[width=0.85\linewidth]{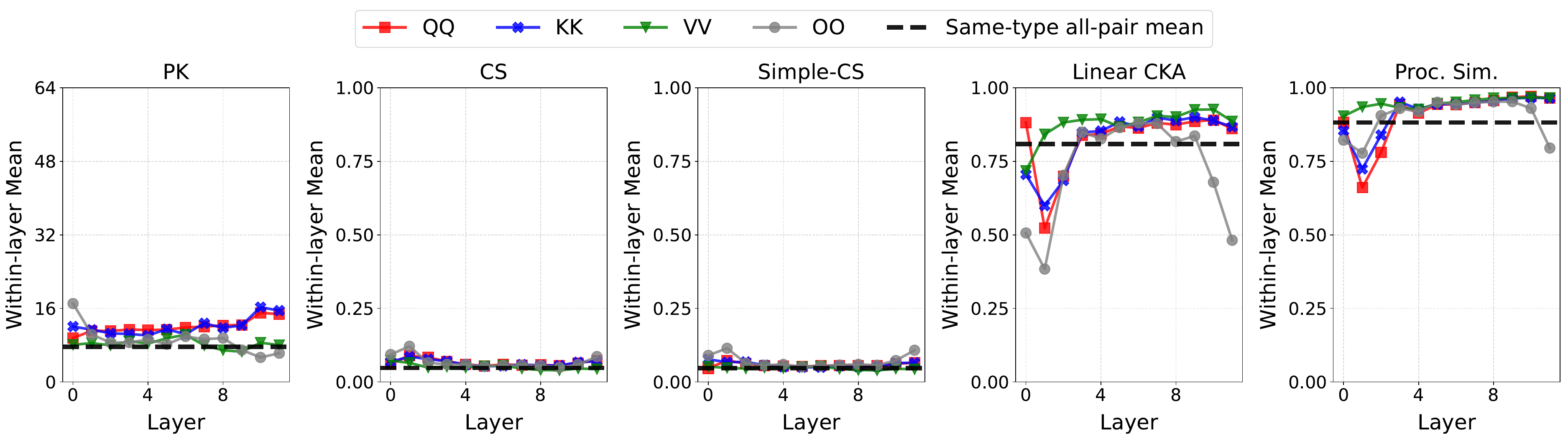}
    \caption{
Layerwise PK, CS, Simple-CS, Linear CKA, and Procrustes Similarity of GPT2-small, where each panel shows the mean over all $66$ head pairs within each layer (all pairwise combinations of the $12$ heads) for the weight pairing types QQ, KK, VV, and OO.
The dashed line marks the mean over all $\binom{144}{2}=10{,}296$ head pairs for each of the four weight pairing types, yielding a total of $4\times\binom{144}{2}=41{,}184$ comparisons: $\operatorname{PK}=7.65$, $\operatorname{CS}=0.05$, $\operatorname{Simple\text{-}CS}=0.05$, $\operatorname{Linear\ CKA}=0.81$, and $\operatorname{Proc.\ Sim.}=0.88$.
}
\label{fig:gpt2-small_layerwise_all_metrics}
\end{figure*}

\begin{figure*}[p!]
    \centering
    \includegraphics[width=0.85\linewidth]{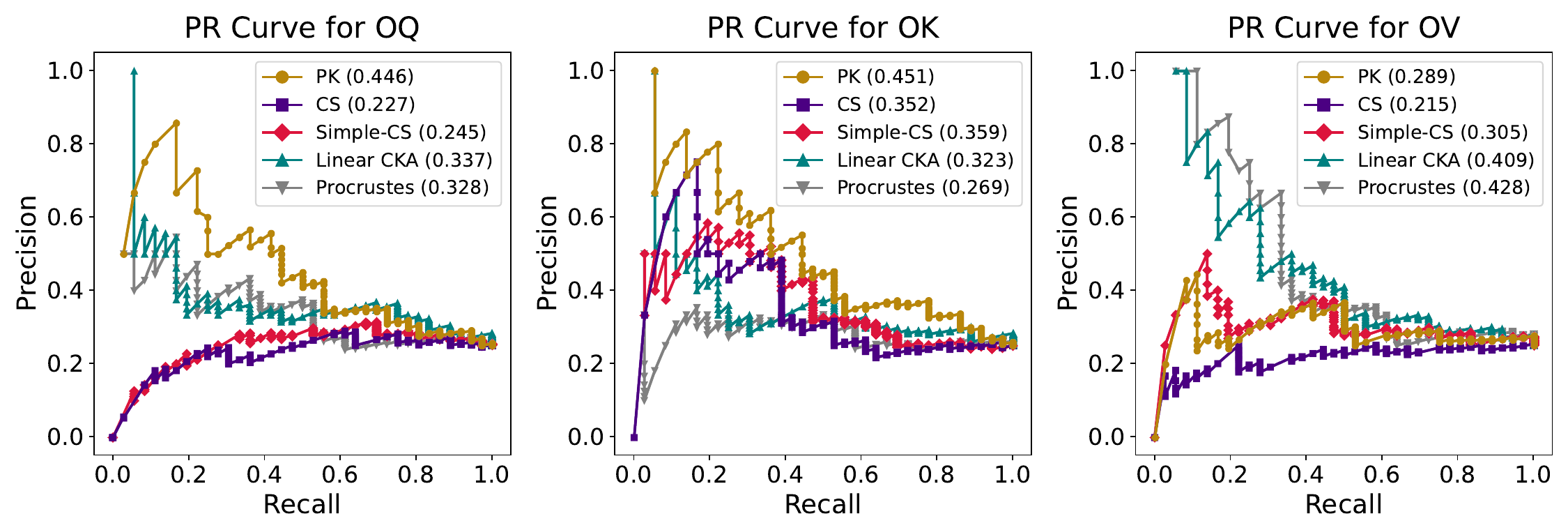}
    \caption{PR curves of head detection for OQ, OK, and OV in GPT2-small, shown for each method.}
\label{fig:gpt2-small_head_detection_OQ_OK_OV_pr_curve}
\end{figure*}

\subsection{Score heatmaps for 16 weight pairing types}\label{app:heatmap-for-all-methods}
In this section, we examine the score patterns between heads for 16 weight pairing types using heatmaps for each method.
Figures~\ref{fig:gpt2-small_heatmap_pk},~\ref{fig:gpt2-small_heatmap_cs},~\ref{fig:gpt2-small_heatmap_simple_cs},~\ref{fig:gpt2-small_heatmap_linear_cka}, and~\ref{fig:gpt2-small_heatmap_procrustes_similarity} show the score heatmaps for PK, CS, Simple-CS, Linear CKA, and Procrustes Similarity, respectively.
The color bar of each heatmap is scaled to the range of the corresponding method\footnote{Specifically, the range of PK is from 0 to $d_\text{head}$, whereas those of the other methods are from 0 to 1. Rescaling the PK range to 0 to 1 does not change the colors in the heatmap.}.

In the early layers, PK, CS, and Simple-CS tend to yield high within-layer scores for same-type weight pairings, namely QQ, KK, VV, and OO.
The PK heatmaps exhibit greater contrast than those of CS and Simple-CS, indicating that PK scores span a broader portion of their range.
Furthermore, for the OQ, OK, and OV heatmaps of PK, scores tend to be high in the bottom-right region, corresponding to pairs of a head in an earlier layer and a head in a later layer.
This pattern is consistent with the relationships being readily visible in the PK wiring diagrams.
In contrast, although CS and Simple-CS have a range from 0 to 1, most of their scores are close to 0 and exhibit limited variation.

Although Linear CKA and Procrustes Similarity also have a range from 0 to 1, their scores are close to 1 in most cases and therefore show similarly limited variation.
However, we find that the scores become extremely small for several specific heads.
A more detailed analysis of these methods and specific heads is left for future work.

\subsection{Layerwise averages for same-type weight pairings}\label{app:layerwise-metric-trend}

Building on the analysis in Appendix~\ref{app:heatmap-for-all-methods}, Fig.~\ref{fig:gpt2-small_layerwise_all_metrics} shows, for each method and each same-type weight pairing (QQ, KK, VV, and OO), the mean score within each layer together with the overall mean across the considered head pairs as a baseline.

For PK, the within-layer mean is higher than the baseline in most cases.
Furthermore, the mean scores for QQ and KK increase with depth, whereas those for VV and OO decrease.
For CS and Simple-CS, the within-layer means remain close to the baseline, although slight variations are observed in the early layers and the final layer.
For Linear CKA and Procrustes Similarity, the scores are generally high, as also observed in the heatmaps.
In the early layers, however, the scores are relatively low, and those for QQ, KK, and OO drop sharply in layer 1, that is, the second layer.
From the middle layers onward, the mean scores for all four pairing types exceed the baseline, except that OO drops sharply again toward the final layers, falling below the baseline.
A more detailed layerwise analysis is left for future work.

\section{Details of the evaluation of each method on GPT2-small}\label{app:eval-details}

\subsection{Details of head detection performance for OQ, OK, and OV}\label{app:eval-detecting}
Figure~\ref{fig:gpt2-small_head_detection_OQ_OK_OV_pr_curve} shows the PR curves used to compute the PR-AUC values in Table~\ref{tab:PRAUC-OQ-OK-OV}.
For OQ and OK, the PK curve lies above the other curves in almost all cases, resulting in larger PR-AUC values.
In contrast, for OV, the Linear CKA and Procrustes Similarity curves lie above the PK curve, resulting in higher PR-AUC values than that of PK.

\subsection{Details of classwise classification performance for QQ, KK, VV, and OO}\label{app:eval-clustering}

Table~\ref{tab:gpt2_small_classification_appendix} reports, for each head class, the mean PR-AUC and ROC-AUC scores used to compute Table~\ref{tab:classification-summary}, with the mean taken over QQ, KK, VV, and OO.
It also reports PR-AUC and ROC-AUC computed separately for each of the four weight pairing types QQ, KK, VV, and OO.

Focusing on the mean scores over QQ, KK, VV, and OO, we find that PK achieves high classification performance for Duplicate Token, Induction, Negative Name Mover, Backup Name Mover, and S-Inhibition Heads, indicating that PK performs well for many head classes.
In contrast, CS outperforms PK for Previous Token Heads, and Procrustes Similarity outperforms PK for Name Mover Heads.

The results for individual weight pairing types further show that classwise classification performance depends strongly on the weight pairing type as well as the method.
For Previous Token Heads, PK underperforms CS for QQ, KK, and VV, whereas PK attains the best performance for OO.
For Name Mover Heads, PK underperforms Procrustes Similarity for VV and OO, whereas PK attains the best performance for QQ and KK.
A more detailed analysis of the relationships among the method, weight pairing type, and head class is left for future work.

\begin{table*}[p!]
\scriptsize
\centering
    
\begin{subtable}{\textwidth}
\centering
\begin{tabular}{l|l|rrrrrrr|r}
\toprule
& Method & Dup (3) & Prev (10) & Ind (6) & NM (3) & N-NM (2) & B-NM (8) & S-Inh (4) & Avg \\
\midrule
\multirow{5}{*}{QQ}  & PK & \textbf{0.006} & 0.012 & \textbf{0.199} & \textbf{0.055} & \underline{0.001} & \textbf{0.090} & \textbf{0.009} & \textbf{0.053} \\
 & CS & 0.001 & \textbf{0.052} & 0.023 & 0.005 & 0.000 & 0.020 & 0.003 & 0.015 \\
 & Simple-CS & \underline{0.002} & \underline{0.021} & 0.045 & 0.016 & 0.001 & \underline{0.041} & \underline{0.005} & \underline{0.019} \\
 & Linear CKA & 0.001 & 0.003 & 0.036 & 0.003 & 0.001 & 0.006 & 0.001 & 0.007 \\
 & Proc. Sim. & 0.001 & 0.003 & \underline{0.049} & \underline{0.019} & \textbf{0.012} & 0.039 & 0.001 & 0.018 \\
\midrule
\multirow{5}{*}{KK}  & PK & \textbf{0.006} & 0.010 & \textbf{0.685} & \textbf{0.031} & 0.003 & \textbf{0.063} & \textbf{0.008} & \textbf{0.115} \\
 & CS & 0.001 & \textbf{0.111} & 0.066 & 0.006 & 0.001 & 0.022 & 0.003 & 0.030 \\
 & Simple-CS & \underline{0.002} & \underline{0.055} & \underline{0.137} & \underline{0.011} & 0.002 & \underline{0.031} & \underline{0.004} & 0.034 \\
 & Linear CKA & 0.001 & 0.003 & 0.002 & 0.001 & \underline{0.008} & 0.003 & 0.003 & 0.003 \\
 & Proc. Sim. & 0.000 & 0.004 & 0.003 & 0.002 & \textbf{0.250} & 0.012 & 0.002 & \underline{0.039} \\
\midrule
\multirow{5}{*}{VV}  & PK & 0.000 & 0.018 & \underline{0.009} & 0.008 & \underline{0.009} & 0.015 & \textbf{0.002} & 0.009 \\
 & CS & \textbf{0.000} & \textbf{0.021} & 0.007 & 0.002 & 0.003 & 0.007 & 0.002 & 0.006 \\
 & Simple-CS & \underline{0.000} & \underline{0.021} & \textbf{0.010} & 0.006 & 0.006 & 0.011 & \underline{0.002} & 0.008 \\
 & Linear CKA & 0.000 & 0.004 & 0.001 & \underline{0.034} & 0.004 & \underline{0.036} & 0.001 & \underline{0.011} \\
 & Proc. Sim. & 0.000 & 0.007 & 0.003 & \textbf{0.070} & \textbf{0.022} & \textbf{0.045} & 0.002 & \textbf{0.021} \\
\midrule
\multirow{5}{*}{OO}  & PK & \textbf{0.002} & \textbf{0.038} & \textbf{0.024} & 0.002 & \textbf{0.019} & 0.002 & \textbf{0.004} & 0.013 \\
 & CS & \underline{0.000} & 0.012 & 0.002 & 0.000 & 0.000 & 0.002 & 0.001 & 0.003 \\
 & Simple-CS & 0.000 & 0.009 & 0.002 & 0.000 & 0.000 & 0.002 & 0.001 & 0.002 \\
 & Linear CKA & 0.000 & 0.006 & 0.004 & \underline{0.430} & 0.000 & \underline{0.010} & 0.001 & \underline{0.064} \\
 & Proc. Sim. & 0.000 & \underline{0.014} & \underline{0.010} & \textbf{0.461} & \underline{0.001} & \textbf{0.018} & \underline{0.002} & \textbf{0.072} \\
\midrule
\multirow{5}{*}{Mean over QQ, KK, VV, OO}  & PK & \textbf{0.003} & 0.019 & \textbf{0.229} & 0.024 & \underline{0.008} & \textbf{0.042} & \textbf{0.006} & \textbf{0.047} \\
 & CS & 0.001 & \textbf{0.049} & 0.024 & 0.003 & 0.001 & 0.013 & 0.002 & 0.013 \\
 & Simple-CS & \underline{0.001} & \underline{0.027} & \underline{0.048} & 0.008 & 0.002 & 0.021 & \underline{0.003} & 0.016 \\
 & Linear CKA & 0.000 & 0.004 & 0.011 & \underline{0.117} & 0.003 & 0.014 & 0.001 & 0.021 \\
 & Proc. Sim. & 0.000 & 0.007 & 0.016 & \textbf{0.138} & \textbf{0.071} & \underline{0.029} & 0.002 & \underline{0.038} \\
\bottomrule
\end{tabular}
\caption{PR-AUC}
\label{tab:gpt2_small_classification_appendix_pr_auc}
\end{subtable}

\par\bigskip

\begin{subtable}{\textwidth}
\centering
\begin{tabular}{l|l|rrrrrrr|r}
\toprule
& Method & Dup (3) & Prev (10) & Ind (6) & NM (3) & N-NM (2) & B-NM (8) & S-Inh (4) & Avg \\
\midrule
\multirow{5}{*}{QQ}  & PK & 0.453 & 0.748 & 0.967 & \textbf{0.997} & \underline{0.952} & \underline{0.922} & \textbf{0.953} & \underline{0.856} \\
 & CS & 0.416 & \textbf{0.898} & 0.898 & 0.973 & 0.902 & 0.895 & 0.911 & 0.842 \\
 & Simple-CS & 0.454 & \underline{0.842} & 0.943 & \underline{0.992} & 0.921 & 0.921 & \underline{0.931} & \textbf{0.858} \\
 & Linear CKA & \textbf{0.800} & 0.265 & \underline{0.971} & 0.923 & 0.945 & 0.779 & 0.730 & 0.773 \\
 & Proc. Sim. & \underline{0.538} & 0.433 & \textbf{0.974} & 0.989 & \textbf{0.996} & \textbf{0.967} & 0.714 & 0.802 \\
\midrule
\multirow{5}{*}{KK}  & PK & 0.360 & 0.732 & \textbf{0.998} & \textbf{0.995} & 0.985 & \textbf{0.926} & \textbf{0.964} & 0.851 \\
 & CS & 0.361 & \textbf{0.932} & 0.977 & 0.977 & 0.945 & 0.891 & 0.902 & \underline{0.855} \\
 & Simple-CS & 0.365 & \underline{0.897} & \underline{0.991} & \underline{0.987} & 0.979 & \underline{0.899} & \underline{0.924} & \textbf{0.863} \\
 & Linear CKA & \textbf{0.739} & 0.252 & 0.702 & 0.791 & \underline{0.994} & 0.643 & 0.893 & 0.716 \\
 & Proc. Sim. & \underline{0.394} & 0.468 & 0.806 & 0.908 & \textbf{1.000} & 0.895 & 0.730 & 0.743 \\
\midrule
\multirow{5}{*}{VV}  & PK & 0.418 & 0.874 & 0.709 & 0.821 & \underline{0.994} & 0.753 & \underline{0.850} & 0.774 \\
 & CS & \textbf{0.531} & \textbf{0.889} & 0.725 & 0.777 & 0.981 & 0.725 & 0.829 & 0.780 \\
 & Simple-CS & \underline{0.520} & \underline{0.885} & \underline{0.734} & 0.792 & 0.992 & 0.749 & 0.847 & \textbf{0.788} \\
 & Linear CKA & 0.287 & 0.458 & 0.454 & \underline{0.996} & 0.989 & \textbf{0.932} & 0.745 & 0.694 \\
 & Proc. Sim. & 0.307 & 0.698 & \textbf{0.746} & \textbf{0.998} & \textbf{0.998} & \underline{0.903} & \textbf{0.856} & \underline{0.786} \\
\midrule
\multirow{5}{*}{OO}  & PK & \underline{0.562} & \textbf{0.914} & \underline{0.842} & 0.712 & \textbf{0.997} & 0.332 & \textbf{0.910} & \underline{0.753} \\
 & CS & \textbf{0.601} & 0.831 & 0.685 & 0.512 & 0.902 & 0.336 & 0.779 & 0.663 \\
 & Simple-CS & 0.557 & 0.789 & 0.714 & 0.506 & 0.881 & 0.364 & 0.794 & 0.658 \\
 & Linear CKA & 0.327 & 0.651 & 0.791 & \underline{0.999} & 0.611 & \underline{0.660} & 0.693 & 0.676 \\
 & Proc. Sim. & 0.369 & \underline{0.851} & \textbf{0.893} & \textbf{1.000} & \underline{0.922} & \textbf{0.802} & \underline{0.877} & \textbf{0.816} \\
\midrule
\multirow{5}{*}{Mean over QQ, KK, VV, OO}  & PK & 0.448 & 0.817 & \textbf{0.879} & 0.881 & \textbf{0.982} & 0.733 & \textbf{0.919} & \textbf{0.809} \\
 & CS & \underline{0.477} & \textbf{0.887} & 0.821 & 0.810 & 0.932 & 0.712 & 0.855 & 0.785 \\
 & Simple-CS & 0.474 & \underline{0.853} & 0.845 & 0.819 & 0.943 & 0.733 & \underline{0.874} & \underline{0.792} \\
 & Linear CKA & \textbf{0.538} & 0.407 & 0.729 & \underline{0.927} & 0.885 & \underline{0.754} & 0.765 & 0.715 \\
 & Proc. Sim. & 0.402 & 0.612 & \underline{0.855} & \textbf{0.974} & \underline{0.979} & \textbf{0.892} & 0.794 & 0.787 \\
\bottomrule
\end{tabular}
\caption{ROC-AUC}
\label{tab:gpt2_small_classification_appendix_auc}
\end{subtable}
\caption{
PR-AUC and ROC-AUC for each head class in GPT2-small, computed separately for the weight pairing types QQ, KK, VV, and OO, together with the mean over the four weight pairing types for each head class.
These values are used to compute the summary reported in Table~\ref{tab:classification-summary}.
The maximum is shown in bold and the second largest is underlined.
We consider the $\binom{144}{2}=10{,}296$ directed head pairs whose source is in an earlier layer than the target; for heads in the same layer, the head with the smaller index is treated as the source.
Dup, Prev, Ind, NM, N-NM, B-NM, and S-Inh denote Duplicate Token, Previous Token, Induction, Name Mover, Negative Name Mover, Backup Name Mover, and S-Inhibition heads, respectively.
}
\label{tab:gpt2_small_classification_appendix}
\end{table*}

\begin{table*}[t!]
\small
\centering
    
\begin{subtable}{\textwidth}
\centering
\begin{tabular}{rlrlrlr}
\toprule
Top & \multicolumn{2}{c}{OQ} & \multicolumn{2}{c}{OK} & \multicolumn{2}{c}{OV}\\
\midrule
1 & L10H7 (N-NM) & 161.02 & L5H1 (Ind) & 140.81 & L4H7 (Id) & 228.05 \\
2 & L5H6 (Prev) & 135.27 & L7H7 & 123.09 & L11H3 (B-NM) & 180.96 \\
3 & L7H1 & 100.09 & L5H5 (Ind) & 105.05 & L9H3 & 145.73 \\
4 & L8H10 (S-Inh) & 65.27 & L11H3 (B-NM) & 79.57 & L5H2 & 135.06 \\
5 & L10H11 & 60.13 & L11H8 & 76.27 & L11H8 & 133.63 \\
6 & L9H0 & 52.52 & L9H4 & 70.04 & L7H1 & 63.83 \\
7 & L7H7 & 51.84 & L10H11 & 65.57 & L8H7 & 62.65 \\
8 & L2H1 & 51.22 & L6H8 (Prev) & 51.08 & L6H8 (Prev) & 55.31 \\
9 & L1H8 & 50.87 & L6H6 & 48.90 & L6H6 & 51.27 \\
10 & L11H10 (N-NM) & 49.37 & L5H0 (Ind) & 47.84 & L7H4 & 47.76 \\
\bottomrule
\end{tabular}
\caption{Inlet Head Scores}
\label{tab:inlet_appendix}
\end{subtable}

\par\bigskip

\begin{subtable}{\textwidth}
\centering
\begin{tabular}{rlrlrlr}
\toprule
Top & \multicolumn{2}{c}{OQ} & \multicolumn{2}{c}{OK} & \multicolumn{2}{c}{OV}\\
\midrule
1 & L7H6 & 243.47 & L4H7 (Id) & 263.86 & L0H1 (Dup, Id) & 201.87 \\
2 & L1H1 & 178.36 & L4H11 (Prev) & 191.91 & L3H0 (Dup, Id) & 91.87 \\
3 & L4H7 (Id) & 129.33 & L0H9 & 142.50 & L4H7 (Id) & 78.90 \\
4 & L8H10 (S-Inh) & 119.35 & L1H11 (Id) & 135.82 & L5H10 & 78.79 \\
5 & L5H2 & 114.06 & L5H10 & 106.17 & L5H11 & 77.93 \\
6 & L5H3 & 109.91 & L1H1 & 95.06 & L3H10 & 71.73 \\
7 & L6H6 & 100.38 & L0H11 & 80.11 & L1H6 & 67.85 \\
8 & L0H9 & 85.10 & L1H9 & 80.02 & L5H5 (Ind) & 60.81 \\
9 & L3H10 & 76.71 & L5H3 & 77.17 & L0H4 (Id) & 53.09 \\
10 & L7H4 & 56.06 & L6H0 & 54.17 & L1H1 & 40.85 \\
\bottomrule
\end{tabular}
\caption{Outlet Head Scores}
\label{tab:outlet_appendix}
\end{subtable}

\caption{
For OQ, OK, and OV in GPT2-small, we report (a) the top 10 heads and their inlet scores in (\ref{eq:inlet}) and (b) the top 10 heads and their outlet scores in (\ref{eq:outlet}).
When a head belongs to the head classes in Table~\ref{tab:head-class} or to the top 10 heads by Identity Head score in Table~\ref{tab:identity}, we also report this information.
Dup, Prev, Ind, NM, N-NM, B-NM, and S-Inh denote Duplicate Token, Previous Token, Induction, Name Mover, Negative Name Mover, Backup Name Mover, and S-Inhibition heads, respectively.
}
\label{tab:inlet_outlet_appendix}
\end{table*}

\section{Details of inlet and outlet scores in GPT2-small}\label{app:inlet-outlet}
Motivated by the wiring diagram based on PK in Fig.~\ref{fig:wd-pk-top20}, where L4H7 acts as a hub, we aim to detect such hub heads in GPT2-small.
To this end, we characterize the hub role in terms of inlet and outlet behavior, and propose the inlet score and outlet score to identify heads that play each role\footnote{Although inlet and outlet scores can be defined for similarity measures other than PK, such as CS, we focus on PK in this paper.}.
This section describes the definitions and results.

First, for each of OQ, OK, and OV, we identify inlet heads by quantifying the extent to which a target head receives the maximum PK score from source heads in earlier layers.
For a head $h$ in layer 1 or later, and for $\omega\in\{\text{``Q''},\text{``K''},\text{``V''}\}$, we define the PK inlet score as
\begin{align}
&\operatorname{PK\text{-}Inlet}_{\,{\text{O}\,\omega}}(h) \notag\\
=\,&\sum_{h'\in \bigcup_{\ell'<\ell_h}\mathcal{H}_{\ell'}}\operatorname{PK}((h',\text{O}),(h, \omega))
\,\mathbb{I}\Biggl(\operatorname{PK}((h',\text{O}),(h, \omega))=
\max_{h''\in \bigcup_{\ell_{h'}<\ell''}\mathcal{H}_{\ell''}}\operatorname{PK}((h',\text{O}),(h'',\omega))\Biggr),\label{eq:inlet}
\end{align}
where
\begin{equation}
    \mathbb{I}(\text{condition}) :=
\begin{cases}
1 & \text{if the condition holds},\\
0 & \text{otherwise},
\end{cases}
\label{eq:cond}
\end{equation}
where $\ell_h$ denotes the layer index of head $h$, and $\mathcal{H}_\ell$ denotes the set of heads in layer $\ell$.
Here and throughout this section, we use shorthand notation for PK\footnote{For example, $\operatorname{PK}((h',\text{O}),(h, \omega))$ abbreviates $\operatorname{PK}(\mathcal{S}^{(h',\text{O})},\mathcal{S}^{(h, \omega)})$.}.
Fix a target head $h$ in layer $\ell_h$, and consider any source head $h'$ in an earlier layer $\ell_{h'}<\ell_h$.
For this fixed $h'$, we compute $\operatorname{PK}((h',\text{O}),(h'',\omega))$ for all target heads $h''$ in later layers $\ell_{h'}<\ell''$.
If $\operatorname{PK}((h',\text{O}),(h, \omega))$ attains this maximum, we add its value to the score.
Summing these contributions over all such source heads $h'$ gives the inlet score for the target head $h$.

Next, for each of OQ, OK, and OV, we identify outlet heads by quantifying the extent to which a source head yields the maximum PK score for target heads in subsequent layers.
For a head $h$ in any layer before the final layer, and for $\omega\in\{\text{``Q''},\text{``K''},\text{``V''}\}$, we define the PK outlet score as
\begin{align}
&\operatorname{PK\text{-}Outlet}_{\,{\text{O}\,\omega}}(h)\notag\\
=\,&\sum_{h'\in \bigcup_{\ell_h <\ell'}\mathcal{H}_{\ell'}}\operatorname{PK}((h, \text{O}),(h', \omega))\,\mathbb{I}\Biggl(\operatorname{PK}((h, \text{O}),(h', \omega))
=\max_{h''\in \bigcup_{\ell''<\ell_{h'}}\mathcal{H}_{\ell''}}\operatorname{PK}((h'',\text{O}),(h',\omega))\Biggr),\label{eq:outlet}
\end{align}
where $\mathbb{I}(\cdot)$, $\ell_h$, and $\mathcal{H}_\ell$ are defined as above.
Fix a source head $h$ in layer $\ell_h$, and consider any target head $h'$ in a later layer $\ell_h<\ell_{h'}$.
For this fixed $h'$, we compute $\operatorname{PK}((h'',\text{O}),(h',\omega))$ for all source heads $h''$ in earlier layers $\ell''<\ell_{h'}$.
If $\operatorname{PK}((h, \text{O}),(h',\omega))$ attains this maximum, we add its value to the score.
Summing these contributions over all such target heads $h'$ gives the outlet score for the source head $h$.

Table~\ref{tab:inlet_outlet_appendix} reports the top 10 heads and their scores for OQ, OK, and OV based on the inlet and outlet scores in (\ref{eq:inlet}) and (\ref{eq:outlet}).
L4H7 attains the largest inlet score for OV, and its value exceeds the largest inlet scores for OQ and OK.
In contrast, no other Identity Head appears among the top-ranked inlet heads.
Moreover, L4H7 also attains the largest outlet score for OK, and its value exceeds the largest outlet scores for OQ and OV.
L4H7 also ranks third for the outlet scores of OQ and OV.
In addition, Identity Heads such as L0H1, L3H0, L1H11, and L0H4 have large outlet scores.
A more detailed study of whether L4H7 is special, and of the relationship between Identity Heads and inlet and outlet scores, is left for future work.

\section{Approximate distribution of PK between random orthogonal matrices}\label{app:pk-random}
In Section~\ref{sec:disc}, we introduced a framework that quantifies the informativeness of PK distributions for weight pairings by using an approximate distribution of PK for random orthogonal matrices and measuring the Kullback--Leibler (KL) divergence between the two distributions.
In this section, we derive the approximate distribution of PK for random orthogonal matrices used in this framework.

Concretely, we sample $\bm{U}_A,\bm{U}_B\in \mathcal{V}_m(\mathbb{R}^d)$ from the uniform distribution\footnote{
Following \citet[Theorem 2.2.1 (iii)]{chikuse2003statistics} and \citet[Proposition 7.2]{eaton2007home}, we define the uniform distribution on $\mathcal{V}_m(\mathbb{R}^d)$ as follows.
We sample $\bm{W}\in\mathbb{R}^{d\times m}$ with i.i.d. entries from $\mathcal{N}(0,1)$, and orthonormalize its columns using the Gram--Schmidt procedure (or equivalently QR decomposition or SVD).
We use the resulting matrix $\bm{U}$ as a sample from the uniform distribution.} on the Stiefel manifold of $m$ orthonormal vectors in $\mathbb{R}^d$,
\begin{equation}
\mathcal{V}_m(\mathbb{R}^d)
 := \{\bm{U}\in\mathbb{R}^{d\times m} : \bm{U}^\top\bm{U} = \bm{I}_m\},
\end{equation}
and derive an approximate theoretical distribution of PK between the subspaces $\mathcal{S}_A$ and $\mathcal{S}_B$ spanned by the columns of $\bm{U}_A$ and $\bm{U}_B$.
When $m=d$, $\mathcal{V}_d(\mathbb{R}^d)$ coincides with the orthogonal group in $\mathbb{R}^d$,
\begin{equation}
    \mathcal{O}(\mathbb{R}^d) = \{\bm{R}\in\mathbb{R}^{d\times d} : \bm{R}^\top\bm{R} = \bm{R}\bm{R}^\top = \bm{I}_d \}.
\end{equation}

\subsection{Simplifying the setting using rotation invariance of PK}\label{app:simplified-random-settings}
By the rotation invariance of PK (see Appendix~\ref{app:pk-rotation-invariance}), letting $\bm{e}_i^\text{Std}\in\mathbb{R}^d$ denote the $i$-th standard basis vector, we may fix
\begin{equation}
\bm{U}_A
= (\bm{e}_1^\text{Std},\ldots,\bm{e}_m^\text{Std}) = 
\begin{pmatrix}
\bm{I}_m \\
\bm{0}
\end{pmatrix}\in \mathcal{V}_m(\mathbb{R}^d),\label{eq:UA}
\end{equation}
without loss of generality, where $\bm{I}_m\in\mathbb{R}^{m\times m}$ is the $m$-dimensional identity matrix.

Next, we sample a random orthogonal matrix $\bm{R}\in \mathcal{O}(\mathbb{R}^d)$ from the uniform distribution on $\mathcal{O}(\mathbb{R}^d)$ and define\footnote{
The resulting $\bm{U}_B$ follows the uniform distribution on $\mathcal{V}_m(\mathbb{R}^d)$; see \citet[Section 2.5.1]{chikuse2003statistics}.}
\begin{equation}
\bm{U}_B = \bm{R}\bm{U}_A = \bm{R}_{[:, 1:m]},\label{eq:UB}
\end{equation}
where $\bm{R}_{[:,\,1:m]}$ denotes the $d\times m$ submatrix of $\bm{R}$ consisting of its first $m$ columns,
so that $\bm{U}_B\in \mathcal{V}_m(\mathbb{R}^d)$.

Since $\bm{R}$ is random, the PK value between the subspaces $\mathcal{S}_A$ and $\mathcal{S}_B$ spanned by $\bm{U}_A$ and $\bm{U}_B$ is a random variable.
Letting this random variable be $Z$, from (\ref{eq:PK-UU}), (\ref{eq:UA}), and (\ref{eq:UB}), we obtain
\begin{align}
  Z = \operatorname{PK}(\mathcal{S}_A,\mathcal{S}_B)
    = \left\|\bm{U}_A^\top \bm{U}_B\right\|_\text{F}^2
    = \left\|\bm{U}_A^\top \bm{R}\bm{U}_A\right\|_\text{F}^2
    = \left\|\bm{R}_{[1:m],\,[1:m]}\right\|_\text{F}^2 = \sum_{i=1}^m \sum_{j=1}^m R_{ij}^2,\label{eq:Z-PK}
\end{align}
where $\bm{R}_{[1:m],\,[1:m]}$ denotes the top-left $m\times m$ block of $\bm{R}$, $R_{ij}$ is its $(i,j)$ entry, and $\|\cdot\|_{\text{F}}$ is the Frobenius norm.

\subsection{Loose approximation: approximating the distribution of \texorpdfstring{$R_{ij}$}{Rij} as Gaussian and applying the central limit theorem}\label{app:loose-approx}
The entries $R_{ij}$ of the random orthogonal matrix $\bm{R}$ sampled in Appendix~\ref{app:simplified-random-settings} are not independent due to orthogonality constraints.
As a loose approximation, we consider an approximate distribution of $Z$ under the assumption that the entries behave independently.

For each entry $R_{ij}$, \citet[Corollary 2.6]{meckes2019random} showed that\footnote{
Here, $\xrightarrow{\mathcal{D}}$ denotes convergence in distribution (weak convergence): for random variables
$X_d$ and $X$, $X_d \xrightarrow{\mathcal{D}} X$ means that the cumulative distribution functions satisfy
$F_{X_d}(\tau)\to F_X(\tau)$ as $d\to\infty$ for every $\tau$ at which $F_X$ is continuous, where
$F_Y(\tau):=\mathbb{P}(Y\le \tau)$~\cite[Section 7.2.4]{PishroNikTextbook}.
Here, $\mathbb{P}(\cdot)$ denotes probability under the underlying probability
measure. 
Equivalently, we write $X_d \xrightarrow{\mathcal{D}} \mathcal{L}$ if there exists a random variable $X$
such that $X_d \xrightarrow{\mathcal{D}} X$ and $X$ has probability distribution $\mathcal{L}$. 
}
\begin{equation}
    \sqrt{d}\,R_{ij} \xrightarrow{\mathcal{D}} \mathcal{N}(0,1)\quad(d\to\infty).
    \label{eq:Rij-clt}
\end{equation}
Consequently, for large $d$, $R_{ij}$ is approximately distributed as
\begin{equation}
    \mathcal{N}\left(0,\,\frac{1}{d}\right).
    \label{eq:Rij-inf}
\end{equation}
We therefore approximate $R_{ij}$ by independent random variables
\begin{equation}
    X_{ij} \sim \mathcal{N}\left(0,\,\frac{1}{d}\right).\label{eq:Xij-N01d}
\end{equation}
For $d=768$ and $m=64$, which match the dimensions of attention weights in GPT2-small, we sample $\bm{U}_B = \bm{R}_{[:, 1:m]}\in \mathcal{V}_m(\mathbb{R}^d)$ and find that the density of $\mathcal{N}(0,1/d)$ provides a good fit to the histogram of $R_{ij}$ (see Fig.~\ref{fig:entry-hist})\footnote{
This $\mathcal{N}(0,1/d)$ distribution also appears in a related high-dimensional limit: when $d$ is sufficiently large, the cosine similarity between two $d$-dimensional random vectors is approximately distributed as $\mathcal{N}(0,1/d)$, provided that each vector has mean-zero components with finite variance. The component variances are shared within each vector but may differ across the two vectors; see \citet{yamagiwa-etal-2024-axis,yamagiwa-etal-2025-revisiting}.
}.

For $X_{ij}^2$, let $Y_{ij} := \sqrt{d}\,X_{ij}$. Then $Y_{ij} \sim \mathcal{N}(0,1)$, and hence $Y_{ij}^2$ follows a chi-square distribution with one degree of freedom, $\chi_1^2$.
Since $X_{ij}^2 = Y_{ij}^2/d$, it follows that $X_{ij}^2$ follows a Gamma distribution\footnote{
In general, $\chi_\nu^2$ coincides with $\text{Gamma}(\nu/2,2)$.
Moreover, if $Y\sim \text{Gamma}(\nu/2,2)$ and $X=cY$ with $c>0$, then $X\sim \text{Gamma}(\nu/2,2c)$.}
with shape $k=1/2$ and scale $\theta=2/d$,
\begin{equation}
X_{ij}^2 \sim \text{Gamma}\left(\frac{1}{2},\frac{2}{d}\right).\label{eq:EX}
\end{equation}
Similarly, for $d=768$ and $m=64$, we sample $\bm{U}_B = \bm{R}_{[:, 1:m]}\in \mathcal{V}_m(\mathbb{R}^d)$ and find that the density of $\text{Gamma}(1/2,\,2/d)$ provides a good fit to the histogram of $R_{ij}^2$ (see Fig.~\ref{fig:squared-entry-hist}).
From the mean and variance formulas of the Gamma distribution~\cite[Section 8.1.6.5]{Heckert2002NISTHandbook151}, the mean and variance of $X_{ij}^2$ are given by
\begin{align}
    \mathbb{E}[X_{ij}^2] &= k\theta = \frac{1}{d},\label{eq:loose-mean}\\
    \operatorname{Var}(X_{ij}^2) &= k\theta^2 = \frac{2}{d^2}.\label{eq:loose-var}
\end{align}

Using the i.i.d. Gaussian variables $X_{ij}$ introduced in (\ref{eq:Xij-N01d}), we define the corresponding $d\to\infty$ approximation to the Projection Kernel by
\begin{equation}
    Z^{(\infty)}
    := \sum_{i=1}^m\sum_{j=1}^m X_{ij}^2.
    \label{eq:ZX}
\end{equation}
Since $\{X_{ij}\}$ are i.i.d., the family $\{X_{ij}^2\}$ is also i.i.d. and satisfies $\operatorname{Var}(X_{ij}^2)<\infty$.
Hence, by the central limit theorem,
\begin{equation}
  \frac{Z^{(\infty)} - m^2\mathbb{E}[X_{ij}^2]}{\sqrt{m^2\,\operatorname{Var}(X_{ij}^2)}}\xrightarrow{\mathcal{D}}\mathcal{N}(0,1)\quad(m\to\infty).\label{eq:ZX-CLT}
\end{equation}
Substituting (\ref{eq:loose-mean}) and (\ref{eq:loose-var}) into (\ref{eq:ZX-CLT}), we obtain
\begin{equation}
  \frac{Z^{(\infty)} - \frac{m^2}{d}}{\sqrt{\frac{2m^2}{d^2}}}\xrightarrow{\mathcal{D}}\mathcal{N}(0,1)\quad(m\to\infty).
\end{equation}
Consequently, for large $m$, the distribution of $Z^{(\infty)}$ is well approximated by
\begin{equation}
\mathcal{N}\left(\frac{m^2}{d},\,\frac{2m^2}{d^2}\right).\label{eq:nd-loose}
\end{equation}
For $d=768$ and $m=64$, sampling\footnote{In the simulation, instead of using the identity-based $\bm{U}_A$ in (\ref{eq:UA}), we generated $\bm{U}_A$ and $\bm{U}_B$ by first sampling $d\times m$ matrices whose entries are i.i.d. standard normal $\mathcal{N}(0,1)$, and then taking their SVD to obtain orthonormal basis matrices.} $\bm{U}_A, \bm{U}_B\in \mathcal{V}_m(\mathbb{R}^d)$ and overlaying the density of $\mathcal{N}\left(m^2/d,\,2m^2/d^2\right)$ on the histogram of $Z$ shows a rough match (Fig.~\ref{fig:pk-hist}).
However, due to both the large-$d$ approximation in (\ref{eq:Rij-inf}) and the large-$m$ approximation in (\ref{eq:nd-loose}), the fit is weaker than for the approximations of $R_{ij}$ and $R_{ij}^2$ (Figs.~\ref{fig:entry-hist} and~\ref{fig:squared-entry-hist}).

\subsection{Tight approximation: deriving the mean and variance of PK and applying the central limit theorem}\label{app:tight-approx}
In Appendix~\ref{app:loose-approx}, we derived an approximate distribution of $Z$ from the approximate mean (\ref{eq:loose-mean}) and approximate variance (\ref{eq:loose-var}) under the independence assumption in (\ref{eq:Xij-N01d}).
In this section, we derive a tighter approximate distribution of $Z$ by computing its mean and variance from the distribution of $R_{ij}$.
To this end, we first formulate $\mathbb{E}[Z]$ and $\operatorname{Var}(Z)$, compute them, and then derive an approximate distribution based on the central limit theorem.

\subsubsection{Formulating the mean \texorpdfstring{$\mathbb{E}[Z]$}{EZ} and variance \texorpdfstring{$\operatorname{Var}(Z)$}{VarZ} of \texorpdfstring{$Z$}{Z}}
From (\ref{eq:Z-PK}), the mean of $Z$ is
\begin{align}
\mathbb{E}[Z] &= \sum_{i=1}^m \sum_{j=1}^m \mathbb{E}\left[R_{ij}^2\right],\label{eq:EZ}
\end{align}
and the variance of $Z$ is
\begin{align}
\operatorname{Var}(Z) &=  \mathbb{E}[(Z-\mathbb{E}[Z])^2] = \mathbb{E}[Z^2] - \mathbb{E}[Z]^2.\label{eq:VarZ}
\end{align}
Thus, to compute $\operatorname{Var}(Z)$, it suffices to compute $\mathbb{E}[Z]$ and $\mathbb{E}[Z^2]$.

Expanding $Z^2$ yields
\begin{align}
    Z^2 = \sum_{i=1}^m \sum_{j=1}^m \sum_{i'=1}^m \sum_{j'=1}^m R_{ij}^2 R_{i'j'}^2,
\end{align}
so
\begin{align}
\mathbb{E}[Z^2] = \sum_{i=1}^m \sum_{j=1}^m \sum_{i'=1}^m \sum_{j'=1}^m \mathbb{E}[R_{ij}^2 R_{i'j'}^2].\label{eq:EZ2}
\end{align}

\subsubsection{Computing the mean \texorpdfstring{$\mathbb{E}[Z]$}{EZ}}
\begin{figure}[p!]
\centering
\begin{subfigure}{\columnwidth}
    \centering
    \includegraphics[width=0.75\textwidth]{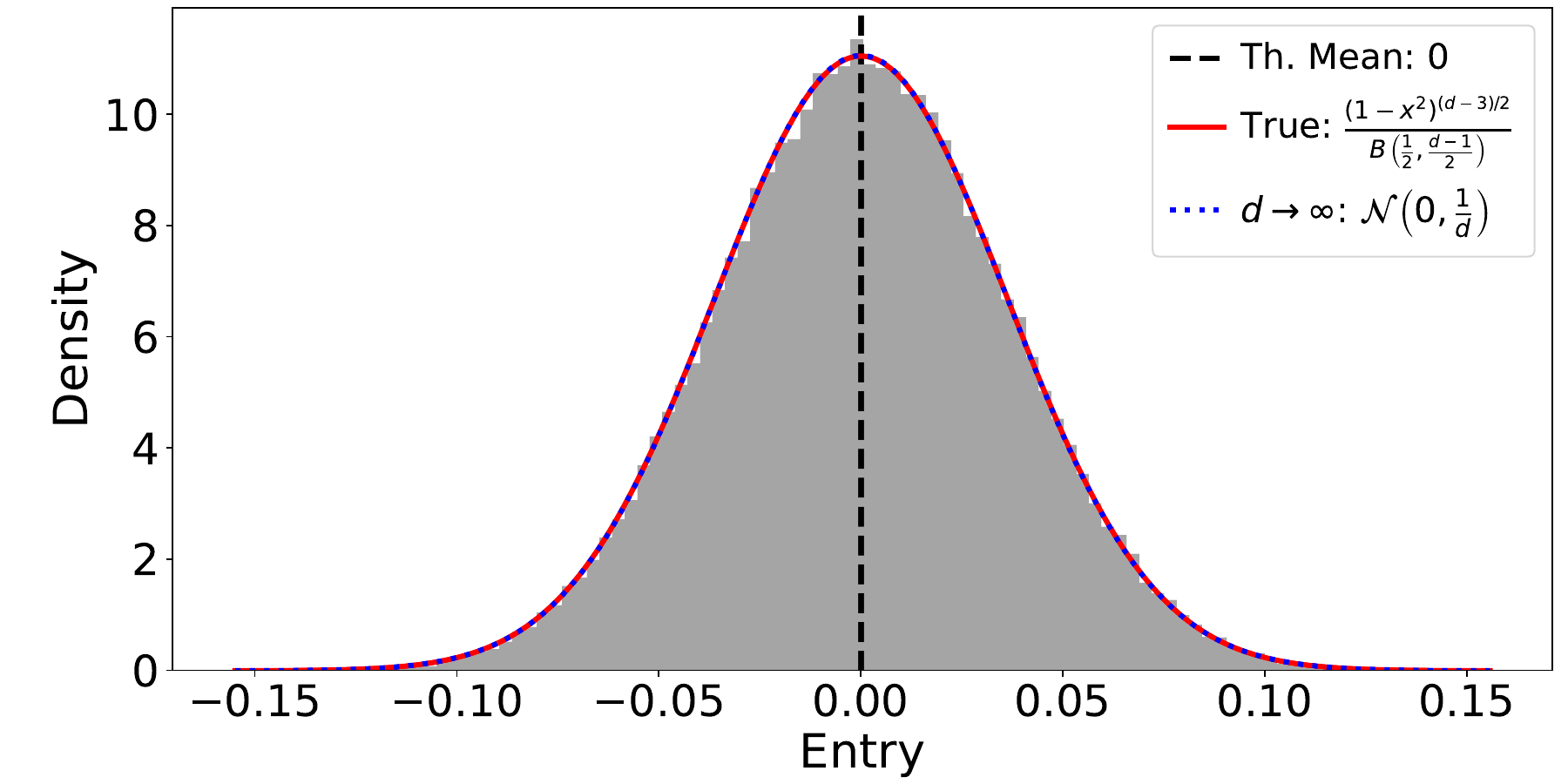}
    \subcaption{Entry $R_{ij}$}
    \label{fig:entry-hist}
\end{subfigure}
\par\bigskip
\begin{subfigure}{\columnwidth}
    \centering
    \includegraphics[width=0.75\textwidth]{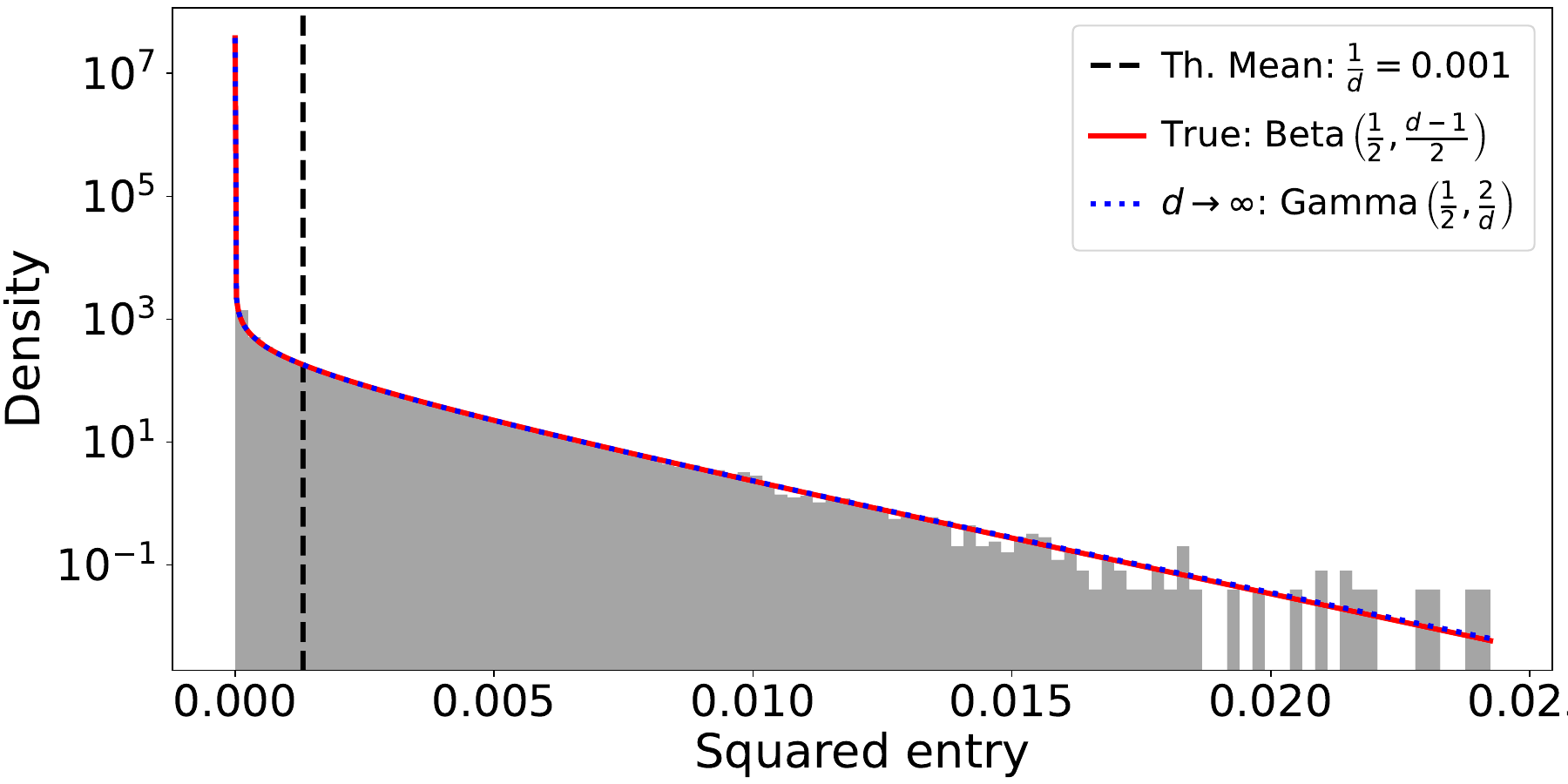}
    \subcaption{Squared entry $R_{ij}^2$}
    \label{fig:squared-entry-hist}
\end{subfigure}
\par\bigskip
\begin{subfigure}{\columnwidth}
    \centering
    \includegraphics[width=0.75\textwidth]{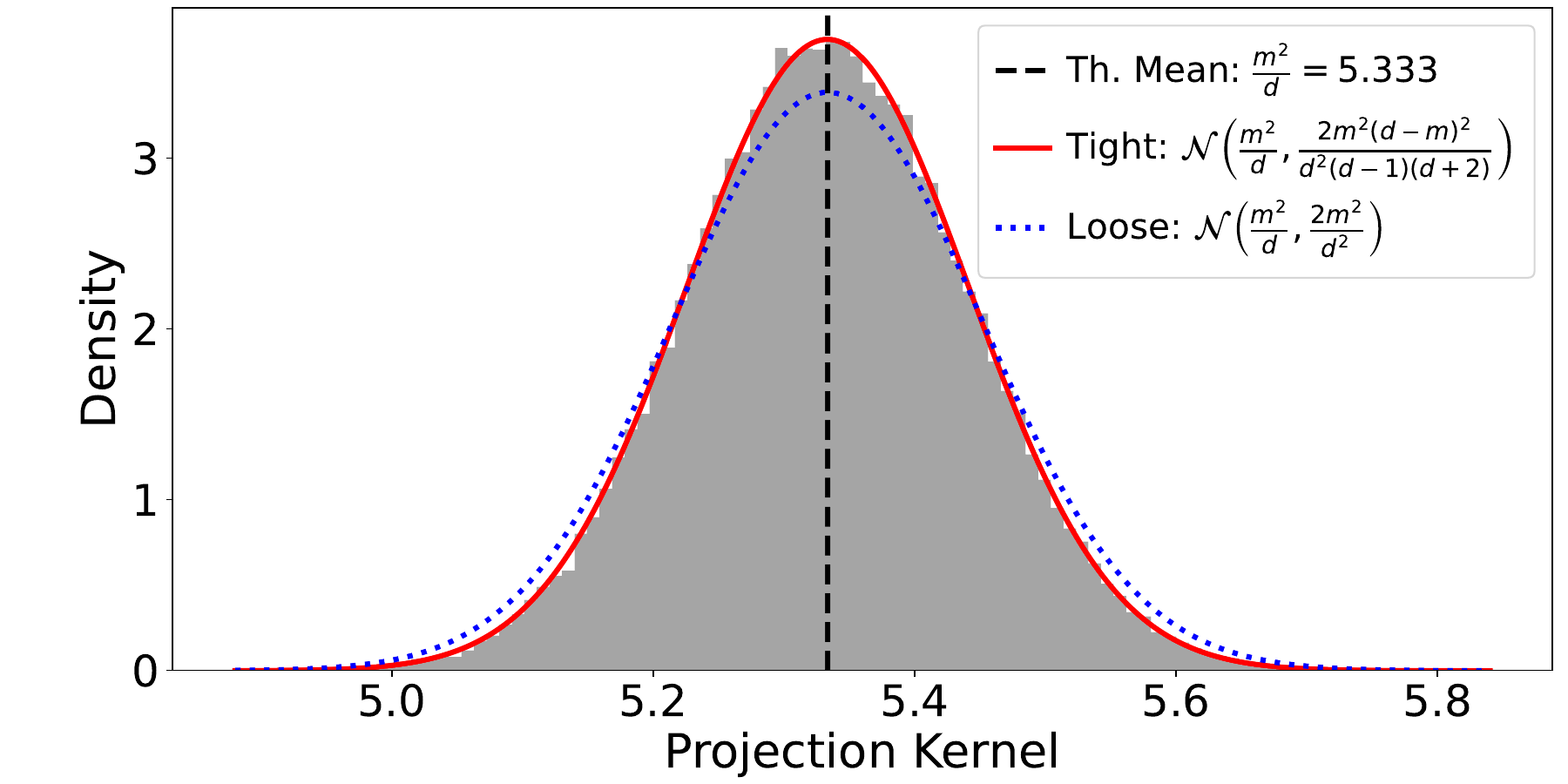}
    \subcaption{Projection Kernel $Z$}
    \label{fig:pk-hist}
\end{subfigure}
\caption{
For random orthogonal matrices in $\mathcal{V}_m(\mathbb{R}^d)$, we show (a) a histogram of entry values, (b) squared entry values, and (c) PK values between pairs of random orthogonal matrices.
We use $d=768$ and $m=64$, following GPT2-small; (a) and (b) are based on $100{,}000$ randomly sampled entries, and (c) on $100{,}000$ randomly sampled pairs.
For (a) and (b), we also plot the corresponding density and its asymptotic approximation as $d\to\infty$.
For (c), we plot two approximations to the PK distribution: the tight approximation in Appendix~\ref{app:tight-approx} and the loose approximation in Appendix~\ref{app:loose-approx}.
The two approximations are nearly indistinguishable for (a) and (b), while for (c) the tight approximation matches the histogram better, especially near the mean.
}
\label{fig:e-se-pk-hist}
\end{figure}

The density and mean of $R_{ij}$ are given by~\citet[Appendix E]{preisendorfer1982data}:
\begin{align}
    f_{R_{ij}}(x)
    &= \frac{1}{B\left(\frac{1}{2},\frac{d-1}{2}\right)}
       (1-x^2)^{\frac{d-3}{2}}
       \quad x\in[-1,1], \label{eq:Oij-density-func}\\
    \mathbb{E}[R_{ij}] &= 0,
\end{align}
where $B$ is the Beta function.
The density in (\ref{eq:Oij-density-func}) closely matches the histogram of $R_{ij}$ in Fig.~\ref{fig:entry-hist}.

The distribution and mean of $R_{ij}^2$ are given by~\citet[Appendix E]{preisendorfer1982data}:
\begin{align}
    &R_{ij}^2\sim\text{Beta}\left(\frac{1}{2},\frac{d-1}{2}\right),\\
    &\mathbb{E}[R_{ij}^2] = \frac{1}{d}.\label{eq:ERij2}
\end{align}
The density of $\text{Beta}\left(1/2,(d-1)/2\right)$ also closely matches the histogram of $R_{ij}^2$ in Fig.~\ref{fig:squared-entry-hist}.
From (\ref{eq:ERij2}), we have $\operatorname{Var}(R_{ij}) = \mathbb{E}[R_{ij}^2] - \mathbb{E}[R_{ij}]^2 = 1/d$.

Using (\ref{eq:ERij2}) in (\ref{eq:EZ}), we obtain\footnote{
The expected Projection Kernel between subspaces spanned by random orthogonal matrices, $m^2/d$, can also be understood naturally through the following discrete analogy.
Uniformly at random, choose $m$ of $d$ boxes and place a red ball in each selected box, and independently repeat the procedure with blue balls.
Each box is selected in both trials with probability $(m/d)^2$, so the expected number of boxes containing both a red and a blue ball is $m^2/d$.
If each box is identified with a vector in a fixed orthonormal basis, selecting boxes corresponds to choosing basis vectors to form subspaces (see Section~\ref{sec:toy-example-of-subspaces}).
}
\begin{equation}
\mathbb{E}[Z]
= \sum_{i=1}^m \sum_{j=1}^m \mathbb{E}\left[R_{ij}^2\right]
= m^2\mathbb{E}\left[R_{ij}^2\right]
= \frac{m^2}{d}.\label{eq:EZ-value}
\end{equation}
This matches the mean of the loose approximation in (\ref{eq:nd-loose}) for $Z^{(\infty)}$ in (\ref{eq:ZX}).
Moreover, when $m=d$, we have $\bm{U}_A,\bm{U}_B\in\mathcal{V}_m(\mathbb{R}^d)=\mathcal{O}(\mathbb{R}^d)$, so both are full-rank orthogonal matrices. In this case, the corresponding subspaces satisfy $\mathcal{S}_A=\mathcal{S}_B=\mathbb{R}^d$, and hence $Z=\operatorname{PK}(\mathcal{S}_A,\mathcal{S}_B)=d$. Consistently, $\,m^2/d=d^2/d=d$.

\subsubsection{Computing \texorpdfstring{$\mathbb{E}[Z^2]$}{EZ2} and \texorpdfstring{$\operatorname{Var}(Z)$}{VarZ}}
Since $\bm{R}\in \mathcal{O}(\mathbb{R}^d)$ is a random orthogonal matrix, the entries $R_{ij}$ are dependent within each row and column.
To compute the summand $\mathbb{E}[R_{ij}^2 R_{i'j'}^2]$ in (\ref{eq:EZ2}), we consider the following four cases\footnote{
These four cases exhaust all $m^4$ pairs $((i,j),(i',j'))$:
$m^2 + 2m^2(m-1) + m^2(m-1)^2 = m^4$.
}:
\begin{enumerate}[label=(\roman*)]
    \item If $i' = i$ and $j'=j$, then $\mathbb{E}[R_{ij}^2 R_{i'j'}^2]=\mathbb{E}[R_{ij}^4]$, and there are $m^2$ such terms.
    \item If $i'=i$ and $j'\neq j$, then $\mathbb{E}[R_{ij}^2 R_{i'j'}^2]=\mathbb{E}[R_{ij}^2 R_{ij'}^2]$, and there are $m^2(m-1)$ such terms.
    \item If $i'\neq i$ and $j' = j$, then $\mathbb{E}[R_{ij}^2 R_{i'j'}^2]=\mathbb{E}[R_{ij}^2 R_{i'j}^2]$, and there are $m^2(m-1)$ such terms.
    \item If $i'\neq i$ and $j' \neq j$, then $\mathbb{E}[R_{ij}^2 R_{i'j'}^2]$ appears, and there are $m^2(m-1)^2$ such terms.
\end{enumerate}
Therefore, we can write
\begin{align}
    \mathbb{E}[Z^2] = m^2 \mathbb{E}[R_{ij}^4] +m^2(m-1) \mathbb{E}[R_{ij}^2 R_{ij'}^2]
    + m^2(m-1)\mathbb{E}[R_{ij}^2 R_{i'j}^2]
    +m^2(m-1)^2 \mathbb{E}[R_{ij}^2 R_{i'j'}^2].\label{eq:EZ2-2}
\end{align}

For entries $R_{ij}, R_{rs}, R_{\alpha\beta}, R_{\lambda\mu}\in\mathbb{R}$ of a random orthogonal matrix $\bm{R}\in \mathcal{O}(\mathbb{R}^d)$, letting $\delta_{ij}$ be the Kronecker delta, the fourth moment
$\mathbb{E}[R_{ij}R_{rs}R_{\alpha\beta}R_{\lambda\mu}]$ is given by~\citet[Lemma 2.2]{meckes2019random}:
\begin{align}
\mathbb{E}[R_{ij}R_{rs}R_{\alpha\beta}R_{\lambda\mu}]
=&\,-\frac{1}{(d-1)d(d+2)}(\delta_{ir}\delta_{\alpha\lambda}\delta_{j\beta}\delta_{s\mu}+\delta_{ir}\delta_{\alpha\lambda}\delta_{j\mu}\delta_{s\beta}+\delta_{i\alpha}\delta_{r\lambda}\delta_{js}\delta_{\beta\mu}\notag\\
&\quad\quad+\delta_{i\alpha}\delta_{r\lambda}\delta_{j\mu}\delta_{\beta s}+\delta_{i\lambda}\delta_{r\alpha}\delta_{js}\delta_{\beta\mu}
+\delta_{i\lambda}\delta_{r\alpha}\delta_{j\beta}\delta_{s \mu})\notag\\
&\quad+\frac{d+1}{(d-1)d(d+2)}(\delta_{ir}\delta_{\alpha\lambda}\delta_{js}\delta_{\beta\mu}+\delta_{i\alpha}\delta_{r\lambda}\delta_{j\beta}\delta_{s\mu}+\delta_{i\lambda}\delta_{r\alpha}\delta_{j\mu}\delta_{s\beta}).\label{eq:Rijrs}
\end{align}
Setting $r=i$, $s=j$, $\alpha=\lambda=i'$, and $\beta=\mu=j'$ in (\ref{eq:Rijrs}) yields
\begin{align}
\mathbb{E}[R_{ij}^2 R_{i'j'}^2]=-\frac{2}{(d-1)d(d+2)}(\delta_{jj'}+\delta_{ii'}+\delta_{ii'}\delta_{jj'})
+\frac{d+1}{(d-1)d(d+2)}(1+2\delta_{ii'}\delta_{jj'}).\label{eq:Rijij}
\end{align}
Here we used $\delta_{\alpha\beta}^2 = \delta_{\alpha\beta}$ and $\delta_{\alpha\beta}=\delta_{\beta\alpha}$.

Setting $i'=i$ and $j'=j$ in (\ref{eq:Rijij}), we obtain
\begin{align}
\mathbb{E}[R_{ij}^4] =\frac{3}{d(d+2)}.\label{eq:ijij}
\end{align}
Similarly, setting $i'=i$ and $j\neq j'$ yields
\begin{align}
\mathbb{E}[R_{ij}^2 R_{ij'}^2]=\frac{1}{d(d+2)}.\label{eq:ijij-}
\end{align}
Setting $i'\neq i$ and $j'=j$ yields
\begin{align}
\mathbb{E}[R_{ij}^2 R_{i'j}^2]
=\frac{1}{d(d+2)}.\label{eq:iji-j}
\end{align}
Finally, setting $i'\neq i$ and $j'\neq j$ yields
\begin{align}
\mathbb{E}[R_{ij}^2 R_{i'j'}^2]
=\frac{d+1}{(d-1)d(d+2)}.\label{eq:iji-j-}
\end{align}
Note that (\ref{eq:ijij-}) and (\ref{eq:iji-j}) imply\footnote{This also follows from symmetry.} $\mathbb{E}[R_{ij}^2 R_{ij'}^2] = \mathbb{E}[R_{ij}^2 R_{i'j}^2]$.

Substituting (\ref{eq:ijij}), (\ref{eq:ijij-}), (\ref{eq:iji-j}), and (\ref{eq:iji-j-}) into (\ref{eq:EZ2-2}), we obtain
\begin{align}
\mathbb{E}[Z^2]
=\frac{m^2}{(d-1)d(d+2)}\left\{(m^2+2)d +m(m-4)\right\}.\label{eq:EZ2-3}
\end{align}
Using (\ref{eq:EZ-value}) and (\ref{eq:EZ2-3}), we compute $\operatorname{Var}(Z)$ in (\ref{eq:VarZ}) as
\begin{align}
\operatorname{Var}(Z) = \frac{2m^2(d-m)^2}{d^2(d-1)(d+2)}.\label{eq:VarZ-value}
\end{align}
In particular, when $m \ll d$, we have $(d-m)^2 \approx d^2\approx (d-1)(d+2)$ and thus $\operatorname{Var}(Z) \approx 2m^2/d^2$, which matches the variance of the loose approximation in (\ref{eq:nd-loose}) for $Z^{(\infty)}$ in (\ref{eq:ZX}).
Moreover, when $m=d$, we have seen that $Z=\operatorname{PK}(\mathcal{S}_A,\mathcal{S}_B)=d$ for any $\bm{U}_A,\bm{U}_B\in\mathcal{V}_m(\mathbb{R}^d)=\mathcal{O}(\mathbb{R}^d)$, and hence $\operatorname{Var}(Z)=0$. 
This also follows from (\ref{eq:VarZ-value}), since substituting $m=d$ yields $\operatorname{Var}(Z)=0$.

\subsubsection{Approximating the distribution of \texorpdfstring{$Z$}{Z} using the central limit theorem}
Although $Z$ can be written as a sum of $m^2$ random variables $R_{ij}^2$, these variables are not independent due to the orthogonality constraints on $R_{ij}$.
Therefore, unlike Appendix~\ref{app:loose-approx}, we cannot neither treat $Z$ as a sum of independent variables nor directly apply the central limit theorem.

To assess the strength of these dependencies, we compare the variance and covariances of $R_{ij}^2$.
From (\ref{eq:ERij2}) and (\ref{eq:ijij}), the variance is
\begin{align}
\operatorname{Var}(R_{ij}^2) =\mathbb{E}[R_{ij}^4] - \mathbb{E}[R_{ij}^2]^2
=\frac{2(d-1)}{d^2(d+2)} = O\left(\frac{1}{d^2}\right).\label{eq:Varijij}
\end{align}
For $j\neq j'$, from (\ref{eq:ERij2}) and (\ref{eq:ijij-}), the covariance is
\begin{align}
\operatorname{Cov}(R_{ij}^2,R_{ij'}^2) =\mathbb{E}[R_{ij}^2R_{ij'}^2] - \mathbb{E}[R_{ij}^2]\mathbb{E}[R_{ij'}^2]
=-\frac{2}{d^2(d+2)} = O\left(\frac{1}{d^3}\right).\label{eq:Covijij-}
\end{align}
The same order holds for $\operatorname{Cov}(R_{ij}^2,R_{i'j}^2)$ when $i\neq i'$.
For $i\neq i'$ and $j\neq j'$, from (\ref{eq:ERij2}) and (\ref{eq:iji-j-}), we have
\begin{align}
\operatorname{Cov}(R_{ij}^2,R_{i'j'}^2) =&\,\mathbb{E}[R_{ij}^2R_{i'j'}^2] - \mathbb{E}[R_{ij}^2]\mathbb{E}[R_{i'j'}^2]
=\frac{2}{d^2(d-1)(d+2)} = O\left(\frac{1}{d^4}\right).\label{eq:Coviji-j-}
\end{align}
Thus, the covariances in (\ref{eq:Covijij-}) and (\ref{eq:Coviji-j-}) are at least a factor of $1/d$ smaller than the variance in (\ref{eq:Varijij}).

These estimates suggest that $Z$ behaves like a sum of $m^2$ nearly independent random variables $R_{ij}^2$.
We therefore approximate the distribution of $Z$ by a Gaussian distribution\footnote{
We do not verify conditions required for a rigorous application of the central limit theorem, such as independence or specific dependence structures, and instead adopt a Gaussian approximation as a practical model for the distribution of $Z$.}.
Using the mean in (\ref{eq:EZ-value}) and the variance in (\ref{eq:VarZ-value}), we adopt
\begin{align}
  \mathcal{N}\left(\mathbb{E}[Z],\,\operatorname{Var}(Z)\right)
  =\mathcal{N}\left(\frac{m^2}{d},
     \frac{2m^2(d-m)^2}{d^2(d-1)(d+2)}\right)\label{eq:nd-tight}
\end{align}
as a tight approximate distribution of $Z$.
Overlaying the density of (\ref{eq:nd-tight}) on the histogram of $Z$ in Fig.~\ref{fig:pk-hist} shows a better fit than the loose approximation in Appendix~\ref{app:loose-approx}.

\section{Details of experiments with BLOOM-560M and Qwen2-0.5B-Instruct}\label{app:bloom-qwen2}
In Section~\ref{sec:other-models}, we used BLOOM-560M and Qwen2-0.5B-Instruct to conduct the same evaluation as in Section~\ref{sec:eval-metric}.
This section provides further details of these experiments.

\begin{table*}[t!]
    \centering
    \small
    \begin{tabular}{lccc}
        \toprule
        & GPT2-small & BLOOM-560M & Qwen2-0.5B-Instruct \\
        \midrule
        \# parameters
            & 124M & 560M & 494M \\
        \# layers
            & 12 & 24 & 24 \\
        \# query heads
            & 12 & 16 & 14 \\
        \# KV heads
            & 12 & 16 & 2 \\
        Attention architecture
            & MHA & MHA & GQA \\
        $d$
            & 768 & 1024 & 896 \\
        $d_{\text{head}}$
            & 64 & 64 & 64 \\
        FFN architecture
            & GELU MLP & GELU MLP & SwiGLU \\
        $d_{\text{FFN}}$
            & 3072 & 4096 & 4864 \\
        Normalization
            & LayerNorm & LayerNorm & RMSNorm \\
        Positional encoding
            & learned absolute embeddings & ALiBi & RoPE \\
        \bottomrule
    \end{tabular}
\caption{
Architectural configurations of the models used in our experiments.
In particular, GPT2-small and BLOOM-560M use multi-head attention (MHA)~\cite{NIPS2017_3f5ee243}, whereas Qwen2-0.5B-Instruct uses grouped-query attention (GQA)~\cite{ainslie-etal-2023-gqa} and thus has fewer KV heads than query heads.
}
    \label{tab:model_configurations}
\end{table*}

\subsection{Differences in model architecture}\label{app:model-comparison}

\begin{table*}[p!]
\centering

\begin{subtable}{.245\textwidth}
  \centering
 \begin{tabular}{lr}
  \toprule
  Head & Score \\
  \midrule
L7H8 & 0.923 \\
L0H5 & 0.867 \\
L21H0 & 0.810 \\
L6H3 & 0.638 \\
L23H1 & 0.620 \\
L20H0 & 0.577 \\
L22H4 & 0.524 \\
L1H3 & 0.511 \\
L12H0 & 0.496 \\
L5H2 & 0.490 \\
\bottomrule
  \end{tabular}
\caption{Identity}
\label{tab:bloom_560m_identity}
\end{subtable}
\begin{subtable}{.245\textwidth}
  \centering
 \begin{tabular}{lr}
  \toprule
  Head & Score \\
  \midrule
L6H14 & 0.323 \\
L11H10 & 0.209 \\
L1H14 & 0.187 \\
L3H12 & 0.161 \\
L11H9 & 0.054 \\
L13H13 & 0.051 \\
L11H11 & 0.046 \\
L4H15 & 0.043 \\
L0H11 & 0.034 \\
L6H13 & 0.034 \\
\bottomrule
  \end{tabular}
\caption{Duplicate Token}
\label{tab:bloom_560m_duplicate}
\end{subtable}
\begin{subtable}{.245\textwidth}
  \centering
 \begin{tabular}{lr}
  \toprule
  Head & Score \\
  \midrule
L6H0 & 0.996 \\
L6H2 & 0.995 \\
L7H0 & 0.982 \\
L5H0 & 0.970 \\
L9H3 & 0.888 \\
L11H0 & 0.875 \\
L4H0 & 0.731 \\
L8H0 & 0.726 \\
L17H0 & 0.696 \\
L11H5 & 0.544 \\
\bottomrule
  \end{tabular}
\caption{Previous Token}
\label{tab:bloom_560m_previous}
\end{subtable}
\begin{subtable}{.245\textwidth}
  \centering
 \begin{tabular}{lr}
  \toprule
  Head & Score \\
  \midrule
L10H10 & 0.972 \\
L8H13 & 0.963 \\
L7H12 & 0.957 \\
L8H11 & 0.955 \\
L10H12 & 0.949 \\
L10H13 & 0.936 \\
L9H14 & 0.878 \\
L13H12 & 0.873 \\
L18H13 & 0.864 \\
L8H14 & 0.858 \\
\bottomrule
  \end{tabular}
\caption{Induction}
\label{tab:bloom_560m_induction}
\end{subtable}
\caption{
For Identity, Duplicate Token, Previous Token, and Induction Heads in BLOOM-560M, we list the top 10 heads and their head scores.
See Appendix~\ref{app:head-scores} for how we compute these scores.
}
\label{tab:bloom_560m_head_scores_top10}
\end{table*}

\begin{table*}[p!]
\centering

\begin{subtable}{.245\textwidth}
  \centering
 \begin{tabular}{lr}
  \toprule
  Head & Score \\
  \midrule
L10H7 & 1.000 \\
L23H0 & 0.998 \\
L16H11 & 0.981 \\
L8H6 & 0.958 \\
L17H10 & 0.935 \\
L17H5 & 0.905 \\
L12H0 & 0.892 \\
L10H0 & 0.863 \\
L6H3 & 0.830 \\
L14H1 & 0.802 \\
\bottomrule
  \end{tabular}
\caption{Identity}
\label{tab:qwen2_0_5b_instruct_identity}
\end{subtable}
\begin{subtable}{.245\textwidth}
  \centering
 \begin{tabular}{lr}
  \toprule
  Head & Score \\
  \midrule
L7H7 & 0.893 \\
L5H1 & 0.863 \\
L1H1 & 0.837 \\
L5H0 & 0.782 \\
L12H2 & 0.617 \\
L9H0 & 0.282 \\
L4H9 & 0.277 \\
L4H2 & 0.220 \\
L5H6 & 0.138 \\
L7H8 & 0.104 \\
\bottomrule
  \end{tabular}
\caption{Duplicate Token}
\label{tab:qwen2_0_5b_instruct_duplicate}
\end{subtable}
\begin{subtable}{.245\textwidth}
  \centering
 \begin{tabular}{lr}
  \toprule
  Head & Score \\
  \midrule
L8H10 & 0.906 \\
L0H5 & 0.786 \\
L10H13 & 0.773 \\
L15H4 & 0.724 \\
L1H5 & 0.717 \\
L5H5 & 0.665 \\
L20H9 & 0.604 \\
L4H0 & 0.545 \\
L2H3 & 0.542 \\
L0H7 & 0.524 \\
\bottomrule
  \end{tabular}
\caption{Previous Token}
\label{tab:qwen2_0_5b_instruct_previous}
\end{subtable}
\begin{subtable}{.245\textwidth}
  \centering
 \begin{tabular}{lr}
  \toprule
  Head & Score \\
  \midrule
L11H11 & 0.988 \\
L9H12 & 0.940 \\
L11H13 & 0.924 \\
L16H1 & 0.912 \\
L11H12 & 0.853 \\
L9H8 & 0.621 \\
L11H9 & 0.597 \\
L16H6 & 0.566 \\
L20H7 & 0.547 \\
L9H9 & 0.497 \\
\bottomrule
  \end{tabular}
\caption{Induction}
\label{tab:qwen2_0_5b_instruct_induction}
\end{subtable}
\caption{
For Identity, Duplicate Token, Previous Token, and Induction Heads in Qwen2-0.5B-Instruct, we list the top 10 heads and their head scores.
See Appendix~\ref{app:head-scores} for how we compute these scores.
}
\label{tab:qwen2_0_5b_instruct_head_scores_top10}
\end{table*}

\begin{table*}[p!]
\scriptsize
\centering
\begin{tabular}{lrl}
\toprule
Head Classes & Count & Heads \\
\midrule
Duplicate Token Heads & 4 & L1H14$^*$, L3H12$^*$, L6H14$^*$, L11H10 \\
Previous Token Heads & 10 & L4H0, L5H0, L6H0, L6H2, L7H0, L8H0$^*$, L9H3$^*$, L11H0$^*$, L11H5$^*$, L17H0$^*$ \\
Induction Heads & 10 & L7H12, L8H11, L8H13, L8H14$^*$, L9H14$^*$, L10H10, L10H12$^*$, L10H13$^*$, L13H12$^*$, L18H13$^*$ \\
Name Mover Heads & 11 & L18H7, L18H8, L19H8, L19H11, L20H8, L20H9, L20H10, L20H11, L20H12, L21H9, L21H10 \\
Negative Name Mover Heads & 3 & L17H9, L19H15, L21H6 \\
S-Inhibition Heads & 6 & L12H6, L14H6, L16H5, L19H4, L21H2, L22H3 \\
\bottomrule
\end{tabular}
\caption{
Head classes and their corresponding heads in our experimental setup for BLOOM-560M.
Heads that are not included in \citet[Fig.~2]{zhang2025the} are marked with ``$*$''.
The total number of annotated heads is 44.
}
\label{tab:bloom_560m_head_class}
\end{table*}

\begin{table*}[p!]
\scriptsize
\centering
\begin{tabular}{lrl}
\toprule
Head Classes & Count & Heads \\
\midrule
Duplicate Token Heads & 5 & L1H1$^*$, L5H0, L5H1$^*$, L7H7, L12H2$^*$ \\
Previous Token Heads & 10 & L0H5$^*$, L0H7$^*$, L1H5, L2H3$^*$, L4H0, L5H5, L8H10$^*$, L10H13$^*$, L15H4$^*$, L20H9$^*$ \\
Induction Heads & 9 & L9H8$^*$, L9H9, L9H12, L11H9$^*$, L11H11, L11H12$^*$, L11H13, L16H1$^*$, L16H6$^*$ \\
Name Mover Heads & 6 & L16H0, L20H11, L21H5, L21H6, L21H11, L21H12 \\
Negative Name Mover Heads & 3 & L20H7, L22H6, L23H12 \\
S-Inhibition Heads & 8 & L13H8, L14H8, L14H10, L15H7, L16H5, L17H6, L21H0, L22H8 \\
\bottomrule
\end{tabular}
\caption{
Head classes and their corresponding heads in our experimental setup for Qwen2-0.5B-Instruct.
Heads that are not included in \citet[Fig.~9]{zhang2025the} are marked with ``$*$''.
The total number of annotated heads is 41.
}
\label{tab:qwen2_0_5b_instruct_head_class}
\end{table*}

Table~\ref{tab:model_configurations} summarizes the architectural differences among GPT2-small~\cite{radford2019language}, which is the primary model used in this work, and BLOOM-560M~\cite{workshop2023bloom176bparameteropenaccessmultilingual} and Qwen2-0.5B-Instruct~\cite{yang2024qwen2technicalreport}, which are used in Section~\ref{sec:other-models}.

GPT2-small and BLOOM-560M share many features of classical Transformer architectures, including multi-head attention (MHA)~\cite{NIPS2017_3f5ee243}, a GELU~\cite{hendrycks2023gaussianerrorlinearunits} MLP as the FFN, and layer normalization~\cite{ba2016layernormalization}.
However, they differ in positional encoding: GPT2-small uses learned absolute embeddings, whereas BLOOM-560M uses ALiBi~\cite{press2022train}.
Qwen2-0.5B-Instruct, on the other hand, has a more modern architecture based on grouped-query attention (GQA)~\cite{ainslie-etal-2023-gqa}, with SwiGLU~\cite{shazeer2020gluvariantsimprovetransformer} as its FFN and root mean square layer normalization (RMSNorm)~\cite{NEURIPS2019_1e8a1942}.
It further uses RoPE~\cite{10.1016/j.neucom.2023.127063} for positional encoding.

\begin{figure*}[t!]
    \centering
    \includegraphics[width=\linewidth]{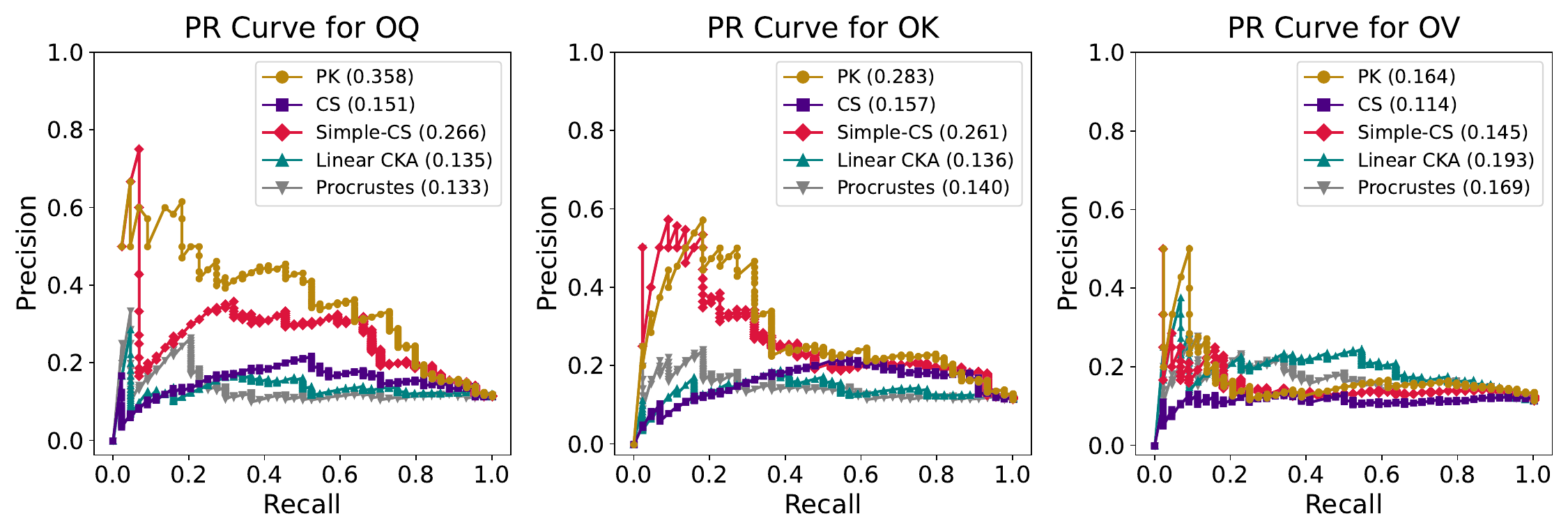}
    \caption{PR curves of head detection for OQ, OK, and OV in BLOOM-560M, shown for each method.}
    \label{fig:bloom-560m_head_detection_OQ_OK_OV_pr_curve}
\end{figure*}

\begin{figure*}[t!]
    \centering
    \includegraphics[width=\linewidth]{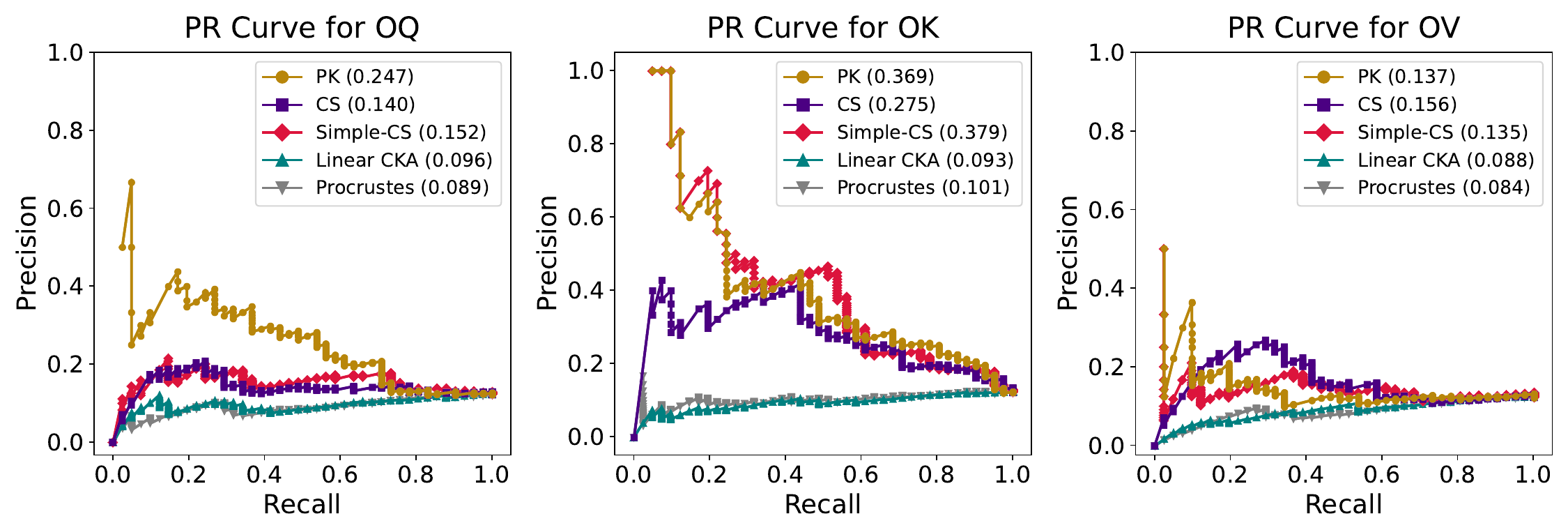}
    \caption{PR curves of head detection for OQ, OK, and OV in Qwen2-0.5B-Instruct, shown for each method.}
    \label{fig:qwen2-0.5b-instruct_head_detection_OQ_OK_OV_pr_curve}
\end{figure*}

A particularly important difference lies in the attention architecture.
While GPT2-small and BLOOM-560M use MHA, Qwen2-0.5B-Instruct uses GQA and has only two key and value heads for 14 query heads in each layer.
For Qwen2-0.5B-Instruct, we analyzed the GQA architecture within the same framework as MHA by simply repeating the shared key and value head weights.

\subsection{Details of head scores and head class annotations}\label{app:bloom-qwen2-head-class}

In this section, as in Appendix~\ref{app:head-scores}, we assign heads in BLOOM-560M and Qwen2-0.5B-Instruct to head classes based on their head scores.
Tables~\ref{tab:bloom_560m_head_scores_top10} and~\ref{tab:qwen2_0_5b_instruct_head_scores_top10} show the top 10 head scores and the corresponding heads for Identity, Duplicate Token, Previous Token, and Induction Heads in BLOOM-560M and Qwen2-0.5B-Instruct, respectively.
As with GPT2-small (Table~\ref{tab:table2_4columns} in Appendix~\ref{app:head-scores-details}), multiple heads achieve high head scores for each head class.
Based on these results and the head class annotations of \citet{zhang2025the}, we define the head classes shown in Tables~\ref{tab:bloom_560m_head_class} and~\ref{tab:qwen2_0_5b_instruct_head_class}.

\subsection{Details of the evaluation of each method}\label{app:bloom-qwen2-eval-details}
This section provides further details of Tables~\ref{tab:PRAUC-OQ-OK-OV-two-models} and~\ref{tab:classification-summary-by-model} in Section~\ref{sec:other-models}.

\begin{table*}[t!]
\small
\centering

\begin{subtable}{\textwidth}
\centering
\begin{tabular}{rlrlrlr}
\toprule
Top & \multicolumn{2}{c}{OQ} & \multicolumn{2}{c}{OK} & \multicolumn{2}{c}{OV}\\
\midrule
1 & L17H9 (N-NM) & 340.46 & L6H12 & 372.52 & L21H5 & 262.06 \\
2 & L6H2 (Prev) & 198.95 & L17H6 & 264.11 & L17H8 & 188.13 \\
3 & L19H10 & 182.18 & L11H5 (Prev) & 234.30 & L5H2 (Id) & 180.28 \\
4 & L19H15 (N-NM) & 179.84 & L5H9 & 180.51 & L18H6 & 151.28 \\
5 & L5H0 (Prev) & 144.04 & L13H15 & 167.87 & L11H4 & 131.25 \\
6 & L22H14 & 109.76 & L7H15 & 137.67 & L6H3 (Id) & 111.10 \\
7 & L18H12 & 105.52 & L23H12 & 107.98 & L10H3 & 107.47 \\
8 & L10H9 & 96.50 & L8H13 (Ind) & 103.30 & L7H3 & 104.10 \\
9 & L10H13 (Ind) & 89.25 & L15H13 & 99.56 & L20H1 & 103.87 \\
10 & L10H8 & 86.51 & L22H9 & 97.46 & L9H4 & 98.08 \\
\bottomrule
\end{tabular}
\caption{Inlet Head Scores}
\label{tab:bloom_560m_inlet_appendix}
\end{subtable}

\par\bigskip

\begin{subtable}{\textwidth}
\centering
\begin{tabular}{rlrlrlr}
\toprule
Top & \multicolumn{2}{c}{OQ} & \multicolumn{2}{c}{OK} & \multicolumn{2}{c}{OV}\\
\midrule
1 & L6H13 & 323.58 & L6H3 (Id) & 627.11 & L0H3 & 248.62 \\
2 & L12H2 & 263.44 & L17H0 (Prev) & 412.18 & L13H13 & 144.67 \\
3 & L17H8 & 206.67 & L0H5 (Id) & 180.32 & L6H13 & 120.39 \\
4 & L15H10 & 189.91 & L13H13 & 128.17 & L11H9 & 114.29 \\
5 & L5H2 (Id) & 140.45 & L9H0 & 106.13 & L6H3 (Id) & 110.92 \\
6 & L15H13 & 131.03 & L11H4 & 100.69 & L11H11 & 108.36 \\
7 & L17H9 (N-NM) & 120.90 & L1H13 & 92.80 & L11H10 (Dup) & 107.61 \\
8 & L11H4 & 117.98 & L6H14 (Dup) & 84.50 & L1H3 (Id) & 97.84 \\
9 & L9H4 & 106.55 & L6H13 & 82.84 & L3H9 & 89.75 \\
10 & L2H1 & 104.56 & L11H10 (Dup) & 81.41 & L6H14 (Dup) & 85.67 \\
\bottomrule
\end{tabular}
\caption{Outlet Head Scores}
\label{tab:bloom_560m_outlet_appendix}
\end{subtable}
\caption{
For OQ, OK, and OV in BLOOM-560M, we report (a) the top 10 heads and their inlet scores in (\ref{eq:inlet}) and (b) the top 10 heads and their outlet scores in (\ref{eq:outlet}).
When a head belongs to the head classes in Table~\ref{tab:bloom_560m_head_class} or to the top 10 heads by Identity Head score in Table~\ref{tab:bloom_560m_identity}, we also report this information.
Dup, Prev, Ind, NM, N-NM, and S-Inh denote Duplicate Token, Previous Token, Induction, Name Mover, Negative Name Mover, and S-Inhibition heads, respectively.
}
\label{tab:bloom_560m_inlet_outlet_appendix}
\end{table*}

\begin{table*}[p!]
\scriptsize
\centering
    
\begin{subtable}{\textwidth}
\centering
\begin{tabular}{l|l|rrrrrr|r}
\toprule
& Method & Dup (4) & Prev (10) & Ind (10) & NM (11) & N-NM (3) & S-Inh (6) & Avg \\
\midrule
\multirow{5}{*}{QQ}  & PK & \textbf{0.001} & \textbf{0.041} & \textbf{0.090} & \textbf{0.183} & \underline{0.001} & \underline{0.001} & \textbf{0.053} \\
 & CS & 0.000 & 0.005 & 0.011 & 0.021 & 0.000 & 0.001 & 0.006 \\
 & Simple-CS & \underline{0.001} & \underline{0.012} & \underline{0.055} & \underline{0.084} & \textbf{0.021} & \textbf{0.001} & \underline{0.029} \\
 & Linear CKA & 0.000 & 0.001 & 0.000 & 0.001 & 0.000 & 0.000 & 0.000 \\
 & Proc. Sim. & 0.000 & 0.001 & 0.001 & 0.001 & 0.000 & 0.000 & 0.000 \\
\midrule
\multirow{5}{*}{KK}  & PK & \textbf{0.001} & \textbf{0.014} & \textbf{0.140} & \textbf{0.095} & \textbf{0.001} & \textbf{0.001} & \textbf{0.042} \\
 & CS & 0.000 & 0.002 & 0.005 & 0.010 & 0.000 & 0.001 & 0.003 \\
 & Simple-CS & \underline{0.001} & \underline{0.010} & \underline{0.091} & \underline{0.062} & \underline{0.001} & \underline{0.001} & \underline{0.027} \\
 & Linear CKA & 0.000 & 0.001 & 0.000 & 0.001 & 0.000 & 0.000 & 0.000 \\
 & Proc. Sim. & 0.000 & 0.001 & 0.000 & 0.001 & 0.000 & 0.000 & 0.000 \\
\midrule
\multirow{5}{*}{VV}  & PK & 0.000 & \textbf{0.012} & \textbf{0.006} & \underline{0.005} & \underline{0.001} & \underline{0.001} & \textbf{0.004} \\
 & CS & \textbf{0.001} & 0.007 & 0.004 & 0.002 & 0.001 & 0.001 & 0.003 \\
 & Simple-CS & \underline{0.000} & \underline{0.011} & \underline{0.006} & 0.003 & \textbf{0.001} & 0.001 & \underline{0.004} \\
 & Linear CKA & 0.000 & 0.000 & 0.001 & \textbf{0.011} & 0.000 & \textbf{0.002} & 0.002 \\
 & Proc. Sim. & 0.000 & 0.001 & 0.001 & 0.004 & 0.000 & 0.001 & 0.001 \\
\midrule
\multirow{5}{*}{OO}  & PK & 0.000 & \textbf{0.035} & \textbf{0.008} & \textbf{0.001} & \textbf{0.001} & \underline{0.000} & \textbf{0.008} \\
 & CS & \textbf{0.001} & \underline{0.005} & 0.002 & 0.001 & 0.000 & 0.000 & 0.002 \\
 & Simple-CS & \underline{0.000} & 0.005 & 0.003 & \underline{0.001} & \underline{0.000} & 0.000 & \underline{0.002} \\
 & Linear CKA & 0.000 & 0.001 & \underline{0.004} & 0.001 & 0.000 & \textbf{0.001} & 0.001 \\
 & Proc. Sim. & 0.000 & 0.003 & 0.003 & 0.001 & 0.000 & 0.000 & 0.001 \\
\midrule
\multirow{5}{*}{Mean over QQ, KK, VV, OO}  & PK & \textbf{0.001} & \textbf{0.026} & \textbf{0.061} & \textbf{0.071} & \underline{0.001} & \textbf{0.001} & \textbf{0.027} \\
 & CS & \underline{0.001} & 0.005 & 0.006 & 0.008 & 0.000 & 0.001 & 0.003 \\
 & Simple-CS & 0.000 & \underline{0.009} & \underline{0.038} & \underline{0.038} & \textbf{0.006} & 0.001 & \underline{0.015} \\
 & Linear CKA & 0.000 & 0.001 & 0.001 & 0.003 & 0.000 & \underline{0.001} & 0.001 \\
 & Proc. Sim. & 0.000 & 0.001 & 0.001 & 0.002 & 0.000 & 0.000 & 0.001 \\
\bottomrule
\end{tabular}
\caption{PR-AUC}
\label{tab:bloom_560m_classification_appendix_pr_auc}
\end{subtable}

\par\bigskip

\begin{subtable}{\textwidth}
\centering
\begin{tabular}{l|l|rrrrrr|r}
\toprule
& Method & Dup (4) & Prev (10) & Ind (10) & NM (11) & N-NM (3) & S-Inh (6) & Avg \\
\midrule
\multirow{5}{*}{QQ}  & PK & \textbf{0.906} & \textbf{0.943} & \textbf{0.987} & \textbf{0.995} & 0.847 & 0.856 & \underline{0.923} \\
 & CS & 0.839 & 0.905 & 0.946 & 0.977 & \underline{0.898} & \textbf{0.876} & 0.907 \\
 & Simple-CS & \underline{0.877} & \underline{0.929} & \underline{0.978} & \underline{0.991} & \textbf{0.978} & \underline{0.875} & \textbf{0.938} \\
 & Linear CKA & 0.492 & 0.500 & 0.382 & 0.542 & 0.025 & 0.685 & 0.437 \\
 & Proc. Sim. & 0.327 & 0.445 & 0.438 & 0.666 & 0.096 & 0.704 & 0.446 \\
\midrule
\multirow{5}{*}{KK}  & PK & \textbf{0.953} & \textbf{0.914} & \textbf{0.995} & \textbf{0.996} & \textbf{0.930} & \textbf{0.896} & \textbf{0.947} \\
 & CS & 0.821 & 0.834 & 0.934 & 0.966 & 0.858 & 0.821 & 0.872 \\
 & Simple-CS & \underline{0.919} & \underline{0.886} & \underline{0.989} & \underline{0.993} & \underline{0.915} & \underline{0.866} & \underline{0.928} \\
 & Linear CKA & 0.690 & 0.497 & 0.398 & 0.492 & 0.288 & 0.748 & 0.519 \\
 & Proc. Sim. & 0.325 & 0.466 & 0.376 & 0.611 & 0.257 & 0.713 & 0.458 \\
\midrule
\multirow{5}{*}{VV}  & PK & 0.820 & 0.865 & \textbf{0.747} & \underline{0.858} & 0.896 & 0.668 & \underline{0.809} \\
 & CS & \textbf{0.852} & \underline{0.880} & 0.739 & 0.814 & \underline{0.910} & 0.674 & \textbf{0.811} \\
 & Simple-CS & \underline{0.828} & \textbf{0.884} & \underline{0.744} & 0.842 & 0.883 & 0.665 & 0.808 \\
 & Linear CKA & 0.197 & 0.307 & 0.594 & \textbf{0.948} & 0.860 & \textbf{0.857} & 0.627 \\
 & Proc. Sim. & 0.468 & 0.657 & 0.609 & 0.778 & \textbf{0.949} & \underline{0.773} & 0.706 \\
\midrule
\multirow{5}{*}{OO}  & PK & 0.681 & \textbf{0.940} & \underline{0.825} & 0.637 & \underline{0.909} & 0.717 & 0.785 \\
 & CS & \textbf{0.793} & \underline{0.910} & 0.764 & 0.619 & 0.902 & \underline{0.734} & \underline{0.787} \\
 & Simple-CS & \underline{0.748} & 0.905 & 0.777 & \textbf{0.681} & \textbf{0.934} & \textbf{0.756} & \textbf{0.800} \\
 & Linear CKA & 0.285 & 0.711 & \textbf{0.864} & 0.597 & 0.223 & 0.613 & 0.549 \\
 & Proc. Sim. & 0.534 & 0.885 & 0.777 & \underline{0.644} & 0.537 & 0.592 & 0.661 \\
\midrule
\multirow{5}{*}{Mean over QQ, KK, VV, OO}  & PK & \underline{0.840} & \textbf{0.916} & \textbf{0.888} & \underline{0.871} & \underline{0.896} & \underline{0.784} & \underline{0.866} \\
 & CS & 0.826 & 0.882 & 0.846 & 0.844 & 0.892 & 0.776 & 0.844 \\
 & Simple-CS & \textbf{0.843} & \underline{0.901} & \underline{0.872} & \textbf{0.877} & \textbf{0.928} & \textbf{0.791} & \textbf{0.869} \\
 & Linear CKA & 0.416 & 0.504 & 0.560 & 0.645 & 0.349 & 0.726 & 0.533 \\
 & Proc. Sim. & 0.413 & 0.613 & 0.550 & 0.675 & 0.460 & 0.695 & 0.568 \\
\bottomrule
\end{tabular}
\caption{ROC-AUC}
\label{tab:bloom_560m_classification_appendix_auc}
\end{subtable}

\caption{
PR-AUC and ROC-AUC for each head class in BLOOM-560M, computed separately for the weight pairing types QQ, KK, VV, and OO, together with the mean over the four weight pairing types for each head class.
These values are used to compute the summary reported in Table~\ref{tab:classification-summary-by-model}.
The maximum is shown in bold and the second largest is underlined.
We consider the $\binom{24\times 16}{2}=\binom{384}{2}=73{,}536$ directed head pairs whose source is in an earlier layer than the target; for heads in the same layer, the head with the smaller index is treated as the source.
Dup, Prev, Ind, NM, N-NM, and S-Inh denote Duplicate Token, Previous Token, Induction, Name Mover, Negative Name Mover, and S-Inhibition heads, respectively.}
\label{tab:bloom_560m_classification_appendix}
\end{table*}

\begin{table*}[p!]
\scriptsize
\centering
    
\begin{subtable}{\textwidth}
\centering
\begin{tabular}{l|l|rrrrrr|r}
\toprule
& Method & Dup (5) & Prev (10) & Ind (9) & NM (6) & N-NM (3) & S-Inh (8) & Avg \\
\midrule
\multirow{5}{*}{QQ}  & PK & \textbf{0.101} & 0.001 & \textbf{0.030} & \textbf{0.023} & \textbf{0.002} & \underline{0.003} & \textbf{0.027} \\
 & CS & 0.003 & \underline{0.001} & \underline{0.020} & \underline{0.013} & \underline{0.001} & 0.002 & \underline{0.007} \\
 & Simple-CS & 0.002 & \textbf{0.003} & 0.012 & 0.011 & 0.001 & \textbf{0.003} & 0.005 \\
 & Linear CKA & \underline{0.004} & 0.000 & 0.009 & 0.003 & 0.000 & 0.000 & 0.003 \\
 & Proc. Sim. & 0.001 & 0.001 & 0.004 & 0.003 & 0.000 & 0.001 & 0.001 \\
\midrule
\multirow{5}{*}{KK}  & PK & 0.050 & 0.001 & \textbf{0.145} & \textbf{0.070} & 0.001 & \textbf{0.020} & \textbf{0.048} \\
 & CS & 0.001 & \underline{0.001} & 0.017 & 0.008 & \underline{0.001} & 0.002 & 0.005 \\
 & Simple-CS & 0.000 & \textbf{0.001} & 0.013 & 0.005 & \textbf{0.001} & 0.002 & 0.004 \\
 & Linear CKA & \textbf{0.051} & 0.001 & 0.141 & 0.067 & 0.000 & 0.019 & 0.046 \\
 & Proc. Sim. & \underline{0.050} & 0.001 & \underline{0.142} & \underline{0.067} & 0.000 & \underline{0.019} & \underline{0.046} \\
\midrule
\multirow{5}{*}{VV}  & PK & \textbf{0.050} & 0.001 & \textbf{0.145} & \textbf{0.069} & \textbf{0.000} & \textbf{0.019} & \textbf{0.047} \\
 & CS & 0.001 & \textbf{0.002} & 0.012 & 0.002 & \underline{0.000} & 0.001 & 0.003 \\
 & Simple-CS & 0.001 & \underline{0.001} & 0.016 & 0.002 & 0.000 & 0.001 & 0.004 \\
 & Linear CKA & 0.050 & 0.000 & 0.141 & 0.067 & 0.000 & 0.018 & 0.046 \\
 & Proc. Sim. & \underline{0.050} & 0.001 & \underline{0.141} & \underline{0.067} & 0.000 & \underline{0.019} & \underline{0.046} \\
\midrule
\multirow{5}{*}{OO}  & PK & 0.003 & 0.003 & \textbf{0.025} & 0.000 & \textbf{0.001} & \textbf{0.001} & \textbf{0.006} \\
 & CS & \underline{0.003} & \underline{0.007} & \underline{0.009} & 0.000 & 0.000 & \underline{0.001} & \underline{0.003} \\
 & Simple-CS & \textbf{0.003} & \textbf{0.007} & 0.007 & 0.000 & \underline{0.001} & 0.001 & 0.003 \\
 & Linear CKA & 0.000 & 0.000 & 0.001 & \underline{0.001} & 0.000 & 0.001 & 0.001 \\
 & Proc. Sim. & 0.000 & 0.000 & 0.001 & \textbf{0.002} & 0.000 & 0.001 & 0.001 \\
\midrule
\multirow{5}{*}{Mean over QQ, KK, VV, OO}  & PK & \textbf{0.051} & 0.001 & \textbf{0.087} & \textbf{0.041} & \textbf{0.001} & \textbf{0.011} & \textbf{0.032} \\
 & CS & 0.002 & \underline{0.003} & 0.015 & 0.006 & \underline{0.001} & 0.001 & 0.005 \\
 & Simple-CS & 0.001 & \textbf{0.003} & 0.012 & 0.005 & 0.001 & 0.002 & 0.004 \\
 & Linear CKA & \underline{0.026} & 0.000 & \underline{0.073} & 0.035 & 0.000 & \underline{0.010} & \underline{0.024} \\
 & Proc. Sim. & 0.025 & 0.001 & 0.072 & \underline{0.035} & 0.000 & 0.010 & 0.024 \\
\bottomrule
\end{tabular}
\caption{PR-AUC}
\label{tab:qwen2_0_5b_instruct_classification_appendix_pr_auc}
\end{subtable}

\par\bigskip

\begin{subtable}{\textwidth}
\centering
\begin{tabular}{l|l|rrrrrr|r}
\toprule
& Method & Dup (5) & Prev (10) & Ind (9) & NM (6) & N-NM (3) & S-Inh (8) & Avg \\
\midrule
\multirow{5}{*}{QQ}  & PK & 0.875 & 0.494 & 0.887 & \textbf{0.980} & \textbf{0.988} & \underline{0.879} & 0.850 \\
 & CS & \textbf{0.935} & \underline{0.611} & \textbf{0.939} & \underline{0.979} & \underline{0.980} & 0.874 & \underline{0.887} \\
 & Simple-CS & \underline{0.920} & \textbf{0.683} & \underline{0.918} & 0.976 & 0.962 & \textbf{0.889} & \textbf{0.891} \\
 & Linear CKA & 0.889 & 0.252 & 0.731 & 0.721 & 0.935 & 0.537 & 0.677 \\
 & Proc. Sim. & 0.713 & 0.381 & 0.593 & 0.759 & 0.560 & 0.553 & 0.593 \\
\midrule
\multirow{5}{*}{KK}  & PK & 0.612 & \underline{0.371} & 0.922 & 0.959 & 0.954 & \underline{0.828} & 0.774 \\
 & CS & 0.717 & \textbf{0.404} & \textbf{0.976} & \textbf{0.975} & \textbf{0.966} & 0.818 & \textbf{0.809} \\
 & Simple-CS & 0.610 & 0.360 & \underline{0.958} & \underline{0.967} & \underline{0.966} & \textbf{0.843} & \underline{0.784} \\
 & Linear CKA & \textbf{0.855} & 0.335 & 0.773 & 0.299 & 0.147 & 0.570 & 0.497 \\
 & Proc. Sim. & \underline{0.767} & 0.359 & 0.867 & 0.494 & 0.399 & 0.581 & 0.578 \\
\midrule
\multirow{5}{*}{VV}  & PK & 0.810 & 0.653 & 0.857 & 0.915 & \underline{0.869} & 0.509 & 0.769 \\
 & CS & \textbf{0.879} & \textbf{0.731} & \underline{0.859} & \textbf{0.935} & \textbf{0.875} & 0.500 & \textbf{0.797} \\
 & Simple-CS & \underline{0.820} & \underline{0.656} & \textbf{0.863} & \underline{0.926} & 0.856 & 0.517 & \underline{0.773} \\
 & Linear CKA & 0.426 & 0.217 & 0.361 & 0.370 & 0.227 & \underline{0.566} & 0.361 \\
 & Proc. Sim. & 0.579 & 0.371 & 0.622 & 0.588 & 0.608 & \textbf{0.610} & 0.563 \\
\midrule
\multirow{5}{*}{OO}  & PK & 0.865 & 0.708 & \textbf{0.922} & 0.375 & \underline{0.846} & \underline{0.655} & \underline{0.729} \\
 & CS & \textbf{0.949} & \textbf{0.781} & 0.860 & 0.308 & 0.804 & 0.642 & 0.724 \\
 & Simple-CS & \underline{0.948} & \underline{0.773} & \underline{0.886} & 0.292 & \textbf{0.858} & 0.617 & \textbf{0.729} \\
 & Linear CKA & 0.209 & 0.118 & 0.658 & \underline{0.750} & 0.567 & \textbf{0.707} & 0.501 \\
 & Proc. Sim. & 0.207 & 0.119 & 0.743 & \textbf{0.763} & 0.619 & 0.525 & 0.496 \\
\midrule
\multirow{5}{*}{Mean over QQ, KK, VV, OO}  & PK & 0.790 & 0.557 & 0.897 & \textbf{0.807} & \textbf{0.914} & \textbf{0.718} & 0.781 \\
 & CS & \textbf{0.870} & \textbf{0.632} & \textbf{0.909} & \underline{0.799} & 0.906 & 0.709 & \textbf{0.804} \\
 & Simple-CS & \underline{0.824} & \underline{0.618} & \underline{0.906} & 0.790 & \underline{0.910} & \underline{0.717} & \underline{0.794} \\
 & Linear CKA & 0.595 & 0.230 & 0.631 & 0.535 & 0.469 & 0.595 & 0.509 \\
 & Proc. Sim. & 0.566 & 0.307 & 0.706 & 0.651 & 0.547 & 0.567 & 0.557 \\
\bottomrule
\end{tabular}
\caption{ROC-AUC}
\label{tab:qwen2_0_5b_instruct_classification_appendix_auc}
\end{subtable}

\caption{
PR-AUC and ROC-AUC for each head class in Qwen2-0.5B-Instruct, computed separately for the weight pairing types QQ, KK, VV, and OO, together with the mean over the four weight pairing types for each head class.
These values are used to compute the summary reported in Table~\ref{tab:classification-summary-by-model}.
The maximum is shown in bold and the second largest is underlined.
We consider the $\binom{24\times 14}{2}=\binom{336}{2}=56{,}280$ directed head pairs whose source is in an earlier layer than the target; for heads in the same layer, the head with the smaller index is treated as the source.
Dup, Prev, Ind, NM, N-NM, and S-Inh denote Duplicate Token, Previous Token, Induction, Name Mover, Negative Name Mover, and S-Inhibition heads, respectively.}
\label{tab:qwen2_0_5b_instruct_classification_appendix}
\end{table*}

\subsubsection{Details of head detection performance for OQ, OK, and OV}
Figures~\ref{fig:bloom-560m_head_detection_OQ_OK_OV_pr_curve} and~\ref{fig:qwen2-0.5b-instruct_head_detection_OQ_OK_OV_pr_curve} show the PR curves corresponding to the PR-AUC values in Table~\ref{tab:PRAUC-OQ-OK-OV-two-models} for BLOOM-560M and Qwen2-0.5B-Instruct, respectively.
As with GPT2-small (Fig.~\ref{fig:gpt2-small_head_detection_OQ_OK_OV_pr_curve} in Appendix~\ref{app:eval-detecting}), for OQ and OK, the PR curve of PK dominates the other curves or achieves a comparable AUC.
For OV, however, the PK curve is dominated by those of other methods, such as Linear CKA for BLOOM-560M and CS for Qwen2-0.5B-Instruct.

\subsubsection{Details of classwise classification performance for QQ, KK, VV, and OO}

Tables~\ref{tab:bloom_560m_classification_appendix} and~\ref{tab:qwen2_0_5b_instruct_classification_appendix} report, for each head class in BLOOM-560M and Qwen2-0.5B-Instruct, respectively, the mean PR-AUC and ROC-AUC scores used to compute Table~\ref{tab:classification-summary-by-model}, with the mean taken over QQ, KK, VV, and OO.
The tables also report PR-AUC and ROC-AUC separately for each of the four weight pairing types, QQ, KK, VV, and OO.

For PR-AUC, PK achieves the best average performance across head classes in Tables~\ref{tab:bloom_560m_classification_appendix_pr_auc} and~\ref{tab:qwen2_0_5b_instruct_classification_appendix_pr_auc} for all five settings, namely QQ, KK, VV, OO, and their mean.
Moreover, unlike for GPT2-small (Table~\ref{tab:gpt2_small_classification_appendix_pr_auc} in Appendix~\ref{app:eval-clustering}), PK achieves the best score or a score comparable to the best for many head classes across the five settings.

For ROC-AUC, Simple-CS outperforms PK for BLOOM-560M in Table~\ref{tab:bloom_560m_classification_appendix_auc}, whereas CS outperforms PK for Qwen2-0.5B-Instruct in Table~\ref{tab:qwen2_0_5b_instruct_classification_appendix_auc}.
However, because the classification setting is extremely imbalanced\footnote{For example, Induction Head pairs account for only 45 of $73{,}536$ pairs in BLOOM-560M and 36 of $56{,}280$ pairs in Qwen2-0.5B-Instruct.}, we place greater emphasis on PR-AUC than on ROC-AUC.

\begin{table*}[t]
\small
\centering
\begingroup
\renewcommand{\arraystretch}{1.2}

\begin{tabular}{lllll}
  \toprule
  Weight preprocessing & Module & Original & Preprocessed & Shape \\
  \midrule
  \multirow{3}{*}{Layer normalization folding} & $\operatorname{LN}(\bm{x})$ & $\bm{\gamma} \odot \frac{\bm{x} - \operatorname{Mean}(\bm{x})\bm{1}_d}{\operatorname{Std}(\bm{x})} + \bm{\beta}$ & $\frac{\bm{x} - \operatorname{Mean}(\bm{x})\bm{1}_d}{\operatorname{Std}(\bm{x})}$ & $\mathbb{R}^{d} \rightarrow \mathbb{R}^{d}$ \\
  & Weight after LN & $\bm{W}_\text{in}$ & $\bm{W}_\text{in}\bm{D}_\gamma\bm{C}_d$ & $\mathbb{R}^{d_\text{mid}\times d}$ \\
  & Bias after LN & $\bm{b}_\text{in}$ & $\bm{W}_\text{in}\bm{\beta} + \bm{b}_\text{in}$ & $\mathbb{R}^{d_\text{mid}}$ \\
  \midrule
  \multirow{2}{*}{Centering writing weights} & Weight before LN & $\bm{W}_\text{out}$ & $\bm{C}_d\bm{W}_\text{out}$ & $\mathbb{R}^{d \times d_\text{mid}}$ \\
  & Bias before LN & $\bm{b}_\text{out}$ & $\bm{C}_d\bm{b}_\text{out}$  & $\mathbb{R}^{d}$ \\
  \midrule
  Centering the unembedding matrix & Unembedding matrix & $\bm{E}_\text{out}$ & $\bm{E}_\text{out}\bm{C}_T$ & $\mathbb{R}^{d\times T}$ \\
  \midrule
  \multirow{2}{*}{Folding attention bias} & Bias in value & $\bm{b}_\text{V}^h$ & $\bm{0}_{d_\text{head}}$ & $\mathbb{R}^{d_\text{head}}$ \\
  & Bias in output & $\bm{b}_\text{O}^h$ & $\bm{W}_\text{O}^h \bm{b}_\text{V}^h + \bm{b}_\text{O}^h$ & $\mathbb{R}^d$ \\
  \bottomrule
\end{tabular}

\endgroup
\caption{
Summary of weight preprocessing~\cite{elhage2021mathematical,nanda2022transformerlens}.
Let $d$ be the model dimension and $d_\text{mid}$ the intermediate dimension.
Here, $\bm{\gamma},\bm{\beta}\in\mathbb{R}^d$ are the LN scaling weight and bias, and $\bm{D}_\gamma\in\mathbb{R}^{d\times d}$ is the diagonal matrix with diagonal entries $\bm{\gamma}$.
The matrices $\bm{C}_d\in\mathbb{R}^{d\times d}$ and $\bm{C}_T\in\mathbb{R}^{T\times T}$ are the centering matrices in (\ref{eq:C_d}) for dimensions $d$ and $T$, where $T$ is the vocabulary size.
Note that only $\operatorname{LN}(\bm{x})$ is a function, whereas the other entries are parameters.
See Appendix~\ref{app:wf-bg} for more details.
}
\label{tab:wf}
\end{table*}

\subsection{Detecting hub heads in BLOOM-560M based on inlet and outlet scores}\label{app:bloom-hub}

For BLOOM-560M, we examine whether, as in GPT2-small, heads that function as hub heads appear among those with high inlet and outlet head scores, and whether these heads are Identity Heads.
Note that the repeated key and value head weights in GQA lead to many ties in the OK and OV inlet scores (\ref{eq:inlet}) in Appendix~\ref{app:inlet-outlet}.
Therefore, we do not consider Qwen2-0.5B-Instruct in this analysis because its GQA architecture makes the hub analysis more difficult to interpret.

Table~\ref{tab:bloom_560m_inlet_outlet_appendix} shows the top 10 heads and their inlet and outlet head scores for OQ, OK, and OV.
As with GPT2-small (Table~\ref{tab:inlet_outlet_identity} in Section~\ref{sec:detecting-hub-heads}), the Identity Heads L5H2 and L6H3 appear among the top-ranked heads for OV inlet scores.
L5H2 also appears among the top-ranked heads for OQ outlet scores, while L6H3 appears among the top-ranked heads for both OQ and OV outlet scores.
Thus, L5H2 and L6H3 in BLOOM-560M are Identity Heads that function as hubs, similarly to L4H7 in GPT2-small.
This result suggests that the hub role of Identity Heads observed in Section~\ref{sec:detecting-hub-heads} is not specific to GPT2-small, but may be a more general property.

Interestingly, as with GPT2-small, other Identity Heads also appear among the top-ranked heads for outlet scores in Table~\ref{tab:bloom_560m_outlet_appendix}.
A more detailed analysis of the functions of Identity Heads is left for future work.

\section{Comparing PK and CS on preprocessed weights in GPT2-small}\label{app:weight-preproc}
Pretrained weights can be simplified without changing the output of a language model~\cite{elhage2021mathematical,nanda2022transformerlens}.
Following \citet{gurnee2024universal}, we call this procedure \emph{weight preprocessing}, and we call the resulting weights \emph{preprocessed weights}.

In TransformerLens~\cite{nanda2022transformerlens}, which is widely used in mechanistic interpretability research, this weight preprocessing is applied by default when loading pretrained model weights from HuggingFace~\cite{wolf-etal-2020-transformers}{\renewcommand{\thefootnote}{\fnsymbol{footnote}}\footnote[3]{Our experiments used TransformerLens v2.16.1 (June 2025;
  \url{https://github.com/TransformerLensOrg/TransformerLens/tree/v2.16.1}), in which
  \texttt{HookedTransformer.from\_pretrained} applies \texttt{fold\_ln}, \texttt{center\_writing\_weights},
  \texttt{center\_unembed}, and \texttt{fold\_value\_biases} by default. TransformerLens v3.0.0 (April 2026;
  \url{https://github.com/TransformerLensOrg/TransformerLens/tree/v3.0.0}) deprecated \texttt{HookedTransformer} in
  favor of the newly introduced \texttt{TransformerBridge}, which preserves the original HuggingFace weights on loading
  and applies this preprocessing only when \texttt{enable\_compatibility\_mode()} is called explicitly.
  \texttt{HookedTransformer} was subsequently removed in v4.0.0 (commit
  \texttt{ffc629caed24516a3338200cf9e2086f2ade7f5e}, September 2026;
  \url{https://github.com/TransformerLensOrg/TransformerLens/pull/1759}), so this preprocessing is no longer applied by
  default in current versions.}}.
Therefore, when analyzing weights with TransformerLens using PK or CS, one should note that the weights may differ from the original ones.

In this section, we first describe the weight preprocessing procedure based on \citet{nanda2022transformerlens}.
We then compare, for GPT2-small, PK and CS computed on the original weights with those computed on the preprocessed weights.

\subsection{Background: weight preprocessing}\label{app:wf-bg}
Based on \citet{nanda2022transformerlens}, we describe four types of weight preprocessing\footnote{\url{https://github.com/TransformerLensOrg/TransformerLens/blob/v2.16.1/further_comments.md}.}.
Table~\ref{tab:wf} summarizes these procedures and compares the original weights and the preprocessed weights.

\subsubsection{Layer normalization folding}\label{app:wf-ln}
By decomposing layer normalization (LN)~\cite{ba2016layernormalization} into four steps, we can fold the LN scaling weight and bias into the weight and bias immediately after LN, thereby simplifying LN.

\paragraph{Layer normalization.}
Using the scaling weight $\bm{\gamma}\in\mathbb{R}^d$ and bias $\bm{\beta}\in\mathbb{R}^d$, LN for an input $\bm{x}=(x_1,\ldots,x_d)^\top\in\mathbb{R}^d$ is
\begin{equation}
    \operatorname{LN}(\bm{x}) = \bm{\gamma} \odot \frac{\bm{x} - \operatorname{Mean}(\bm{x})\bm{1}_d}{\operatorname{Std}(\bm{x})} + \bm{\beta},\label{eq:LN}
\end{equation}
where $\odot$ denotes the Hadamard product, $\bm{1}_d\in\mathbb{R}^d$ is the all-ones vector, and $\operatorname{Mean}(\cdot)$ and $\operatorname{Std}(\cdot)$ are defined as
\begin{align}
    \operatorname{Mean}(\bm{x}) &= \frac{1}{d}{\bm{1}_d}^\top\bm{x},\label{eq:Mean}\\
    \operatorname{Std}(\bm{x}) &= \sqrt{\frac{1}{d}\sum_{i=1}^d \left(x_i-\operatorname{Mean}(\bm{x})\right)^2}.\label{eq:Std}
\end{align}

\paragraph{Centering matrix.}
To simplify algebraic manipulations of LN, we use the following centering matrix~\cite{borg2005modern}:
\begin{align}
\bm{C}_d=\bm{I}_d - \frac{1}{d}\bm{1}_d{\bm{1}_d}^\top\in\mathbb{R}^{d\times d},\label{eq:C_d}
\end{align}
where $\bm{I}_d\in\mathbb{R}^{d\times d}$ is the identity matrix.
From the definition of $\bm{C}_d$ in (\ref{eq:C_d}) and $\operatorname{Mean}(\cdot)$ in (\ref{eq:Mean}), we have
\begin{align}
\bm{C}_d\bm{x}
= \bm{x} - \frac{1}{d}\bm{1}_d{\bm{1}_d}^\top \bm{x}
= \bm{x} - \operatorname{Mean}(\bm{x})\bm{1}_d.\label{eq:Cd-x-MeanI}
\end{align}
The matrix $\bm{C}_d$ is an orthogonal projection matrix since it satisfies
\begin{align}
    \bm{C}_d\bm{C}_d &= \bm{C}_d,\label{eq:CdCd-Cd}\\
    {\bm{C}_d}^\top &= \bm{C}_d.\label{eq:Cdtop-Cd}
\end{align}

\paragraph{Decomposing LN into four steps.}
LN can be decomposed into the following four steps\footnote{
Since ${\bm{1}_d}^\top \bm{C}_d
    = {\bm{1}_d}^\top \left(\bm{I}_d - \bm{1}_d{\bm{1}_d}^\top/d\right) = \bm{0}^\top$,
we have ${\bm{1}_d}^\top \bm{C}_d\bm{x}=0$ for any $\bm{x}\in\mathbb{R}^d$.
Thus, $\bm{C}_d$ is the orthogonal projection onto the orthogonal complement of the one-dimensional subspace spanned by $\bm{1}_d$.
In particular, for (\ref{eq:LN-x1}), we have
$\bm{x}^{(1)}=\bm{C}_d\bm{x}
= \bm{x} - \bm{x}^\top(\bm{1}_d/\sqrt{d})\bm{1}_d/\sqrt{d}$,
so \citet{nanda2022transformerlens} view centering as removing the component of $\bm{x}$ in the $\bm{1}_d$ direction using the unit vector $\bm{1}_d/\sqrt{d}$.
For the resulting $\bm{x}^{(1)}$, they obtain $\bm{x}^{(2)}$ in (\ref{eq:LN-x2}) by projecting it onto the sphere of radius $\sqrt{d}$ while preserving its direction.
}:
\begin{align}
  \bm{x}^{(1)}
    &= \bm{x} - \operatorname{Mean}(\bm{x})\bm{1}_d = \bm{C}_d\bm{x}\label{eq:LN-x1}\quad(\because(\ref{eq:Cd-x-MeanI})),\\
  \bm{x}^{(2)}
    &= \frac{\bm{x}^{(1)}}
      {\sqrt{\frac{1}{d}\sum_{i=1}^d \left(x_i-\operatorname{Mean}(\bm{x})\right)^2}}
     = \sqrt{d}\,\frac{\bm{x}^{(1)}}
      {\sqrt{\sum_{i=1}^d \left(x_i^{(1)}\right)^2}} 
     = \sqrt{d}\,\frac{\bm{x}^{(1)}}{\|\bm{x}^{(1)}\|},\label{eq:LN-x2}\\
  \bm{x}^{(3)}
    &= \bm{\gamma} \odot \bm{x}^{(2)}
    =  \bm{D}_\gamma \bm{x}^{(2)},\label{eq:LN-x3}\\
  \bm{x}^{(4)} & = \bm{x}^{(3)} + \bm{\beta}.\label{eq:LN-x4}
\end{align}
Here, $x_i^{(1)}$ is the $i$-th component of $\bm{x}^{(1)}$ in (\ref{eq:LN-x1}), and $\bm{D}_\gamma\in\mathbb{R}^{d\times d}$ is the diagonal matrix whose diagonal entries are $\bm{\gamma}$.
By definition, $\operatorname{LN}(\bm{x})=\bm{x}^{(4)}$.

\paragraph{Folding into the weight and bias after LN.}
Based on the four-step decomposition, consider the affine map immediately after LN that takes $\operatorname{LN}(\bm{x})=\bm{x}^{(4)}$ as input.
Let $\bm{W}_\text{in}\in\mathbb{R}^{d_\text{mid}\times d}$ and $\bm{b}_\text{in}\in\mathbb{R}^{d_\text{mid}}$ be its weight and bias\footnote{For example, $d_\text{mid}=d_\text{head}$ for an attention layer and $d_\text{mid}=4d$ for a MLP layer.}.
Then
\begin{align}
    \bm{W}_\text{in} \bm{x}^{(4)} + \bm{b}_\text{in}
    &=\bm{W}_\text{in} \bm{D}_\gamma \frac{\sqrt{d}\bm{C}_d \bm{x}}{\|\bm{x}^{(1)}\|} + \bm{W}_\text{in}\bm{\beta} + \bm{b}_\text{in}\quad(\because (\ref{eq:LN-x4}), (\ref{eq:LN-x3}), (\ref{eq:LN-x2}), (\ref{eq:LN-x1})) \notag \\
    &=\bm{W}_\text{in} \bm{D}_\gamma \bm{C}_d \bm{C}_d \frac{\sqrt{d}\bm{x}}{\|\bm{x}^{(1)}\|} + \bm{W}_\text{in}\bm{\beta} + \bm{b}_\text{in}\quad(\because(\ref{eq:CdCd-Cd})) \notag \\
    &=\bm{W}_\text{in} \bm{D}_\gamma \bm{C}_d\bm{x}^{(2)} + \bm{W}_\text{in}\bm{\beta} + \bm{b}_\text{in}
    \quad(\because(\ref{eq:LN-x1}), (\ref{eq:LN-x2})).\label{eq:lnwf}
\end{align}
Define
\begin{align}
    &\operatorname{LN}^{(2)}(\bm{x}) = \bm{x}^{(2)}=\frac{\bm{x} - \operatorname{Mean}(\bm{x})\bm{1}_d}{\operatorname{Std}(\bm{x})},\label{eq:LN2}\\
    &\bm{W}_\text{in}' = \bm{W}_\text{in} \bm{D}_\gamma\bm{C}_d,\,
    \bm{b}_\text{in}' = \bm{W}_\text{in}\bm{\beta} + \bm{b}_\text{in}.\label{eq:W_in-fold}
\end{align}
Then, from $\operatorname{LN}(\bm{x})=\bm{x}^{(4)}$ and (\ref{eq:lnwf}), we obtain
\begin{equation}
    \bm{W}_\text{in} \operatorname{LN}(\bm{x}) + \bm{b}_\text{in}
    = \bm{W}_\text{in}' \operatorname{LN}^{(2)}(\bm{x}) + \bm{b}_\text{in}'.\label{eq:WinLNbin-WinLN2bin}
\end{equation}
From (\ref{eq:WinLNbin-WinLN2bin}), replacing LN in (\ref{eq:LN}) with $\operatorname{LN}^{(2)}(\bm{x})$ in (\ref{eq:LN2}), and replacing $\bm{W}_\text{in}$ and $\bm{b}_\text{in}$ with $\bm{W}_\text{in}'$ and $\bm{b}_\text{in}'$, does not change the output of the affine map.
In particular, $\operatorname{LN}^{(2)}(\bm{x})$ in (\ref{eq:LN2}) performs only centering and standard-deviation normalization, and is simpler than the original LN in (\ref{eq:LN}).
For attention heads, this folding can be applied to the reading weights $\bm{W}_\text{Q}^h$, $\bm{W}_\text{K}^h$, and $\bm{W}_\text{V}^h$.

\subsubsection{Centering weights that write to the residual stream}\label{app:wf-centering-writing}
Let $\bm{\zeta}\in\mathbb{R}^{d_\text{mid}}$ be an intermediate representation, and suppose the affine output $\bm{W}_\text{out}\bm{\zeta}+\bm{b}_\text{out}\in\mathbb{R}^d$ is written to the residual stream, where $\bm{W}_\text{out}\in\mathbb{R}^{d\times d_\text{mid}}$ and $\bm{b}_\text{out}\in\mathbb{R}^d$.
Consider applying LN to $\bm{W}_\text{out}\bm{\zeta}+\bm{b}_\text{out}$.
As seen in (\ref{eq:LN-x1}), LN performs centering in the first step.
Therefore, centering gives
\begin{align}
\bm{C}_d(\bm{W}_\text{out}\bm{\zeta}+\bm{b}_\text{out})
&=\bm{C}_d\bm{C}_d(\bm{W}_\text{out}\bm{\zeta}+\bm{b}_\text{out}) \quad(\because(\ref{eq:CdCd-Cd}))\notag\\
&=\bm{C}_d(\bm{C}_d\bm{W}_\text{out}\bm{\zeta}+\bm{C}_d\bm{b}_\text{out}).\label{eq:CWDzbCCWzCb}
\end{align}
Thus, defining
\begin{equation}
    \bm{W}_\text{out}' = \bm{C}_d\bm{W}_\text{out},\,
    \bm{b}_\text{out}' =\bm{C}_d\bm{b}_\text{out},\label{eq:W_out-fold}
\end{equation}
we can replace $\bm{W}_\text{out}$ and $\bm{b}_\text{out}$ with $\bm{W}_\text{out}'$ and $\bm{b}_\text{out}'$ without changing the result of centering.
For attention heads, this centering can be applied to the writing weights $\bm{W}_\text{O}^h$.

\subsubsection{Other preprocessing steps}\label{app:wf-others}
When pretrained model weights are loaded using the default settings in TransformerLens~\cite{nanda2022transformerlens}{\renewcommand{\thefootnote}{\fnsymbol{footnote}}\footnote[3]{Our experiments used TransformerLens v2.16.1 (June 2025;
  \url{https://github.com/TransformerLensOrg/TransformerLens/tree/v2.16.1}), in which
  \texttt{HookedTransformer.from\_pretrained} applies \texttt{fold\_ln}, \texttt{center\_writing\_weights},
  \texttt{center\_unembed}, and \texttt{fold\_value\_biases} by default. TransformerLens v3.0.0 (April 2026;
  \url{https://github.com/TransformerLensOrg/TransformerLens/tree/v3.0.0}) deprecated \texttt{HookedTransformer} in
  favor of the newly introduced \texttt{TransformerBridge}, which preserves the original HuggingFace weights on loading
  and applies this preprocessing only when \texttt{enable\_compatibility\_mode()} is called explicitly.
  \texttt{HookedTransformer} was subsequently removed in v4.0.0 (commit
  \texttt{ffc629caed24516a3338200cf9e2086f2ade7f5e}, September 2026;
  \url{https://github.com/TransformerLensOrg/TransformerLens/pull/1759}), so this preprocessing is no longer applied by
  default in current versions.}}, the unembedding matrix is also centered\footnote{
Let $\bm{E}_\text{out}\in\mathbb{R}^{d\times T}$ be the unembedding matrix and let $\widetilde{\bm{x}}\in\mathbb{R}^d$ be the vector after the final LN in (\ref{eq:LN-final}), where $T = |\mathcal{T}|$.
Logits are computed as ${\bm{E}_\text{out}}^\top\widetilde{\bm{x}}\in\mathbb{R}^{T}$, and token probabilities are computed by applying softmax.
By definition of softmax, adding the same constant to all components of ${\bm{E}_\text{out}}^\top\widetilde{\bm{x}}$ does not change the output.
Let $\bm{C}_T\in\mathbb{R}^{T\times T}$ be the $T$-dimensional centering matrix, analogous to (\ref{eq:C_d}).
Then centering logits using $\bm{C}_T$ does not affect the softmax output, so
\begin{align*}
    \operatorname{softmax}({\bm{E}_\text{out}}^\top\widetilde{\bm{x}}) =\operatorname{softmax}(\bm{C}_T{\bm{E}_\text{out}}^\top\widetilde{\bm{x}})
    =\operatorname{softmax}((\bm{E}_\text{out}{\bm{C}_T}^\top)^\top\widetilde{\bm{x}})
    =\operatorname{softmax}((\bm{E}_\text{out}\bm{C}_T)^\top\widetilde{\bm{x}}) \in [0, 1]^T \quad(\because (\ref{eq:Cdtop-Cd})).
\end{align*}
Thus, defining $\bm{E}_\text{out}'=\bm{E}_\text{out}\bm{C}_T$, we can replace $\bm{E}_\text{out}$ with $\bm{E}_\text{out}'$ without changing the softmax output.
}, and the bias term in the output computation of attention is also folded\footnote{
In $o^{h}(\bm{X})$ in (\ref{eq:ohX}), we omitted bias terms.
Including the bias terms $\bm{b}_{\text{V}}^{h}\in\mathbb{R}^{d_\text{head}}$ and $\bm{b}_{\text{O}}^{h}\in\mathbb{R}^d$ in addition to $\bm{W}_{\text{V}}^{h}$ and $\bm{W}_{\text{O}}^{h}$ yields
\begin{align*}
o^{h}(\bm{X})
&=\bm{W}_{\text{O}}^{h}(\bm{W}_{\text{V}}^{h}\bm{X}{\bm{A}^h}^\top + \bm{b}_{\text{V}}^{h}) + \bm{b}_{\text{O}}^{h}
=\bm{W}_{\text{O}}^{h}\bm{W}_{\text{V}}^{h}\bm{X}{\bm{A}^h}^\top  + \bm{W}_{\text{O}}^{h}\bm{b}_{\text{V}}^{h} + \bm{b}_{\text{O}}^{h}\notag\\
&=\bm{W}_{\text{O}}^{h}(\bm{W}_{\text{V}}^{h}\bm{X}{\bm{A}^h}^\top + \bm{0}_{d_\text{head}}) + \bm{W}_{\text{O}}^{h}\bm{b}_{\text{V}}^{h} + \bm{b}_{\text{O}}^{h}.
\end{align*}
Thus, defining ${\bm{b}_\text{V}^h}' = \bm{0}_{d_\text{head}}$ and ${\bm{b}_{\text{O}}^{h}}' = \bm{W}_{\text{O}}^{h}\bm{b}_{\text{V}}^{h} + \bm{b}_{\text{O}}^{h}$, we can replace $\bm{b}_{\text{V}}^{h}$ and $\bm{b}_{\text{O}}^{h}$ with ${\bm{b}_\text{V}^h}'$ and ${\bm{b}_{\text{O}}^{h}}'$ without changing the attention output.
}.
These steps are unrelated to PK and CS.

\subsection{Wiring diagrams based on PK and CS between preprocessed weights}\label{app:wp-pk-vs-cs}
We compare the top 20 wiring diagrams from the original GPT2-small weights (PK: Fig.~\ref{fig:wd-pk-top20}, CS: Fig.~\ref{fig:wd-cs-top20}) with those from the preprocessed weights described in Appendix~\ref{app:wf-bg}, with LN folding applied as in (\ref{eq:W_in-fold}) and writing weights centered as in (\ref{eq:W_out-fold}).

\begin{figure*}[p!]
    \centering
    \includegraphics[width=0.96\linewidth]{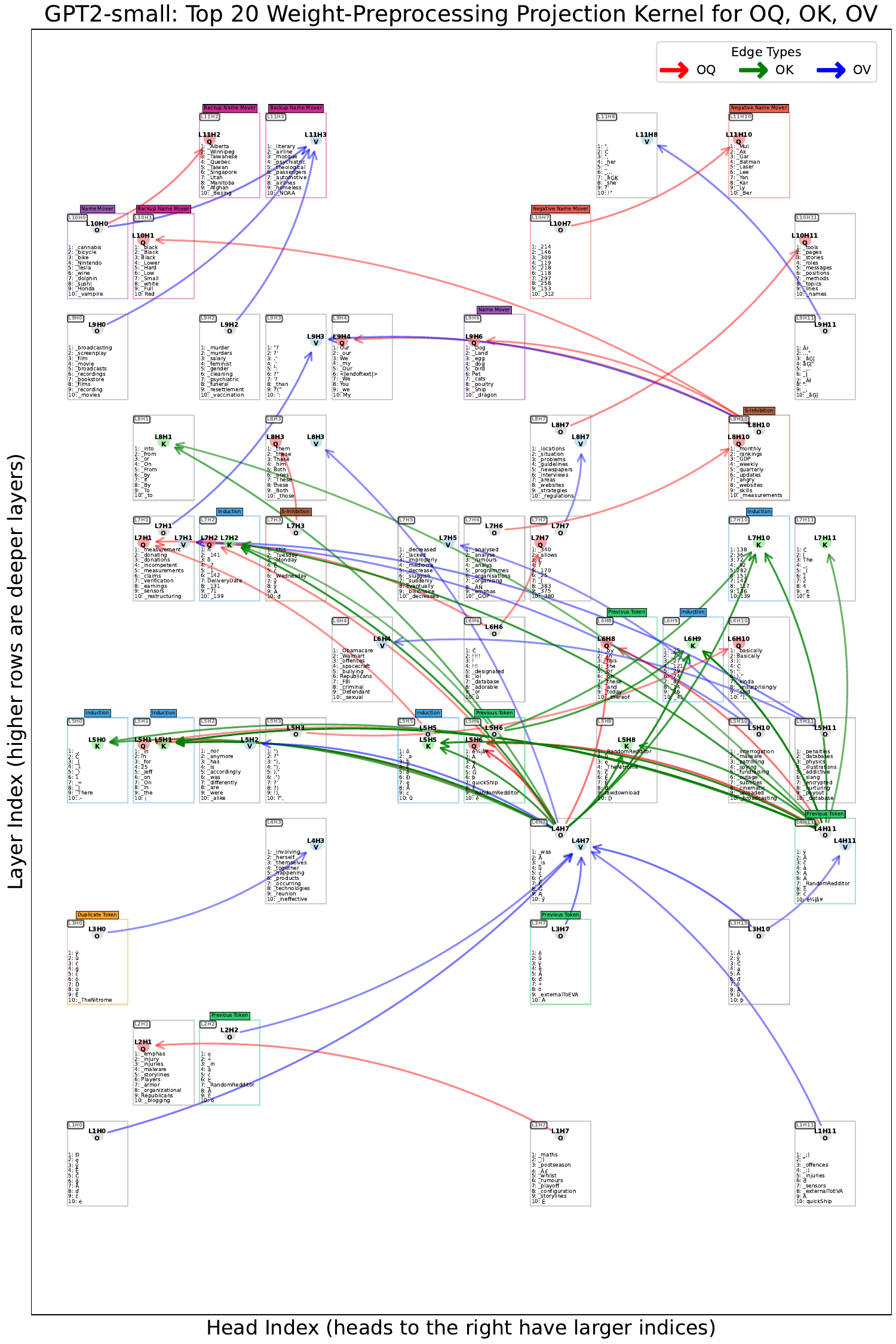}
    \caption{
For GPT2-small, to compare with the wiring diagram constructed from PK on the original weights in Fig.~\ref{fig:wd-pk-top20}, we visualize the top 20 scores of $\operatorname{PK}_{\,\text{OQ}}$, $\operatorname{PK}_{\,\text{OK}}$, and $\operatorname{PK}_{\,\text{OV}}$ computed on the preprocessed weights.
}
    \label{fig:wd-pk_wp-top20}
\end{figure*}

Figure~\ref{fig:wd-pk_wp-top20} shows the top 20 PK edges computed from preprocessed weights for each of the OQ, OK, and OV pairings.
This wiring diagram is almost unchanged from the original one in Fig.~\ref{fig:wd-pk-top20}.
In fact, among the 60 edges, 57 are shared (19 for OQ, 20 for OK, and 18 for OV)\footnote{
Edges that appear only in the original wiring diagram are the OQ edge between L8H10 and L10H6, the OV edge between L5H5 and L6H6, and the OV edge between L7H7 and L8H5.
Edges that appear only in the preprocessed wiring diagram are the OQ edge between L6H6 and L7H7, the OV edge between L3H10 and L4H11, and the OV edge between L5H10 and L7H5.
}.
This result indicates that PK is robust to weight changes induced by weight preprocessing.

\begin{figure*}[p!]
    \centering
    \includegraphics[width=0.96\linewidth]{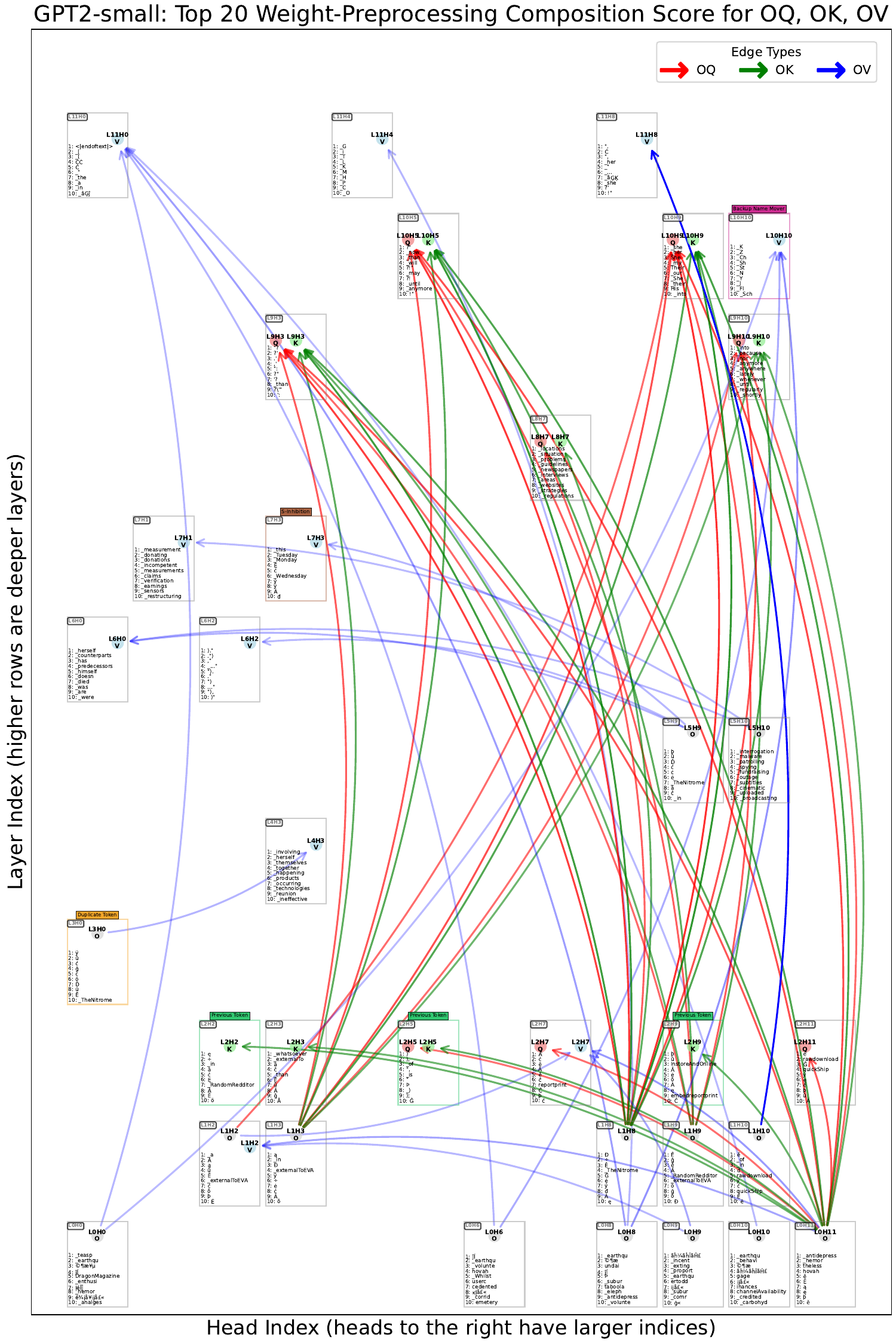}
    \caption{
For GPT2-small, to compare with the wiring diagram constructed from CS on the original weights in Fig.~\ref{fig:wd-cs-top20}, we visualize the top 20 scores of $\operatorname{CS}_{\,\text{OQ}}$, $\operatorname{CS}_{\,\text{OK}}$, and $\operatorname{CS}_{\,\text{OV}}$ computed on the preprocessed weights.
}
    \label{fig:wd-cs_wp-top20}
\end{figure*}

Figure~\ref{fig:wd-cs_wp-top20} shows the top 20 CS edges computed from preprocessed weights for each of the OQ, OK, and OV pairings.
This wiring diagram differs substantially from the original one in Fig.~\ref{fig:wd-cs-top20}.
In fact, among the 60 edges, only 16 are shared (1 for OQ, 4 for OK, and 11 for OV).
This result indicates that CS is sensitive to weight changes induced by weight preprocessing.
In particular, the change is largest for OQ and OK, and many head pairs now have edges in both pairings, which suggests that the preprocessing reduces the difference between the query and key weights as measured by CS.
Moreover, edges between distant layers appear only for OV in the original wiring diagram, whereas they also appear for OQ and OK after weight preprocessing.

\subsection{Quantitative comparison of scores computed from original and preprocessed weights for PK and CS}\label{app:eval-wp-vs-org}

In this subsection, based on the results in Appendix~\ref{app:wp-pk-vs-cs}, where weight preprocessing changes the wiring diagrams of GPT2-small, we quantitatively compare these differences.
Specifically, for PK and CS, we compare the mean squared error (MSE) and Pearson's $r$ between the values computed from the original weights and those computed from the preprocessed weights for each weight pairing type.
We consider all $\binom{144}{2} = 10{,}296$ pairs among the 144 heads in GPT2-small, which has 12 layers with 12 heads per layer.

\subsubsection{MSE comparison}\label{app:mse-org-vs-wp}
In this section, we compare PK and CS in terms of the MSE between scores computed from the original weights and those computed from the preprocessed weights.
To make PK comparable to CS, we normalize PK by $d_\text{head}$ so that it lies on the same $[0,1]$ scale as CS (see Appendix~\ref{app:cs-range} for details), and we refer to this quantity as Normalized PK.

\begin{figure}[p!]
    \centering
    \includegraphics[width=0.75\columnwidth]{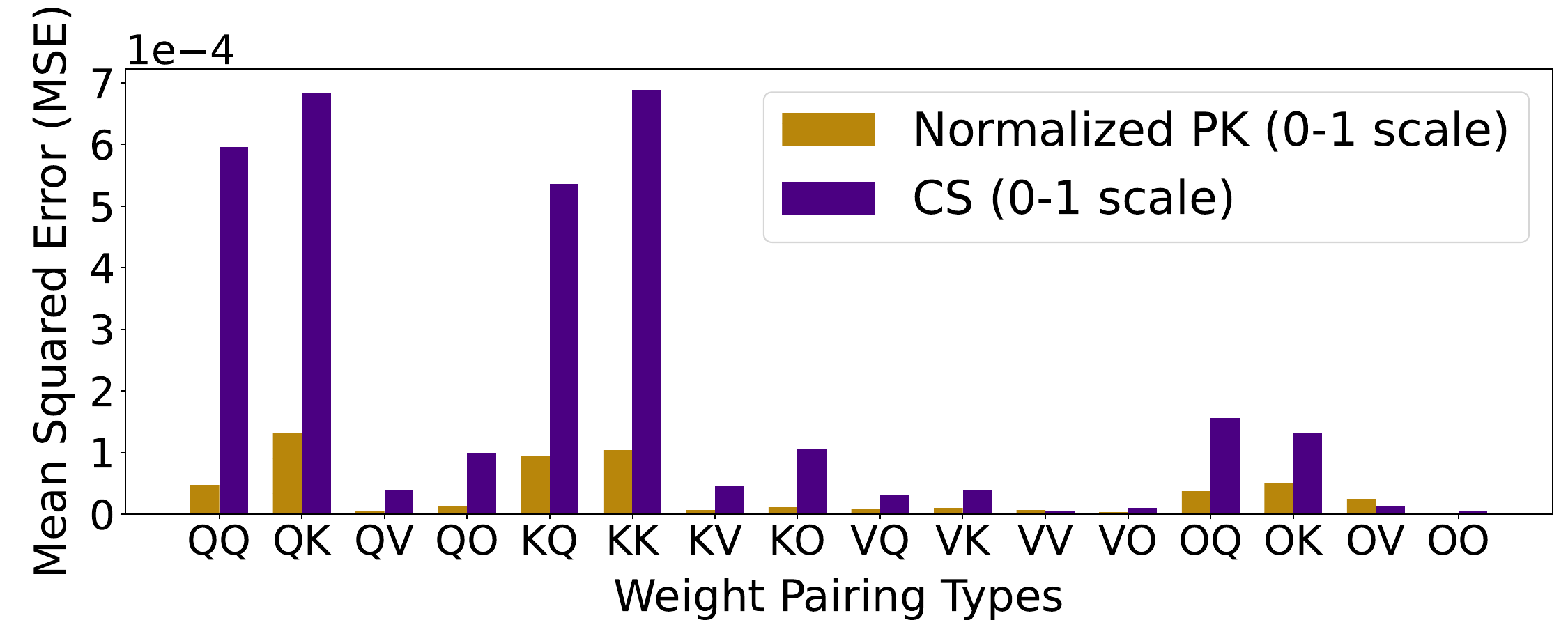}
    \caption{
For GPT2-small, we report the MSE between scores from the original and preprocessed weights for PK and CS, by weight pairing type.
Normalized PK denotes PK normalized by $d_\text{head}$, which maps it to the same $[0,1]$ scale as CS.
}
    \label{fig:pk-mse-vs-cs-mse}
\end{figure}

Figure~\ref{fig:pk-mse-vs-cs-mse} reports the MSE for Normalized PK and CS across weight pairing types.
Across the 16 weight pairings, Normalized PK yields MSE values that are smaller than or comparable to those of CS.
This result indicates that PK is robust to changes induced by weight preprocessing, whereas CS is more sensitive to these changes, consistent with the findings in Appendix~\ref{app:wp-pk-vs-cs}.

We also find a common trend in both Normalized PK and CS.
Pairings that involve only query and key weights (QQ, QK, KQ, and KK) exhibit larger MSE than pairings that involve only value or output weights (VV, VO, OV, and OO).
Interestingly, although the query, key, and value weights all undergo the same layer normalization folding, the MSE for QQ or KK is larger than that for VV, and the difference is especially large for CS.
A more detailed analysis of these results is left for future work.

\begin{figure}[p!]
    \centering
    \includegraphics[width=0.75\columnwidth]{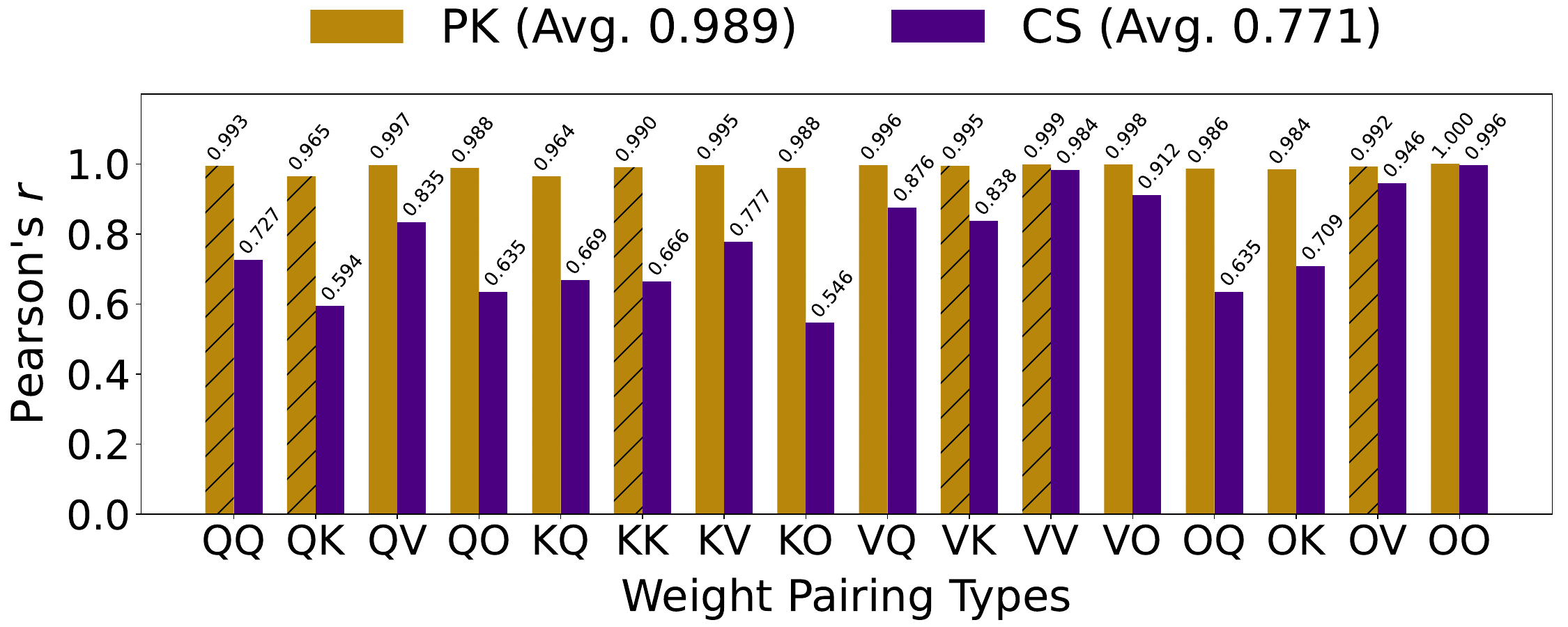}
    \caption{
For GPT2-small, we report Pearson's $r$ between scores from the original and preprocessed weights for PK and CS, by weight pairing type.
}
    \label{fig:wp-pk-vs-cs-pearsonr}
\end{figure}

\begin{figure}[p!]
    \centering
    \includegraphics[width=0.75\linewidth]{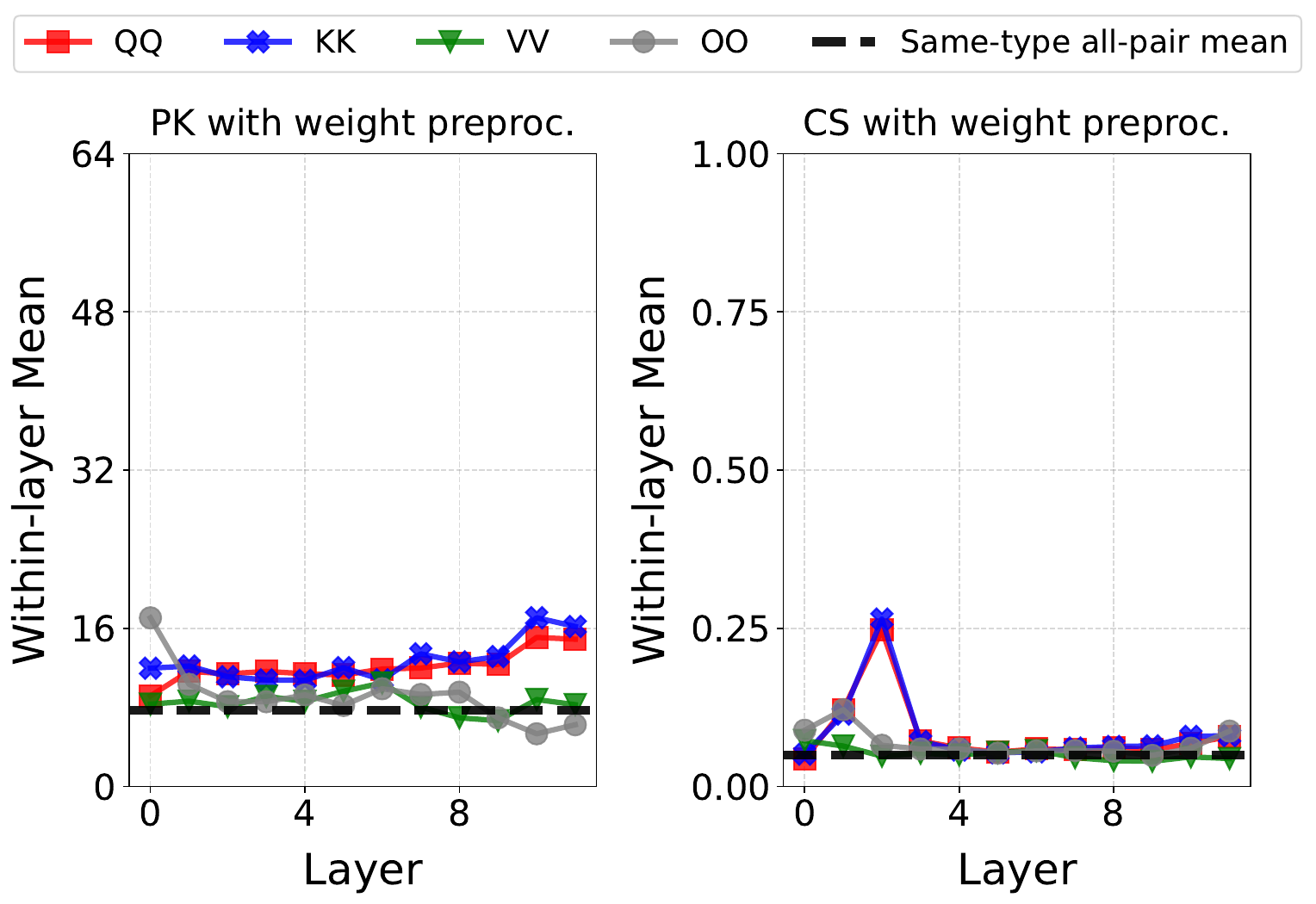}
\caption{
Layerwise PK and CS with weight preprocessing of GPT2-small, where each panel shows the mean over all $66$ head pairs within each layer (all pairwise combinations of the $12$ heads) for the weight pairing types QQ, KK, VV, and OO.
The dashed line marks the mean over all $\binom{144}{2}=10{,}296$ head pairs for each of the four weight pairing types, yielding a total of $4\times\binom{144}{2}=41{,}184$ comparisons: $\operatorname{PK}=7.71$ and $\operatorname{CS}=0.05$.
}
    \label{fig:gpt2-small_layerwise_weight_preproc_metrics}
\end{figure}

\subsubsection{Correlation comparison}\label{app:wp-pearsonr}
In this section, we compare PK and CS in terms of Pearson's $r$ between scores computed from the original weights and those computed from the preprocessed weights.

Figure~\ref{fig:wp-pk-vs-cs-pearsonr} reports Pearson's $r$ for PK and CS across the 16 weight pairing types.
For PK, the correlation coefficient is close to $1$ for all weight pairing types, with a mean of $0.989$.
In contrast, for CS, the correlation is lower for weight pairing types involving query or key weights.
For example, KO exhibits the lowest correlation coefficient, with a value of $0.546$.
These results are also consistent with the findings in Appendix~\ref{app:wp-pk-vs-cs} and Appendix~\ref{app:mse-org-vs-wp}.

\subsection{Score heatmaps for PK and CS after weight preprocessing}\label{app:wp-heatmap}

\begin{figure*}[p!]
    \centering
    \includegraphics[width=0.8\linewidth]{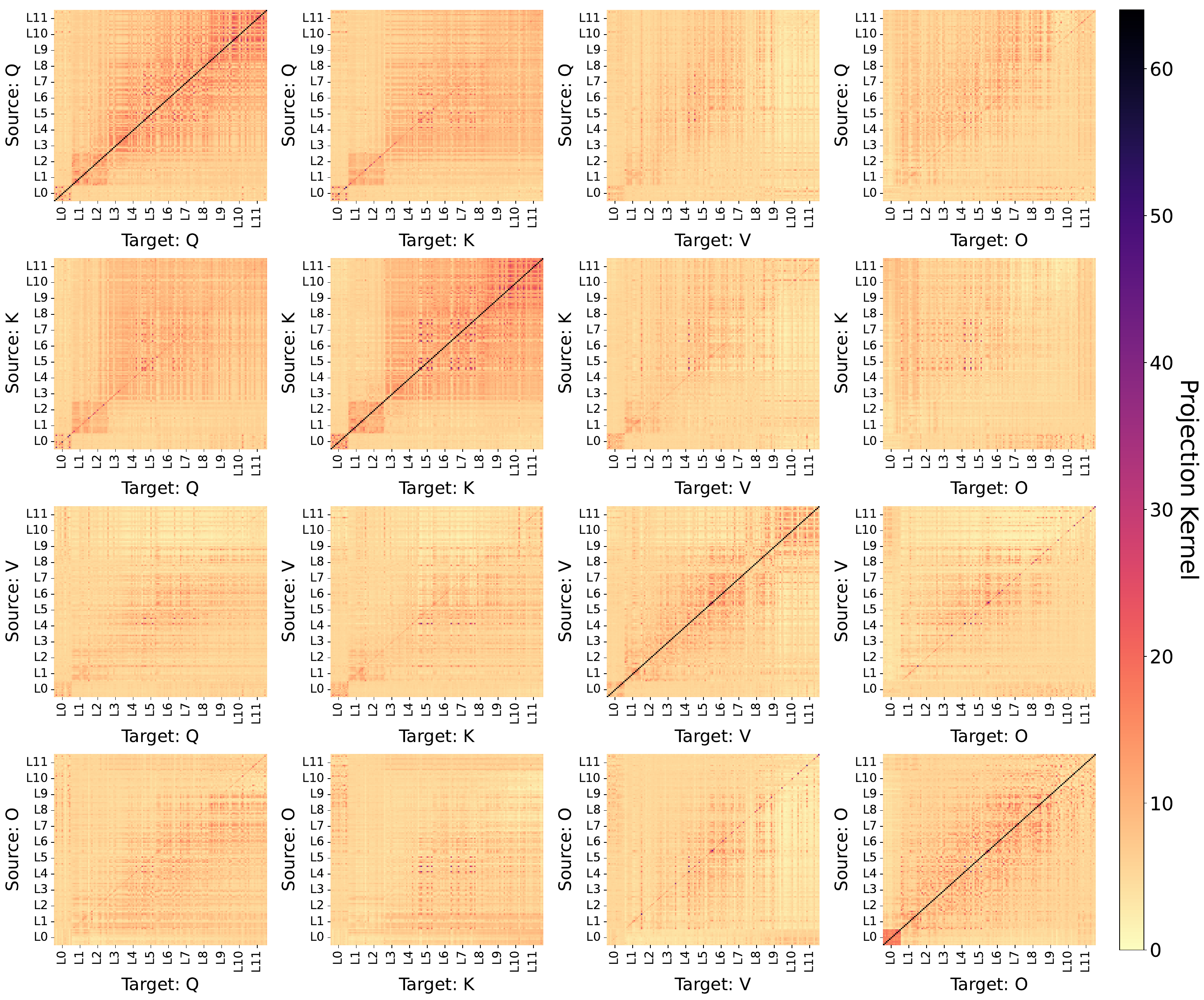}
    \caption{Projection Kernel (PK) with weight preprocessing between all head pairs of GPT2-small for each weight pairing type (source, target) of Q, K, V, and O, where the tick label L$\ell$ denotes layer $\ell$.}
    \label{fig:gpt2-small_heatmap_pk_weight_preproc}
\end{figure*}

\begin{figure*}[p!]
    \centering
    \includegraphics[width=0.8\linewidth]{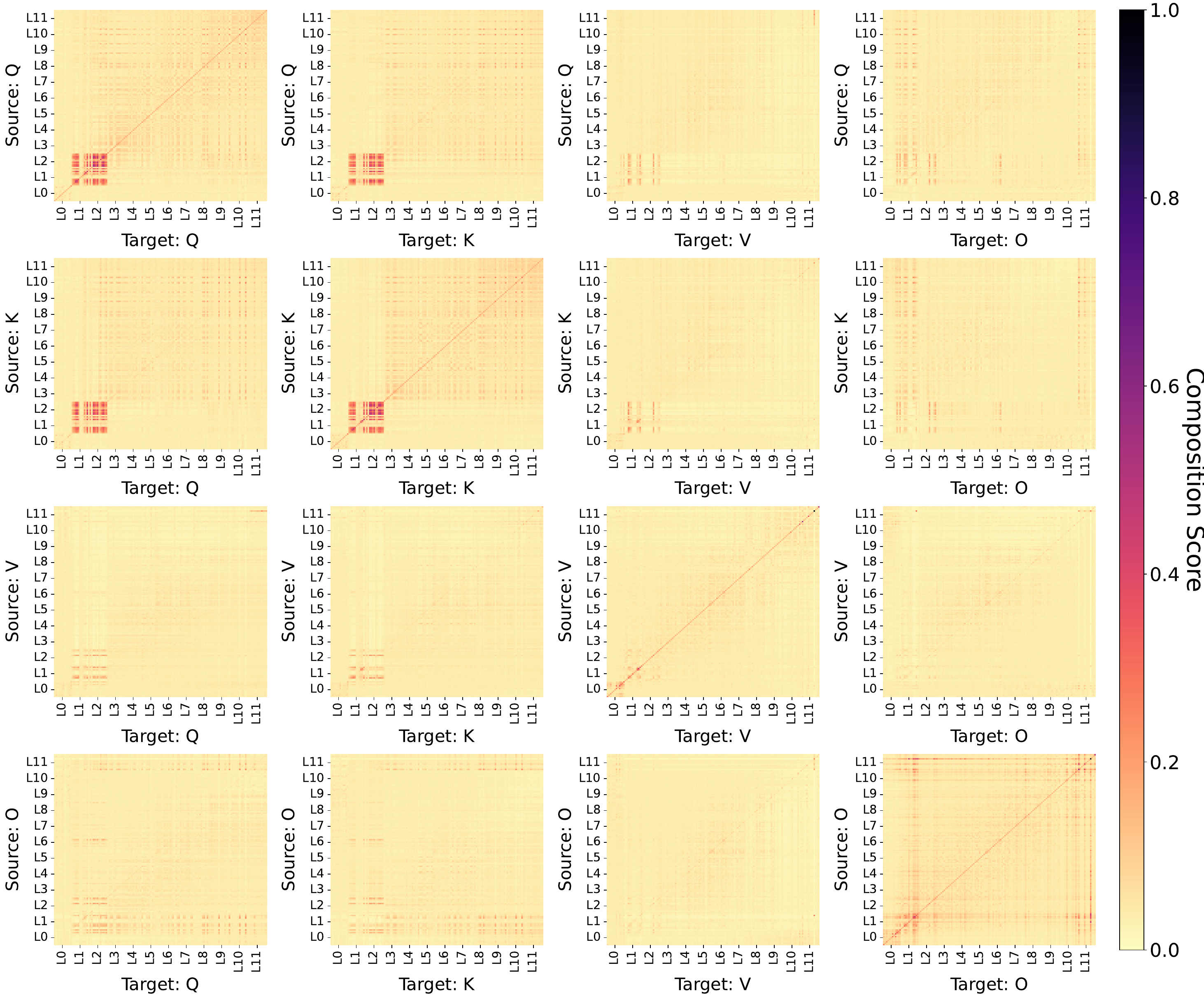}
    \caption{Composition Score (CS) with weight preprocessing between all head pairs of GPT2-small for each weight pairing type (source, target) of Q, K, V, and O, where the tick label L$\ell$ denotes layer $\ell$.}
    \label{fig:gpt2-small_heatmap_cs_weight_preproc}
\end{figure*}

As with the score heatmaps in Appendix~\ref{app:heatmap-for-all-methods}, we examine heatmaps of scores between heads for PK and CS across the 16 weight pairing types after weight preprocessing.
Figures~\ref{fig:gpt2-small_heatmap_pk_weight_preproc} and~\ref{fig:gpt2-small_heatmap_cs_weight_preproc} show the score heatmaps computed from the preprocessed weights for PK and CS, respectively.

For PK, the heatmap patterns are nearly identical to those computed from the original weights in Fig.~\ref{fig:gpt2-small_heatmap_pk} in Appendix~\ref{app:heatmap-for-all-methods}.
This observation is consistent with the correlations between the corresponding PK scores being close to $1$ (Fig.~\ref{fig:wp-pk-vs-cs-pearsonr} in Appendix~\ref{app:wp-pearsonr}).

In contrast, for CS, the heatmaps computed from the preprocessed weights differ from those computed from the original weights in Fig.~\ref{fig:gpt2-small_heatmap_cs} in Appendix~\ref{app:heatmap-for-all-methods}.
In particular, for QQ, QK, KQ, and KK, the CS scores in layers 1 and 2 become larger after preprocessing, resulting in different heatmap patterns.
This observation is also consistent with the relatively large MSE between the corresponding CS scores (Fig.~\ref{fig:pk-mse-vs-cs-mse} in Appendix~\ref{app:mse-org-vs-wp}).

\subsection{Layerwise averages of PK and CS for same-type weight pairings after weight preprocessing}\label{app:wp-layerwise-metric-trend}

Motivated by the results in Appendix~\ref{app:wp-heatmap}, we analyze the layerwise behavior of PK and CS after weight preprocessing, following Appendix~\ref{app:layerwise-metric-trend}.
Specifically, for the same-type weight pairings (QQ, KK, VV, and OO), we compute the layerwise mean scores and report them together with the overall mean across the considered head pairs as a baseline in Fig.~\ref{fig:gpt2-small_layerwise_weight_preproc_metrics}.

Compared with the results for the original weights in Fig.~\ref{fig:gpt2-small_layerwise_all_metrics}, PK exhibits nearly the same trends.
In contrast, for CS, the QQ and KK scores become larger in layers 1 and 2, reflecting the heatmap patterns observed in Fig.~\ref{fig:gpt2-small_heatmap_cs_weight_preproc} in Appendix~\ref{app:wp-heatmap}.

Although these results are limited to GPT2-small, they show that CS can change substantially with weight preprocessing for specific weight types, such as query and key weights, and in specific layers, even though the preprocessing does not change the model output.
In contrast, PK remains nearly unchanged after weight preprocessing, suggesting that it may capture intrinsic properties of the heads and produce results nearly identical to those obtained from the original weights.

\section{Revisiting whether the attention-head weights of GPT2-small have full column rank}\label{app:check-d-head}

In Section~\ref{sec:proposed}, we assumed that the column space of each attention weight $\bm{W}^r\in\mathcal{W}\subset\mathbb{R}^{d\times d_\text{head}}$ in (\ref{eq:mathcalW}) has dimension $d_\text{head}$, implying that $\bm{W}^r$ has full column rank.
In this section, we revisit this assumption.

First, Figs.~\ref{fig:gpt2-small_singular_values_W_Q},~\ref{fig:gpt2-small_singular_values_W_K},~\ref{fig:gpt2-small_singular_values_W_V}, and~\ref{fig:gpt2-small_singular_values_W_O} show the singular values of $\bm{W}_\text{Q}^h$, $\bm{W}_\text{K}^h$, $\bm{W}_\text{V}^h$, and $\bm{W}_\text{O}^h$, respectively, for each head in each layer of GPT2-small.
For almost all heads, the singular values are visibly greater than zero, although a few are close to zero.
Nevertheless, using \texttt{np.linalg.matrix\_rank}\footnote{Note that \texttt{np.linalg.matrix\_rank} determines the rank using a numerical tolerance on the singular values (see \url{https://numpy.org/doc/stable/reference/generated/numpy.linalg.matrix_rank.html} for more details).}, we confirmed that all of these weight matrices have rank $d_\text{head} = 64$.

Next, to account for differences in the magnitudes of the singular values, we measure the effective rank~\cite{DBLP:conf/eusipco/RoyV07} as a continuous-valued notion of rank, following \citet{wang2026dimensional}\footnote{
Note that their study focuses on the dimensionality of activations output by Transformer blocks rather than on weights.
}.
For $\bm{W}^r\in\mathcal{W}\subset\mathbb{R}^{d\times d_\text{head}}$, let its singular values be $\sigma_1^r\geq\ldots\geq\sigma_{d_\text{head}}^r>0$.
The effective rank is defined as follows\footnote{
As examples, we compute the effective rank when the singular values are uniform and when the largest singular value is sufficiently larger than the others.
First, consider the case $\sigma_1^r=\cdots=\sigma_{d_\text{head}}^r>0$.
Then, since $\sigma_i^r / \sum_{i'=1}^{d_\text{head}}\sigma_{i'}^r = 1 / d_\text{head}$,
\begin{align*}
\operatorname{erank}(\bm{W}^r)
=
\exp\left(
-\sum_{i=1}^{d_\text{head}}
\frac{1}{d_\text{head}}
\log\frac{1}{d_\text{head}}
\right)
=
\exp\left(
-\log\frac{1}{d_\text{head}}
\right)
= d_\text{head}.
\end{align*}
In contrast, consider the limit in which the largest singular value dominates the others, that is, $\sigma_i^r/\sigma_1^r\to 0$ for $i\neq 1$.
In this limit,
$\sigma_1^r/\sum_{i'=1}^{d_\text{head}}\sigma_{i'}^r\to 1$
and
$\sigma_i^r/\sum_{i'=1}^{d_\text{head}}\sigma_{i'}^r\to 0$
for $i\neq 1$.
Therefore,
\begin{align*}
\frac{\sigma_i^r}
{\sum_{i'=1}^{d_\text{head}}\sigma_{i'}^r}
\log\left(
\frac{\sigma_i^r}
{\sum_{i'=1}^{d_\text{head}}\sigma_{i'}^r}
\right)
\to
\begin{cases}
1\log 1 = 0, & i=1,\\
\displaystyle\lim_{x\to 0^+}x\log x = 0, & i\neq 1.
\end{cases}
\end{align*}
Thus,
\begin{align*}
\operatorname{erank}(\bm{W}^r)
\to
\exp\left(
-\sum_{i=1}^{d_\text{head}}0
\right)
=
1.
\end{align*}
Hence, the effective rank generally satisfies
\begin{align*}
1
\leq
\operatorname{erank}(\bm{W}^r)
\leq
d_\text{head}.
\end{align*}
It equals $d_\text{head}$ when the singular values are uniform, whereas it approaches $1$ in the limit where a single singular value dominates.
Thus, the effective rank continuously interpolates between these cases and can be interpreted as a measure of the effective dimensionality of the singular-value distribution.
}
\begin{align}
    \operatorname{erank}(\bm{W}^r) = \exp\left(-\sum_{i=1}^{d_\text{head}} \frac{\sigma_i^r}{\sum_{i'=1}^{d_\text{head}}\sigma_{i'}^r} \log \left(\frac{\sigma_i^r}{\sum_{i'=1}^{d_\text{head}}\sigma_{i'}^r}\right) \right).
\end{align}

\begin{figure}[t!]
    \centering
    \includegraphics[width=0.63\linewidth]{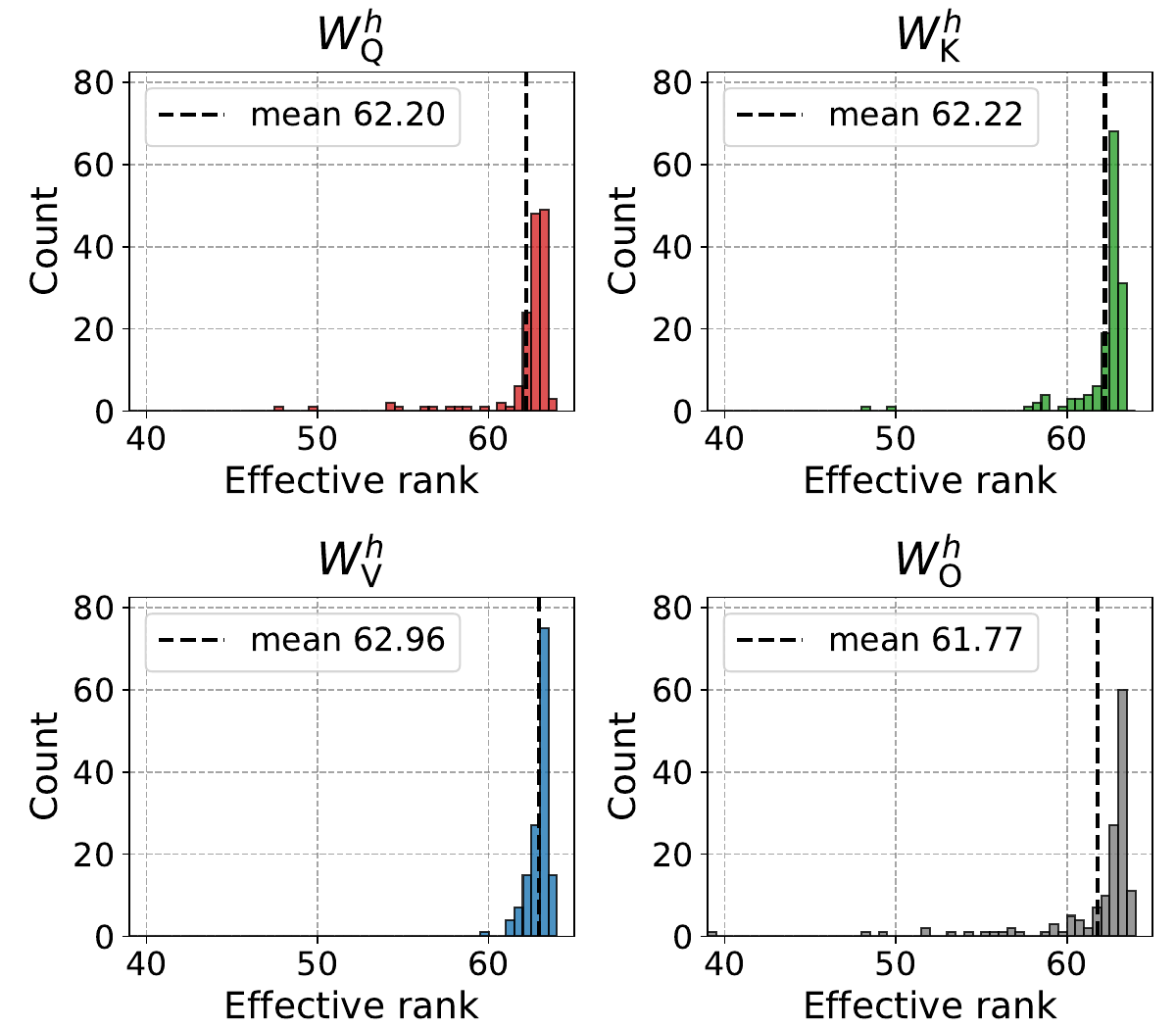}
    \caption{
Effective ranks of $\bm{W}_\text{Q}^h$, $\bm{W}_\text{K}^h$, $\bm{W}_\text{V}^h$, and $\bm{W}_\text{O}^h$ across all 144 heads of GPT2-small. 
The maximum possible effective rank is $d_{\text{head}} = 64$.
The black dashed line marks the mean of each panel.}
    \label{fig:gpt2-small_effective_rank_hist}
\end{figure}

Figure~\ref{fig:gpt2-small_effective_rank_hist} shows the results.
For each weight type, some heads have effective ranks noticeably below $d_\text{head}$, particularly for $\bm{W}_\text{O}^h$, whereas the mean effective rank remains relatively close to $d_\text{head}$.

Thus, although the assumption in Section~\ref{sec:proposed} may not hold exactly when interpreted in terms of effective dimensionality, these results provide some empirical support for its validity.

\begin{figure*}[p!]
    \centering
    \includegraphics[width=\linewidth]{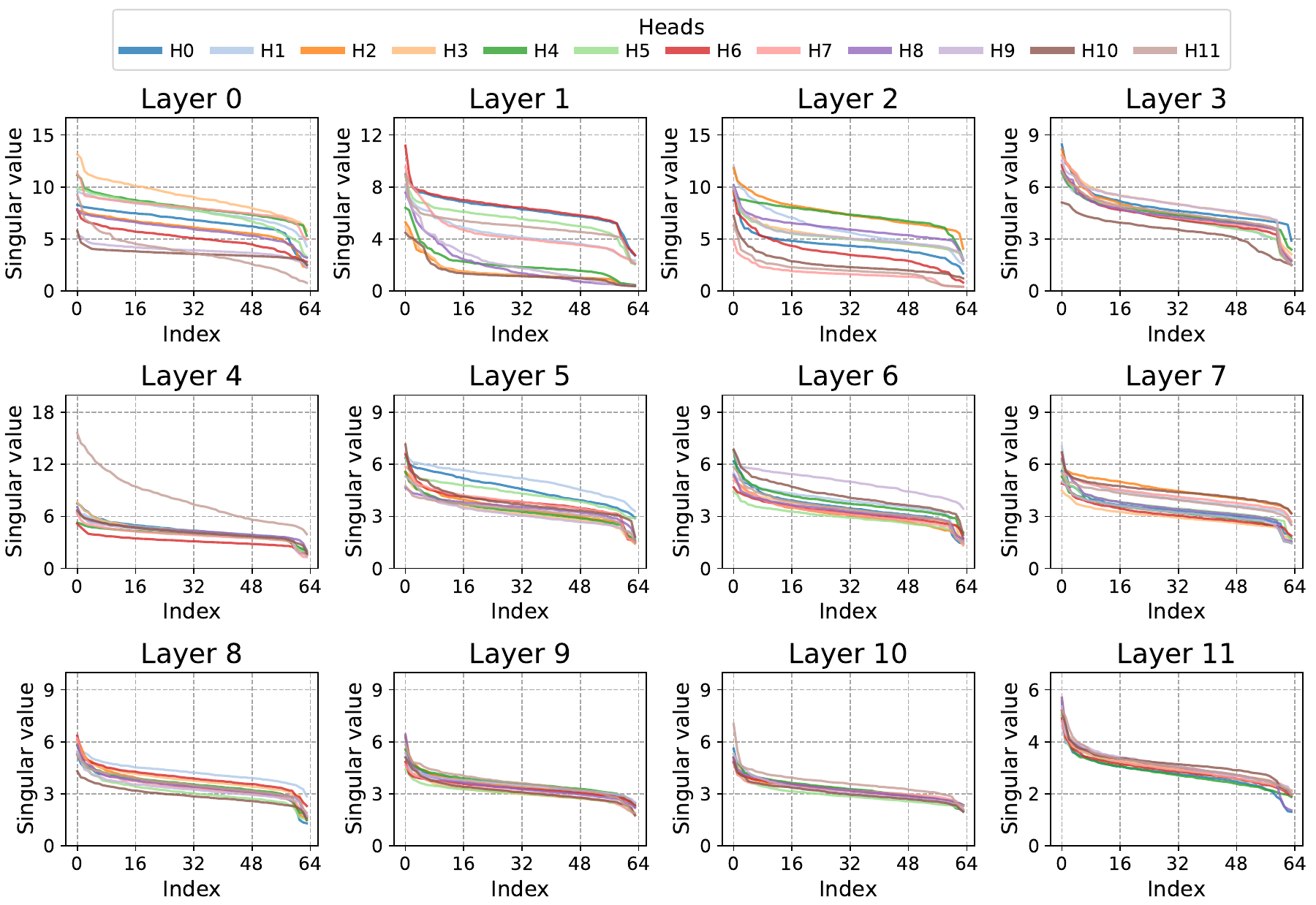}
    \caption{Singular values of GPT2-small attention-head weights ${\bm{W}_\text{Q}^h}$.}
    \label{fig:gpt2-small_singular_values_W_Q}
\end{figure*}

\begin{figure*}[p!]
    \centering
    \includegraphics[width=\linewidth]{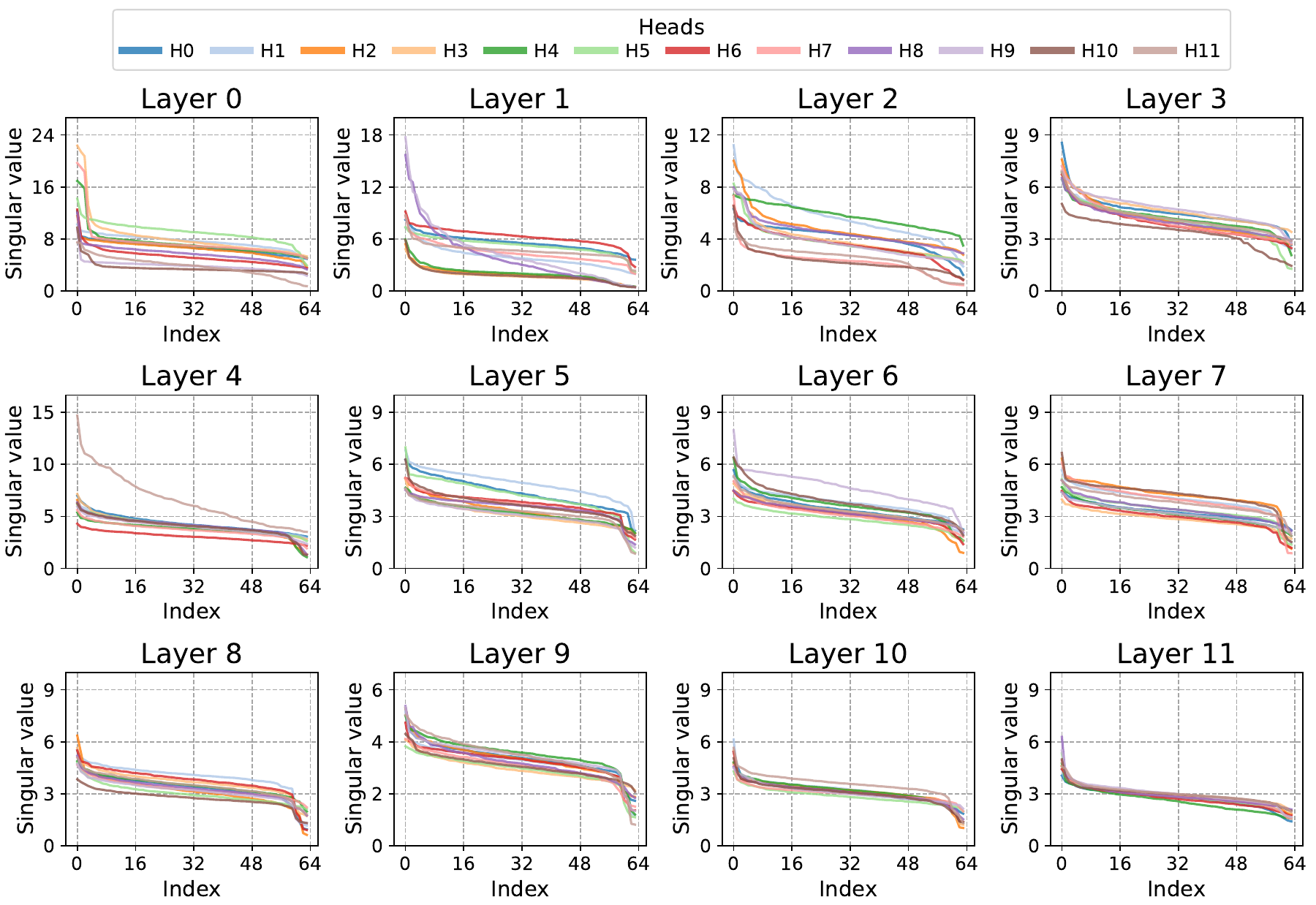}
    \caption{Singular values of GPT2-small attention-head weights ${\bm{W}_\text{K}^h}$.}
    \label{fig:gpt2-small_singular_values_W_K}
\end{figure*}

\begin{figure*}[p!]
    \centering
    \includegraphics[width=\linewidth]{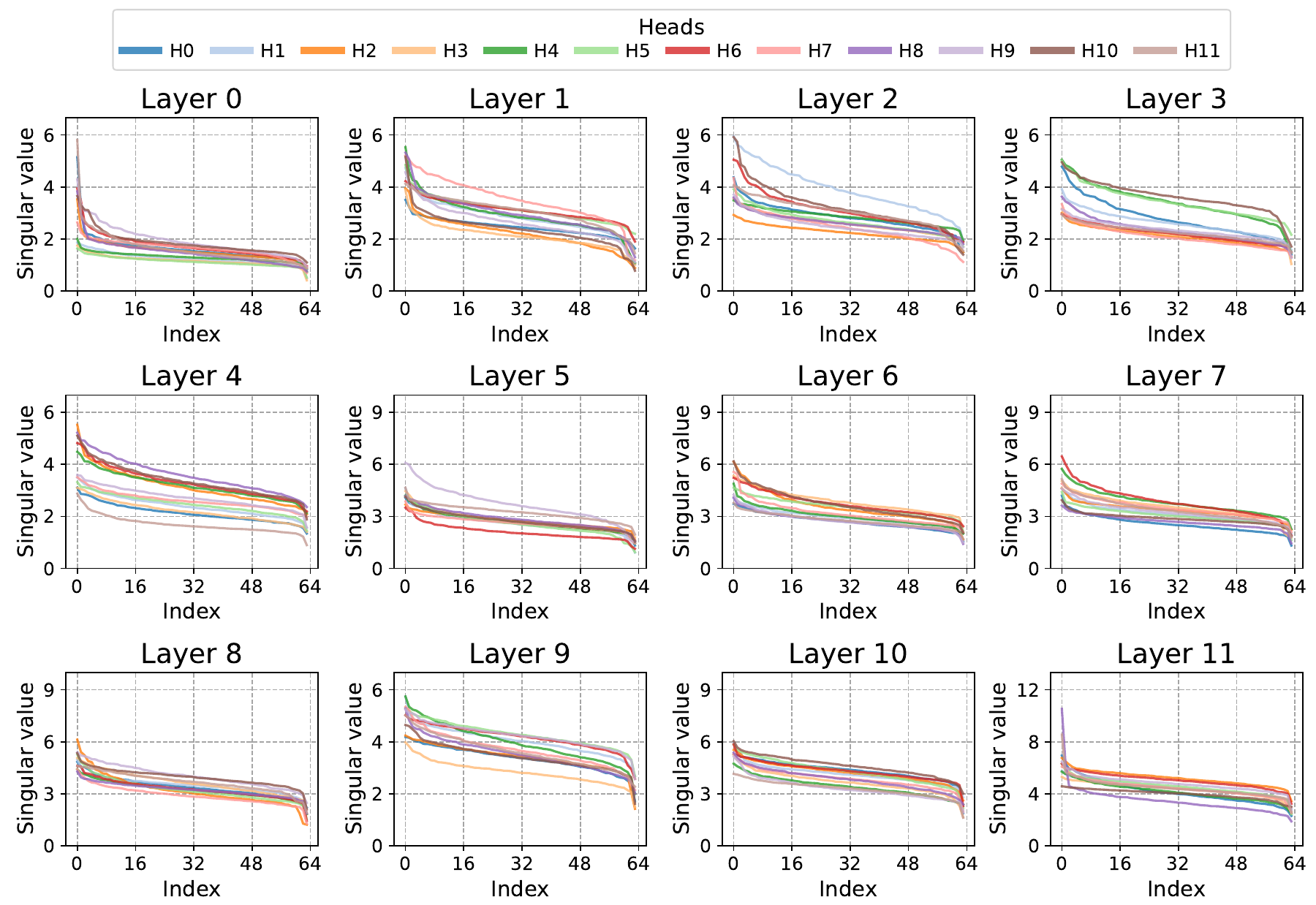}
    \caption{Singular values of GPT2-small attention-head weights ${\bm{W}_\text{V}^h}$.}
    \label{fig:gpt2-small_singular_values_W_V}
\end{figure*}

\begin{figure*}[p!]
    \centering
    \includegraphics[width=\linewidth]{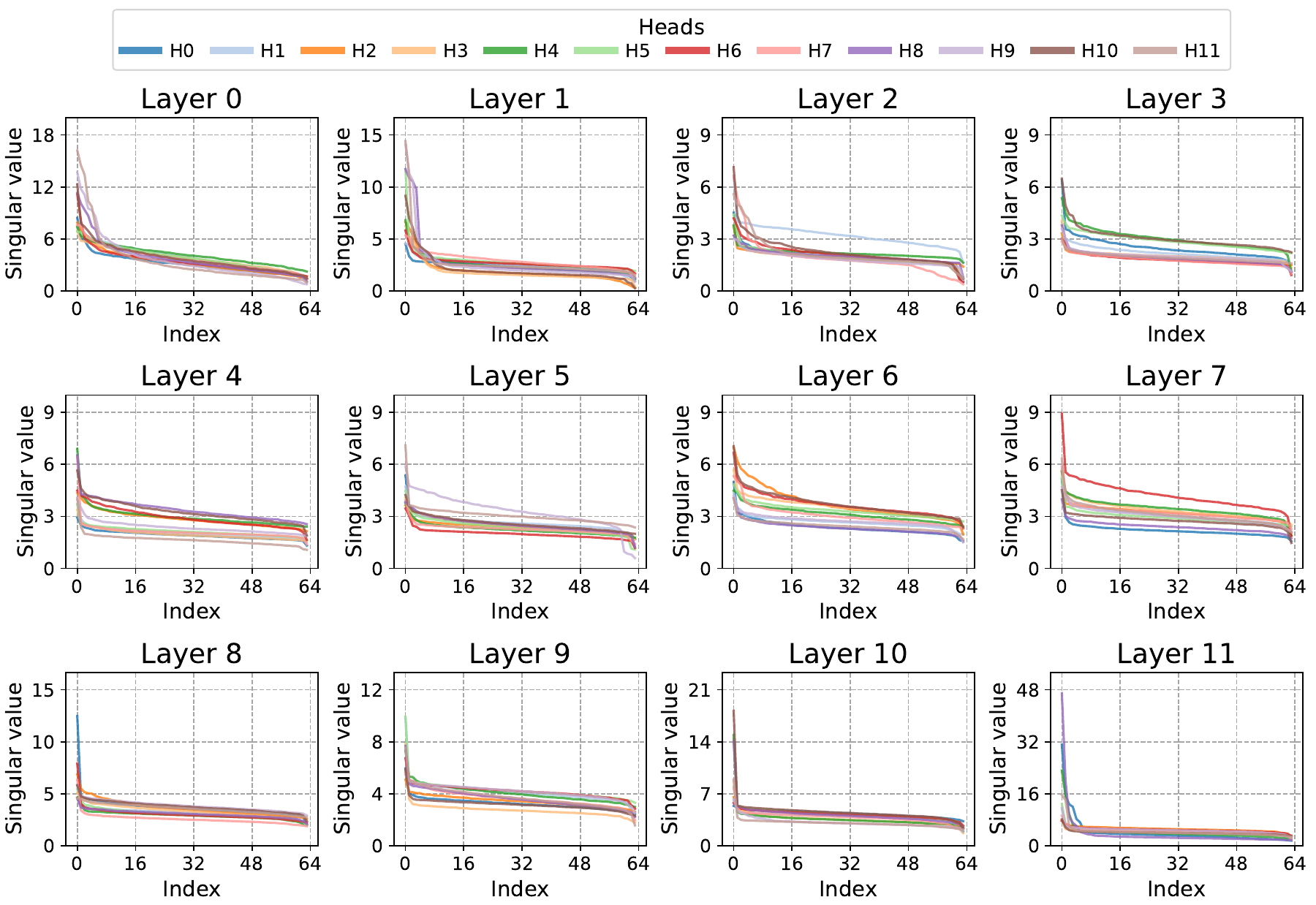}
    \caption{Singular values of GPT2-small attention-head weights $\bm{W}_\text{O}^h$.}
    \label{fig:gpt2-small_singular_values_W_O}
\end{figure*}

\end{document}